\PassOptionsToPackage{table}{xcolor}
\PassOptionsToPackage{nopatch=footnote}{microtype}
\documentclass[11pt, a4paper, copyright, nonumbering]{mll_style}

\usepackage{graphicx}
\usepackage{xcolor}
\usepackage{booktabs}
\usepackage{tabularx}
\usepackage{tabularray}
\usepackage{multirow}
\usepackage{diagbox}
\usepackage{hhline}
\usepackage{longtable}
\usepackage{adjustbox}
\usepackage{array}
\usepackage{makecell}
\usepackage{wrapfig}
\usepackage{float}
\usepackage{placeins}
\usepackage{needspace}
\usepackage{capt-of}

\newcolumntype{C}{>{\centering\arraybackslash}X}

\renewcommand{\topfraction}{0.9}
\renewcommand{\bottomfraction}{0.8}
\renewcommand{\textfraction}{0.07}
\renewcommand{\floatpagefraction}{0.8}

\usepackage{amsmath,amsfonts,amssymb}
\usepackage{bm}
\usepackage{nicefrac}
\usepackage{extarrows}
\usepackage{amsthm}

\usepackage{xspace}
\usepackage{enumitem}
\usepackage{pifont}
\usepackage{listings}
\usepackage[most]{tcolorbox}
\usepackage{fontawesome5}
\usepackage{scalefnt}
\usepackage{titletoc}
\usepackage{silence}
\usepackage{etoolbox}

\usepackage{hyperref}
\usepackage{xurl}

\providecommand{\TaskStrut}{\rule[-0.6ex]{0pt}{2.8ex}}
\providecommand{\resulttablefont}{\fontsize{6.5}{7.5}\selectfont}
\providecommand{\resulttaskheader}[1]{\TaskStrut\bfseries #1}

\providecommand{\blockheadcols}[2]{%
  \addlinespace[2pt]\multicolumn{#1}{l}{\textit{#2}}\\[1pt]%
}
\newcolumntype{L}[1]{>{\raggedright\let\newline\\\arraybackslash\hspace{0pt}}m{#1}}
\newcolumntype{R}[1]{>{\raggedleft\let\newline\\\arraybackslash\hspace{0pt}}m{#1}}

\newcommand{\ignore}[1]{}

\makeatletter
\DeclareRobustCommand\onedot{\futurelet\@let@token\@onedot}
\def\@onedot{\ifx\@let@token.\else.\null\fi\xspace}

\makeatother

\definecolor{MyBlue}{rgb}{0.46, 0.50, 0.61}
\definecolor{MyDarkBlue}{rgb}{0,0,0}
\definecolor{MyDarkGreen}{RGB}{0,0,0}
\definecolor{MyDarkRed}{rgb}{0.8,0.02,0.02}
\definecolor{MyOrange}{rgb}{1.0, 0.4, 0.2}
\definecolor{MyPurple}{RGB}{111,0,255}
\definecolor{MyRed}{rgb}{0,0,0}
\definecolor{MyGold}{rgb}{0.75,0.6,0.12}
\definecolor{MyDarkgray}{rgb}{0.66, 0.66, 0.66}
\definecolor{MyBrown}{rgb}{0.65, 0.16, 0.16}
\definecolor{MyMutedRose}{rgb}{0.58, 0.29, 0.35}

\definecolor{citecolor}{HTML}{696FAD}

\makeatletter
\DeclareRobustCommand{\benchmarkname@shape}{%
  \edef\@tempa{\f@series}\edef\@tempb{\mddefault}%
  \ifx\@tempa\@tempb\textsc{MindTopo}\else MindTopo\fi}
\newcommand{\benchmarkname}{\benchmarkname@shape\xspace}
\makeatother
\AtBeginDocument{\pdfstringdefDisableCommands{\def\benchmarkname{MindTopo}}}

\newif\ifpropositionfirstitem
\propositionfirstitemtrue

\newcommand{\myparagraph}[1]{\noindent\textbf{#1}}

\theoremstyle{definition}
\newtheorem{definition}{Definition}
\newtheoremstyle{defnbreak}{\topsep}{\topsep}{}{0pt}{\bfseries}{.}{\newline}{}
\theoremstyle{defnbreak}
\newtheorem{definitionbr}[definition]{Definition}

\definecolor{bggray}{HTML}{F5F5F5}
\definecolor{pvdblue}{HTML}{DAE8FC}
\definecolor{RoseQuartzBg}{HTML}{F7CAC9}
\definecolor{RoseQuartz}{HTML}{F5A798}
\definecolor{Serenity}{HTML}{92A8D1}
\definecolor{OrangeRed}{rgb}{1.0, 0.27, 0.0}
\definecolor{RoyalBlue}{cmyk}{1, 0.50, 0, 0}
\definecolor{Turquoise}{HTML}{0F4C81}
\definecolor{mint}{rgb}{0.24, 0.71, 0.54}
\definecolor{green}{rgb}{0.0, 0.120, 0.0}

\definecolor{takeawayblue}{HTML}{4F86E8}
\definecolor{takeawaybg}{HTML}{EEF5FD}
\newtcolorbox{keytakeaways}[1]{
    enhanced,
    colback=takeawaybg,
    colframe=takeawayblue,
    boxrule=0pt,
    leftrule=1.8pt,
    arc=1.5mm,
    left=6pt,
    right=6pt,
    top=7pt,
    bottom=3pt,
    before skip=6pt,
    after skip=6pt,
    attach boxed title to top left={xshift=8pt,yshift=-1.6mm},
    boxed title style={
        colback=takeawayblue,
        colframe=takeawayblue,
        boxrule=0pt,
        arc=1.2mm,
        left=3pt,
        right=3pt,
        top=0.5pt,
        bottom=0.5pt
    },
    coltitle=white,
    fonttitle=\small\fontseries{b}\selectfont,
    fontupper=\small,
    title={\faLightbulb\ Key Takeaways: #1}
}

\newdimen\abovecrulesep
\newdimen\belowcrulesep
\makeatletter
\patchcmd{\@@@cmidrule}{\aboverulesep}{\abovecrulesep}{}{}
\patchcmd{\@xcmidrule}{\belowrulesep}{\belowcrulesep}{}{}
\makeatother

\definecolor{mybluetitle}{HTML}{4B527E}

\definecolor{codegreen}{HTML}{478058}%
\definecolor{codegray}{rgb}{0.5,0.5,0.5}
\definecolor{codepurple}{HTML}{4F5E80}
\definecolor{backcolour}{rgb}{0.95,0.95,0.92}
\lstdefinestyle{mystyle}{
    backgroundcolor=\color{backcolour},
    commentstyle=\color{codegreen},
    keywordstyle=\color{magenta},
    numberstyle=\tiny\color{codegray},
    stringstyle=\color{codepurple},
    basicstyle=\ttfamily\scriptsize,
    breakatwhitespace=false,
    breaklines=true,
    captionpos=b,
    keepspaces=true,
    frame=none,
    numbersep=5pt,
    showspaces=false,
    showstringspaces=false,
    showtabs=false,
    tabsize=2
}

\newtcolorbox{promptbox}[2][]{
    enhanced, 
    breakable,
    center title,
    left*=0pt, right*=0pt,
    boxsep=2pt, left=5pt, right=5pt,
    skin first=enhanced,
    skin middle=enhanced,
    skin last=enhanced,
    colback  = backcolour,
    fonttitle=\bfseries\rmfamily,
    fontupper=\scriptsize,
    title={\footnotesize\strut{#2}},
    #1
    }

\newtcolorbox{onebox}[2][]{
    enhanced, 
    center title,
    left*=0pt, right*=0pt,
    boxsep=2pt, left=5pt, right=5pt,
    skin first=enhanced,
    skin middle=enhanced,
    skin last=enhanced,
    colframe = mybluetitle!90,
  colback  = mybluetitle!10,
    fonttitle=\bfseries\rmfamily\fontfamily{phv}\selectfont,
    title={\footnotesize\strut{#2}  \refstepcounter{subsubsection} \addcontentsline{toc}{subsubsection}{\string\numberline{\thesubsubsection}#2}
    },
    #1
    }

\newcommand{\name}{\benchmarkname}

\usepackage[numbers]{natbib}
\usepackage{setspace}
\usepackage{xltabular}
\usepackage{threeparttable}
\usepackage{siunitx}

\usepackage{mathtools}
\usepackage{dsfont}

\usepackage{subcaption}
\usepackage{tikz}
\usepackage{pgfplots}
\pgfplotsset{compat=1.16}

\usepackage{cleveref}
\hypersetup{colorlinks=true}

\usepackage{CJKutf8}

\usepackage[toc,page,header]{appendix}

\titlespacing*{\section}{0pt}{2ex plus 2pt minus 2pt}{4pt}
\titlespacing*{\subsection}{0pt}{1.5ex plus 2pt minus 1pt}{2pt}

\definecolor{tableblue}{RGB}{201,226,239}

\providecommand{\resulttablefont}{\fontsize{6.5}{7.5}\selectfont}
\providecommand{\resulttaskheader}[1]{\TaskStrut\bfseries #1}

\makeatletter
\def\@BTrule[#1]{%
  \ifx\longtable\undefined
    \let\@BTswitch\@BTnormal
  \else\ifx\hline\LT@hline
    \nobreak
    \let\@BTswitch\@BLTrule
  \else
     \let\@BTswitch\@BTnormal
  \fi\fi
  \global\@thisrulewidth=#1\relax
  \ifnum\@thisruleclass=\tw@\vskip\@aboverulesep\else
  \ifnum\@lastruleclass=\z@\vskip\@aboverulesep\else
  \ifnum\@lastruleclass=\@ne\vskip\doublerulesep\fi\fi\fi
  \@BTswitch}
\makeatother

\addto\extrasenglish{
}

 {\begin{list}{}%
         {\setlength{\leftmrargin}{#1}}%
         \item[]%
 }
 {\end{list}}

\reportnumber{001}

\definecolor{firstgrey}{gray}{0.66}
\definecolor{secondgrey}{gray}{0.88}

\fancypagestyle{firststyle}{
    \fancyhead[L]{%
        \raisebox{-2pt}{\includegraphics[height=34pt]{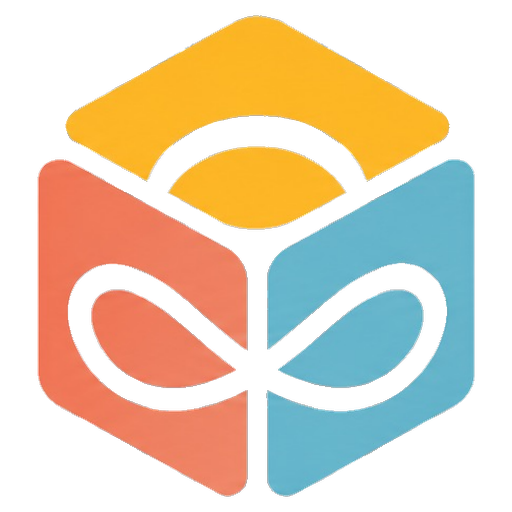}}%
        \hspace{6pt}{\sffamily\bfseries\LARGE MindTopo}%
    }
    \fancyhead[C]{}
    \fancyhead[R]{%
        \includegraphics[height=30pt]{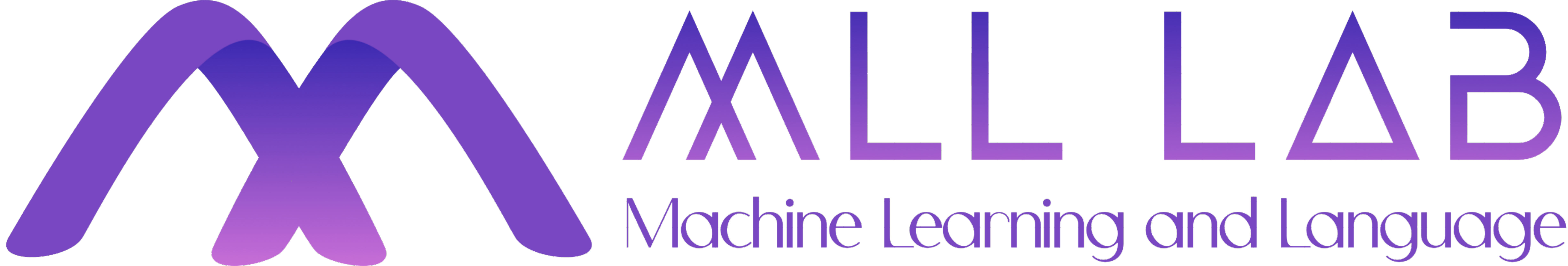}%
    }
}

\title{\centering \name: Can Foundation Models Reason in Topological Space?}

\date{}

\author{
Yunfei Ge*$^{1}$, Anbang Liu*$^{1}$, Qineng Wang*$^{1\dag}$, Johnalbert Garnica*$^{1}$, Jianwen Lyu$^{1}$,
Zihan Wang$^{1}$, Reuben Tan$^{2}$, Jianfeng Gao$^{2}$, Ruohan Zhang$^{1,3}$, Yining Hong$^{3}$,
Jiajun Wu$^{3}$, Manling Li$^{1}$
\\
\small $^1$Northwestern University~~~$^2$Microsoft Research~~~$^3$Stanford University
\\
\footnotesize \emph{* Equal contribution. $\dag$ Project lead.}
\\
{
    \footnotesize 
    \href{https://mind-topo.github.io}{\faGlobe~Website}
    \quad 
    \href{https://github.com/mll-lab-nu/MindTopo}{\faGithub~Code}
    \quad
    \href{https://huggingface.co/datasets/MLL-Lab/MindTopo}{\faDatabase~Dataset}
}
\vspace{-10pt}
}

\begin{document}
\begin{abstract}
Spatial reasoning depends on metric properties such as distance, angle, and shape, and on topological relations that remain invariant under continuous deformation. Cognitive science identifies these relations as foundational to spatial understanding, yet foundation-model evaluations largely focus on metric or viewpoint-dependent relations. We introduce \textbf{\benchmarkname}, a benchmark of \textbf{topological intuition} across five properties grounded in cognitive science and formal topology, namely continuity, separation, order, enclosure, and knots. \benchmarkname{} evaluates each property at two cognitive levels. \textbf{Reasoning} asks a model to identify topological relations or infer how they change. \textbf{Planning} instantiates a foundation model as a closed-loop agent whose policy selects environment actions. \benchmarkname{} provides a Core Set of 3{,}900 instances across 13 procedurally generated task types, split into an Easy and a Hard level, and an Extended Set of 11{,}035 original instances. On the Core Set, we benchmark 13 multimodal large language models (MLLMs), and every MLLM performs better on reasoning than on planning. On the Easy level, the strongest model trails humans in both reasoning and planning. On Qwen3-VL-2B-Instruct, supervised fine-tuning followed by reinforcement learning raises average reasoning accuracy from 14.24\% to 51.53\% and open-loop planning success from 0.20\% to 6.33\%. We also study agent configurations augmented with image and video generation and evaluate rollouts from seven video generative models in planning environments. Generated videos achieve final-state success in up to 35.2\% of rollouts, yet none of the 1{,}890 evaluated video rollouts passes all process-validity checks.
\end{abstract}

\maketitle
\begin{figure*}[htbp]
    \centering
    \includegraphics[width=\linewidth]{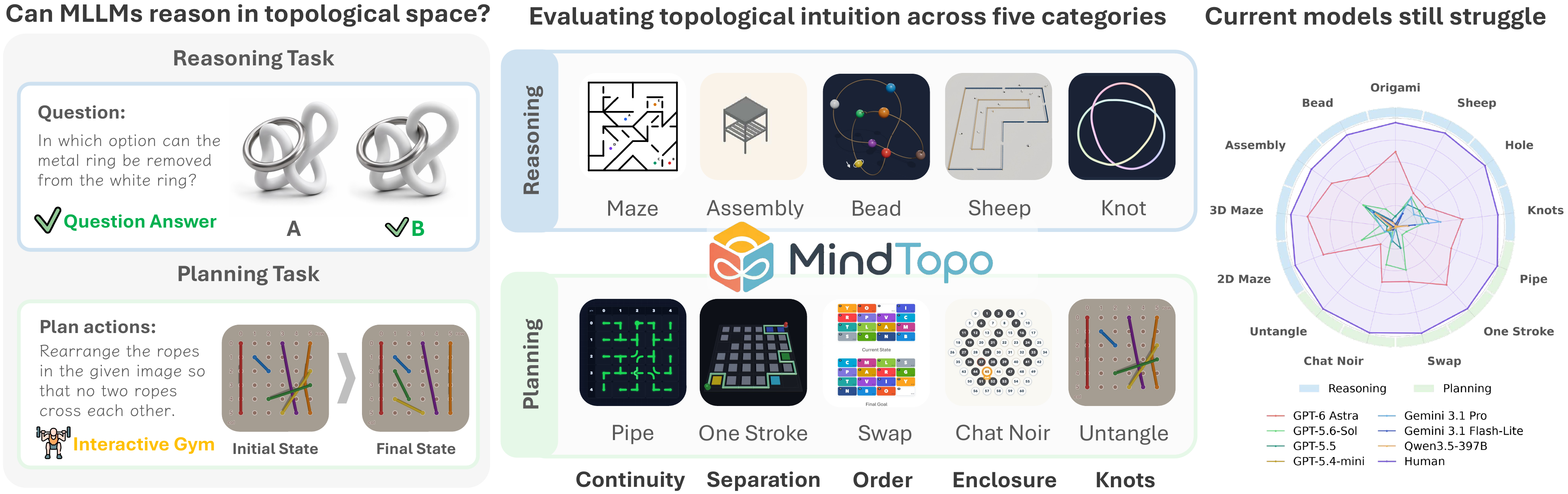}
    \caption{\textbf{Overview of \benchmarkname.} Five topological properties (\textbf{continuity}, \textbf{separation}, \textbf{order}, \textbf{enclosure}, \textbf{knots}) probed through \emph{reasoning} and \emph{planning}. The radar shows selected models on the full Core Set and humans on the Easy level.}
    \label{fig:teaser}
\end{figure*}

\section{Introduction}
\label{sec:intro}
\begin{minipage}{\textwidth}
\raggedright
\small\textit{``Representational space has to reconstruct, on its own plane and in the same order of succession, the elementary spatial relationships — first topological, then euclidean and projective.''}\\[0.3em]
{\raggedleft\footnotesize Jean Piaget \& B\"arbel Inhelder, \textit{The Child's Conception of Space} (1956)\par}
\end{minipage}

Spatial reasoning in the physical world involves more than Euclidean or metric properties, such as distance, angle, and shape. Consider bending, stretching, or coiling a closed rope loop without cutting it or passing one segment through another. Its distances, angles, and shape can change greatly while its knot type remains the same. Topology studies spatial relations of this kind, which remain invariant under continuous deformation~\cite{hatcher2002algebraic,rolfsen1976knots}. Recognizing the invariant is only the beginning. When a foundation model must act, it must predict when an intervention will alter the relation and select actions that produce a desired one. We use \textbf{topological intuition} for this ability to perceive, infer, and operate on topological relations.

Cognitive science places these relations at the foundation of spatial understanding. Piaget argued that topological relations precede Euclidean and projective relations in spatial development~\cite{piaget2013child}. Studies in adult vision similarly found that the visual system extracts topological invariants before Euclidean features~\cite{chen1982topological,chen2005topological}. Together, these findings motivate the evaluation of topology as a distinct layer of spatial reasoning. Yet spatial benchmarks for foundation models largely test metric or viewpoint-dependent relations such as distance, direction, size, shape, and three-dimensional location~\cite{chen2024spatialvlm,yang2025thinking,wang2025mindcube}. Recent work isolates path connectivity, curve containment, topological grid puzzles, or knot reasoning~\cite{dao2025alphamaze,mohseni2026curvebench,maniparambil2026topobench,chen2025knot}, but does not show whether models can reason across a broader set of topological relations and act on them. We therefore ask \textit{to what extent current foundation models can reason about topological structure and plan actions that transform it}.

To answer this question, we introduce \benchmarkname{} (Figure~\ref{fig:teaser}), a benchmark organized around five properties informed by Piaget's classification and later cognitive work~\cite{piaget2013child,strohecker1991knot,martin1976analysis}. \textbf{Continuity} and \textbf{separation} describe how a scene divides into connected parts. \textbf{Order} records the sequence of marked elements along a path or boundary. \textbf{Enclosure} describes the division between inside and outside, including holes. \textbf{Knots} capture entanglement that persists under deformation. Formal topology supplies the invariants and related structural targets used to instantiate these properties~\cite{hatcher2002algebraic,rolfsen1976knots,huntington1916cyclic}. The cognitive taxonomy determines what \benchmarkname{} tests, while the formal targets determine how its tasks are generated and evaluated.

The same property can pose different cognitive demands depending on what a model must do with it. \benchmarkname{} therefore evaluates each property at two cognitive levels. \textbf{Reasoning} tasks present one or more rendered scenes and ask the model to identify a topological relation or infer how a specified change affects it. \textbf{Planning} tasks instantiate a foundation model as a closed-loop agent in an interactive environment. At each step, the agent receives a rendered observation. Its \textbf{policy} selects an action that builds, preserves, or alters the corresponding structure. A foundation model may recognize that two ropes cross yet fail to plan the moves that remove every crossing. The two cognitive levels thus test whether reasoning about a topological relation transfers to action. Each task type comes from a parametric generator with ground truth computed from the scene state. The Core Set contains 3{,}900 instances across 13 task types, comprising 2{,}400 reasoning questions and 1{,}500 planning episodes, split evenly between an Easy and a Hard level. The Extended Set preserves the 11{,}035 original instances and partially overlaps Core. We report Extended results on an 8{,}780-instance evaluation subset after removing exact-input overlaps with Core and within-set duplicates.

This design lets us compare reasoning with action. We benchmark 13 multimodal large language models (MLLMs) on the Core Set, and every MLLM scores higher on reasoning than on planning. On the Easy level, the strongest model trails humans in both reasoning and planning. We also examine whether supervised fine-tuning and reinforcement learning can reduce these deficits on Qwen3-VL-2B-Instruct. Supervised fine-tuning followed by reinforcement learning gives the strongest average performance, but gains are much larger for reasoning than for open-loop planning. Reinforcement learning across reasoning tasks produces uneven gains on held-out tasks. The remaining gap between reasoning and planning raises a further question, namely whether visual prediction can help a foundation-model agent preserve topology while acting. We study agent configurations augmented with image and video generation and evaluate rollouts from seven video generative models in planning environments. Generated videos achieve final-state success in up to 35.2\% of rollouts, yet none of the 1{,}890 evaluated video rollouts passes all process-validity checks.

Overall, our contributions are fourfold. First, we formulate topological intuition as an evaluation target for spatial reasoning, grounding five properties in cognitive science and formal topology. Second, we introduce a benchmark of 13 procedurally generated task types, with a 3{,}900-instance Core Set and an 11{,}035-instance Extended Set, that probes these properties through reasoning and planning at two difficulty levels. Third, we benchmark 13 MLLMs on the Core Set and evaluate agent configurations augmented with image and video generation, with rollouts from seven video generative models, revealing a gap between reasoning about topological structure and acting on it. Fourth, we examine supervised fine-tuning and reinforcement learning on Qwen3-VL-2B-Instruct, showing that training improves reasoning more than planning and produces uneven gains on held-out tasks.

\begin{figure}
    \centering
    \includegraphics[width=\linewidth]{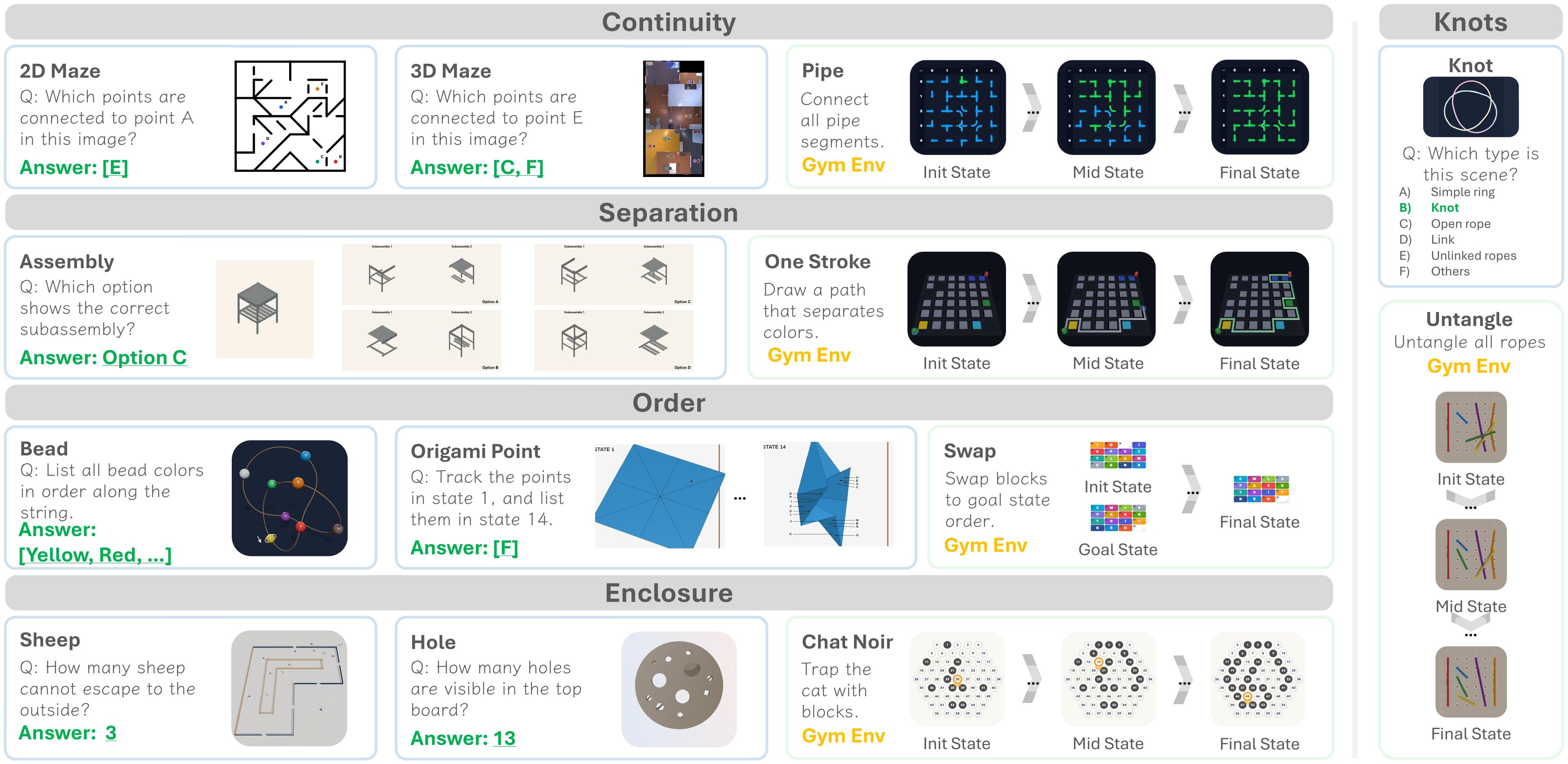}
    \caption{\textbf{\benchmarkname~task overview.} The 13 tasks by topological property and cognitive level. \textbf{Gym Env} marks the interactive planning environments, and the remaining tasks pair a rendered scene with a visual question.}
    \label{fig:overview}
\end{figure}

\section{\benchmarkname Benchmark}
\label{sec:benchmark}

\subsection{Problem Formulation}
\label{sec:formulation}

Following Piaget's account of topological space~\cite{piaget2013child}, we study topological intuition as one part of spatial reasoning. It is the ability to recognize and use spatial relations that remain unchanged when a scene is continuously deformed. \benchmarkname{} evaluates this ability at two cognitive levels~\cite{piaget2013child,martin1976analysis}, \emph{reasoning} and \emph{planning}.

\myparagraph{Scenes and properties.}
Each instance begins with a parametric scene state $s$. The state determines the spatial structure and the marked elements relevant to the task. A renderer produces one or more visual observations $o=\mathrm{render}(s)$.
We write $X(s) \in \mathcal{X}$ for the resulting embedded structure, including its task-relevant marks, and $\mathcal{V}$ for the task-dependent value space of a property $\Phi\colon\mathcal{X}\to\mathcal{V}$.

A \textbf{topological property} $\Phi$ assigns a value to an embedded structure and remains unchanged under ambient isotopy. In other words,
{\color{MyDarkBlue}
\begin{equation}
X \simeq X' \quad \Longrightarrow \quad \Phi(X)=\Phi(X').
\end{equation}
}%
Here $X \simeq X'$ means that one structure can be continuously deformed into the other without cutting, joining, or passing one part through another~\citep[Ch.\,1]{rolfsen1976knots}. Appendix~\ref{app:formal_defs} gives the exact target of each task.

\myparagraph{Reasoning.}
A reasoning instance is a triple $(o,q,a^*)$. The question $q$ asks a foundation model to identify a topological relation in the observation or predict how it changes after a stated edit. The model returns
{\color{MyDarkBlue}
\begin{equation}
\hat a=M(o,q),
\end{equation}
}%
and the prediction is correct when $\hat a=a^*$. The reference answer $a^*$ is computed from the scene state. Scalar and categorical answers are scored by exact match, while unordered answers are compared as sets.

\myparagraph{Planning.}
A planning episode starts from a scene state $s_0$, a task instruction $u$, and a horizon $H$. At step $t$, the environment renders $o_t=\mathrm{render}(s_t)$.
Let $h_t$ denote the interaction history made available to the evaluated configuration through step $t$, including the current observation and any retained previous observations and actions. An evaluated configuration $C$ built around a foundation model induces the policy
{\color{MyDarkBlue}
\begin{equation}
a_t \sim \pi_C(\,\cdot \mid u,h_t).
\end{equation}
}%
The simulator then applies the selected action and updates the scene state according to
{\color{MyDarkBlue}
\begin{equation}
s_{t+1} \sim P(\,\cdot \mid s_t,a_t).
\end{equation}
}%
Together, the configuration and the environment interface form the agent evaluated by \benchmarkname{}, which acts in a closed loop. The episode succeeds if its goal predicate $g$ is satisfied in some state $s_t$ with $t\leq H$. The policy belongs to the complete configuration, whether it uses an MLLM alone or also uses image or video generation.

Reasoning tests whether a foundation model can identify or predict a topological relation. Planning tests whether that relation can guide a valid sequence of actions.

\subsection{Topological Properties}
\label{sec:properties}

Figure~\ref{fig:overview} shows representative scenes, questions, and gym environments for every task in the suite. Each property below describes its cognitive grounding, our reasoning task, and the matched planning task. Appendix~\ref{app:cogsci} expands the cognitive-science grounding of the five properties, and Appendix~\ref{app:benchmark} gives the full specification of every task, including scene generation, difficulty parameterization, and metrics.

\noindent\textbf{Continuity.} Continuity captures whether a path or surface forms an unbroken whole, and is among the earliest spatial concepts a child acquires~\cite{piaget2013child}. It underlies many practical reasoning tasks, such as navigating a maze, deciding whether a wire is severed, or threading a cable through an opening. Our two reasoning tasks, 2D Maze and 3D Maze, ask which of several marked points in a rendered maze or house are reachable from a designated target point, so the answer hinges on topological cuts rather than on metric layout. We additionally include \emph{what-if} sub-questions that ask which wall removals or door openings reconnect two disconnected points. The matched planning task places the agent in a grid of rotatable pipe segments. Through $90^{\circ}$ rotations, it must connect every pipe segment back to the source, probing whether it can \emph{construct} continuity as well as recognize it.

\noindent\textbf{Separation.} Separation is the complement of proximity~\cite{martin1976analysis} and undergirds object individuation~\cite{piaget2013child}. Our reasoning task, \emph{Assembly}, is a furniture subassembly judgment. Given a fully assembled object and five candidate two-part decompositions, the model selects the one whose two subassemblies together reconstruct the object. The planning counterpart is the One Stroke partitioning environment, in which the agent draws a single corner-to-corner path through a colored grid that keeps same-colored cells in a shared region and separates different colors, operationalizing the act of \emph{creating} separation between previously contiguous regions.

\noindent\textbf{Order.} Order captures the sequential arrangement of elements along a path or boundary~\cite{piaget2013child}. It is essential whenever a model must track \emph{which comes before which} under a transformation of the scene. Our first reasoning task, Bead, adapts Piaget's bead-replication paradigm. Given a curved or twisted string of colored beads, the model enumerates the sequence from a designated starting bead, or decides whether two strings are identical, reversed, cyclically rotated, or different. \emph{Origami Point} asks which lettered dots in the final image of a folding sheet are the dots shown on the flat sheet. The accompanying planning environment, Swap, asks the agent to reach a target arrangement of lettered blocks by repeatedly swapping a chosen block with the single empty cell.

\noindent\textbf{Enclosure.} Enclosure is the inside/outside relation induced by a closed boundary, and is identified by Piaget as the topological origin of three-dimensional ``insideness''~\cite{piaget2013child}. We instantiate two reasoning tasks. \emph{Sheep} shows numbered sheep in a tilted view of nested or partitioned fences and asks how many cannot escape, which can escape, which cell holds the most sheep, or how many gaps must be repaired. \emph{Hole} asks the model to count the through-holes in a solid board. The matched planning task is \emph{Chat Noir}. On a hexagonal grid, the agent places one block per turn so as to encircle a moving cat before it escapes to the boundary, requiring forward reasoning about whether a partial boundary can still be closed.

\noindent\textbf{Knots.} Piaget originally subsumed knots under enclosure, but subsequent cognitive evidence supports treating them as a distinct ability. Strohecker characterizes knots as the ``mother structure'' that coordinates all other topological relations~\cite{strohecker1991knot}. Croom and Firestone show that humans reason about knots far worse than they perceive them, dissociating knot understanding from domain-general physical reasoning~\cite{croom2024tangled}. Our reasoning task presents one or more rendered ropes and loops and asks a battery of topology questions. They ask whether a single loop is truly knotted or just visually tangled, whether two loops are linked, and, in \emph{what-if} variants, which rings become free after a specified ring is cut and removed. The matched planning environment is \emph{Untangle}. Given a board of plug positions joined by ropes, the agent moves plugs across a discrete grid until no two ropes cross.

\subsection{Data Collection and Statistics}
\label{sec:data_collection}

\begin{figure}[t]
    \centering
    \includegraphics[width=\linewidth]{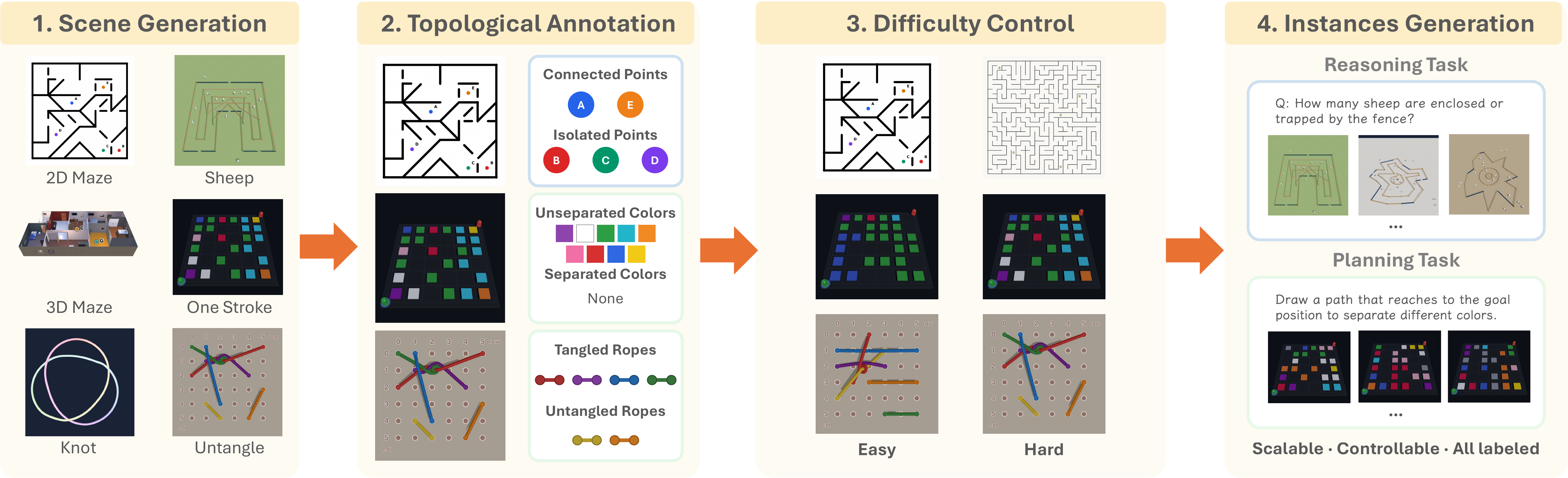}
    \caption{\textbf{Fully automated data collection pipeline for \benchmarkname.}
    Acceptance checks specific to each task filter invalid or
    ambiguous instances before export.}
    \label{fig:collection_pipeline}
\end{figure}

Figure~\ref{fig:collection_pipeline} shows the four-stage pipeline that
builds every task in \benchmarkname. All four stages are automated and use fixed
seeds. \textit{Scene Generation} samples scene states and renders them using
simulators or rendering scripts specific to each task.
\textit{Topological Annotation} refers to programmatic labeling rather than
human annotation. Code specific to each task derives the reference answer or
success condition directly from the generated state and metadata. Depending
on the task, this computation uses graph search, geometric membership tests,
known construction metadata, or simulator predicates.
\textit{Difficulty Control} gives each task an Easy and a Hard level of 150
instances, and the two levels differ in their generator settings
(Table~\ref{tab:app_level_construction}). For 11 tasks, the Easy level keeps
150 items of the hard tier of the previous release. \textit{Instance Generation} converts each
accepted state into either a templated reasoning question or a planning
episode with an initial state, goal, and success condition. Reasoning and
planning tasks are matched by topological property, but their scenes
are produced separately by their respective task generators.

\noindent\textbf{Quality control.}
Before export, each generator applies construction and acceptance criteria
specific to its task. Depending on the task, these criteria enforce valid
scene structure, reject ambiguous or unsolvable configurations, check
visibility when the task depends on visible marks or components, and confirm
that the generated answer or success condition agrees with the accepted
state. Each exported record retains its generation seed and configuration
metadata. Planning records also retain the initial state and action budget.
Appendix~\ref{app:benchmark} describes the scene generation, difficulty
settings, output formats, scoring rules, and applicable acceptance criteria
for every task.

\noindent\textbf{Shortcut and camera controls.}
Appendix~\ref{app:shortcut_controls} compares full input with text-only,
answer-prior, appearance-only, and symbolic-input controls, and
Appendix~\ref{app:camera} describes the controlled camera and viewpoint factors.

\Needspace{0.28\textheight}
\begin{wrapfigure}{R}{0.49\linewidth}
    \vspace{-0.8em}
    \centering
    \includegraphics[width=\linewidth]{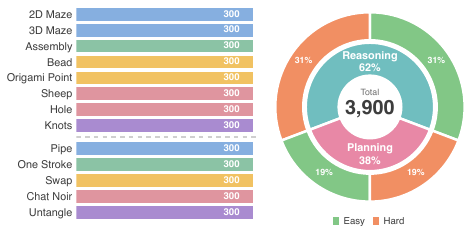}
    \caption{Core Set statistics for \benchmarkname.}
    \label{fig:data_stats}
    \vspace{-0.8em}
\end{wrapfigure}
Figure~\ref{fig:data_stats} summarizes the Core Set. The \textbf{Core Set} contains 3{,}900 instances across the 13 task types and the five topological properties. Each of the eight reasoning tasks (2D Maze, 3D Maze, Assembly, Bead, Origami Point, Sheep, Hole, and Knots) and the five planning tasks (Pipe, One Stroke, Swap, Chat Noir, and Untangle) contributes 300 instances, 150 at each level. The \textbf{Extended Set} preserves the complete original dataset of 11{,}035 instances in the code repository. It shares 2{,}250 exact model inputs with Core. For the Extended evaluation in Table~\ref{tab:extended_summary}, we exclude these overlaps and five repeated Swap inputs, yielding an 8{,}780-instance evaluation subset. Table~\ref{tab:extended_summary} reports the sizes and scores of this subset, and Appendix~\ref{app:extended_data} defines the filtering rule and records traceability. Per-task counts appear in Appendix~\ref{app:task_catalog}, and the evaluation design in Appendix~\ref{app:eval_design}.

\section{Experiments}
\label{sec:experiments}

\subsection{Experimental Setups}
\label{sec:experimental_setups}

\noindent\textbf{Models.} On the Core Set, we benchmark 13 hosted MLLMs, seven proprietary and six open-weight. The proprietary models are GPT-6 Astra~\cite{openai2026gpt6astra}, GPT-5.6-Sol~\cite{openai2026gpt56}, GPT-5.5~\cite{openai2026gpt55}, GPT-5.4-mini~\cite{openai2026gpt54mini}, Gemini-3.1-Flash-Lite~\cite{google2026gemini31flashlite}, Gemini-3.1-Pro~\cite{google2026gemini31pro}, and Claude-Haiku-5.5~\cite{anthropic2026claudehaiku55}. The open-weight models are Nemotron-Nano-12B-VL-v2~\cite{nvidia2025nemotronnanov2vl}, Qwen3.5-397B-A17B~\cite{qwen2026qwen35}, Llama-4-Maverick-17B-128E~\cite{meta2025llama4}, Ministral-3-14B-Instruct-2512~\cite{mistral2025ministral3}, InternVL3.5-241B~\cite{wang2025internvl35}, and Gemma-4-31B-IT~\cite{google2026gemma4}. We additionally report a random-chance baseline, the expected accuracy of a uniformly random legal answer for reasoning tasks and the success rate of uniformly random legal actions under the same step budget for planning tasks. Model identifiers, decoding configuration, and prompt templates appear in Appendix~\ref{app:models_prompts}.

\noindent\textbf{Human evaluation.} Five professional annotators evaluated the previous release of the benchmark under the same instructions as the models, with one completed annotation per example. Figure~\ref{fig:teaser} plots human accuracy from this study with model scores on the full Core Set, and Table~\ref{tab:easy_level_human} compares it with model scores on the Easy level. Human scores range from 83.33\% to 100\% over all tasks and tiers (Table~\ref{tab:app_human_difficulty}). Three annotators also independently evaluated a shared 200-example subset, yielding an agreement of $0.89$. Full details appear in Appendix~\ref{app:human_interface} and Appendix~\ref{app:iaa}.

\noindent\textbf{Protocol.} All models are evaluated with deterministic decoding (temperature 0). Reasoning tasks are scored by exact match for scalar and multiple-choice answers and by set equality for unordered list answers. Planning tasks are scored by executing the predicted actions in the corresponding environment and counting an episode as successful only when it reaches the task-defined terminal condition. Primary metrics are task accuracy for reasoning and episode success rate for planning. The primary benchmark evaluation uses the 3{,}900 Core instances. Table~\ref{tab:extended_summary} separately reports the 8{,}780-instance Extended evaluation subset, with exact-input overlaps with Core and within-set duplicates excluded. Training and generative probing use the separately specified evaluation protocols below.

\definecolor{proprietary}{HTML}{FFF2E6}
\definecolor{openweight}{HTML}{FFFDED}

\definecolor{perceptiontask}{HTML}{EEF3FF}
\definecolor{interactivetask}{HTML}{E9F7EF}

\definecolor{bestscorepurple}{HTML}{D2C8FD}
\definecolor{secondscorepurple}{HTML}{E9E4FE}
\newcommand{\bestscore}[1]{\cellcolor{bestscorepurple}#1}
\newcommand{\secondscore}[1]{\cellcolor{secondscorepurple}#1}
\definecolor{rankscoreblue}{HTML}{5B8DB8}
\newcommand{\thirdscore}[1]{\cellcolor{rankscoreblue!13}#1}
\newcommand{\fourthscore}[1]{\cellcolor{rankscoreblue!10}#1}

\newcommand{\tblank}{\phantom{00.00}}
\newcommand{\topoblanks}{& \tblank & \tblank & \tblank & \tblank & \tblank & \tblank & \tblank & \tblank & \tblank & \tblank & \tblank & \tblank & \tblank}

\providecommand{\TaskStrut}{\rule[-0.6ex]{0pt}{2.8ex}}

\newcommand{\PTask}[1]{%
  \cellcolor{perceptiontask}{\TaskStrut\footnotesize\bfseries #1}%
}

\newcommand{\ITask}[1]{%
  \cellcolor{interactivetask}{\TaskStrut\footnotesize\bfseries #1}%
}

\begin{table*}[tbp]
\centering
\setlength{\tabcolsep}{2pt}
\renewcommand{\arraystretch}{1.15}
\resizebox{\textwidth}{!}{%
\begin{tabular}{@{}p{3mm} l *{15}{c}@{}}
\toprule

& \multirow{2}{*}{\textbf{Model}}
& \multicolumn{3}{c}{\textbf{Continuity}}
& \multicolumn{2}{c}{\textbf{Separation}}
& \multicolumn{3}{c}{\textbf{Order}}
& \multicolumn{3}{c}{\textbf{Enclosure}}
& \multicolumn{2}{c}{\textbf{Knots}}
& \multicolumn{2}{c}{\textbf{Avg.}} \\

\cmidrule(lr){3-5}
\cmidrule(lr){6-7}
\cmidrule(lr){8-10}
\cmidrule(lr){11-13}
\cmidrule(lr){14-15}
\cmidrule(l){16-17}

&
& \PTask{2D Maze}
& \PTask{3D Maze}
& \ITask{Pipe}
& \PTask{Assembly}
& \ITask{One Stroke}
& \PTask{Bead}
& \PTask{Origami Point}
& \ITask{Swap}
& \PTask{Sheep}
& \PTask{Hole}
& \ITask{Chat Noir}
& \PTask{Knots}
& \ITask{Untangle}
& \PTask{Reason}
& \ITask{Plan} \\

\midrule

\multicolumn{17}{>{\columncolor{proprietary}}l}{\textit{\footnotesize Proprietary Models}} \\
& GPT-6 Astra
& \bestscore{70.67} & \bestscore{81.33} & \bestscore{54.67}
& \bestscore{71.00} & \bestscore{70.67}
& \bestscore{53.33} & \bestscore{69.33} & \bestscore{51.33}
& \secondscore{35.33} & \bestscore{33.33} & \bestscore{51.67}
& \bestscore{62.00} & \bestscore{20.67}
& \bestscore{59.54} & \bestscore{49.80} \\
& GPT-5.6-Sol
& \secondscore{33.00} & \secondscore{14.00} & \secondscore{11.00}
& \secondscore{36.00} & \secondscore{9.33}
& \secondscore{18.67} & \thirdscore{9.67} & \secondscore{40.33}
& \thirdscore{31.33} & \thirdscore{26.33} & \secondscore{35.33}
& 25.33 & 3.00
& \secondscore{24.29} & \secondscore{19.80} \\
& GPT-5.5
& 8.67 & 7.67 & 0.00
& \thirdscore{33.00} & 0.00
& 7.33 & \secondscore{10.33} & \thirdscore{19.67}
& 23.67 & \secondscore{27.33} & 12.67
& 30.67 & \secondscore{7.67}
& 18.58 & \thirdscore{8.00} \\
& GPT-5.4-mini
& 7.67 & 2.67 & 0.00
& 22.67 & 0.00
& 5.33 & 2.33 & 0.00
& 5.00 & 0.33 & 0.00
& 12.33 & 3.33
& 7.29 & 0.67 \\
& Gemini-3.1-Flash-Lite
& 6.67 & 1.33 & 0.00
& 24.33 & 0.00
& 6.33 & 2.67 & 0.00
& 14.33 & 1.67 & 0.33
& 18.00 & 3.00
& 9.42 & 0.67 \\
& Gemini-3.1-Pro
& \thirdscore{10.67} & \secondscore{14.00} & 0.00
& 27.00 & 0.00
& 6.00 & 3.00 & 1.00
& 30.33 & 20.67 & 13.67
& \secondscore{41.67} & \thirdscore{6.33}
& 19.17 & 4.20 \\
& Claude-Haiku-5.5
& 1.00 & \thirdscore{13.33} & 0.00
& 30.00 & 0.00
& \thirdscore{11.00} & 4.33 & 0.00
& \bestscore{42.33} & 25.33 & \thirdscore{22.00}
& \thirdscore{31.67} & 5.00
& \thirdscore{19.87} & 5.40 \\

\midrule

\multicolumn{17}{>{\columncolor{openweight}}l}{\textit{\footnotesize Hosted Open-Weight Models}} \\
& Nemotron-Nano-12B-VL-v2
& 4.00 & 1.33 & 0.00
& 21.67 & 0.00
& 2.67 & 0.33 & 0.00
& 3.33 & 0.33 & 0.00
& 11.00 & 0.67
& 5.58 & 0.13 \\
& Qwen3.5-397B-A17B
& 6.67 & 4.67 & 0.00
& 22.33 & 0.00
& 4.67 & 1.33 & 0.00
& 0.33 & 1.67 & 0.67
& 11.00 & 0.33
& 6.58 & 0.20 \\
& Llama-4-Maverick-17B-128E
& 8.00 & 3.67 & 0.00
& 24.00 & 0.00
& 4.67 & 3.00 & 0.00
& 2.00 & 0.00 & 0.00
& 10.00 & 2.67
& 6.92 & 0.53 \\
& Ministral-3-14B-Instruct-2512
& 5.00 & 2.00 & 0.00
& 24.00 & 0.00
& 7.00 & 1.67 & 0.00
& 1.33 & 0.00 & 0.00
& 8.67 & 3.00
& 6.21 & 0.60 \\
& InternVL3.5-241B
& 2.00 & 3.33 & 0.00
& 20.33 & 0.00
& 3.00 & 3.67 & 0.00
& 4.33 & 0.33 & 0.00
& 14.67 & 0.00
& 6.46 & 0.00 \\
& Gemma-4-31B-IT
& 2.33 & 3.33 & 0.00
& 13.00 & 0.00
& 0.00 & 0.00 & 0.00
& 7.00 & 0.00 & 1.33
& 8.67 & 0.67
& 4.29 & 0.40 \\

\midrule

& Random Chance
& 4.30 & 3.70 & 0.00
& 20.00 & 0.00
& 7.08 & 2.18 & 0.00
& 5.52 & 6.67 & 0.00
& 13.97 & 2.67
& 7.93 & 0.53 \\

\bottomrule
\end{tabular}%
}
\caption{
\textbf{Performance on the \benchmarkname{} Core Set (\%).}
Each task has 300 instances, 150 per level. Reason and Plan average the
reasoning and planning tasks, and shading marks the top three models per column.
Human results on the Easy level appear in Figure~\ref{fig:teaser} and
Table~\ref{tab:easy_level_human}.
{\setlength{\fboxsep}{2pt}\protect\colorbox{perceptiontask}{\rule{0pt}{1.1ex}Reasoning task}}
and
{\setlength{\fboxsep}{2pt}\protect\colorbox{interactivetask}{\rule{0pt}{1.1ex}Planning task}}
denote task type.
}
\label{tab:topology_benchmark_refreshed}
\end{table*}

\subsection{Benchmark Results}
\label{sec:benchmark_results}

\noindent\textbf{Planning performance lags behind reasoning.} Table~\ref{tab:topology_benchmark_refreshed} reports accuracy or episode success rate on the 300 instances of each of the 13 task types. GPT-6 Astra achieves the highest reasoning and planning averages, 59.54\% and 49.80\%. On the full Core Set, every model scores lower on planning than on reasoning. GPT-5.6-Sol drops from 24.29\% to 19.80\%, Claude-Haiku-5.5 from 19.87\% to 5.40\%, Gemini-3.1-Pro from 19.17\% to 4.20\%, and Qwen3.5-397B-A17B from 6.58\% to 0.20\%. On the Easy level, GPT-6 Astra reaches 76.33\% on reasoning and 81.87\% on planning, against 94.23\% and 99.87\% for humans. It matches humans on 2D Maze, Pipe, and Swap but stays far behind on Bead, Sheep, Hole, and Untangle, and every other model trails humans on every task. On the Hard level, GPT-6 Astra solves 17.73\% of planning episodes, against 81.87\% on Easy.

\begin{table*}[!t]
\centering
\begin{subtable}{\textwidth}
\centering
\fontsize{7}{8.2}\selectfont
\begin{tabular*}{0.495\textwidth}{@{\extracolsep{\fill}}lrr@{}}
\toprule
Task & Core Set & Extended subset \\
\midrule
2D Maze & 300 & 850 \\
3D Maze & 300 & 846 \\
Pipe & 300 & 450 \\
Assembly & 300 & 890 \\
One Stroke & 300 & 450 \\
Bead & 300 & 850 \\
Origami Point & 300 & 1,000 \\
\bottomrule
\end{tabular*}%
\hfill
\begin{tabular*}{0.495\textwidth}{@{\extracolsep{\fill}}lrr@{}}
\toprule
Task & Core Set & Extended subset \\
\midrule
Swap & 300 & 445 \\
Sheep & 300 & 700 \\
Hole & 300 & 699 \\
Chat Noir & 300 & 450 \\
Knots & 300 & 850 \\
Untangle & 300 & 300 \\
\midrule
Total & 3,900 & 8,780 \\
\bottomrule
\end{tabular*}

\caption{Per-task sizes of Core and the filtered Extended evaluation subset.}
\label{tab:release_composition}
\end{subtable}

\vspace{0.8em}

\begin{subtable}{\textwidth}
\centering
\setlength{\tabcolsep}{2pt}
\renewcommand{\arraystretch}{1.15}
\resizebox{\textwidth}{!}{%
\begin{tabular}{@{}p{3mm} l *{15}{c}@{}}
\toprule
& \multirow{2}{*}{\textbf{Model}}
& \multicolumn{3}{c}{\textbf{Continuity}}
& \multicolumn{2}{c}{\textbf{Separation}}
& \multicolumn{3}{c}{\textbf{Order}}
& \multicolumn{3}{c}{\textbf{Enclosure}}
& \multicolumn{2}{c}{\textbf{Knots}}
& \multicolumn{2}{c}{\textbf{Avg.}} \\
\cmidrule(lr){3-5}\cmidrule(lr){6-7}\cmidrule(lr){8-10}
\cmidrule(lr){11-13}\cmidrule(lr){14-15}\cmidrule(l){16-17}
& & \PTask{2D Maze} & \PTask{3D Maze} & \ITask{Pipe}
& \PTask{Assembly} & \ITask{One Stroke}
& \PTask{Bead} & \PTask{Origami Point} & \ITask{Swap}
& \PTask{Sheep} & \PTask{Hole} & \ITask{Chat Noir}
& \PTask{Knots} & \ITask{Untangle}
& \PTask{Reason} & \ITask{Plan} \\
\midrule
\multicolumn{17}{>{\columncolor{proprietary}}l}{\textit{\footnotesize Proprietary Models}} \\
& GPT-5.6-Sol
& \bestscore{88.71}
& \bestscore{70.92}
& \bestscore{60.00}
& \bestscore{60.67}
& \bestscore{41.11}
& \bestscore{64.71}
& \bestscore{77.00}
& \bestscore{86.74}
& \secondscore{78.29}
& \bestscore{80.11}
& \bestscore{70.67}
& \thirdscore{66.71}
& \thirdscore{40.33}
& \bestscore{73.39}
& \bestscore{59.77} \\
& GPT-5.5
& \secondscore{67.29}
& \thirdscore{51.18}
& \thirdscore{2.89}
& \secondscore{57.42}
& \secondscore{13.11}
& \thirdscore{39.88}
& \secondscore{60.00}
& \secondscore{50.11}
& \thirdscore{67.29}
& \secondscore{79.54}
& \thirdscore{39.56}
& \fourthscore{64.82}
& \secondscore{44.33}
& \secondscore{60.93}
& \secondscore{30.00} \\
& GPT-5.4-mini
& 17.53
& 22.34
& \fourthscore{0.22}
& 36.18
& 0.00
& 32.47
& 27.20
& \fourthscore{13.71}
& 25.71
& 23.75
& 8.44
& 28.59
& \secondscore{44.33}
& 26.72
& \fourthscore{13.34} \\
& Gemini-3.1-Flash-Lite
& \fourthscore{33.76}
& 36.05
& 0.00
& \fourthscore{45.73}
& 0.00
& \fourthscore{39.76}
& \fourthscore{42.20}
& 11.69
& \fourthscore{63.14}
& 31.62
& 8.44
& \secondscore{71.53}
& \fourthscore{36.67}
& \fourthscore{45.48}
& 11.36 \\
& Gemini-3.1-Pro
& \thirdscore{36.12}
& \secondscore{58.27}
& \secondscore{7.33}
& \thirdscore{53.82}
& \thirdscore{4.44}
& \secondscore{51.88}
& 41.10
& \thirdscore{22.47}
& \bestscore{78.43}
& \thirdscore{76.11}
& \secondscore{45.11}
& \bestscore{75.29}
& \bestscore{52.00}
& \thirdscore{58.88}
& \thirdscore{26.27} \\
\midrule
\multicolumn{17}{>{\columncolor{openweight}}l}{\textit{\footnotesize Hosted Open-Weight Models}} \\
& Nemotron-Nano-12B-VL-v2
& 14.12
& 11.82
& 0.00
& 28.88
& 0.00
& 20.47
& 26.70
& 1.57
& 20.43
& 9.87
& 0.22
& 28.35
& 6.00
& 20.08
& 1.56 \\
& Qwen3.5-397B-A17B
& 14.00
& \fourthscore{38.65}
& 0.00
& 44.83
& 0.00
& 21.88
& \thirdscore{42.60}
& 12.36
& 9.43
& \fourthscore{34.91}
& 11.78
& 36.59
& 15.00
& 30.36
& 7.83 \\
& Llama-4-Maverick-17B-128E
& 25.65
& 17.97
& 0.00
& 35.73
& 0.00
& 30.94
& 17.10
& 7.42
& 15.57
& 3.72
& 2.44
& 35.88
& 12.67
& 22.82
& 4.51 \\
& Ministral-3-14B-Instruct-2512
& 20.35
& 12.41
& 0.00
& 28.76
& 0.00
& 16.47
& 18.40
& 5.17
& 16.29
& 7.30
& 5.33
& 47.88
& 20.67
& 20.98
& 6.23 \\
& InternVL3.5-241B
& 17.18
& 11.82
& 0.00
& 12.70
& 0.00
& 27.29
& 19.50
& 5.62
& 22.86
& 22.03
& 1.33
& 49.88
& 7.33
& 22.91
& 2.86 \\
& Gemma-4-31B-IT
& 30.35
& 18.68
& 0.00
& 19.10
& 0.00
& 9.41
& 1.60
& 1.35
& 9.29
& 0.00
& \fourthscore{17.33}
& 6.24
& 19.67
& 11.83
& 7.67 \\
\midrule
& \textbf{Human}
& 98.33 & 96.67 & 100.00
& 93.67 & 99.67
& 95.67 & 96.00 & 100.00
& 97.67 & 99.67 & 99.67
& 94.67 & 100.00
& 96.54 & 99.87 \\
\bottomrule
\end{tabular}%
}

\caption{Performance on the Extended evaluation subset (\%). Reason and
Plan average the reasoning and planning tasks. Shading marks the four highest
distinct positive scores per task among the 11 models. Human results by tier
appear in Table~\ref{tab:app_human_difficulty}.}
\label{tab:extended_results}
\end{subtable}

\caption{
\textbf{Composition and performance of the Extended evaluation subset.}
The code repository preserves the complete 11{,}035-instance Extended Set.
This table uses the 8{,}780 instances remaining after excluding 2{,}250 Core
overlaps and five repeated Swap inputs.
{\setlength{\fboxsep}{2pt}\protect\colorbox{perceptiontask}{\rule{0pt}{1.1ex}Reasoning task}}
and
{\setlength{\fboxsep}{2pt}\protect\colorbox{interactivetask}{\rule{0pt}{1.1ex}Planning task}}
denote task type.
}
\label{tab:extended_summary}
\end{table*}

\noindent\textbf{The Extended evaluation subset shows the same reasoning--planning gap.}
Table~\ref{tab:extended_summary} reports the composition and results of the
8{,}780-instance Extended evaluation subset. It excludes exact-input overlaps
with Core and repeated inputs within the original Extended Set. Among the 11
models evaluated on it, GPT-5.6-Sol has the highest
reasoning and planning averages, at 73.39\% and 59.77\%, against 96.54\% and
99.87\% for humans. Every model again
scores lower on planning than on reasoning.

\noindent\textbf{Planning retains less of reasoning below the frontier.} GPT-6 Astra and GPT-5.6-Sol retain more than 80\% of their reasoning average when they plan. GPT-5.5 retains 43\%, Claude-Haiku-5.5 27\%, and Gemini-3.1-Pro 22\%. Claude-Haiku-5.5 outperforms GPT-5.5 on several reasoning tasks, including Sheep (42.33\% versus 23.67\%), and on the reasoning average (19.87\% versus 18.58\%), yet its planning average is lower (5.40\% versus 8.00\%). No hosted open-weight model exceeds 1\% average planning success, close to the 0.53\% of a random legal policy.

\noindent\textbf{The runner-up varies across topological properties.} GPT-6 Astra leads every property average, with 68.89\% on continuity, 70.84\% on separation, 58.00\% on order, 40.11\% on enclosure, and 41.34\% on knots. GPT-5.6-Sol is second on continuity, separation, order, and enclosure, and Gemini-3.1-Pro is second on knots with 24.00\%. Enclosure shows the smallest margin among the top three models (40.11\%, 31.00\%, and 29.89\% for GPT-6 Astra, GPT-5.6-Sol, and Claude-Haiku-5.5), and Claude-Haiku-5.5 achieves the best Sheep score at 42.33\%.

\noindent\textbf{Untangle is the hardest planning environment.} The best Untangle score is 20.67\% by GPT-6 Astra, and no other model exceeds 7.67\%. Pipe and One Stroke separate GPT-6 Astra (54.67\% and 70.67\%) from every other model, which reaches at most 11.00\% and 9.33\%. Each environment requires a sequence of legal actions toward a structural goal, namely a connected pipe network, color-consistent regions, or zero projected rope crossings.

\noindent\textbf{Swap and Chat Noir are the most accessible planning tasks below the frontier.} Below GPT-6 Astra, GPT-5.6-Sol reaches 40.33\% on Swap and 35.33\% on Chat Noir, and GPT-5.5 reaches 19.67\% and 12.67\%, while every other model scores at most 1.00\% on Swap. Claude-Haiku-5.5 and Gemini-3.1-Pro reach 22.00\% and 13.67\% on Chat Noir. Results for the Easy and Hard levels appear in Appendix~\ref{app:full_results_difficulty}.

\begin{keytakeaways}{Planning Lags Behind Topology Recognition}
\begin{itemize}[leftmargin=1.2em,itemsep=0pt,topsep=0pt,parsep=0pt]
    \item Planning exposes failures that reasoning scores do not reveal.
    \item Planning success concentrates in a few frontier models, and open-weight models stay near random.
    \item Aggregate scores hide uneven performance across topological properties.
\end{itemize}
\end{keytakeaways}

\subsection{Training}
\label{sec:training}

\begin{table*}[t]
\centering
\resulttablefont
\setlength{\tabcolsep}{3pt}
\renewcommand{\arraystretch}{1.12}
\begin{tabularx}{\textwidth}{@{}l *{5}{>{\hsize=0.95\hsize}C} >{\hsize=1.4\hsize}C *{3}{>{\hsize=0.95\hsize}C}@{}}
\toprule
\textbf{Training}
& \cellcolor{perceptiontask}\resulttaskheader{2D Maze}
& \cellcolor{perceptiontask}\resulttaskheader{Knots}
& \cellcolor{perceptiontask}\resulttaskheader{Assembly}
& \cellcolor{perceptiontask}\resulttaskheader{Bead}
& \cellcolor{perceptiontask}\resulttaskheader{Sheep}
& \cellcolor{interactivetask}\resulttaskheader{One Stroke}
& \cellcolor{interactivetask}\resulttaskheader{Untangle}
& \cellcolor{interactivetask}\resulttaskheader{Pipe}
& \cellcolor{interactivetask}\resulttaskheader{Swap} \\
\midrule
Base & 9.20 & 11.20 & 32.08 & 9.31 & 9.40 & 0.00 & 0.80 & 0.00 & 0.00 \\
\midrule
SFT & \textbf{14.10} & 75.50 & \textbf{57.35} & 41.74 & 51.00 & \textbf{0.80} & 11.81 & 0.00 & 7.21 \\
RL & 10.10 & 42.70 & 38.52 & 34.73 & 25.30 & 0.00 & 15.42 & 0.00 & 1.40 \\
Leave-one-task-out RL & 8.20 & 15.50 & 31.32 & 26.53 & 8.90 & -- & -- & -- & -- \\
SFT + RL & 13.50 & \textbf{79.60} & 53.41 & \textbf{52.95} & \textbf{58.20} & 0.50 & \textbf{17.42} & 0.00 & \textbf{7.41} \\
\bottomrule
\end{tabularx}
\caption{Qwen3-VL-2B-Instruct performance (\%) across selected reasoning and planning
tasks under the Base, supervised fine-tuning (SFT), reinforcement
learning (RL), leave-one-task-out RL, and SFT + RL settings. For
leave-one-task-out RL, each reasoning-task result comes from a separate policy
trained on the other four reasoning tasks and evaluated on the excluded task.
All policies are trained only on the Easy level. Bold marks the best score per task.
{\setlength{\fboxsep}{2pt}\protect\colorbox{perceptiontask}{\rule{0pt}{1.1ex}Reasoning task}}
and
{\setlength{\fboxsep}{2pt}\protect\colorbox{interactivetask}{\rule{0pt}{1.1ex}Planning task}}
denote task type.}
\label{tab:rl_sft_results}
\end{table*}

\noindent\textbf{Methodology.}
We test whether the deficits exposed by \benchmarkname{} can be improved through
task-specific training. Starting from Qwen3-VL-2B-Instruct~\cite{bai2025qwen3vl}, we compare the
frozen base policy with answer-only supervised fine-tuning (SFT), reinforcement
learning (RL) with Group Relative Policy Optimization (GRPO), and SFT followed
by RL. The comparison covers five reasoning tasks and four planning tasks. For
planning, the model predicts a complete open-loop action sequence from the
initial observation. The sequence
is executed in the corresponding environment and receives credit only if it
reaches the task-defined success state. We additionally train five
leave-one-task-out RL
policies, each trained on four reasoning tasks and evaluated on the fifth.
All policies are trained only on the Easy level. Each task's Easy-level examples
are split 8:1:1 into training, validation, and test sets, and Bead uses a
variant with fewer beads. Full data,
optimization, and evaluation details appear in Appendix~\ref{app:training}.

\noindent\textbf{Training teaches recurring visual patterns.}
The base policy averages 8.00\% across the nine tasks in
Table~\ref{tab:rl_sft_results}. SFT raises this average to 28.83\%, RL alone
to 18.69\%, and SFT followed by RL to 31.44\%. Most of the gain comes from
four reasoning tasks. Under SFT followed by RL, Knots rises from 11.20\% to 79.60\%, Sheep from 9.40\% to
58.20\%, Bead from 9.31\% to 52.95\%, and Assembly from 32.08\% to 53.41\%, with 57.35\% under SFT alone. In
each of these tasks the answer rests on a visual pattern that recurs across
instances, such as the crossings of a closed curve, the fence layer around a
sheep, or the color order along a single string. Answer supervision is enough
to teach such patterns, and RL after SFT adds further gains on Knots, Bead,
Sheep, and Untangle.

\noindent\textbf{Global connectivity and long action sequences remain hard.}
Three difficulties persist under every training condition. First, tracing
connectivity across a whole scene barely improves. 2D Maze reaches at most
14.10\% from a base of 9.20\%, although it is a reasoning task like the four
above. Second, planning stays far behind reasoning. SFT followed by RL lifts
average reasoning accuracy from 14.24\% to 51.53\% but average planning success
only from 0.20\% to 6.33\%. The planning gains come from Untangle (17.42\%) and
Swap (7.41\%), while Pipe stays at 0\% and One Stroke never exceeds 0.80\%. Both
of these tasks require every action in a long sequence to respect a constraint
over the whole board, which means connecting all pipes in Pipe and separating
every color region with one stroke in One Stroke. Third, what a policy learns
stays largely within the trained task. Leave-one-task-out policies improve
held-out Bead from 9.31\% to 26.53\% and held-out Knots from 11.20\% to 15.50\%,
but leave 2D Maze, Assembly,
and Sheep at or below their base scores, and
Appendix~\ref{app:training} shows the same uneven pattern on three further
tasks.

\begin{keytakeaways}{Training Teaches Patterns More Readily Than Global Structure}
\begin{itemize}[leftmargin=1.2em,itemsep=0pt,topsep=0pt,parsep=0pt]
    \item SFT and RL sharply improve reasoning tasks whose answers rest on recurring visual patterns, namely Knots, Sheep, Bead, and Assembly.
    \item Connectivity across the whole scene in 2D Maze and constraint-preserving action sequences in Pipe and One Stroke barely improve.
    \item Held-out gains appear on two of five tasks, Bead and Knots.
\end{itemize}
\end{keytakeaways}

\subsection{Error Analysis}
\label{sec:error_analysis}

\begin{figure}[t]
    \centering
    \includegraphics[width=\linewidth]{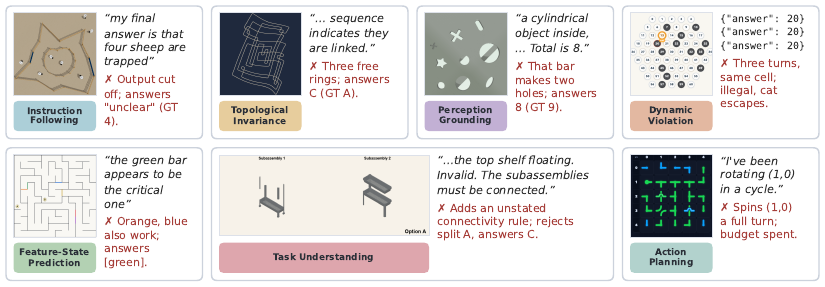}
    \caption{One labeled failure per error category. Each panel pairs the frame the model saw with a verbatim line from its response.}
    \label{fig:error_demos}
\end{figure}

\noindent\textbf{Methodology.} We analyze Gemini-3.1-Pro and InternVL3.5-241B as representatives of the proprietary and open-weight models. For each of the 26 pairs of model and task, we sample 35 failures uniformly from the pair's incorrect predictions on the 300 benchmark instances, yielding 910 labeled failures (560 reasoning and 350 planning, 415 Easy and 495 Hard). Each failure receives one primary category from a seven-class taxonomy that runs from instruction compliance to action planning. Categories are assigned in the fixed causal priority order of Appendix~\ref{app:error_distributions}, and a failure receives the first category in that order that explains it. All percentages are pooled shares of the labeled failures. Figure~\ref{fig:error_demos} shows one representative example per category. Full definitions, sampling details, and per-model distributions appear in Appendix~\ref{app:error_methodology}.

\begin{wrapfigure}{R}{0.49\linewidth}
    \vspace{-0.8em}
    \centering
    \includegraphics[width=\linewidth]{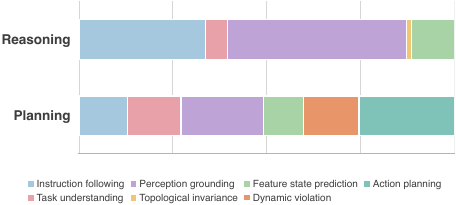}
    \caption{Category distribution of the labeled failures on reasoning and planning tasks, pooled over Gemini-3.1-Pro and InternVL3.5-241B.}
    \label{fig:error_breakdown}
    \vspace{-0.8em}
\end{wrapfigure}
\noindent\textbf{Planning failures sit further downstream than reasoning failures.} Figure~\ref{fig:error_breakdown} shows that reasoning failures concentrate in perception grounding (47.5\%) and instruction following (33.8\%). In planning, feature-state prediction, dynamic violations, and action planning together account for 50.9\% of failures. In reasoning, these three categories account for only 11.4\% of failures, all of them feature-state prediction. Perception grounding still accounts for 22.0\% of planning failures, but 63 of these 77 come from Pipe. Reasoning is thus gated by reading the scene and returning a usable answer, whereas planning adds failures in choosing and executing valid transitions.

\noindent\textbf{Reasoning failures combine visual grounding and answer delivery.} Perception grounding accounts for 58.9\% of Gemini-3.1-Pro's reasoning failures, followed by instruction following (21.1\%) and feature-state prediction (10.4\%). InternVL3.5-241B fails most often on instruction following (46.4\%), mostly responses truncated before a committed answer, and then on perception grounding (36.1\%). Origami Point stands apart, with 78.6\% of its failures assigned to feature-state prediction. Explicit topological-invariance errors are rare in this single-primary-label analysis (1.8\% and 1.1\%).

\noindent\textbf{Planning failures depend on the environment.} Gemini-3.1-Pro's largest planning category is action planning (41.1\%), followed by perception grounding (24.0\%) and feature-state prediction (21.1\%), while dynamic violations are almost absent (0.6\%). InternVL3.5-241B is led by dynamic violations (29.1\%), with instruction following, task understanding, and perception grounding each near 20\%. Pipe failures are 90.0\% perception grounding, as models misread pipe ends and the source direction. Swap failures are 67.1\% task understanding, since on Easy boards both models play an adjacent-only sliding puzzle although any tile may swap with the blank. Chat Noir and One Stroke split by model. Gemini-3.1-Pro fails mainly through action planning on Chat Noir (97.1\%) and feature-state prediction on One Stroke (62.9\%). InternVL3.5-241B fails mainly through dynamic violations on both (68.6\% and 77.1\%), and every one of its failures on Untangle is a truncated response. Full distributions appear in Appendix~\ref{app:error_distributions}, and Appendix~\ref{app:biases} summarizes the error signature of each environment.

\begin{keytakeaways}{Reasoning and Planning Fail at Different Stages}
\begin{itemize}[leftmargin=1.2em,itemsep=0pt,topsep=0pt,parsep=0pt]
    \item Reasoning failures come mainly from visual grounding and unusable answers.
    \item About half of planning failures lie in state prediction, dynamics, and action choice.
    \item The dominant planning failure varies strongly with the environment.
\end{itemize}
\end{keytakeaways}

\subsection{Probing Experiment}
\label{sec:probing_experiment}

\begin{table*}[!t]
\centering
\resulttablefont
\setlength{\tabcolsep}{1.7pt}
\renewcommand{\arraystretch}{1.323}
\begin{tabularx}{0.597\textwidth}[t]{l *{6}{C}}
\toprule
\textbf{Planner}
& \multicolumn{1}{>{\columncolor{perceptiontask}}c}{\resulttaskheader{Knots}}
& \multicolumn{1}{>{\columncolor{perceptiontask}}c}{\resulttaskheader{Sheep}}
& \multicolumn{1}{>{\columncolor{perceptiontask}}c}{\resulttaskheader{2D Maze}}
& \multicolumn{1}{>{\columncolor{interactivetask}}c}{\resulttaskheader{Untangle}}
& \multicolumn{1}{>{\columncolor{interactivetask}}c}{\resulttaskheader{One Stroke}}
& \multicolumn{1}{>{\columncolor{interactivetask}}c}{\resulttaskheader{Pipe}} \\
\midrule
\rowcolor{black!6}
\blockheadcols{7}{Planner only}
GPT-5.6-Luna     & 26.00 & 30.33 & \secondscore{22.17} & 19.05 & \bestscore{2.86} & \bestscore{29.52} \\
GPT-5.4-mini     & 25.71 & 19.05 & 15.24 & 23.81 & 0.00 & 0.00 \\
InternVL3.5-241B & 42.86 & 17.14 & 15.24 & 3.81  & 0.00 & 0.00 \\

\addlinespace[2pt]
\rowcolor{black!6}
\blockheadcols{7}{+ GPT-Image-2 (image)}
GPT-5.6-Luna     & \secondscore{57.14} & \bestscore{68.57} & \bestscore{61.90} & 24.76 & 0.00 & \secondscore{20.00} \\
GPT-5.4-mini     & \bestscore{60.00} & \secondscore{57.14} & 12.38 & 6.67 & 0.00 & 0.00 \\
InternVL3.5-241B & 56.19 & 30.48 & 13.33 & 0.00 & 0.00 & 0.00 \\
\bottomrule
\end{tabularx}%
\hfill
\renewcommand{\arraystretch}{1.12}%
\begin{tabularx}{0.383\textwidth}[t]{l *{3}{C}}
\toprule
\textbf{Planner}
& \multicolumn{1}{>{\columncolor{interactivetask}}c}{\resulttaskheader{Untangle}}
& \multicolumn{1}{>{\columncolor{interactivetask}}c}{\resulttaskheader{One Stroke}}
& \multicolumn{1}{>{\columncolor{interactivetask}}c}{\resulttaskheader{Pipe}} \\
\midrule
\rowcolor{black!6}
\blockheadcols{4}{+ Wan2.2-I2V-A14B (video)}
GPT-5.6-Luna     & 24.76 & 0.00 & 9.52 \\
InternVL3.5-241B & 4.76  & 0.00 & 0.00 \\

\addlinespace[2pt]
\rowcolor{black!6}
\blockheadcols{4}{+ Seedance-2.0-Mini (video)}
GPT-5.6-Luna     & \secondscore{25.71} & 0.00 & 6.67 \\
InternVL3.5-241B & \bestscore{27.62} & 0.00 & 0.00 \\

\addlinespace[2pt]
\rowcolor{black!6}
\blockheadcols{4}{+ Veo-3.1-Lite (video)}
GPT-5.6-Luna     & \bestscore{27.62} & 0.00 & 5.71 \\
InternVL3.5-241B & 5.71  & 0.00 & 0.00 \\
\bottomrule
\end{tabularx}

\caption{Probing success rate (\%). Each gray band names what the planner is paired with, and each row is a planner. The left panel reports planner-only and image-interleaved runs on all six tasks, and the right panel reports video-interleaved runs, which cover the planning tasks only. Lavender marks the best and second-best result per task across both panels.
{\setlength{\fboxsep}{2pt}\protect\colorbox{perceptiontask}{\rule{0pt}{1.1ex}Reasoning task}}
and
{\setlength{\fboxsep}{2pt}\protect\colorbox{interactivetask}{\rule{0pt}{1.1ex}Planning task}}
denote task type.}
\label{tab:interleaved_results}
\end{table*}

\begin{wrapfigure}[15]{R}{0.39\linewidth}
    \vspace{-0.8em}
    \centering
    \includegraphics[width=\linewidth]{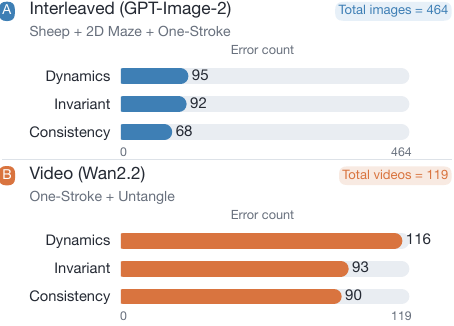}
    \captionsetup{justification=raggedright,singlelinecheck=false}
    \caption{\small Per-generation errors in 464 GPT-Image-2 images (top) and 119 Wan2.2-I2V-A14B videos (bottom). Each panel labels its evaluated tasks.}
    \label{fig:probing_gen_errors}
    \vspace{-0.8em}
\end{wrapfigure}

\noindent\textbf{Methodology.} Section~\ref{sec:error_analysis} shows that, for Gemini-3.1-Pro and InternVL3.5-241B, about half of the labeled planning failures arise in action choice, dynamics, and state prediction, and that outside Pipe few come from reading the scene. We probe whether explicit visual prediction can supply the topological state that models lose across actions. The success-rate configurations use three planners, GPT-5.6-Luna~\cite{openai2026gpt56}, GPT-5.4-mini, and InternVL3.5-241B, on six tasks (Knots, Sheep, 2D Maze, Untangle, One Stroke, and Pipe). The \textbf{baseline} acts directly from the rendered observation. The \textbf{interleaved} variant lets the planner call GPT-Image-2~\cite{openai2026images2} at any step, up to ten times per episode, to render the predicted next state. The \textbf{video} variants use Wan2.2-I2V-A14B~\cite{wan2025wanopenadvancedlargescale}, Seedance-2.0-Mini~\cite{bytedance2026seedance2mini}, and Veo-3.1-Lite~\cite{google2026veo31lite} to roll out candidate plans. Figure~\ref{fig:probing_gen_errors} audits images and videos from a separate run with GPT-5.4-mini as the planner. Each probing condition uses 105 episodes per task, 35 from each of three generator difficulty settings, so the episodes extend beyond the two benchmark levels. The full imagined-rollout protocol and failure-mode analysis appear in Appendix~\ref{app:generative_world_model}.

\begin{table*}[!t]
\centering
\resulttablefont
\setlength{\tabcolsep}{2pt}
\renewcommand{\arraystretch}{0.98}
\newcommand{\cvhead}[1]{\shortstack[c]{#1}}

\begin{tabular*}{\textwidth}{@{\extracolsep{\fill}}l *{11}{c}@{}}
\toprule
& \multicolumn{6}{c}{\textbf{Static checks}}
& \multicolumn{3}{c}{\textbf{Dynamic checks}}
& \multicolumn{2}{c}{\textbf{Outcome}} \\
\cmidrule(lr){2-7}\cmidrule(lr){8-10}\cmidrule(l){11-12}
\addlinespace[1pt]
\textbf{Video generator}
& \cvhead{Color\\sep.}
& \cvhead{Direction\\valid.}
& \cvhead{Path\\cont.}
& \cvhead{Path\\valid.}
& \cvhead{Start/end\\valid.}
& \cvhead{\textbf{Static}\\\textbf{(all)}}
& \cvhead{Grid-cell\\temp.}
& \cvhead{Path\\temp.}
& \cvhead{\textbf{Dynamic}\\\textbf{(all)}}
& \cvhead{Final-state\\success}
& \cvhead{\textbf{Process-}\\\textbf{valid}} \\
\midrule
\multicolumn{12}{>{\columncolor{interactivetask}}l}{\textit{One Stroke}} \\
LTX-2.3
 & $1.0$ & $7.6$ & $2.9$ & $7.6$ & $18.1$ & $0.0$
 & $1.0$ & $0.0$ & $0.0$
 & $0.0$ & $0.0$ \\

LTX-2.5
 & $0.0$ & $0.0$ & $0.0$ & $0.0$ & $0.0$ & $0.0$
 & $1.9$ & $0.0$ & $0.0$
 & $0.0$ & $0.0$ \\
Veo-3.1-Lite
 & $0.0$ & $0.0$ & $0.0$ & $0.0$ & $0.0$ & $0.0$
 & $3.8$ & $1.0$ & $0.0$
 & $0.0$ & $0.0$ \\
Wan2.2-I2V-A14B
 & $0.0$ & $52.4$ & $27.6$ & $52.4$ & $56.2$ & $0.0$
 & $18.1$ & $16.2$ & $1.9$
 & $0.0$ & $0.0$ \\
MiniMax-H3
 & $0.0$ & $0.0$ & $0.0$ & $0.0$ & $0.0$ & $0.0$
 & $1.9$ & $0.0$ & $0.0$
 & $0.0$ & $0.0$ \\
Seedance-2.5
 & $0.0$ & $0.0$ & $0.0$ & $1.0$ & $1.9$ & $0.0$
 & $2.9$ & $0.0$ & $0.0$
 & $0.0$ & $0.0$ \\
\end{tabular*}

\begin{tabular*}{\textwidth}{@{\extracolsep{\fill}}l *{8}{c}@{}}
\midrule
& \multicolumn{2}{c}{\textbf{Static checks}}
& \multicolumn{4}{c}{\textbf{Dynamic checks}}
& \multicolumn{2}{c}{\textbf{Outcome}} \\
\cmidrule(lr){2-3}\cmidrule(lr){4-7}\cmidrule(l){8-9}
\addlinespace[1pt]
\textbf{Video generator}
& \cvhead{Connectivity--\\color match}
& \cvhead{\textbf{Static}\\\textbf{(all)}}
& \cvhead{Cell\\occupancy}
& \cvhead{Pipe-type\\consist.}
& \cvhead{Rotation\\valid.}
& \cvhead{\textbf{Dynamic}\\\textbf{(all)}}
& \cvhead{Final-state\\success}
& \cvhead{\textbf{Process-}\\\textbf{valid}} \\
\midrule
\multicolumn{9}{>{\columncolor{interactivetask}}l}{\textit{Pipe}} \\
LTX-2.3
 & $0.0$ & $0.0$
 & $0.0$ & $0.0$ & $0.0$ & $0.0$
 & $3.8$ & $0.0$ \\

LTX-2.5
 & $0.0$ & $0.0$
 & $1.9$ & $1.0$ & $0.0$ & $0.0$
 & $35.2$ & $0.0$ \\
Veo-3.1-Lite
 & $1.0$ & $1.0$
 & $2.9$ & $0.0$ & $0.0$ & $0.0$
 & $5.7$ & $0.0$ \\
Wan2.2-I2V-A14B
 & $0.0$ & $0.0$
 & $5.7$ & $0.0$ & $0.0$ & $0.0$
 & $13.3$ & $0.0$ \\
MiniMax-H3
 & $0.0$ & $0.0$
 & $2.9$ & $0.0$ & $0.0$ & $0.0$
 & $19.0$ & $0.0$ \\
Seedance-2.5
 & $0.0$ & $0.0$
 & $33.3$ & $0.0$ & $0.0$ & $0.0$
 & $16.2$ & $0.0$ \\
\end{tabular*}

\begin{tabular*}{\textwidth}{@{\extracolsep{\fill}}l *{8}{c}@{}}
\midrule
& \multicolumn{3}{c}{\textbf{Static checks}}
& \multicolumn{3}{c}{\textbf{Dynamic checks}}
& \multicolumn{2}{c}{\textbf{Outcome}} \\
\cmidrule(lr){2-4}\cmidrule(lr){5-7}\cmidrule(l){8-9}
\addlinespace[1pt]
\textbf{Video generator}
& \cvhead{Endpoint\\config.}
& \cvhead{Rope\\integrity}
& \cvhead{\textbf{Static}\\\textbf{(all)}}
& \cvhead{Endpoint\\tracking}
& \cvhead{Topology\\flicker}
& \cvhead{\textbf{Dynamic}\\\textbf{(all)}}
& \cvhead{Final-state\\success}
& \cvhead{\textbf{Process-}\\\textbf{valid}} \\
\midrule
\multicolumn{9}{>{\columncolor{interactivetask}}l}{\textit{Untangle}} \\
LTX-2.3
 & $0.0$ & $1.0$ & $0.0$
 & $0.0$ & $66.7$ & $0.0$
 & $32.4$ & $0.0$ \\

LTX-2.5
 & $1.0$ & $7.6$ & $1.0$
 & $1.0$ & $73.3$ & $1.0$
 & $6.7$ & $0.0$ \\
Veo-3.1-Lite
 & $0.0$ & $5.7$ & $0.0$
 & $0.0$ & $32.4$ & $0.0$
 & $1.0$ & $0.0$ \\
Wan2.2-I2V-A14B
 & $0.0$ & $21.0$ & $0.0$
 & $0.0$ & $46.7$ & $0.0$
 & $2.9$ & $0.0$ \\
MiniMax-H3
 & $4.8$ & $42.9$ & $3.8$
 & $2.9$ & $57.1$ & $1.0$
 & $5.7$ & $0.0$ \\
Seedance-2.5
 & $20.0$ & $48.6$ & $9.5$
 & $0.0$ & $70.5$ & $0.0$
 & $3.8$ & $0.0$ \\
\bottomrule
\end{tabular*}

\caption{\textbf{Strict video-level computer-vision (CV) pass rates (\%).}
The generators are LTX-2.3~\cite{lightricks2026ltx23}, LTX-2.5~\cite{lightricks2026ltx25}, Veo-3.1-Lite~\cite{google2026veo31lite}, Wan2.2-I2V-A14B~\cite{wan2025wanopenadvancedlargescale}, MiniMax-H3~\cite{minimax2026h3}, and Seedance-2.5~\cite{bytedance2026seedance25}. Appendix~\ref{app:cv_video_diagnostics} defines every check.}
\label{tab:cv_metric_results}
\end{table*}

\noindent\textbf{Interleaved image prediction helps on Knots and Sheep but not on Pipe or One Stroke.} Adding GPT-Image-2 raises Knots and Sheep for GPT-5.4-mini (25.71\% to 60.00\% and 19.05\% to 57.14\%) and for InternVL3.5-241B (42.86\% to 56.19\% and 17.14\% to 30.48\%) (Table~\ref{tab:interleaved_results}). With GPT-5.6-Luna, the GPT-Image-2 condition reaches 68.57\% on Sheep and 61.90\% on 2D Maze but 0.00\% on One Stroke.

\noindent\textbf{Final-state success and process validity are different criteria.} A rollout achieves \textbf{final-state success} when its final frame shows a solved task state. It is \textbf{process valid} only if, in addition, every checked frame shows a legal task state and every pair of adjacent sampled frames forms a legal transition. Table~\ref{tab:cv_metric_results} reports both criteria for every generator and task, and Table~\ref{tab:interleaved_results} reports whether the planner solves the task.

\noindent\textbf{None of the 1{,}890 evaluated video rollouts passes all process-validity checks.} A manual audit of 119 Wan2.2-I2V-A14B rollouts finds dynamics violations in 116, topological-invariance errors in 93, and consistency errors in 90 (Figure~\ref{fig:probing_gen_errors}). An automatic verifier confirms this pattern at scale on 1{,}890 rollouts, 105 for each of six video generators on each of One Stroke, Pipe, and Untangle (Table~\ref{tab:cv_metric_results}). Final-state success reaches up to 35.2\% of the rollouts in a task, yet the process-valid rate is 0.0\% for every generator on every task. Single checks often pass, as the Untangle topology readout stays free of flicker in 32.4\% to 73.3\% of videos, yet no generator passes all static checks in more than 9.5\% of the videos of any task.

\noindent\textbf{Verifier validation.} The verifier parses 41 evenly spaced frames of each video into task states. It applies static checks to four sampled frames and dynamic checks to every adjacent pair. A video passes only when its final state is readable and every check succeeds. A human audit of 180 LTX-2.3 and Wan2.2-I2V-A14B videos agrees with all 180 video-level metric decisions. We use these checks to screen for invalid rollouts. Appendix~\ref{app:cv_video_diagnostics} details the audit and parser errors.

\WFclear
\vspace{0.5\baselineskip}

\begin{keytakeaways}{Final-State Success Does Not Certify Topological Rollouts}
\begin{itemize}[leftmargin=1.2em,itemsep=0pt,topsep=0pt,parsep=0pt]
    \item Generated images help on Knots and Sheep but not on One Stroke or Pipe.
    \item Video generators achieve final-state success in up to 35.2\% of rollouts, yet none of the evaluated videos passes all process-validity checks.
    \item Rollout evaluation must verify topology at every step as well as at the final state.
\end{itemize}
\end{keytakeaways}

\section{Related Work}
\label{sec:related}

\noindent\textbf{Cognitive and Topological Evaluation.}
Cognitive studies motivate evaluating topological relations as a foundation of spatial understanding~\cite{piaget2013child,chen1982topological,chen2005topological}. BabyVision tests basic visual abilities through both language and generated visual outputs~\cite{chen2026babyvision}, while related work examines vision models' sensitivity to geometric and topological concepts~\cite{wang2025computer}. Among benchmarks that target topology directly, CurveBench asks models to recover containment trees from images of nested curves~\cite{mohseni2026curvebench}, whereas TopoBench requires complete solutions to symbolic grid puzzles governed by global spatial constraints~\cite{maniparambil2026topobench}. KnotGym brings topology into interactive evaluation through rope manipulation from visual observations~\cite{chen2025knot}. \benchmarkname{} broadens this coverage by organizing five topological properties within one suite and evaluating each through both reasoning questions and interactive planning tasks.

\noindent\textbf{Spatial Reasoning and Interaction.}
Spatial benchmarks evaluate relations within scenes and the integration of information across viewpoints~\cite{ma20253dsrbench,wang2025site,jia2025omnispatial,yang2025thinking,wang2025mindcube}. Theory of Space extends this perspective to constructing and revising spatial beliefs through active exploration~\cite{zhang2026theory}. Planning with the Views examines whether models can compose individual camera movements into longer plans~\cite{wang2026planningviews}. Interactive evaluations include sliding puzzles in iVISPAR~\cite{mayer2025ivispar} and tasks across simulation environments in SpatialWorld~\cite{gao2026spatialworld}, alongside broader studies of embodied decision interfaces and visual action selection~\cite{li2024embodiedagent,yang2025embodiedbench}. ESI-Bench grounds active perception and manipulation in Spelke's core knowledge systems~\cite{hong2026esibench}. \benchmarkname{} organizes reasoning and planning around topological properties, providing the complementary coverage summarized in Table~\ref{tab:comparison} of Appendix~\ref{app:extended_rw}.

\noindent\textbf{World Modeling and Agent Training.}
Predicting the consequences of actions connects spatial evaluation with world modeling. CausalSpatial uses generated visual evidence to answer questions about specified object motions~\cite{ma2026causalspatial}, while ENACT evaluates forward and inverse world modeling through observation and action sequence reordering~\cite{wang2025enact}. For agents acting across multiple turns, \mbox{RAGEN}~\cite{wang2025ragen} and \mbox{RAGEN-2}~\cite{wang2026ragen2} study training stability, while \mbox{VAGEN}~\cite{wang2025vagen} reinforces world-model reasoning. Our rollout diagnostics complement these approaches by checking whether predicted transitions preserve task topology.

\section{Conclusion}
\label{sec:conclusion}

We presented \benchmarkname, a benchmark of topological intuition with 13 procedurally generated task types. It covers continuity, separation, order, enclosure, and knots, five properties drawn from Piaget's classification and later cognitive work, at two cognitive levels. Every MLLM evaluated on the Core Set scores lower on planning than on reasoning. On the Easy level, the strongest model trails humans in both reasoning and planning. Training on Qwen3-VL-2B-Instruct improves task performance, yet planning success stays low and gains on held-out tasks are uneven. Generated videos achieve final-state success in up to 35.2\% of rollouts, yet none of the 1{,}890 evaluated video rollouts passes all process-validity checks.

\section*{Acknowledgments}
This work used Delta at NCSA through ACCESS allocation
CIS250698, supported by NSF grants \#2138259, \#2138286, \#2138307, \#2137603,
and \#2138296~\citep{boerner2023access}, and the Quest high performance computing
facility at Northwestern University.

\bibliographystyle{unsrt}
\providecommand{\arxivbibdatabase}{references}
\bibliography{\arxivbibdatabase}
\clearpage
\appendix
\clearpage
\renewcommand{\topfraction}{0.92}
\renewcommand{\bottomfraction}{0.7}
\renewcommand{\textfraction}{0.05}
\renewcommand{\floatpagefraction}{0.85}
\WarningFilter*{latex}{Text page \thepage\space contains only floats}
\makeatletter
\setlength{\@fptop}{0pt}
\setlength{\@fpbot}{0pt plus 1fil}
\makeatother

\section*{\centering Appendix}

\startcontents[appendix]

\vspace{0.5em}
\noindent\textbf{Table of Contents}\\[-0.6em]
{\color{black!40}\hrule}
\vspace{0.4em}

{\small
\printcontents[appendix]{}{1}{\setcounter{tocdepth}{2}}
}

\vspace{0.5em}
{\color{black!40}\hrule}
\vspace{1em}

\section{Benchmark Details}
\label{app:benchmark}

\subsection{Cognitive-Science Grounding of the Five Properties}
\label{app:cogsci}

Our taxonomy is grounded in the developmental psychology account of
topological cognition initiated by Piaget and Inhelder
\cite{piaget2013child}, refined by subsequent mathematical and
educational analyses
\cite{martin1976analysis,strohecker1991knot}, and
formalized in standard algebraic topology
\cite{hatcher2002algebraic}. Piaget's original five classes are proximity,
separation, order, enclosure, and continuity. We depart from them in two
respects. Proximity, which develops in tandem with separation and is
constitutive of it~\cite{martin1976analysis}, is absorbed into our
Separation property, and Knots is promoted to a property of its own
for the reasons given below.

\paragraph{Continuity.}
Continuity, in Piaget's phrasing, is the perception of a line, surface,
or spatial field as an unbroken whole~\cite{piaget2013child}.
Poincar\'e captures its minimal form as a relation in which adjacent
elements are indistinguishable ($A{=}B$, $B{=}C$) while distant ones are
not ($A{\neq}C$)~\cite{poincare1905science}. Piaget himself treats continuity as the
synthesis of the other four primitives rather than a primitive on the
same level, a view that Martin reinforces by
showing the four to be mutually constitutive \cite{martin1976analysis}. The standard
mathematical formalization is path-connectedness, the existence of a
continuous map from $[0,1]$ joining two points
\citep[Ch.\,1]{hatcher2002algebraic}. Our Continuity tasks
(\textsc{2D Maze}, \textsc{3D Maze}, \textsc{Pipe}) ask whether two
locations lie in the same path-component of a region given only its
visual depiction.

\paragraph{Separation.}
Separation is the ability to distinguish neighboring elements. 
Piaget illustrates the lack of it
with the syncretic infant percept of an object leaning against a
wall, perceived as a single ill-defined patch until separation
analyses it into two units~\cite{piaget2013child}.
Martin argues that proximity and separation
develop in tandem. Finer separation does not displace proximity but
allows the child to perceive different degrees of it over larger
fields \cite{martin1976analysis}. The topological correspondent is the decomposition of a
space into its connected components, the most basic invariant
preserved under homeomorphism \citep[Ch.\,0]{hatcher2002algebraic}.
Our Separation tasks (\textsc{Assembly}, \textsc{One Stroke}) ask the
model to decide which substructures are independent units and which
belong to a single connected whole.

\paragraph{Order.}
Order is the relation of spatial succession, by which elements are
arranged one after another along a direction. Piaget documents it in
the infant's gaze across the rungs of a cot and in the way young
children seriate objects along a line~\cite{piaget2013child}.
Piaget treats seriation as one of the deep
epistemic structures whose combination underwrites mathematical
thought, and Martin notes that, among Piaget's
primitives, order admits the cleanest mathematical formalization \cite{martin1976analysis}.
Its Euclidean refinement gives oriented coordinate systems and
similarity transformations that preserve relative position. Our
Order tasks test sequence read-out along a one-dimensional substrate
(\textsc{Bead}), point-tracking through ordered folds
(\textsc{Origami Point}), and recovery of permutations under
elementary operations (\textsc{Swap}).

\paragraph{Enclosure.}
Enclosure, which Piaget calls \emph{surrounding}, is the relation by
which a boundary partitions space into an interior and an
exterior~\cite{piaget2013child}. In three dimensions it takes the
form of \emph{insideness}, exemplified by an object inside a closed
box or by the contrast between a surface with a hole and one
without. Algebraic topology formalizes both sides of this intuition.
The Jordan Curve Theorem \citep[\S 2.B]{hatcher2002algebraic} shows
that any subspace of $S^{2}$ homeomorphic to $S^{1}$ separates
$S^{2}$ into two components, and a hole in a higher-dimensional
space is detected by a spherical cycle bounding a missing interior
\citep[Ch.\,2]{hatcher2002algebraic}. Our Enclosure tasks cover three
faces of this property. \textsc{Sheep} tests interior versus exterior
judgment under fence-induced partition, \textsc{Hole} tests the
enumeration of bounded missing regions, and \textsc{Chat Noir} tests
planning under a closing-boundary constraint.

\paragraph{Knots.}
Knots occupy an ambiguous place in Piaget's own framework. Chapter 4
of \textit{The Child's Conception of Space} uses them to probe
surrounding, arguing that distinguishing a true knot from a deceptive
one requires grasping how a strand encloses
itself~\cite{piaget2013child}. Strohecker later
re-positions knot cognition as a comprehensive ability that draws on
all of Piaget's substructures at once, together with set- and
group-theoretic intuitions that are absent from the original
primitives~\cite{strohecker1991knot}.
Recent empirical work supports this dissociation, showing that knot
reasoning fails to track other physical-intuition
abilities~\cite{croom2024tangled}. We therefore give Knots its own
property column, since it captures phenomena such as crossing parity,
linking number, and isotopy that no single Piagetian primitive
accounts for. Our tasks
\textsc{Knots} and \textsc{Untangle} test the perceptual and
interventional sides of this composite ability.

\subsection{Mathematical Foundations of the Five Properties}
\label{app:formal_defs}

This section defines each property mathematically and explains
how the benchmark tasks use the resulting invariants and related
structural targets. A scene
specification $s$ determines an embedded structure
$X(s) = (E, K, \mu)$. Here
$E \in \{\mathbb{R}^2, \mathbb{R}^3\}$ is the ambient space,
$K \subset E$ is the compact embedded structure, and $\mu$ contains
the marked elements. When $K$ denotes a free region, we use its closure
inside the compact board domain $D \subset E$. This convention makes
$K$ compact in every task.

Two marked embedded structures $X=(E,K,\mu)$ and
$X'=(E,K',\mu')$ are \emph{ambient-isotopic}, written $X\simeq X'$,
if there is a continuous map $h\colon E\times[0,1]\to E$ such that,
for every $t\in[0,1]$, the map $h_t(\cdot)=h(\cdot,t)$ is a
homeomorphism, $h_0=\operatorname{id}_E$, $h_1(K)=K'$, and $h_1$
carries every marked element in $\mu$ to its corresponding marked
element in $\mu'$~\citep[Ch.\,1]{rolfsen1976knots}.

These definitions draw on four sources. Hatcher provides the algebraic
topology used for Continuity, Separation, and
Enclosure~\cite{hatcher2002algebraic}. Rolfsen provides the knot theory
used for Knots~\cite{rolfsen1976knots}. Order follows Huntington's
postulates for cyclic order~\cite{huntington1916cyclic}. Martin's
mathematical analysis of Piaget's primitives connects the cognitive
taxonomy to these formal notions~\cite{martin1976analysis}.

The structures produced by our generators are locally path-connected.
Therefore, connected components and path components coincide. We write
$\pi_0(K)$ for the set of components of $K$. Some tasks use a derived
structural target, and we identify each such case below.

\begin{definitionbr}[Continuity]
\label{def:continuity}
Let $p, q \in K$ be marked points. The continuity invariant is
$\Phi_{\mathrm{cont}}(X) = \mathbf{1}\!\left[\,[p] = [q] \text{ in } \pi_0(K)\,\right]$,
that is, whether some continuous path
$\gamma\colon [0,1] \to K$ joins $p$ to
$q$~\citep[Ch.\,1]{hatcher2002algebraic}.
\end{definitionbr}

\textsc{2D Maze} and \textsc{3D Maze} query $\Phi_{\mathrm{cont}}$ with
$K$ the closed free region of the board. A what-if edit removes a wall
or opens a door, which changes $K$ and hence possibly $\pi_0(K)$. \textsc{Pipe}
uses the invariant as its goal predicate. Writing $K(\theta)$ for the
union of pipe segments in rotation state $\theta$, the goal set
contains exactly the states in which every segment lies in the
component of the source.

\begin{definitionbr}[Separation]
\label{def:separation}
The separation invariant $\Phi_{\mathrm{sep}}(X)$ is the partition of
the marked elements induced by the components of $K$, in which two marks are
identified exactly when they lie in the same element of
$\pi_0(K)$~\citep[Ch.\,0]{hatcher2002algebraic}.
\end{definitionbr}

Piaget defines separation as mere disjointness of two sets, which
Martin shows to be strictly weaker than the mathematical notion and to
be preserved by arbitrary injections~\cite{martin1976analysis}. We
therefore formalize the object individuation his tasks target by the
component partition instead. \textsc{Assembly} asks whether the two
candidate subassemblies together reconstruct the complete object,
without missing, extra, or substituted parts. \textsc{One Stroke} evaluates
the invariant on the complement. The drawn stroke $P$ partitions the
board interior, and an episode succeeds when the components of the
board minus $P$ induce exactly the color classes of the marked cells.

\begin{definitionbr}[Order]
\label{def:order}
Let $\gamma \subset E$ be an embedded arc or circle carrying marked
points $\mu = (p_1, \dots, p_n)$ together with a decoration that fixes
a start mark and a traversal direction. The order invariant
$\Phi_{\mathrm{ord}}(X)$ is the sequence in which the marks are met
when $\gamma$ is traversed from the start mark in the given direction.
The underlying linear or cyclic order type is preserved by
orientation-preserving homeomorphisms of $\gamma$ and reversed by
orientation-reversing ones~\cite{huntington1916cyclic}, and the
decoration selects one representative of this orbit under rotations
and reversals.
\end{definitionbr}

In \textsc{Bead}, the decoration is rendered explicitly as a white
start marker and a tangential direction arrow, so the sequence answer
is unique. The relationship queries ignore the decoration and compare
the color sequences of two strings inside the orbit, with \texttt{IDENTICAL}, \texttt{REVERSED},
\texttt{CYCLIC\_ROTATION}, and \texttt{DIFFERENT} naming the orbit
relations. Martin verifies that these order relations are topological
invariants and that arc, circle, and figure eight are pairwise
non-homeomorphic~\cite{martin1976analysis}. \textsc{Origami Point}
and \textsc{Swap} use derived targets. \textsc{Origami Point} tracks
marks through a fold trajectory, a sequence of embeddings of the sheet
into $\mathbb{R}^3$, conserving each mark's intrinsic position on the
sheet while its Euclidean position varies, and scores the recovered
correspondence between marks and labels by set
equality~\cite{piaget2013child}. \textsc{Swap}'s goal predicate matches
the full labeled grid arrangement against a target. It
operationalizes ordered rearrangement in the planning setting.

\begin{definitionbr}[Enclosure]
\label{def:enclosure}
For a closed structure $K$ in the plane, the enclosure invariant
labels each marked point $p \in E \setminus K$ by whether it lies in a
bounded component of $E \setminus K$. For a Jordan curve, the
complement has exactly one bounded component, its
interior~\citep[\S 2.B]{hatcher2002algebraic}. For a solid
$K \subset \mathbb{R}^3$, the hole-count invariant is the first Betti
number
$b_1(K) = \operatorname{rank} H_1(K)$~\citep[Ch.\,2]{hatcher2002algebraic}.
\end{definitionbr}

\textsc{Sheep} queries the planar labeling with $K$ the fence layout,
so a sheep is enclosed exactly when its position has no path to the
unbounded component of the complement. \textsc{Hole} queries $b_1$ of
the rendered board. The generator produces solids that are
handlebodies with no internal cavities, so $H_1(K)$ is free and $b_1$
equals the number of through-handles, since each through-hole contributes
one independent cycle while pits and shallow depressions contribute
none.
\textsc{Chat Noir} uses enclosure as its goal predicate. An episode
succeeds when the cat's cell lies in a component of unblocked cells
that contains no boundary cell.

\begin{definitionbr}[Knots]
\label{def:knots}
Let $K$ be one or more disjoint embedded circles in $\mathbb{R}^3$.
The knot invariant is the ambient isotopy class of
$K$~\citep[Ch.\,1]{rolfsen1976knots}. The tasks query it coarsely. A
single loop is \emph{unknotted} if it is ambient-isotopic to the
standard circle, and two loops are \emph{split} if an embedded sphere
separates them. With both components oriented, the linking number
$\operatorname{lk}$ is the sum of crossing signs where one loop passes
under the other in a diagram~\citep[\S 5.D]{rolfsen1976knots}.
$\operatorname{lk} \neq 0$ certifies a non-split link, while
$\operatorname{lk} = 0$ does not certify splitness, as the Whitehead
link shows.
\end{definitionbr}

\textsc{Knots} asks for these classifications from a single rendering,
together with component counts and what-if queries about removing one
loop. Link relation ground truth comes from the scene's link graph,
recorded at generation time. \textsc{Untangle} works on the projected
diagram. Its goal set contains the configurations whose overhead
projection has zero crossings between rope paths.

\subsection{Task Catalog Overview}
\label{app:task_catalog}

\benchmarkname{} comprises 13 task types organized by the five topological
properties and two cognitive levels (Reasoning, Planning).
Table~\ref{tab:app_task_catalog} lists every task with its property
slot, cognitive level, dataset size, and the appendix subsection in
which it is described in detail. The remainder of this section (Appendices~\ref{app:tasks_continuity} to~\ref{app:tasks_knots})
provides the per-task description.

\begin{table*}[!htbp]
\centering
\scriptsize
\begin{tabular*}{\textwidth}{@{\extracolsep{\fill}}lllrrrl@{}}
\toprule
Task & Property & Cognitive level & Easy & Hard & Total & Section \\
\midrule
2D Maze & Continuity & Reasoning & $150$ & $150$ & $300$ & \S\ref{app:task_2d_maze} \\
3D Maze & Continuity & Reasoning & $150$ & $150$ & $300$ & \S\ref{app:task_3d_maze} \\
Pipe & Continuity & Planning & $150$ & $150$ & $300$ & \S\ref{app:task_pipe} \\
\midrule
Assembly & Separation & Reasoning & $150$ & $150$ & $300$ & \S\ref{app:task_separation_objects} \\
One Stroke & Separation & Planning & $150$ & $150$ & $300$ & \S\ref{app:task_one_stroke} \\
\midrule
Bead & Order & Reasoning & $150$ & $150$ & $300$ & \S\ref{app:task_bead_string} \\
Origami Point & Order & Reasoning & $150$ & $150$ & $300$ & \S\ref{app:task_origami} \\
Swap & Order & Planning & $150$ & $150$ & $300$ & \S\ref{app:task_swap_puzzle} \\
\midrule
Hole & Enclosure & Reasoning & $150$ & $150$ & $300$ & \S\ref{app:task_hole_detection} \\
Sheep & Enclosure & Reasoning & $150$ & $150$ & $300$ & \S\ref{app:task_sheep} \\
Chat Noir & Enclosure & Planning & $150$ & $150$ & $300$ & \S\ref{app:task_chat_noir} \\
\midrule
Knots & Knots & Reasoning & $150$ & $150$ & $300$ & \S\ref{app:task_knot_detection} \\
Untangle & Knots & Planning & $150$ & $150$ & $300$ & \S\ref{app:task_knots_untangle} \\
\midrule
Reasoning total & & & $1{,}200$ & $1{,}200$ & $2{,}400$ & \\
Planning total & & & $750$ & $750$ & $1{,}500$ & \\
\textbf{Total} & & & $\mathbf{1{,}950}$ & $\mathbf{1{,}950}$ & $\mathbf{3{,}900}$ & \\
\bottomrule
\end{tabular*}
\caption{\benchmarkname{} task catalog. Reasoning instances are single-shot questions
and planning instances are multi-step episodes. Section points to the subsection
that describes each task's scene generator, difficulty levels, and scoring rule.}
\label{tab:app_task_catalog}
\end{table*}

\subsection{Shared Generation and Quality Control}
\label{app:generation_qc}

All benchmark instances are produced automatically by generators specific to
each task that use fixed seeds. Each generator samples a state, renders the
model-visible observation, and derives the reference answer or terminal
success condition from the generated state and metadata, as described in
Section~\ref{sec:data_collection}. Human annotators are not used
to create benchmark labels.

Every exported record stores a generation seed and configuration metadata.
Reasoning records contain the rendered inputs, question, reference answer,
and difficulty. Planning records contain the initial state, difficulty, and
action budget, while the success condition is implemented by the environment.

Quality control differs by task because the validity criteria differ.
Depending on the task, the generator enforces structural constraints, rejects
ambiguous or unsolvable states, verifies visibility when visibility is
required, or checks the requested difficulty constraints. An instance is
exported only after its applicable conditions hold. Planning environments
validate submitted actions and evaluate success from the resulting simulator
state.

\subsection{Easy and Hard Levels}
\label{app:level_construction}

Table~\ref{tab:app_level_construction} lists the generator settings that each
level contains. For $11$ tasks, the Easy level keeps $150$ items of the hard
tier of the previous release. The Hard levels of Sheep and Untangle take $50$
items from each of the three tiers of the previous release. Hole takes $50$
items from each tier at the Easy level and $150$ items of the hard tier at the
Hard level. The two levels of a task share no item. The Hard levels of the
remaining $10$ tasks are newly generated, and so is the Easy level of Origami
Point.

\begin{table*}[!htbp]
\centering
\scriptsize
\renewcommand{\arraystretch}{1.1}
\begin{tabular*}{\textwidth}{@{\extracolsep{\fill}}lll@{}}
\toprule
Task & Easy level ($150$ items) & Hard level ($150$ items) \\
\midrule
2D Maze & hard tier, $6{\times}6$ mazes & new, $12{\times}12$, $18{\times}18$, and $24{\times}24$ mazes ($50$ each) \\
3D Maze & hard tier, $6$ or $7$ rooms & new, $8$ rooms \\
Pipe & hard tier, no four-port pieces & new, full boards with four-port pieces \\
Assembly & hard tier, mixed distractors & new, aligned similar-part distractors \\
One Stroke & hard tier, $5$ or $6$ colors & new, $9$ or $10$ colors \\
Bead & hard tier, torus-knot loops & new, three curve settings ($50$ each) \\
Origami Point & new, $10$-image sequences & new, $14$-image sequences \\
Swap & hard tier, $4{\times}4$ grids & new, $5{\times}5$ grids with row and column moves \\
Hole & easy, medium, and hard tiers ($50$ each) & hard tier \\
Sheep & hard tier & easy, medium, and hard tiers ($50$ each) \\
Chat Noir & hard tier, greedy cat & new, expert cat \\
Knots & hard tier & new, three crossing settings ($50$ each) \\
Untangle & hard tier, six ropes & easy, medium, and hard tiers ($50$ each) \\
\bottomrule
\end{tabular*}
\caption{Generator settings in the Easy and Hard levels of each task. Tier
entries come from the previous release, and ``new'' items were generated for
this release. The task subsections below give each setting.}
\label{tab:app_level_construction}
\end{table*}

\subsection{Continuity Tasks}
\label{app:tasks_continuity}

\subsubsection{2D Maze \textnormal{(Continuity, Reasoning)}}
\label{app:task_2d_maze}

\paragraph{Targeted ability.}
The 2D Maze environment isolates a model's ability to reason about
\emph{global connectivity} purely from a top-down rendering, with no
symbolic graph input.

\paragraph{Task formulation.}
Two question types are posed over rendered mazes. The reachability
question (\texttt{reachability\_set}) marks several labeled points and
asks which of them are connected to the target point $A$, with the
answer returned as a bracketed name list such as \texttt{[B, D]}. The
bar removal question (\texttt{bar\_removal}) marks two points $A$ and
$B$ that are disconnected in the shown maze and paints a subset of the
walls as colored bars. The model must list every bar whose removal on
its own reconnects $A$ and $B$, for example \texttt{[purple, red]}.
Both types share \texttt{answer\_type = name\_list} and are scored by
set equality. The benchmark contains $300$ questions, $150$ at each
difficulty level. The Easy level has $61$ reachability and $89$ bar
removal questions, and the Hard level has $75$ of each type.

\paragraph{Scene generation at the Easy level.}
Easy scenes are small mazes with mixed wall styles, produced in three
deterministic, seed-driven stages with ground truth supplied by a
connectivity oracle.

\textit{Skeleton.} Recursive-backtracking depth-first search (DFS) on the
$N\!\times\!N$ grid with $N=6$ yields a spanning tree, after which
walls are closed or opened until $W = \mathrm{round}(\rho\,M) = 36$ of
the $M = 2N(N{-}1) = 60$ internal wall slots are occupied, which
corresponds to a wall density of $\rho = 0.6$.

\textit{Wall taxonomy.} A fraction of the closed walls is converted
into full diagonal walls (Mode-dA), and a further fraction is demoted
to half-length variants (Mode-B/C for horizontal and vertical edges,
Mode-dB/dC for diagonals). Across the Easy scenes, diagonal walls make
up $39\%$ of all walls in reachability scenes and $25\%$ in bar removal
scenes, and partial walls make up $31\%$ in both. The renderer admits
five wall types in three families (Table~\ref{tab:2d_maze_walls}).
Full walls (Mode-A on a shared cell edge, Mode-dA across a cell
diagonal) block passage, while every partial wall (Mode-B/C/dB/dC)
leaves an unobstructed sub-segment of its host edge and therefore does
not separate the adjacent regions in the underlying maze graph. The
task prompt states this rule explicitly. Mode-dA walls partition their
host cell into two halves through the cell center, so labeled points
at cell centers are never placed in diagonal-bearing cells.

\textit{Annotation.} Reachability scenes place five labeled points
($A$ and four candidates) in diagonal-free cells. Bar removal scenes
place $A$ and $B$ and paint four bars (purple, red, green, and blue) on
blocking walls, which are either full horizontal or vertical edges
(Mode-A) or full diagonals (Mode-dA). Across the $89$ bar removal
questions, $260$ bars lie on axis-aligned walls and $96$ on diagonal
walls. A scene is rejected unless $A$ and $B$ are disconnected, at least
two bars reconnect them when removed individually, and each of the two
components containing $A$ and $B$ covers at least one third of the
board area. The last two constraints rule out answers that can be
guessed from a single bar and lopsided components in which the bars
around the smaller side dominate the correct set.

\textit{Connectivity oracle.} Ground truth is computed by a breadth-first search (BFS) over a
\emph{4-triangle decomposition} of each cell ($T_N, T_E, T_S, T_W$
sharing the cell center). Each horizontal or vertical edge is owned by
one triangle and each cell-internal diagonal separates exactly two of
them, so a single BFS uniformly handles axial walls and diagonal
walls. Because labeled points sit at cell centers (the shared vertex
of all four triangles), the start cell's four triangles are all seeded
at hop~$0$. The same routine drives both the answer key and the
\textsc{oracle} sanity baseline, which by construction achieves
$\mathrm{acc.}=1.000$.

\begin{table*}[!htbp]
\centering
\setlength{\tabcolsep}{6pt}
\scriptsize
\begin{tabular*}{\textwidth}{@{\extracolsep{\fill}}lcl@{}}
\toprule
\textbf{Wall type} & \textbf{Has passage?} & \textbf{Geometry (cell edge length $=1$)} \\
\midrule
Mode-A       & $\times$   & full horizontal/vertical edge, length $1$  \\
Mode-B       & \checkmark & centered half-edge, length $1/2$           \\
Mode-C       & \checkmark & endpoint-anchored half-edge, length $1/2$  \\
Mode-dA      & $\times$   & full cell diagonal, length $\sqrt{2}$      \\
Mode-dB / dC & \checkmark & partial cell diagonal, length $\sqrt{2}/2$ \\
\bottomrule
\end{tabular*}
\caption{Wall taxonomy of the 2D Maze renderer, assuming unit cell
edge length. ``Has passage?'' indicates whether the two regions
adjacent to the wall remain connected in the underlying maze graph.
Easy scenes use all five types, while Hard scenes use only Mode-A
walls.}
\label{tab:2d_maze_walls}
\end{table*}

\paragraph{Scene generation at the Hard level.}
Hard scenes are large mazes built from full edge walls only (Mode-A),
so the difficulty comes from long, winding routes rather than from wall
styles. The level contains $50$ questions on each of the
$12{\times}12$, $18{\times}18$, and $24{\times}24$ grids, split evenly
between the two question types. Each scene starts from a random DFS
spanning tree of the grid, in which every pair of cells is joined by
exactly one corridor route. Two tree edges are then cut, which splits
the maze into three connected regions. A scene is kept only if every
region covers at least $\max(10, 0.1N^2)$ cells and some pair of
regions shares at least $N$ boundary walls, so the regions are large
and irregular rather than small rooms. After the cuts, between
$46.6\%$ and $48.1\%$ of the internal wall slots are closed, depending
on the grid size.

Every answer is tied to a long route. For a route through the maze we
measure its length in cell moves, its number of turns, and its number
of branch entrances, that is, side corridors that open off the route at
its interior cells. On the $12{\times}12$, $18{\times}18$, and
$24{\times}24$ grids, a qualifying route must have at least $25$, $60$,
and $110$ moves, at least $8$, $18$, and $30$ turns, and at least $4$,
$8$, and $12$ branch entrances. In reachability scenes, $A$ lies in
the largest region, every reachable candidate is joined to $A$ by a
qualifying route, and the remaining candidates lie in the other two
regions. Each scene shows six labeled points ($A$ and five
candidates), and the number of reachable candidates is balanced over
$0$ to $5$ ($13$, $13$, $13$, $12$, $12$, and $12$ questions). In bar
removal scenes, $A$ and $B$ lie in two neighboring regions and six
bars are painted in purple, red, green, blue, yellow, and orange. A
correct bar sits on the boundary between the regions of $A$ and $B$,
and the route it opens from $A$ to $B$ must qualify, with at least a
quarter of the minimum length on each side of the bar. Distractor bars
sit on boundaries between other pairs of regions, at a Manhattan
distance of at least $\lfloor N/4\rfloor$ cells from the marked points. The number of
correct bars is $1$, $2$, or $3$ in $25$ questions each, and every
color is correct in exactly $25$ questions. Because every wall is a
full edge, ground truth is a plain BFS over the cell graph. All Hard
answers were rechecked with an independent union-find solver, and
each scene admits a single-wall edit that changes its answer, so the
answer depends on the specific wall layout.

\paragraph{Difficulty levels.}
The Easy level tests connectivity in
a compact maze whose wall styles must be told apart, and the Hard
level tests whether connectivity can be tracked along long routes in
much larger mazes (Table~\ref{tab:2d_maze_tiers}).

\begin{table*}[!htbp]
\centering
\setlength{\tabcolsep}{6pt}
\scriptsize
\begin{tabular*}{\textwidth}{@{\extracolsep{\fill}}lcc@{}}
\toprule
\textbf{Parameter}                         & \textbf{Easy}                         & \textbf{Hard} \\
\midrule
Grid size                                  & $6{\times}6$                          & $12{\times}12$, $18{\times}18$, $24{\times}24$ \\
Wall styles                                & full, partial, diagonal               & full edge only \\
Closed internal wall slots                 & $36$ of $60$ ($\rho=0.6$)              & $46.6\%$ to $48.1\%$ \\
Minimum route length (moves)               & not constrained                       & $25$, $60$, $110$ \\
Minimum route turns                        & not constrained                       & $8$, $18$, $30$ \\
Minimum branch entrances                   & not constrained                       & $4$, $8$, $12$ \\
\midrule
Reachability labeled points                & $5$                                   & $6$ \\
Reachability correct set size              & $0$ to $4$                            & $0$ to $5$ \\
Bar removal bars                           & $4$                                   & $6$ \\
Bar removal correct set size               & $2$ to $4$                            & $1$ to $3$ \\
\midrule
Reachability questions                     & $61$                                  & $75$ \\
Bar removal questions                      & $89$                                  & $75$ \\
\bottomrule
\end{tabular*}
\caption{Difficulty levels of the 2D Maze task. Hard entries with
three values refer to the $12{\times}12$, $18{\times}18$, and
$24{\times}24$ grids, which contribute $50$ questions each. Route
constraints apply to every route that defines a Hard answer. Easy bar
removal scenes additionally require at least two correct bars and
components covering at least one third of the board.}
\label{tab:2d_maze_tiers}
\end{table*}

\paragraph{Output format and scoring.}
Both question types share the \texttt{name\_list} answer contract. The
prompt instructs the model to return a single JSON object whose
\texttt{answer} field holds a list of names, with \texttt{[]} reserved
for the empty case. The legal vocabulary is sample specific, namely the
scene's labeled point names for reachability questions and its bar
colors for bar removal questions. Predictions and ground truth are
coerced to uppercased, whitespace-stripped frozensets and compared by
\emph{set equality}, so order, case, and duplicates do not affect the
score. Parsing and aggregation follow the shared pipeline of
Appendix~\ref{app:eval_design}.
The complete task prompt and qualitative examples appear in
Figures~\ref{fig:app_task_2d_maze_examples}
and \ref{fig:app_task_2d_maze_examples_bar_removal}
(Appendix~\ref{app:task_prompt_cards}).

\subsubsection{3D Maze \textnormal{(Continuity, Reasoning)}}
\label{app:task_3d_maze}

\paragraph{Targeted ability.}
This task evaluates whether a model can infer global connectivity in a
metric 3D environment from visual evidence alone. Unlike the 2D Maze
task, the scene contains furnished rooms, occlusions, and door states,
so the model must integrate multiple views before deciding whether two
marked locations belong to the same connected component.

\paragraph{Task formulation.}
Each instance provides five rendered views of the same
ProcTHOR~\cite{deitke2022procthor} house at $1920{\times}1080$
resolution, namely a top-down view and four oblique views in the front,
right, rear, and left directions relative to the sampled house
orientation. Labeled navigation points are placed on valid floor
locations. We use two reasoning question types. In the target-point
connectivity question, the model is given a source point and must list
every other labeled point reachable from it under the shown door
states. In the door-opening question, two points $A$ and $B$ are
disconnected, every currently closed controllable door is marked with
a distinct color, and the model must list the unique minimum set of
closed doors that must be opened to connect the two points. The
benchmark contains $300$ questions, each with its own rendered house,
$150$ at each difficulty level. The Easy level has $79$ connectivity
and $71$ door-opening questions, and the Hard level has $75$ of each.

\paragraph{Scene generation.}
We instantiate furnished indoor layouts with the ProcTHOR/AI2-THOR
renderer~\cite{deitke2022procthor,kolve2017ai2thor} and treat each house
as a 3D maze over valid navigation locations. For each sampled scene,
the generator fixes the requested number of rooms, controllable doors,
and labeled points, then samples door states and point locations with
a deterministic seed. The five views are rendered from the same state
and stored with the question row. Ground truth is computed by a
navigation connectivity oracle that respects walls, closed doors,
furniture, and other solid obstacles, while open doors and open floor
or corridor spaces remain passable.

At the Easy level, houses have $6$ rooms with $5$ controllable doors or
$7$ rooms with $6$ controllable doors, and every scene shows five
labeled points. Connectivity questions use random door states, with
between $0$ and $6$ doors open, and the correct set contains $0$ to $4$
of the four candidate points. Door-opening questions start with all
five doors closed and colored, and the unique minimum set contains $1$
to $4$ doors.

At the Hard level, every house has $8$ rooms, and the generator adds
constraints that require tracking connectivity across many rooms. In
connectivity questions the house has $8$ controllable doors, of which
$5$ to $7$ are open, and seven labeled points are shown (the source
and six candidates). The source's component spans $6$ or $7$ of the
$8$ rooms, and the correct set contains $2$ to $5$ candidates. At least
two of the connected candidates are reachable only along long routes
through several rooms, and at least one disconnected candidate lies
close to the source's component, typically separated from it by a
single closed door. One
disconnected candidate also serves as a deliberate distractor, reached
by a route whose first doors are open but which is blocked by a closed
door near its end ($37$ questions) or by several such detours that are
each blocked ($38$ questions). In door-opening questions the house has
$7$ or $8$ controllable doors, of which $2$ or $3$ are already open,
and only the $4$ to $6$ closed doors are colored. The prompt therefore
asks the model to count only currently closed doors and to minimize
the number of doors opened rather than walking distance. The unique
minimum set contains $3$ or $4$ doors. In $37$ questions a single
route reaches $B$, and in the other $38$ two candidate routes compete,
so the model must compare how many closed doors each route crosses.
The answer route is the one with more doors in total in $19$ of these
questions and the one with fewer doors in the other $19$, so route
length alone does not reveal the answer. Every Hard answer is
verified against the navigation oracle.

\paragraph{Difficulty levels.}
The two levels differ in the room, door, and point setup and in the
door-state constraints above (Table~\ref{tab:3d_maze_tiers}).

\begin{table*}[!htbp]
  \centering
  \scriptsize
  \begin{tabular*}{\textwidth}{@{\extracolsep{\fill}}llcccccc@{}}
    \toprule
    Level & Question type & Rooms & Doors & Points & Open doors & Answer size & Questions \\
    \midrule
    Easy & Connectivity  & $6$ or $7$ & $5$ or $6$ & $5$ & $0$ to $6$ & $0$ to $4$ & $79$ \\
    Easy & Door opening  & $6$        & $5$        & $5$ & $0$        & $1$ to $4$ & $71$ \\
    Hard & Connectivity  & $8$        & $8$        & $7$ & $5$ to $7$ & $2$ to $5$ & $75$ \\
    Hard & Door opening  & $8$        & $7$ or $8$ & $5$ & $2$ or $3$ & $3$ or $4$ & $75$ \\
    \bottomrule
  \end{tabular*}
  \caption{Difficulty levels of the 3D Maze task. Doors counts the
  controllable doors in the house, and answer size counts the points
  or doors in the correct set.}
  \label{tab:3d_maze_tiers}
\end{table*}

\paragraph{Output format and scoring.}
For target-point connectivity questions, the expected answer is a JSON
object whose \texttt{answer} field lists point names in alphabetical
order, such as \texttt{["B","D"]}, and the empty set is written as
\texttt{[]}. Door-opening questions use the same schema with door color
names. Both are scored by exact set equality after normalization.
The complete task prompt and qualitative examples appear in
Figures~\ref{fig:app_task_3d_maze_examples}
and \ref{fig:app_task_3d_maze_examples_door_open}
(Appendix~\ref{app:task_prompt_cards}).

\subsubsection{Pipe \textnormal{(Continuity, Planning)}}
\label{app:task_pipe}

\paragraph{Targeted ability.}
Pipe evaluates sequential continuity reasoning under local
rotations. A successful model must identify disconnected pipe
components, anticipate how 90-degree rotations change openings, and
plan a sequence that connects every pipe segment to the source.

\paragraph{Task formulation.}
The model observes a $5{\times}5$ pipe board at each step. The green
source marks the root of the network, connected pipes are rendered in
green, and disconnected pipes in blue. At every turn, the model chooses
one non-empty cell and the environment rotates that pipe clockwise by
$90^\circ$. An episode succeeds when all non-empty pipe cells are
connected to the source through matching openings before the action
budget is exhausted. The benchmark contains $300$ episodes, $150$ at
each difficulty level.

\paragraph{Scene generation.}
Each puzzle is generated from a connected tree-shaped pipe network
whose solved state is stored with the episode. The generator samples
the source cell, the active pipe count, and the number of three-way
junctions, rejects loops, and then scrambles the pipes by seeded
random rotations. This construction preserves a known solved state
while requiring the model to recover it through local rotations from
the rendered board. The action budget is
$\lceil 1.3\,R\rceil$, where $R$ is the number of clockwise rotations
in the reference solution.

At the Easy level, $17$ to $23$ of the $25$ cells hold a pipe and the
network has $5$ to $7$ three-way junctions. Four-way junctions are
excluded, the reference solution needs $23$ to $27$ rotations, and
between $10$ and $20$ pipes receive a nonzero scramble rotation.

At the Hard level, every one of the $25$ cells holds a pipe and the
network again has $5$ to $7$ three-way junctions, which leaves $7$ to
$9$ single-port endpoints. The board also contains two kinds of
four-port pieces, each carrying two separate channels rather than a
four-way junction. A raised overpass connects north to south and east
to west without joining the two channels, and its connections do not
change under rotation. A double bend shows two separate quarter-circle
curves that pair north with east and south with west, or north with
west and south with east, and one clockwise quarter-turn switches
between the two pairings. Every Hard board contains $1$ to $3$
overpasses and $1$ to $3$ double bends, $2$ to $4$ four-port pieces in
total, all placed in interior cells, and the per-turn prompt explains
these rules. The solved state is a tree over channels, each channel
turns green independently, and both channels of every four-port piece
must reach the source. Hard boards are also constrained to be hard to
read off locally. Only $1$ to $3$ channels are connected to the source
at the start, counting the source itself, at least $20$ pipes start in
a wrong orientation, and at least one double bend starts with the
wrong pairing. The generator further proves that the solved
orientation is unique. After propagating the constraints imposed by
the board boundary and by neighboring pipes, at least $14$ pipes must
still admit several orientations, and at least $6$ must remain
ambiguous after every single-pipe trial assignment has been tested and
its consequences propagated. Because the solution is unique, its
$40$ to $48$ rotations are the minimum number of actions needed.

\paragraph{Difficulty levels.}
The two levels differ in network density, the presence of two-channel
pieces, the scrambling constraints, and the reference solution length
(Table~\ref{tab:pipe_tiers}).

\begin{table*}[!htbp]
  \centering
  \scriptsize
  \begin{tabular*}{\textwidth}{@{\extracolsep{\fill}}lccccccc@{}}
    \toprule
    Level & Grid & Active pipes & Junctions & Four-port pieces & Reference rotations & Budget & Episodes \\
    \midrule
    Easy & $5{\times}5$ & $17$ to $23$ & $5$ to $7$ & $0$        & $23$ to $27$ & $30$ to $36$ & $150$ \\
    Hard & $5{\times}5$ & $25$         & $5$ to $7$ & $2$ to $4$ & $40$ to $48$ & $52$ to $63$ & $150$ \\
    \bottomrule
  \end{tabular*}
  \caption{Difficulty levels of the Pipe task. Junctions counts
  three-way junctions, and four-port pieces are the overpasses and
  double bends of the Hard level. The budget equals $\lceil 1.3\,R\rceil$
  for a reference solution of $R$ rotations.}
  \label{tab:pipe_tiers}
\end{table*}

\paragraph{Output format and scoring.}
Each action is a JSON object whose \texttt{answer} field holds the
integer coordinates \texttt{x} (column) and \texttt{y} (row) of the
cell to rotate, following the visual labels on the board. We evaluate
the full trajectory rather than individual moves. An episode is
correct if the environment reaches a state in which every non-empty
pipe, and at the Hard level every channel of every pipe, belongs to
the source-connected component within the prescribed step budget.
The complete task prompt and qualitative examples appear in
Figure~\ref{fig:app_task_pipe_examples} (Appendix~\ref{app:task_prompt_cards}).

\subsection{Separation Tasks}
\label{app:tasks_separation}

\subsubsection{Assembly \textnormal{(Separation, Reasoning)}}
\label{app:task_separation_objects}

\paragraph{Targeted ability.}
Assembly tests whether a model can recover object decomposition from a
rendered 3D assembly. The task targets separation reasoning. The
complete object must be mentally partitioned into valid subassemblies,
while distractors preserve plausible shape and category cues.

\paragraph{Task formulation.}
Each instance contains one image of a complete object and five
candidate decomposition options. The complete-object image combines
two oblique views, while each option shows a proposed split into two
subassemblies. The model must select the option that exactly matches a
valid decomposition of the original object.

\paragraph{Scene generation.}
We build the task from a catalog of assembly-style objects with known
primitive parts and object categories. An object enters the pool only
if every primitive part is visible in the two fixed complete-object
views, with at least $\min(5\,\text{px}, 5\%)$ of the part's solo
rendered area in view. For each instance, the generator selects a
balanced two-part split as the correct option, in which the two
subassemblies differ in size by at most one part, and constructs four
distractors. Each distractor is built on a different two-part
partition of the same object, so no option can be rejected from its
partition alone. The five options are shuffled deterministically.
After rendering, a scene is kept only if every primitive part shown in
every option keeps at least $\min(13\,\text{px}, 13\%)$ of its solo
rendered area visible in that option view. Otherwise it is regenerated
from the next scene seed. Each question therefore consists of one
complete-object rendering and five option renderings.

\paragraph{Difficulty levels.}
The benchmark contains $300$ reasoning questions, $150$ per level,
built from $17$ catalog objects in six categories. Every object in
both levels has at least $11$ primitive parts, and the two levels
differ mainly in how the distractors are constructed
(Table~\ref{tab:assembly_tiers}). The Easy level uses $16$ objects
with $11$ to $19$ parts from all six categories. Its four distractors
are structurally heterogeneous. One adds an extra same-category
component, two replace one target component with a similar component
from another object of the same category, and one removes a target
component. Part counting or a single mismatched component can
therefore expose some of these distractors. The Hard level uses $12$
objects with $11$ to $15$ parts from four categories. All four of its
distractors are similar-part replacements. Each one swaps exactly one
target part for a part taken from a different object of the same
category, scaled and rotated to the bounding box of the removed part
(per-axis size ratios between $0.61$ and $1.59$). Every Hard option
thus has the same number of parts as the complete object and differs
from it only in the shape of one component. A per-option visibility
check additionally requires the substituted part to stay visible in
the option rendering, so the deciding evidence is always in view.
Each object is sampled with several seeds so that option ordering and
distractor selection vary while the underlying decomposition remains
exact.

\begin{table*}[!htbp]
  \centering
  \scriptsize
  \begin{tabular*}{\textwidth}{@{\extracolsep{\fill}}lccccc@{}}
    \toprule
    Level & Part count & Objects & Object categories & Distractors & Questions \\
    \midrule
    Easy & $11$ to $19$ & $16$ & Bench, Chair, Desk, Misc, Shelf, Table & $1$ extra, $1$ missing, $2$ replaced & $150$ \\
    Hard & $11$ to $15$ & $12$ & Bench, Chair, Misc, Table & $4$ aligned similar-part replacements & $150$ \\
    \bottomrule
  \end{tabular*}
  \caption{Difficulty levels for the Assembly task.}
  \label{tab:assembly_tiers}
\end{table*}

\paragraph{Output format and scoring.}
The canonical response is a multiple-choice JSON answer such as
\texttt{\{"answer":"E"\}}. Predictions are normalized to the option
letter and scored by exact match against the shuffled correct option.
The complete task prompt and qualitative examples appear in
Figures~\ref{fig:app_task_separation_objects_examples_bench_sialland}
to \ref{fig:app_task_separation_objects_examples_chair_applaro}
(Appendix~\ref{app:task_prompt_cards}).

\subsubsection{One Stroke \textnormal{(Separation, Planning)}}
\label{app:task_one_stroke}

\paragraph{Targeted ability.}
One Stroke evaluates separation-aware path planning. The model must
draw a single continuous stroke that separates differently colored
regions while keeping cells of the same color connected on the same
side of the stroke.

\paragraph{Task formulation.}
The environment presents a colored grid with a cursor starting at the
bottom-left corner and a target at the top-right corner. At every turn,
the model issues one of four moves, \texttt{U}, \texttt{D},
\texttt{L}, or \texttt{R}. The stroke cannot leave the board, reuse an
edge, or create a closed loop. Moving back over the most recent edge is
legal and undoes that edge. The episode succeeds only when the completed
path reaches the target while satisfying the color-region separation
constraint.

\paragraph{Scene generation.}
Every puzzle is played on a $6{\times}6$ cell board, which corresponds
to a $7{\times}7$ grid of stroke vertices. The generator first samples
a legal construction path, a simple lattice path from the bottom-left
to the top-right vertex, and computes the regions into which this path
cuts the board. It then gives each of several regions its own color
and fills part of that region's cells with it, leaving the remaining
cells uncolored, so that the construction path satisfies the grouping
and separation constraints by design. Candidates are rejected when the
path has too few turns, cuts the board into too few regions, or does
not realize the required number of colors. The generator then searches
for a shorter valid stroke and stores the shortest one it finds as the
reference solution, which is therefore never longer than the
construction path. The per-instance action budget is $1.2$ times the
construction-path length, rounded up.

\paragraph{Difficulty levels.}
Both levels share the board size and differ in the number of colors,
the number of uncolored cells, and the length of the stroke the model
must plan (Table~\ref{tab:one_stroke_tiers}). Easy
puzzles use $5$ colors ($6$ in a single instance) from a seven-color
palette, leave $0$ to $13$ of the $36$ cells uncolored, and are
accepted only if an exhaustive breadth-first search finds a shortest
valid solution of exactly $20$ moves. Their construction paths have
$20$ to $28$ edges. Hard puzzles use $9$ or $10$ colors from a
ten-color palette and leave $11$ to $14$ cells uncolored. Their
reference solutions have $40$ to $44$ moves with $25$ to $36$ turns
and visit $41$ to $45$ of the $49$ vertices, so the stroke has to
weave through nearly the entire board while every turn must respect
the color constraints. The benchmark contains $300$ planning episodes,
$150$ per level.

\begin{table*}[!htbp]
  \centering
  \scriptsize
  \begin{tabular*}{\textwidth}{@{\extracolsep{\fill}}lccccccc@{}}
    \toprule
    Level & Board & Vertex grid & Colors & Uncolored cells & Reference solution & Budget & Episodes \\
    \midrule
    Easy & $6{\times}6$ & $7{\times}7$ & $5$ to $6$  & $0$ to $13$  & $20$         & $24$ to $34$ & $150$ \\
    Hard & $6{\times}6$ & $7{\times}7$ & $9$ to $10$ & $11$ to $14$ & $40$ to $44$ & $48$ to $53$ & $150$ \\
    \bottomrule
  \end{tabular*}
  \caption{Difficulty levels for the One Stroke planning task.}
  \label{tab:one_stroke_tiers}
\end{table*}

\paragraph{Output format and scoring.}
The action contract is \texttt{\{"answer":"U"\}} with the answer drawn
from \texttt{\{U,D,L,R\}}. We score the full trajectory. An episode is
successful if the model reaches the target within the action budget and
the final stroke satisfies the same-color grouping and different-color
separation constraints.
The complete task prompt and qualitative examples appear in
Figure~\ref{fig:app_task_one_stroke_examples} (Appendix~\ref{app:task_prompt_cards}).

\subsection{Order Tasks}
\label{app:tasks_order}

\subsubsection{Bead \textnormal{(Order, Reasoning)}}
\label{app:task_bead_string}

\paragraph{Targeted ability.}
Bead tests whether a model can recover color order along a
string embedded as a complex 3D curve, including curves that close
into a loop. Each bead is identified by its color. A white marker
designates the first bead and a tangential arrow specifies the
traversal direction, so the required structure is the ordered color
sequence encountered from that marked starting point.

\paragraph{Task formulation.}
The benchmark contains two question types. The sequence task
(\texttt{T\_BS01}) presents one image and asks the model to list every
bead color in traversal order. The image marks the start bead and the
traversal direction, and at the Hard level the prompt also tells the
model to start at the marked bead and follow the indicated direction. The colors are separated by commas, as in
\texttt{"RED, BLUE, GREEN"}. The relationship task
(\texttt{T\_BS02}) presents two images that each show a bead string and
asks whether their full color sequences are \texttt{IDENTICAL},
\texttt{REVERSED}, a \texttt{CYCLIC\_ROTATION} of one another, or
\texttt{DIFFERENT}.

\paragraph{Scene generation.}
Scenes are rendered with a Vite and Playwright pipeline over a
Three.js renderer, and bead colors come from a fixed eight-color
palette. Each level has its own scene generator. The Easy generator
samples a tangled torus-knot loop and places beads at equal
arc-length intervals along it. Its color sequences include
palindromic and periodic patterns, and scenes are rendered at
$1024{\times}1024$, mostly from isometric-front or oblique viewpoints.
The Hard generator has three settings. The first renders tangled
open and closed curves with few crossings. The other two build a
single rope from a random braid in which $7$ or $8$ strands exchange
positions over $38$ to $46$ crossing rows, or $9$ or $10$ strands over
$58$ to $68$ rows. Exterior connectors join the strand ends into one
component, which is cut once when an open string is required. Beads
sit on the connectors and, for some strands, inside the braid passes,
which leaves long bead-free stretches of rope that must be traced
through many crossings. Hard sequence questions use exactly six
colors, never repeat a color on adjacent beads, and have no rotational
or reflective symmetry. For relationship questions, the two strings
share the same color multiset and the same first and last colors (the
first two and last two in braid scenes), so the relation cannot be
read off from color counts or endpoints. Hard scenes are rendered at
$2560{\times}2560$ without ground grid or cast shadows, and every
image passes geometric clearance checks, a minimum crossing angle of
$12^{\circ}$, and a raster check that every bead is fully visible. Their
ground truth is additionally checked against the bead encounter order
along the rendered rope. In both levels, ground truth is the generated
color sequence and is exactly reproducible from the deterministic
scene seed.

\paragraph{Difficulty levels.}
The two levels differ in bead count, curve topology, and the number of
projected rope crossings (Table~\ref{tab:bead_tiers}). The benchmark contains $300$ reasoning
questions, $150$ per level, with $215$ sequence questions and $85$
relationship questions in total. The Easy level has $137$ sequence
and $13$ relationship questions on torus-knot loops with $6$ to $8$
beads. Of its relationship questions, $12$ are \texttt{IDENTICAL} and
one is \texttt{REVERSED}. The Hard level has $78$ sequence and $72$
relationship questions, with $18$ questions for each relation label.
It combines three generator settings of $50$ questions each ($26$
sequence and $24$ relationship questions), namely tangled curves with
$11$ to $14$ beads, $7$ or $8$ strand braids with $14$ to $16$ beads,
and $9$ or $10$ strand braids with $18$ to $20$ beads. Half of the
Hard sequence images show closed loops and half show open strings, and
every relation label receives the same mix of open and closed image
pairs.

\begin{table*}[!htbp]
  \centering
  \scriptsize
  \begin{tabular*}{\textwidth}{@{\extracolsep{\fill}}llcccc@{}}
    \toprule
    Level & Curve family & Bead count & Projected crossings & Resolution & Questions (seq./rel.) \\
    \midrule
    Easy & torus-knot loops & $6$ to $8$ & n/a & $1024{\times}1024$ & $150$ ($137$/$13$) \\
    \midrule
    Hard & tangled open and closed curves & $11$ to $14$ & $3$ to $10$ & $2560{\times}2560$ & $50$ ($26$/$24$) \\
    Hard & $7$ or $8$ strand braids & $14$ to $16$ & $54$ to $72$ & $2560{\times}2560$ & $50$ ($26$/$24$) \\
    Hard & $9$ or $10$ strand braids & $18$ to $20$ & $93$ to $120$ & $2560{\times}2560$ & $50$ ($26$/$24$) \\
    \bottomrule
  \end{tabular*}
  \caption{Difficulty levels for the Bead task. Projected crossings
  count rope crossings in the rendered image and are recorded for the
  Hard scenes only.}
  \label{tab:bead_tiers}
\end{table*}

\paragraph{Output format and scoring.}
\texttt{T\_BS01} returns a JSON string such as
\texttt{\{"answer":"RED, BLUE, GREEN"\}}. The parser converts this
string into an ordered list of color names from the eight color
palette. A prediction is correct only when both order and multiplicity
match the ground truth exactly. \texttt{T\_BS02} returns one of
\texttt{IDENTICAL},
\texttt{REVERSED}, \texttt{CYCLIC\_ROTATION}, or \texttt{DIFFERENT}
and is scored by exact match after case normalization.
The complete task prompt and qualitative examples appear in
Figures~\ref{fig:app_task_bead_string_examples_description}
and \ref{fig:app_task_bead_string_examples_pair_relationship}
(Appendix~\ref{app:task_prompt_cards}).

\subsubsection{Origami Point \textnormal{(Order, Reasoning)}}
\label{app:task_origami}

\paragraph{Targeted ability.}
Under Piaget's framework of conservation under transformation~\cite{piaget2013child}, this task tests whether current multimodal large language models are capable of \emph{distinguishing the invariant} (the identity of marked points on a piece of paper) from \emph{the variant} (their Euclidean location in 3D space) as the paper is folded.

\paragraph{Task formulation.}
We modify the Origami Simulator~\cite{ghassaei2018origami} to predefine points on the paper and to play a preset folding and rotation trajectory in front of a fixed camera. The task contains $300$ instances built on six origami bases (bird, boat, open-sink, pinwheel, square, and waterbomb), with $150$ instances at each of the two difficulty levels. Each instance is a sequence of rendered images in which one base folds from a flat sheet to $70\%$ of its folded state while the paper rotates between images. The first image shows the flat sheet with one to three unmarked dots. The intermediate images show the paper without any dots, and the final image shows the folded paper with every dot labeled by an upper-case letter. The model must report the letters in the final image that belong to the dots shown on the flat sheet. Dots in the final image are drawn at randomized sizes, and the prompt states that size carries no identity information, so the dots can only be matched by following the paper through the fold sequence.

\paragraph{Scene generation.}
Each dot lies at a fixed barycentric coordinate on a face of the crease pattern, so its position in every frame follows from the simulated mesh. Besides the tracked dots, the generator places decoy dots in clusters around them, and every tracked dot has a twin decoy on a different face, placed under one of the base's rotational symmetries ($90^\circ$, $180^\circ$, or $270^\circ$), so that local appearance alone does not identify the target. For each preset, the generator searches over sequences of (fold amount, model rotation) pairs and accepts a trajectory only if every tracked dot remains geometrically visible at every step and the labeled dots in the final image are separated by a minimum distance in pixels. Visibility is computed in closed form by casting a ray from each dot to the camera and testing it against the full mesh, since folded panels block sight from either side. Every label is a geometric property of the simulated mesh rather than a human annotation, so ground truth is exact and reproducible from a fixed seed.

\paragraph{Difficulty levels.}
The two levels differ in sequence length, rotation, and distractor design (Table~\ref{tab:order_origami}). Easy instances show $10$ images, carry two or three tracked dots among $20$ lettered dots, and place every dot on the front side of the paper. The paper turns gradually and mostly in one direction, by about $13^\circ$ per image in the median step, accumulating between $26^\circ$ and $308^\circ$ of rotation (median $118^\circ$). Hard instances show $14$ images and carry one or two tracked dots among $16$ to $22$ lettered dots, of which up to eight lie on the underside of the paper and become visible only after the paper turns over. Rotation in Hard is both larger and non-monotone, as the paper swings back and forth by about $30^\circ$ per image and accumulates between $288^\circ$ and $419^\circ$ of rotation (median $380^\circ$), although its final pose is at most $179^\circ$ from the initial one. Hard also uses a smaller minimum separation between lettered dots, so the final image is more crowded.

\begin{table*}[!htbp]
  \centering
  \scriptsize
  \begin{tabular*}{\textwidth}{@{\extracolsep{\fill}}lcc@{}}
    \toprule
    Parameter & Easy & Hard \\
    \midrule
    Origami bases (instances) & \shortstack{bird $30$, boat $29$, pinwheel $29$,\\ waterbomb $29$, open-sink $23$, square $10$} & \shortstack{pinwheel $73$, bird $53$,\\ waterbomb $23$, boat $1$} \\
    Images per instance & $10$ & $14$ \\
    Fold amount & $0\%$ to $70\%$ & $0\%$ to $70\%$ \\
    Unmarked dots on the flat sheet & $2$ ($103$), $3$ ($47$) & $1$ ($121$), $2$ ($29$) \\
    Lettered dots in the final image & $20$ & $16$ to $22$ \\
    Dots on the back side & $0$ & $0$ to $8$ \\
    Accumulated rotation & $26^\circ$ to $308^\circ$ & $288^\circ$ to $419^\circ$ \\
    Largest rotation between consecutive images & $3^\circ$ to $34^\circ$ & $23^\circ$ to $37^\circ$ \\
    Minimum dot separation (px) & $50$ & $39$ \\
    Final dot size scale & $0.6$ to $1.4$ & $0.75$ to $1.25$ \\
    Questions & $150$ & $150$ \\
    \bottomrule
  \end{tabular*}
  \caption{Easy and Hard levels of the Origami Point task. Counts in parentheses give the number of instances.}
  \label{tab:order_origami}
\end{table*}

\paragraph{Output format and scoring.}
The expected response is a JSON list of upper-case letters drawn from the instance-specific label set, e.g.\ \texttt{\{"answer":["A","C"]\}}. The ground truth always contains between one and three letters. Predictions and ground truth are normalized to sets and compared by set equality, so the answer is order-insensitive but identity-sensitive. Every emitted letter must belong to the label set of the final image, and a missing or spurious letter makes the answer incorrect.
The complete task prompt and qualitative examples appear in
Figures~\ref{fig:app_task_origami_examples_bird_base} to~\ref{fig:app_task_origami_examples_waterbomb_base}
(Appendix~\ref{app:task_prompt_cards}).

\subsubsection{Swap \textnormal{(Order, Planning)}}
\label{app:task_swap_puzzle}

\paragraph{Targeted ability.}
Swap evaluates whether a model can reason about order through a
sequence of constrained permutations. The model must transform an
initial grid into a target grid by using the blank cell as the only
exchange medium.

\paragraph{Task formulation.}
Each episode provides two images at every turn, the current
arrangement and the goal arrangement. The grid contains one blank cell,
and every other cell holds a colored block marked with a unique letter.
At each step, the model selects one non-empty cell, and the chosen block
is swapped with the blank. At the Easy level any non-empty cell may be
selected. At the Hard level the selected cell must share a row or a
column with the blank. It need not be adjacent to the blank, and the
blocks between the two cells do not move. An illegal action leaves the
grid unchanged but still consumes a step. The task is solved when the
current arrangement exactly matches the goal arrangement. The benchmark
contains $300$ episodes, $150$ at each level.

\paragraph{Scene generation.}
Easy episodes use a $4{\times}4$ grid with $15$ blocks. Under a
deterministic seed, the generator samples both the initial and the goal
arrangement and computes the exact shortest-path distance in the state
graph where any non-empty cell may swap with the blank. The step budget
is this theoretical minimum multiplied by $1.2$ and rounded up.
Hard episodes use a $5{\times}5$ grid with $24$ blocks, and the goal
places the blocks in alphabetical order with the blank in the bottom
right corner. The seed selects a target distance between $40$ and $44$.
The blank starts at its goal position and all $24$ blocks start
misplaced, each with a wrong row, a wrong column, or both, so that the
total number of wrong coordinates equals the target distance. A bounded
beam search then looks for a legal solution with exactly that many
moves, and boards without such a solution are rejected. Because a
same-row or same-column swap changes only one coordinate of the moved
block, each move corrects at most one wrong coordinate, so the number of
wrong coordinates is a lower bound on the solution length and the
solution found is provably shortest. Every Hard episode receives a fixed
budget of $75$ actions.

\paragraph{Difficulty levels.}
The two levels differ in grid size, move rule, and solution length
(Table~\ref{tab:swap_tiers}). Hard episodes require roughly three times
as many moves as Easy ones, and the row and column restriction forces
the model to route the blank through intermediate cells instead of
fetching any block directly.

\begin{table*}[!htbp]
  \centering
  \scriptsize
  \begin{tabular*}{\textwidth}{@{\extracolsep{\fill}}lcc@{}}
    \toprule
    Parameter & Easy & Hard \\
    \midrule
    Grid shape & $4{\times}4$ & $5{\times}5$ \\
    Colored blocks & $15$ & $24$ \\
    Legal cells & any non-empty cell & same row or column as the blank \\
    Goal arrangement & sampled & alphabetical, blank in the corner \\
    Misplaced blocks at start & $9$ to $15$ & $24$ \\
    Blank starts at its goal position & $12$ of $150$ & $150$ of $150$ \\
    Shortest solution (steps) & $11$ to $18$ & $40$ to $44$ \\
    Step budget & $14$ to $22$ & $75$ \\
    Episodes & $150$ & $150$ \\
    \bottomrule
  \end{tabular*}
  \caption{Easy and Hard levels of the Swap task.}
  \label{tab:swap_tiers}
\end{table*}

\paragraph{Output format and scoring.}
Actions specify a grid position, e.g.,
\texttt{\{"answer":\allowbreak\{"row":1,\allowbreak"col":2\}\}}.
We evaluate the executed episode, and the prediction is correct only if the model reaches the
goal arrangement within the step budget.
The complete task prompt and qualitative examples appear in
Figure~\ref{fig:app_task_swap_puzzle_examples} (Appendix~\ref{app:task_prompt_cards}).

\subsection{Enclosure Tasks}
\label{app:tasks_enclosure}

\subsubsection{Hole \textnormal{(Enclosure, Reasoning)}}
\label{app:task_hole_detection}

\paragraph{Targeted ability.}
Hole evaluates enclosure reasoning over solid 3D objects. The model
must count only through-holes that connect the top surface to open
space, while ignoring pits, shadows, and shallow depressions that do
not pass through the board.

\paragraph{Task formulation.}
Each instance shows a single top-down rendering of a procedurally
generated board. The prompt asks for the number of visible holes on
the top board. If a lower board is visible in the scene, the prompt
instructs the model to ignore it and evaluate only the top board.

\paragraph{Scene generation.}
The renderer builds a $10{\times}10$ board whose outline is drawn from
rectangles, circles, and polygons, and views it from almost directly
above. It then cuts four kinds of openings. A plain hole passes through
the board and counts once. A pit is a blind depression that does not
pass through the board and does not count. A partially floored hole
passes through the board but has a shelf at half the board thickness
covering $25\%$ to $50\%$ of its extent along one edge, so part of it
looks like a pit, and it counts once. A split hole has a band at half
the board thickness across its middle that divides it into two through
openings, so it counts twice. For half of
the seeds, a second board is placed partly beneath the top board with a
small offset, so that open space under some holes shows another surface.
Ground truth is generated directly from the procedural layout as the
number of through openings that remain visible on the top board.

\paragraph{Difficulty levels.}
The generator has three settings, the easy, medium, and hard tiers of the
previous release, which widen both the answer range and the set of opening
types. The first uses plain holes and one to three pits, the second adds
one to three partially floored holes, and the third uses all four opening
types with at least one of each. The Easy level contains $50$ instances
from each tier, and the Hard level contains $150$ instances from the hard
tier.
Each instance has one rendered image, and the lower
board appears in $74$ instances at each level.
Table~\ref{tab:hole_tiers} reports the composition and the ground-truth
range of both levels.

\begin{table*}[!htbp]
  \centering
  \scriptsize
  \begin{tabular*}{\textwidth}{@{\extracolsep{\fill}}lcc@{}}
    \toprule
    Parameter & Easy & Hard \\
    \midrule
    Previous-release tiers & easy, medium, hard ($50$ each) & hard ($150$) \\
    Board shapes & circle $56$, rect $50$, polygon $44$ & circle $57$, polygon $50$, rect $43$ \\
    Lower board present & $74$ & $74$ \\
    Ground-truth hole count & $4$ to $18$ & $7$ to $17$ \\
    Mean ground-truth count & $10.2$ & $12.3$ \\
    Questions & $150$ & $150$ \\
    \bottomrule
  \end{tabular*}
  \caption{Easy and Hard levels of the Hole reasoning task. Within the Easy level, ground-truth counts range from $4$ to $10$, $7$ to $13$, and $9$ to $18$ for the easy, medium, and hard tiers.}
  \label{tab:hole_tiers}
\end{table*}

\paragraph{Output format and scoring.}
The expected answer is a scalar integer, e.g.,
\texttt{\{"answer":6\}}. Predictions are scored by exact integer match
after parsing, so off-by-one counts and counts that include non-through
pits are marked incorrect.
The complete task prompt and qualitative examples appear in
Figure~\ref{fig:app_task_hole_detection_examples} (Appendix~\ref{app:task_prompt_cards}).

\subsubsection{Sheep \textnormal{(Enclosure, Reasoning)}}
\label{app:task_sheep}

\paragraph{Targeted ability.}
Sheep probes inside and outside judgments and bounded-region reasoning
in nested and partitioned fence layouts. The model must recover each
labeled sheep's enclosing layer or partition cell and determine
whether an opening in the outer fence leaves a route to the outside.
These judgments operationalize the enclosure relation induced by
closed planar boundaries \citep[\S 2.B]{hatcher2002algebraic}.

\paragraph{Task formulation.}
Each instance shows an elevated 3D rendering of a pasture in which
numbered sheep are scattered across a fence layout. The benchmark
contains four question types. Q1, asked for both scene families,
requires counting the sheep that cannot reach the outside without
crossing a fence. For nested-fence scenes, Q2 returns the sheep IDs
that can escape through outer-fence gaps, and Q4 asks how many such
gaps must be repaired to close the outermost fence. For partitioned
scenes, Q3 asks which labeled cell contains the most sheep.

\paragraph{Scene generation.}
Scenes are produced by a Vite and Playwright renderer that supports two
layout families. \emph{Nested-fence} scenes contain two to four nested
convex, concave, or irregular boundaries, with zero or more gaps placed
only on the outermost fence. \emph{Partitioned} scenes divide one
closed enclosure into labeled cells using one of five layouts, namely
grid, hex-cross, nested polygon, polygon star, or radial. For each
scene, the generator samples its shape or partition parameters, the
camera tilt away from the overhead view, and the sheep coordinates from
a deterministic seed, then derives ground truth from geometric layer
and cell membership. In a nested-fence scene, a sheep in the outermost
layer can escape whenever the outer fence has at least one gap, whereas
sheep in deeper layers cannot. In a partitioned scene, every sheep
inside the closed outer boundary is counted as unable to escape. Scenes
are resampled until all requested sheep satisfy spacing constraints,
and partitioned scenes must have a unique most-populated cell
containing at least two sheep.

\paragraph{Difficulty levels.}
Sheep contains $300$ questions, $150$ at each of the Easy and Hard
levels (Table~\ref{tab:sheep_tiers}). The Easy level uses the hard
tier of the previous release, its densest setting, throughout. Its nested-fence scenes have
three or four concave or irregular layers, $13$ to $18$ sheep, and one
to five gaps in the outer fence. Its partitioned scenes cover all five
layouts and place $12$ to $16$ sheep in $4$ to $16$ cells. Every Easy
scene is viewed with a camera tilt of $30^\circ$ or $45^\circ$. The
level contains $56$ Q1 questions ($24$ nested and $32$ partitioned),
$45$ Q2, $4$ Q3, and $45$ Q4 questions. The Hard level takes $50$
questions from each of the three tiers of the previous release. They range from two or
three layers with $7$ to $10$ sheep, through two to four layers with
$10$ to $14$ sheep, to three or four layers with $13$ to $18$ sheep,
and together cover zero to five outer gaps and camera tilts from
$15^\circ$ to $45^\circ$. Only $3$ Hard scenes are partitioned, each
with $11$ sheep in $6$ to $8$ cells. The Hard level is weighted toward
escape reasoning, with $117$ Q2, $21$ Q4, and $12$ Q1 questions ($9$
nested and $3$ partitioned) and no Q3 question. Twelve of its Q2
questions have the answer \texttt{"NONE"}, and eleven of its Q4
questions require no repair. The $300$ questions are posed on
$214$ rendered scenes, and the $30$ scenes that appear at both levels
are asked with a different question type at each level.

\begin{table*}[!htbp]
  \centering
  \scriptsize
  \setlength{\tabcolsep}{5pt}
  \begin{tabular*}{\textwidth}{@{\extracolsep{\fill}}llcccccc@{}}
    \toprule
    Level & Tier & Nested layers & Outer gaps & Sheep (nested/partitioned) & Partition cells & Camera tilt & Q1/Q2/Q3/Q4 \\
    \midrule
    Easy & hard & $3$ to $4$ & $1$ to $5$ & $13$ to $18$ / $12$ to $16$ & $4$ to $16$ & $30^\circ$, $45^\circ$ & $56$/$45$/$4$/$45$ \\
    \midrule
    Hard & easy & $2$ to $3$ & $0$ to $2$ & $7$ to $10$ / $11$ & $6$ to $8$ & $15^\circ$, $30^\circ$ & $5$/$33$/$0$/$12$ \\
         & medium & $2$ to $4$ & $0$ to $3$ & $10$ to $14$ / n/a & n/a & $15^\circ$ to $45^\circ$ & $5$/$40$/$0$/$5$ \\
         & hard & $3$ to $4$ & $1$ to $5$ & $13$ to $18$ / n/a & n/a & $30^\circ$, $45^\circ$ & $2$/$44$/$0$/$4$ \\
    \bottomrule
  \end{tabular*}
  \caption{Composition of the Sheep task. Each level has $150$
  questions, and each Hard tier contributes $50$. Ranges are taken
  over the scenes used at each level.}
  \label{tab:sheep_tiers}
\end{table*}

\paragraph{Output format and scoring.}
Q1 and Q4 return JSON integers. Q2 returns a list of sheep IDs separated
by commas or \texttt{"NONE"}. Q3 returns a region label. Integer
answers and region labels are scored by exact match after parsing. Q2
answers are compared as sets of IDs, so their order does not affect
correctness.
The complete task prompt and qualitative examples appear in
Figures~\ref{fig:app_task_sheep_examples_count_inside_and_escape_probability}
to~\ref{fig:app_task_sheep_examples_fence_repair}
(Appendix~\ref{app:task_prompt_cards}).

\subsubsection{Chat Noir \textnormal{(Enclosure, Planning)}}
\label{app:task_chat_noir}

\paragraph{Targeted ability.}
Chat Noir evaluates adversarial enclosure planning. The model must
progressively block a hex-grid board so that the moving cat loses every
path to the boundary before it can escape.

\paragraph{Task formulation.}
At each turn, the model observes the current board and selects one open
non-cat cell to block. The cat then moves according to the level's
policy. The episode succeeds if the cat has no remaining path from its
current cell to any boundary cell, and fails if the cat reaches the
boundary or the action budget is exhausted. The budget equals the
number of open non-cat cells at the start of the episode.

\paragraph{Scene generation.}
The generator samples the board radius and the initial blockers from a
deterministic seed and places the cat on the center cell, so its
initial distance to the boundary equals the board radius. Initial
blockers are drawn uniformly from non-cat cells. A candidate layout is
rejected if the cat is already trapped, has no legal move, or lacks a
path to the boundary. A search filter then verifies, against the
deterministic policy of the cat used at that level, that the player has
a winning opening. The two levels apply different versions of this
filter, described below, so that every episode is solvable but still
requires multi-step enclosure planning.

\paragraph{Difficulty levels.}
The two levels differ mainly in the cat policy and the opening filter
(Table~\ref{tab:chat_noir_tiers}). Chat Noir contains $300$
episodes, $150$ per level, and every episode starts from a distinct
layout. The Easy level uses a connectivity-aware greedy cat that
avoids immediate one-move traps when possible and otherwise prefers a
shorter boundary distance while keeping more escape routes open.
Boards have radius $3$ ($62$ episodes, $12$ to $14$ initial blockers)
or radius $4$ ($88$ episodes, $16$ to $19$ initial blockers), which
gives budgets of $22$ to $24$ and $41$ to $44$ actions. Each Easy layout
admits at least five winning first blocks within a bounded search of
depth six or seven that expands at most six candidate blocks per turn,
and at least three of these blocks are adjacent to the cat. The Hard level uses an expert cat
that, before each move, checks every legal move for a forced capture
within three complete player turns. It first takes an immediate escape
when one exists, then avoids moves that allow capture within one, two,
or three turns, and finally prefers shorter escape paths and more
shortest escape routes. All Hard boards have radius $4$ with $10$ to
$13$ initial blockers, which gives budgets of $47$ to $50$ actions.
Each Hard layout must have at least one certified winning first block
and provably no winning first block adjacent to the cat. The
certification searches against the expert policy over the full action
budget, under a limit of $200{,}000$ search nodes and $30$ seconds, and
an unfinished search never counts as a proof. The winning lines found
by this search require $6$ to $24$ player blocks. Blocking a cell next
to the cat therefore never starts a win, and the model must open with a
block away from the cat whose value only appears several moves later.

\begin{table*}[!htbp]
  \centering
  \scriptsize
  \begin{tabular*}{\textwidth}{@{\extracolsep{\fill}}llcccc@{}}
    \toprule
    Level & Cat policy & Radius & Initial blockers & Budget & Episodes \\
    \midrule
    Easy & connectivity-aware greedy & $3$ / $4$ & $12$ to $14$ / $16$ to $19$ & $22$ to $24$ / $41$ to $44$ & $62$ / $88$ \\
    Hard & expert three-turn lookahead & $4$ & $10$ to $13$ & $47$ to $50$ & $150$ \\
    \bottomrule
  \end{tabular*}
  \caption{Difficulty levels for the Chat Noir planning task. Easy
  entries separated by a slash refer to radius $3$ and radius $4$
  boards. Easy layouts admit at least five winning first blocks, at
  least three of them adjacent to the cat, whereas Hard layouts admit
  no adjacent winning first block.}
  \label{tab:chat_noir_tiers}
\end{table*}

\paragraph{Output format and scoring.}
The canonical action is a blocked cell index.
An example response is \texttt{\{"answer":12\}}. We score the resulting interaction, and an
episode is correct only if the model traps the cat before escape and
within the maximum number of allowed actions.
The complete task prompt and qualitative examples appear in
Figure~\ref{fig:app_task_chat_noir_examples} (Appendix~\ref{app:task_prompt_cards}).

\subsection{Knots Tasks}
\label{app:tasks_knots}

\subsubsection{Knots \textnormal{(Knots, Reasoning)}}
\label{app:task_knot_detection}

\paragraph{Targeted ability.}
The Knots task evaluates recognition of global knot and link structure
from a single rendered image. It tests whether a model can distinguish
knots from visually deceptive unknots and open ropes, count continuous
rope components, recover link relations among multiple components, and
reason counterfactually about removing one component.

\paragraph{Task formulation.}
Each question presents a single rendered image of one or more ropes.
\texttt{T01} assigns one of six categories to the whole scene, namely
A (simple closed ring), B (knot), C (open rope with no knot), D (link),
E (unlinked multiple ropes), and F (other or mixed). \texttt{T02}
counts the number of distinct ropes. \texttt{T03} classifies a closed
ring scene as A (not linked), B (chain, including a two-ring paired
link), C (all interlocked), or D (mixed). \texttt{T04} asks which
remaining rings become free after a specified colored ring is cut and
removed. This removal question is generated only for supported
configurations with three to six distinctly colored rings and metadata
that records the link graph. Finally, \texttt{T05} counts how many
ropes are linked to at least one other rope.

\paragraph{Scene generation.}
Both levels draw their structures from a catalog of $24$ scene types.
Its $16$ types with one rope cover closed and open unknots, visually
deceptive unknots, trefoil and figure eight knots, torus knots with
more crossings, and loose or occluded variants. The other $8$ types
cover Hopf links, unlinked rings, chains, Borromean rings, multiple
link groups, and mixed scenes containing both linked and free rings.
The two levels present these structures in different ways. Easy
scenes are rendered as 3D ropes seen from an oblique camera, with rope
color, curve slackness, and deformation controlled by a deterministic
seed. Loose knots, near-miss unlinks, and mixed connectivity provide
explicit visual traps. Hard scenes start from a planar diagram of the
same known knot or link, built with the Spherogram library, and expand
it through recorded Reidemeister II and III moves until the projection
reaches a target number of crossings. These moves preserve the
isotopy class \citep{rolfsen1976knots}, so the ground truth is that of
the original structure, while the drawing no longer resembles its
canonical silhouette. Every move is checked for legality, and the
rendered centerlines are checked against the complete signed crossing
order of the expanded diagram. Each Hard image is a $2560{\times}2560$
rendering of rounded tubes with explicit over and under bridges, a
rope diameter of at least $11$ pixels, and crossing angles of at least
$31.9^\circ$. Open ropes end at visible fixed posts. All ropes in a
Hard image share one color except in \texttt{T04}, where each ring has
its own color, and six Hard scenes used for \texttt{T02} and
\texttt{T05} adopt a more strongly wrapped projection.

\paragraph{Difficulty levels.}
Knots contains $300$ questions, $150$ per level
(Table~\ref{tab:knot_tiers}). The Easy level comes from the hard tier of
the previous release, its densest scene setting, which favors dense or mixed
link structures and component counts above six. It contains $47$
\texttt{T01}, $49$ \texttt{T02}, $3$ \texttt{T03}, $6$ \texttt{T04},
and $45$ \texttt{T05} questions, posed on $82$ distinct images.
Its \texttt{T01} scenes are either single knotted ropes, such as the
trefoil, a loose cinquefoil, and torus knots up to $T(3,5)$ and
$T(2,9)$, or multi-component scenes with $2$ to $10$ ropes, and their
answers are $20$ B, $21$ D, $4$ E, and $2$ F. The \texttt{T02} and
\texttt{T05} scenes contain $8$ to $10$ ropes, with $8$ to $10$ ropes to
count in \texttt{T02} and $7$ to $10$ linked ropes in \texttt{T05}. The
\texttt{T04} scenes contain $4$ to $6$ rings, and three or four rings
become free after each removal. Every Hard question uses its own image.
The Hard level spans three settings of projected crossings, $36$ to $50$, $66$ to $90$, and $112$ to $160$,
with $50$ questions each and $10$ per task in every setting. Its scenes
contain $1$ to $8$ ropes, and all $24$ scene types appear. The answers
are balanced by construction. Each setting contains $2$ A, $4$ B, and
one each of C, D, E, and F among its \texttt{T01} questions.
\texttt{T02} answers range from $3$ to $8$. \texttt{T03} has $8$, $8$,
$7$, and $7$ answers for A to D. \texttt{T04} has $10$ questions each
in which no ring, one ring, or several rings become free.
\texttt{T05} always shows six ropes, and its answers $0$, $2$, $3$,
$4$, $5$, and $6$ each occur five times. Borromean rings are excluded
from \texttt{T05}, which asks about pairwise linkage.

\begin{table*}[!htbp]
  \centering
  \scriptsize
  \begin{tabular*}{\textwidth}{@{\extracolsep{\fill}}llcccc@{}}
    \toprule
    Level & Setting & Presentation & Ropes & Projected crossings & T01/T02/T03/T04/T05 \\
    \midrule
    Easy & single & oblique 3D view & $1$ to $10$ & not controlled & $47$/$49$/$3$/$6$/$45$ \\
    \midrule
    Hard & a & expanded planar diagram & $1$ to $8$ & $36$ to $50$ & $10$/$10$/$10$/$10$/$10$ \\
         & b & expanded planar diagram & $1$ to $8$ & $66$ to $90$ & $10$/$10$/$10$/$10$/$10$ \\
         & c & expanded planar diagram & $1$ to $8$ & $112$ to $160$ & $10$/$10$/$10$/$10$/$10$ \\
    \bottomrule
  \end{tabular*}
  \caption{Composition of the Knots task. Each level has $150$
  questions, and each Hard setting contributes $50$. Projected
  crossings count the crossings of the expanded diagram that is
  rendered.}
  \label{tab:knot_tiers}
\end{table*}

\paragraph{Output format and scoring.}
\texttt{T01} returns one option letter from A through F.
\texttt{T03} returns one option letter from A through D.
\texttt{T02} and \texttt{T05} return zero or positive integers.
\texttt{T04} returns a JSON list of the colors
of all rings that become free, or \texttt{["none"]} when none do.
Primary correctness requires an exact match after parsing. The
\texttt{T04} color list is compared as a set, so list order does not
affect correctness.
The complete task prompt and qualitative examples appear in
Figures~\ref{fig:app_task_knots_examples_structure_classification}
to~\ref{fig:app_task_knots_examples_link_property}
(Appendix~\ref{app:task_prompt_cards}).

\subsubsection{Untangle \textnormal{(Knots, Planning)}}
\label{app:task_knots_untangle}

\paragraph{Targeted ability.}
The Untangle task evaluates sequential spatial planning for eliminating
crossings among simulated ropes on a pegboard. The model must track rope
endpoints, anticipate how relocating a lifted endpoint changes the
projected rope layout, and select a sequence of legal moves that removes
all crossings between rope paths in the projection from above.

\paragraph{Task formulation.}
The environment presents an overhead view of a $G{\times}G$ grid of
holes through which $R$ colored ropes are threaded, with each rope's
two endpoints occupying distinct holes. Following the renderer's
coordinate convention, row indices are shown along the top edge and
increase from left to right, while column indices are shown along the
left edge and increase from top to bottom. At every turn, the model
observes the rendered grid and selects an occupied endpoint and an
unoccupied target hole. The environment lifts the endpoint above the
other ropes, moves it to the target hole, lowers it, and advances the
rope physics for $60$ frames. After the ropes settle, the crossing
count is recomputed as the number of rope pairs whose simulated
centerline paths intersect in the projection from above. An episode
succeeds when this count reaches zero before the action budget is
exhausted. Invalid or illegal actions consume one step without
changing the state.

\paragraph{Scene generation.}
For each generator setting, the generator starts from a bank of
noncrossing endpoint templates formed by parallel rows or columns, then
performs multiple seeded scramble searches using legal endpoint
relocations. Immediate reversals are excluded. Intermediate moves are
weighted by the target crossing range, the number and connectivity of
involved ropes, crossing separation, and visual readability. The
generator samples a candidate that completes the setting's scramble
depth and passes its final structural constraints. In the six-rope
setting, the generator also seeds a compact central tangle before
physics relaxation, and candidates must contain a crossing component
of three or four ropes. At least three ropes must each cross more than
one other rope, and the crossing graph must contain at least one
cycle. Evaluation also computes auxiliary metadata with a symbolic
endpoint solver. This solver represents each rope by the straight
segment joining its two holes. It runs breadth first search to remove
all segment overlaps defined using the rope width, subject to a time
limit of $10$ seconds and an expansion limit of $200{,}000$ states. The
resulting plan length serves as a proxy for the minimum plan length and
drives the local oracle policy, and it is not used for scoring.

\paragraph{Difficulty levels.}
Table~\ref{tab:untangle_tiers} lists the generator settings of each
level. Untangle contains
$300$ episodes, $150$ per level, and all of them share a budget of $15$
actions. The Easy level uses the hard tier of the previous release
throughout, with six ropes, $6{\times}6$ grids, scrambles of $7$ to $9$ moves, and $5$
to $7$ initial crossings. The Hard level takes $50$ episodes from each of the
three tiers of the previous release, namely four ropes on a $5{\times}5$ grid,
five ropes on a $6{\times}6$ grid, and six ropes on a $6{\times}6$ grid. The
table reports the initial crossing counts under the
same physics metric that decides success.

\begin{table*}[!htbp]
  \centering
  \scriptsize
  \begin{tabular*}{\textwidth}{@{\extracolsep{\fill}}lcccccc@{}}
    \toprule
    Level & Grid & Ropes & Scramble & Crossings & Budget & Episodes \\
    \midrule
    Easy   & $6{\times}6$ & $6$ & $7$ to $9$ & $5$ to $7$ & $15$ & $150$ \\
    \midrule
    Hard   & $5{\times}5$ & $4$ & $2$ to $3$ & $2$ to $3$ & $15$ & $50$ \\
           & $6{\times}6$ & $5$ & $5$ to $6$ & $4$ to $6$ & $15$ & $50$ \\
           & $6{\times}6$ & $6$ & $7$ to $9$ & $5$ to $7$ & $15$ & $50$ \\
    \bottomrule
  \end{tabular*}
  \caption{Difficulty levels for the Untangle task. The Hard level
  contains one row per tier of the previous release.}
  \label{tab:untangle_tiers}
\end{table*}

\paragraph{Output format and scoring.}
Each turn returns a nested JSON action with four integer fields named
\texttt{src\_row}, \texttt{src\_col}, \texttt{tgt\_row}, and
\texttt{tgt\_col}. These fields identify the occupied source hole and
the empty target hole. An episode is correct only when the executed
actions reduce the physics crossing count defined above to zero within
the action budget.
The complete task prompt and qualitative examples appear in
Figure~\ref{fig:app_task_knots_untangle_examples} (Appendix~\ref{app:task_prompt_cards}).

\subsection{Dataset Statistics}
\label{app:dataset_stats}

Table~\ref{tab:app_dataset_stats} breaks the benchmark down by
topological property, cognitive level, and difficulty level.

\begin{table*}[!htbp]
\centering
\scriptsize
\begin{tabular*}{\textwidth}{@{\extracolsep{\fill}}lccccc@{}}
\toprule
Property & Reason.\ \#Q & Plan.\ inst. & Easy & Hard & Total \\
\midrule
Continuity  & $600$ & $300$ & $450$ & $450$ & $900$ \\
Separation  & $300$ & $300$ & $300$ & $300$ & $600$ \\
Order       & $600$ & $300$ & $450$ & $450$ & $900$ \\
Enclosure   & $600$ & $300$ & $450$ & $450$ & $900$ \\
Knots       & $300$ & $300$ & $300$ & $300$ & $600$ \\
\midrule
\textbf{Total} & $\mathbf{2{,}400}$ & $\mathbf{1{,}500}$ & $\mathbf{1{,}950}$ & $\mathbf{1{,}950}$ & $\mathbf{3{,}900}$ \\
\bottomrule
\end{tabular*}
\caption{\benchmarkname{} dataset composition across the five topological
properties, the Reasoning and Planning cognitive levels,
and the Easy and Hard difficulty levels, with 150 instances per task at each level.}
\label{tab:app_dataset_stats}
\end{table*}

\subsection{Evaluation Design}
\label{app:eval_design}

\paragraph{Reasoning tasks.}
Reasoning tasks share a single response contract. The model receives
one rendered scene or a fixed bundle of views, emits one JSON
response, and is scored without environment interaction. Across the eight
reasoning tasks, the answers fall into four families. \emph{Set} answers occur
in the \textsc{2D Maze} reachability and bar-removal questions, \textsc{3D Maze}
connectivity and door-opening, \textsc{Origami Point}, \textsc{Sheep} escape IDs,
and \textsc{Knots}~T04 free ring colors. \emph{Sequence} answers occur
in the \textsc{Bead} sequence description task. \emph{Scalar
integer} answers occur in \textsc{Hole}, \textsc{Sheep}
Q1/Q4, and \textsc{Knots}~T02/T05. \emph{Categorical label} answers
occur in \textsc{Assembly}, \textsc{Knots}~T01/T03, the \textsc{Bead}
sequence relationship task, and \textsc{Sheep}~Q3.
All four families are parsed by the same layered extractor in our
released evaluation code. It first attempts a strict
\texttt{\{"answer":\,\ldots\}} JSON extraction. If that fails, it tries
final answer phrase matching, scans the text for legal values, and
examines the response tail before declaring a parse failure.
Set answers are scored by frozenset equality, sequences by ordered
list equality, scalars by exact integer match, and labels by exact
match against the legal vocabulary for each instance.

\paragraph{Planning tasks.}
Planning tasks expose an action interface and are scored on the
executed trajectory rather than on the textual plan. At every turn
the model receives the current rendered state and emits one JSON
action. The environment validates legality, applies the action or
charges the budget for an illegal one, and renders the next state.
When a model returns a thought summary before its answer, only the text
after the summary is parsed, so a move mentioned while thinking is never
executed in place of the final answer.
Each task ships with a per-instance step budget. \textsc{Pipe} allows
$1.3$ times the reference number of rotations, rounded up. \textsc{Swap}
allows $1.2$ times the shortest solution, rounded up, at the Easy level and a fixed
$75$ actions at the Hard level. \textsc{One Stroke} allows $1.2$ times the
construction path length, rounded up (Table~\ref{tab:one_stroke_tiers}).
\textsc{Untangle} allows a fixed $15$ actions, and \textsc{Chat Noir}
allows one action per open cell other than the cat's cell. There is no retry, so budget exhaustion or
violation of the success predicate ends the episode and counts as a
failure. Each planning task additionally ships with two reference
policies, an \textsc{oracle} solver and a \textsc{random} policy that
samples legal actions uniformly under the same budget. The reference is exact where an environment
exposes a precomputed optimal plan. For \textsc{Untangle}, the reference is
the symbolic endpoint solver described above.

\paragraph{Metrics.}
The primary per-task metric is \emph{accuracy}, the per-instance $0/1$
correctness averaged over the evaluation split. For planning tasks, this
metric is the \emph{episode success rate}, the fraction of episodes whose
executed trajectory satisfies the success predicate within the step budget. For diagnosis, the
pipeline also computes a \emph{JSON parse rate} (fraction recovered by
the strict-JSON layer) and an \emph{answer-format hit rate} (fraction
whose strict JSON additionally passes the per-instance shape and
legal-value validator), isolating format-following failures from
reasoning failures, and, for planning tasks, a \emph{mean
over-optimality}, the average gap between the executed trajectory
length and the task's reference plan length on solved episodes. The
tables report accuracy. The \emph{per cognitive level mean} averages the
reasoning tasks and the planning tasks separately, and a property average is
the mean over that property's tasks.

\section{Experiments and Analysis}
\label{app:experiments}

\subsection{Models and Setup}
\label{app:models_prompts}

\paragraph{Models.}
Table~\ref{tab:app_model_list} lists every foundation model and generative
backend assessed in this study, together with its access route, evaluated
identifier, and the pipeline each one enters.

\paragraph{Evaluation protocol.}
For reasoning tasks, each model answers a single-turn visual question.
Decoding uses \texttt{temperature=0} and \texttt{do\_sample=false}, with at most
$2048$ newly generated tokens. Each prompt asks for exactly one JSON object,
and the evaluator applies the shared fallback parser only after the model has
finished. Planning tasks use the same deterministic decoding settings and run
through the Hydra+Playwright environment runner. Each turn includes the current
rendered observation, the task rules, and the legal answer schema. In the main
evaluation, no model receives symbolic state graphs, oracle connectivity, or
hidden generator metadata. Failed API calls are retried according to the per-provider
configuration, after which the episode or sample is marked invalid.

All static images are supplied at the renderer's native resolution and
preserved as PNG inputs. Planning screenshots are captured from the browser
frontend. Environments that need a fixed viewport use the configured
$1440{\times}960$ or $1440{\times}1080$ viewport before scene-only cropping.

\begin{table*}[!htbp]
\centering
\scriptsize
\setlength{\tabcolsep}{4pt}
\begin{tabularx}{\textwidth}{@{}>{\raggedright\arraybackslash}p{0.10\textwidth}>{\raggedright\arraybackslash}p{0.16\textwidth}>{\raggedright\arraybackslash}p{0.15\textwidth}X>{\raggedright\arraybackslash}p{0.18\textwidth}@{}}
\toprule
Organization & Model Name & Access & Full Name & Evaluation Pipeline \\
\midrule
\multicolumn{5}{l}{\textit{Proprietary Models}} \\
OpenAI & GPT-6 Astra~\cite{openai2026gpt6astra} & OpenAI API & \nolinkurl{gpt-6-astra} & Reasoning + planning \\
OpenAI & GPT-5.6-Sol~\cite{openai2026gpt56} & OpenAI API & \nolinkurl{gpt-5.6-sol} & Reasoning + planning \\
OpenAI & GPT-5.5~\cite{openai2026gpt55} & OpenAI API & \nolinkurl{gpt-5.5} & Reasoning + planning \\
OpenAI & GPT-5.4-mini~\cite{openai2026gpt54mini} & OpenAI API & \nolinkurl{gpt-5.4-mini-2026-03-17} & Reasoning + planning, probing planner \\
OpenAI & GPT-5.6-Luna~\cite{openai2026gpt56} & OpenAI API & \nolinkurl{gpt-5.6-luna} & Probing planner \\
OpenAI & GPT-5.4-mini~\cite{openai2026gpt54mini} & OpenAI API & \nolinkurl{gpt-5.4-mini} & Generation error audit \\
Google & Gemini-3.1-Pro~\cite{google2026gemini31pro} & Google AI Studio & \nolinkurl{gemini-3.1-pro-preview} & Reasoning + planning \\
Google & Gemini-3.1-Flash-Lite~\cite{google2026gemini31flashlite} & Google AI Studio & \nolinkurl{gemini-3.1-flash-lite-preview} & Reasoning + planning \\
Anthropic & Claude-Haiku-5.5~\cite{anthropic2026claudehaiku55} & Anthropic API & \nolinkurl{claude-haiku-5-5} & Reasoning + planning \\
\midrule
\multicolumn{5}{l}{\textit{Hosted Open-Weight Models}} \\
InternLM & InternVL3.5-241B~\cite{wang2025internvl35} & Intern-AI API & \nolinkurl{OpenGVLab/InternVL3_5-241B-A28B-Instruct} & Reasoning + planning, probing planner \\
NVIDIA & Nemotron-Nano-12B-VL-v2~\cite{nvidia2025nemotronnanov2vl} & DigitalOcean & \nolinkurl{nvidia/NVIDIA-Nemotron-Nano-12B-v2-VL-BF16} & Reasoning + planning \\
Google & Gemma-4-31B-IT~\cite{google2026gemma4} & Google AI Studio & \nolinkurl{google/gemma-4-31B-it} & Reasoning + planning \\
Alibaba & Qwen3.5-397B-A17B~\cite{qwen2026qwen35} & Alibaba Cloud & \nolinkurl{Qwen/Qwen3.5-397B-A17B} & Reasoning + planning \\
Meta & Llama-4-Maverick-17B-128E~\cite{meta2025llama4} & DigitalOcean & \nolinkurl{meta-llama/Llama-4-Maverick-17B-128E-Instruct} & Reasoning + planning \\
Mistral AI & Ministral-3-14B-Instruct-2512~\cite{mistral2025ministral3} & Mistral API & \nolinkurl{mistralai/Ministral-3-14B-Instruct-2512} & Reasoning + planning \\
\midrule
\multicolumn{5}{l}{\textit{Image Generative Models}} \\
OpenAI & GPT-Image-2~\cite{openai2026images2} & OpenAI API & \nolinkurl{gpt-image-2} & Image-edited imagined rollouts  \\
\midrule
\multicolumn{5}{l}{\textit{Video Generative Models}} \\
Lightricks & LTX-2.3~\cite{lightricks2026ltx23} & self-hosted & \nolinkurl{Lightricks/LTX-2.3} & CV-audited image-to-video rollouts \\
Lightricks & LTX-2.5~\cite{lightricks2026ltx25} & self-hosted & \nolinkurl{Lightricks/LTX-2.5} & CV-audited image-to-video rollouts \\
Wan-AI & Wan2.2-I2V-A14B~\cite{wan2025wanopenadvancedlargescale} & self-hosted & \nolinkurl{Wan-AI/Wan2.2-I2V-A14B} & Probing and CV-audited rollouts \\
ByteDance & Seedance-2.0-Mini~\cite{bytedance2026seedance2mini} & API & \nolinkurl{seedance-2.0-mini} & Probing video rollouts \\
Google & Veo-3.1-Lite~\cite{google2026veo31lite} & API & \nolinkurl{veo-3.1-lite-generate-preview} & Probing and CV-audited rollouts \\
MiniMax & MiniMax-H3~\cite{minimax2026h3} & self-hosted & \nolinkurl{MiniMaxAI/MiniMax-H3} & CV-audited image-to-video rollouts \\
ByteDance & Seedance-2.5~\cite{bytedance2026seedance25} & API & \nolinkurl{doubao-seedance-2-5-260628} & CV-audited image-to-video rollouts \\
\bottomrule
\end{tabularx}
\caption{Details of foundation models and generative backends assessed in this
study. Access gives the API or hosting used for each model, and Full Name gives
the evaluated model identifier.}
\label{tab:app_model_list}
\end{table*}

\paragraph{Prompts.}
We use one prompt template per task, kept fixed across all models,
with no task-specific prompt tuning after seeing model outputs.
Representative property-level prompt templates appear at the end of the paper
in Appendix~\ref{app:prompt_templates},
Figures~\ref{fig:app_prompt_continuity} to~\ref{fig:app_prompt_knots}.

\subsection{Training}
\label{app:training}

\paragraph{Training tasks and data splits.}
All training experiments start from
\nolinkurl{Qwen/Qwen3-VL-2B-Instruct}. The reasoning suite contains 2D Maze,
Assembly, Bead, Sheep, and Knots, while the planning suite contains Pipe, One
Stroke, Swap, and Untangle. For each single-task experiment, we split the Easy level data into training,
validation, and test sets in proportions of $80\%$, $10\%$, and $10\%$, using
the fixed seed \texttt{20260705}. Reasoning examples are stratified by question
type. The held-out experiments train on all Easy level examples from four
reasoning tasks and assign one fifth of the fifth task to validation and the
remaining four fifths to test.
For Bead, we use a simplified variant with fewer beads than in the standard
benchmark.

\paragraph{Supervised fine-tuning.}
SFT uses answer-only supervision. Reasoning targets contain only the JSON answer
required by the benchmark, with no chain-of-thought or explanatory text. For
planning, the target is one complete JSON action sequence from the initial
observation. One Stroke uses exact shortest-path search, Pipe uses the minimum
required clockwise rotations, Swap uses the environment's shortest action
sequence, and Untangle uses exact breadth-first search under the same
width-aware overlap criterion as the environment. Every planning target is
replayed and must solve the corresponding environment before it enters the
training set.

We train completion-only LoRA adapters for one epoch with a global batch size
of 5. The prompt and assistant-prefix tokens are masked from the loss. LoRA is
applied to all linear modules with rank 32, alpha 64, and dropout 0.05. We use
AdamW with learning rate $10^{-4}$, cosine decay, a 0.03 warmup ratio, weight
decay 0.01, gradient clipping at 1.0, BF16, and gradient checkpointing. The
maximum completion length is 128 tokens for reasoning and 2048 tokens for
planning.

\paragraph{Reinforcement learning.}
RL uses GRPO through the VAGEN/VERL training stack~\cite{wang2025vagen}. Each
update contains five prompts. We sample eight completions per reasoning prompt
and 32 per planning prompt, giving 40 and 160 sampled completions per update,
respectively. Sampled completions use temperature 0.8 and top-$p$ 0.95. We train for one
epoch with learning rate $10^{-6}$ and KL coefficient 0.001. A correct answer
or successful action sequence receives reward 1.0, a parseable but incorrect
answer receives 0.0, and an invalid output receives $-0.1$. Planning reward is
computed by executing the predicted sequence in the actual environment. No
credit is assigned for intermediate actions that do not reach the success
state. The RL runs do not enable LoRA and update the model from either the base
checkpoint or the merged SFT checkpoint.

\paragraph{SFT followed by RL and leave-one-task-out RL.}
For SFT + RL, we merge the task-specific LoRA adapter into the base checkpoint
and initialize GRPO from the merged model. The subsequent data split, reward,
sampling, and optimization settings are identical to the corresponding
RL-only run. For leave-one-task-out RL, we train five separate policies. Each policy sees
four reasoning environments during training and is evaluated on the excluded
environment, so every value in the Leave-one-task-out RL row of
Table~\ref{tab:rl_sft_results} comes from a different held-out policy.

\paragraph{Evaluation protocol.}
Decoding is deterministic with one greedy completion per example. Reasoning
predictions use the benchmark's answer parser and task-specific scorer. Planning
predictions are parsed as complete action sequences and executed from a fresh
environment reset. A prediction is correct only if execution reaches the
task-defined success state. The single-task comparisons keep the input format,
scorer, and test examples fixed across Base, SFT, RL, and SFT + RL, using the
same test set. Leave-one-task-out RL uses the same input format and scorer and
is evaluated on the test portion of the excluded task.

\begin{table}[H]
\centering
\scriptsize
\setlength{\tabcolsep}{3pt}
\renewcommand{\arraystretch}{1.12}
\begin{tabular*}{\textwidth}{@{\extracolsep{\fill}}lccc@{}}
\toprule
\textbf{Training}
& \PTask{3D Maze} & \PTask{Origami Point} & \PTask{Hole} \\
\midrule
Base & 18.37 & 19.60 & 0.10 \\
Held-out 2D Maze & -- & 0.90 & 4.80 \\
Held-out Sheep & 18.78 & 0.40 & -- \\
Held-out Knots & 18.57 & 0.90 & 8.41 \\
Held-out Bead & 20.28 & -- & 4.20 \\
Held-out Assembly & 18.78 & 6.50 & 0.00 \\
\bottomrule
\end{tabular*}
\caption{Transfer performance (\%) for the Base policy and RL policies trained
while holding out one reasoning task.}
\label{tab:rl_heldout_transfer_results}
\end{table}

\paragraph{Held-out transfer.}
Table~\ref{tab:rl_heldout_transfer_results} evaluates the five
leave-one-task-out RL
policies on three additional reasoning tasks. Performance on 3D Maze remains
close to the base policy, ranging from 18.57\% to 20.28\% across the applicable
held-out policies. All policies fall below the 19.60\% base result on Origami
Point, while Hole ranges from 0.00\% to 8.41\% against a base of 0.10\% and
remains low in absolute terms. Together with the held-out results in the main table, these
results show that cross-task transfer is uneven and depends on the target
relation rather than following automatically from multi-environment training.

\subsection{Illegal Actions}
\label{app:illegal_actions}

\paragraph{Legality rules.}
An action is illegal when it falls outside the environment-specific legal-action set. The precise rejection conditions differ by environment and are summarized in Table~\ref{tab:app_illegal_actions}. These checks are applied by the environment after parsing the model output and before any state transition is executed.

\begin{table*}[!htbp]
\centering
\scriptsize
\renewcommand{\arraystretch}{1.12}
\begin{tabular*}{\textwidth}{
  @{\extracolsep{\fill}}
  >{\raggedright\arraybackslash}p{0.22\textwidth}
  >{\raggedright\arraybackslash}p{0.70\textwidth}
  @{}
}
\toprule
\textbf{Environment} & \textbf{Illegal-action condition} \\
\midrule
\texttt{knots\_untangle} &
The source is not a movable occupied endpoint, the target is not an empty
in-bounds hole, or the pair of source and target is not a permitted endpoint move. \\
\texttt{continuity\_pipe} &
The selected coordinate is out of bounds or points to an empty cell. \\
\texttt{separation\_one\_stroke} &
The direction is invalid, leaves the board, reuses a prohibited edge, or
creates a closed loop. Immediate backtracking over the previous edge remains
legal. \\
\texttt{order\_swap\_2d\_puzzle} &
The selected coordinate is out of bounds or does not identify a movable
non-empty cell. At the Hard level, selecting a cell that shares neither
a row nor a column with the blank is also illegal. \\
\texttt{enclosure\_chat\_noir} &
The selected index is out of bounds, is already blocked, or identifies the
cat's current cell. \\
\bottomrule
\end{tabular*}
\caption{Environment-specific conditions under which a parsed action is
rejected as illegal.}
\label{tab:app_illegal_actions}
\end{table*}

\paragraph{Invalid responses and API failures.}
Malformed model output is recorded as an invalid response and converted to an invalid-action sentinel before environment execution. API failures are tracked separately and do not reach the environment, so they are not counted as environment-level illegal actions.

\paragraph{Effect of rejection.}
An illegal action leaves the environment state unchanged and consumes one interaction step. The harness records both the rejection flag and its associated reason, but under the default evaluation protocol the model receives only the resulting unchanged observation rather than an explicit rejection message. The harness can also list the legal actions in the prompt through the option \texttt{run.include\_legal\_moves\_in\_prompt}, which is disabled by default.

\subsection{Human Evaluation}
\label{app:human}

\subsubsection{Annotation Interface and Human Performance Evaluation}
\label{app:human_interface}

\begin{figure}[H]
\centering
\begin{minipage}[t]{0.49\linewidth}
\centering
\includegraphics[width=\linewidth]{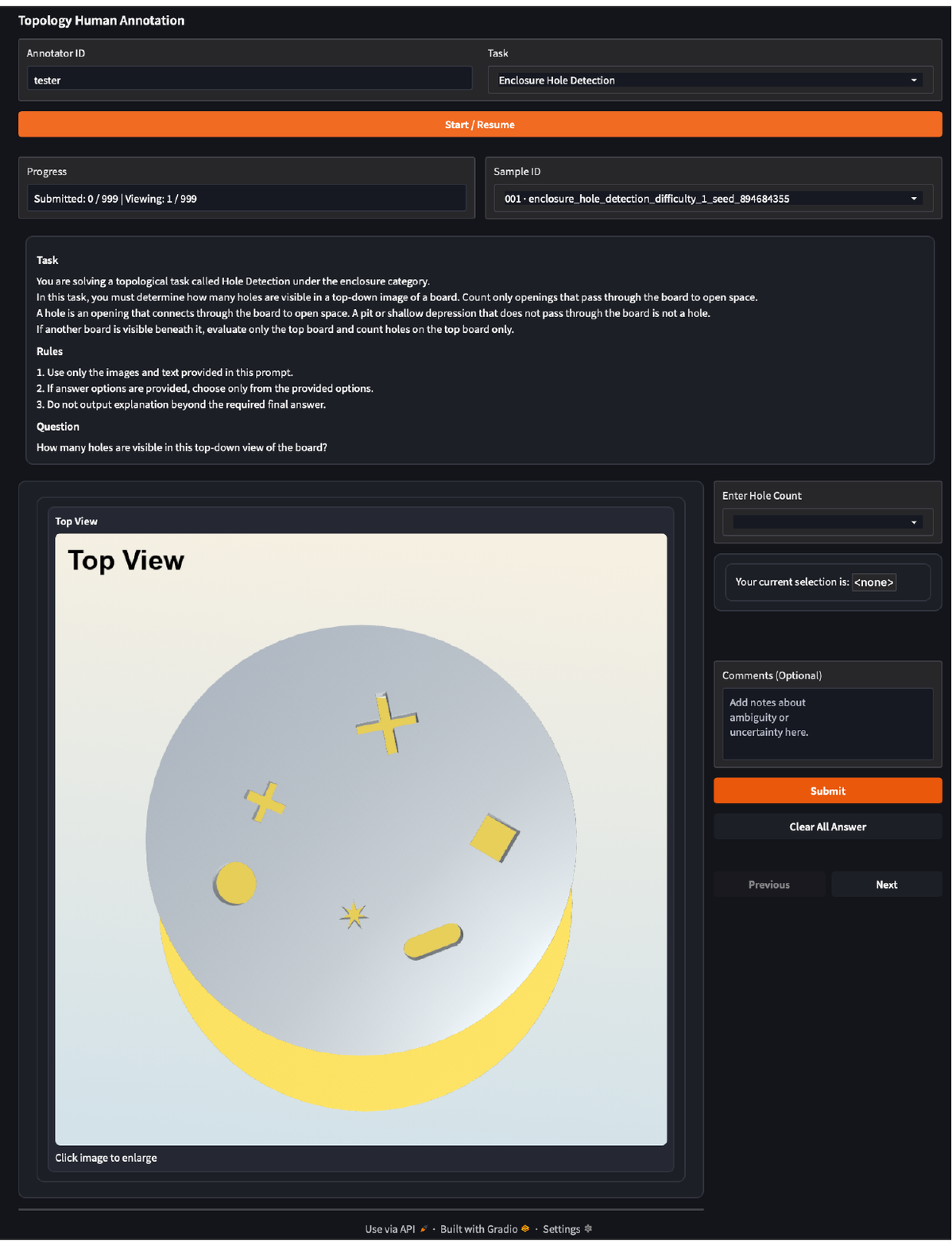}
\caption{Human annotation interface for reasoning tasks. Annotators are shown the same visual inputs and task instructions used in model evaluation, together with a task-specific answer widget and an optional comment field for ambiguous cases.}
\label{fig:human_annotator_interface_reasoning}
\end{minipage}\hfill
\begin{minipage}[t]{0.49\linewidth}
\centering
\includegraphics[width=\linewidth]{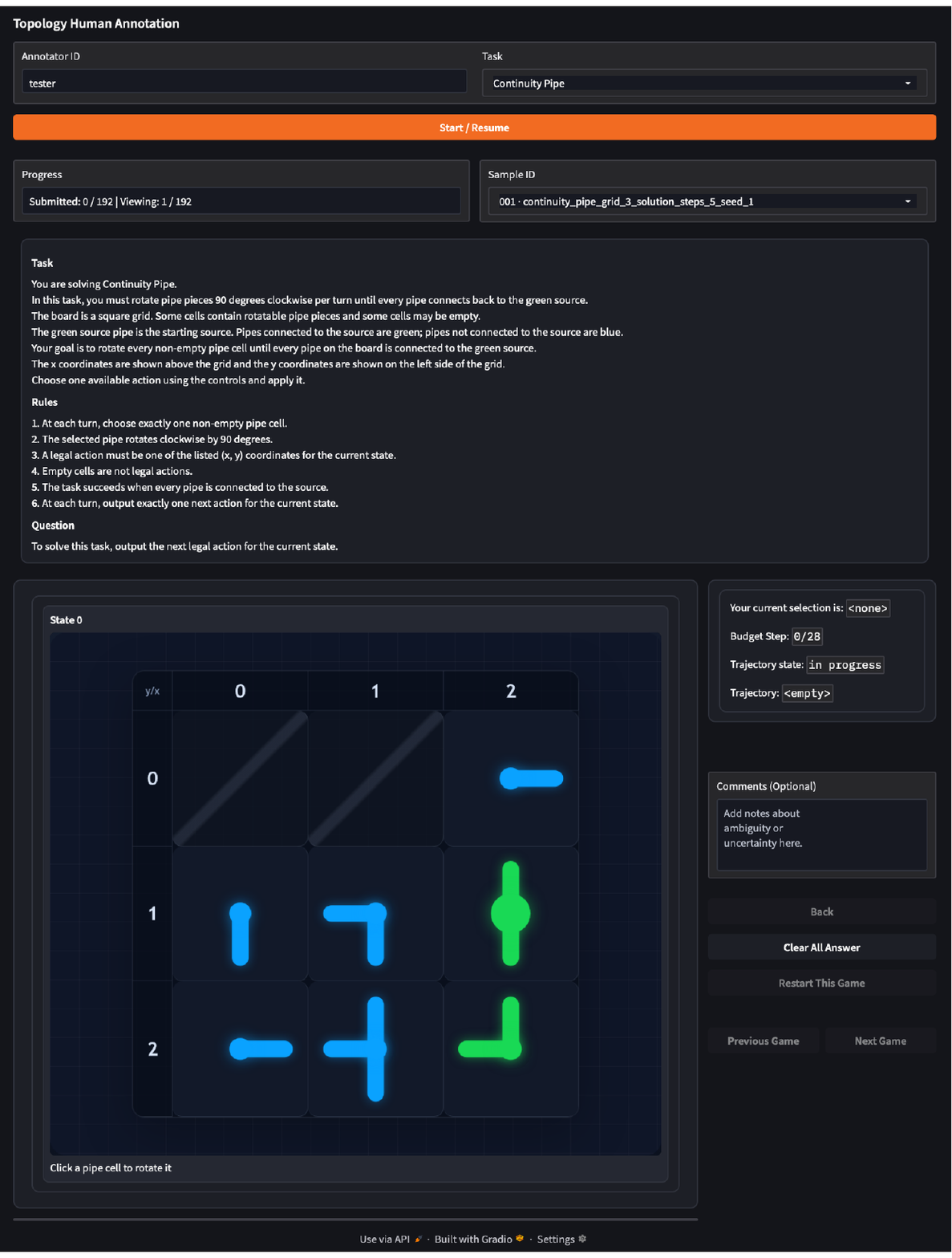}
\caption{Human annotation interface for planning tasks. Annotators interact with the same browser-based environment used by the model runner, and success is determined from the environment terminal state.}
\label{fig:human_annotator_interface_planning}
\end{minipage}
\end{figure}

Human annotators use the same visual inputs and answer schemas as model
evaluations. For reasoning tasks, the interface loads each JSONL sample,
renders the image panel(s), exposes a task-specific answer widget
(single choice, integer field, set/list entry, or sequence entry), and saves an
optional comment field for ambiguous cases. For planning tasks, annotators play
the same browser-based environment as the model runner. The system resets the
episode from the JSONL \texttt{reset\_config}, records every selected action,
and scores success from the environment terminal state rather than from
self-reported answers.
Figures~\ref{fig:human_annotator_interface_reasoning}
and~\ref{fig:human_annotator_interface_planning} show the two interfaces.

Five annotators were recruited through a professional annotation company and
had substantial experience in visual data annotation. Before production, each
annotator received written instructions and video training, completed 300 trial
examples, and passed a qualification round. Two additional auditors monitored
annotation quality throughout the evaluation. Annotators were paid more than
1.5 times the applicable local minimum hourly wage and agreed that their answers
and performance results could be used for research. The evaluation used only
synthetic puzzle data and collected no personal or sensitive information. It did
not involve institutional review.

The five annotators completed one annotation per example on the previous
release of the benchmark. Annotations were kept as per-tier accuracies
(Table~\ref{tab:app_human_difficulty}), and Table~\ref{tab:easy_level_human}
reports them next to the Easy-level model scores. Humans average
$94.23\%$ on the reasoning tasks and $99.87\%$ on the planning tasks, against
$76.33\%$ and $81.87\%$ for GPT-6 Astra, the strongest model, on the Easy
level.
Human accuracies are computed with
the same parser and scorer used for models. Set-valued answers are canonicalized
by sorting and de-duplicating labels, scalar answers are normalized through the
shared answer parser, and planning episodes are marked correct only when the
environment terminates successfully. Ground-truth labels come from the
generators, so these accuracies are a human reference and not a source of
labels.

\begin{table*}[!htbp]
\centering
\setlength{\tabcolsep}{2pt}
\renewcommand{\arraystretch}{1.15}
\resizebox{\textwidth}{!}{%
\begin{tabular}{@{}p{3mm} l *{15}{c}@{}}
\toprule
& \multirow{2}{*}{\textbf{Model}}
& \multicolumn{3}{c}{\textbf{Continuity}}
& \multicolumn{2}{c}{\textbf{Separation}}
& \multicolumn{3}{c}{\textbf{Order}}
& \multicolumn{3}{c}{\textbf{Enclosure}}
& \multicolumn{2}{c}{\textbf{Knots}}
& \multicolumn{2}{c}{\textbf{Avg.}} \\
\cmidrule(lr){3-5}\cmidrule(lr){6-7}\cmidrule(lr){8-10}\cmidrule(lr){11-13}\cmidrule(lr){14-15}\cmidrule(l){16-17}
&
& \PTask{2D Maze} & \PTask{3D Maze} & \ITask{Pipe}
& \PTask{Assembly} & \ITask{One Stroke}
& \PTask{Bead} & \PTask{Origami Point} & \ITask{Swap}
& \PTask{Sheep} & \PTask{Hole} & \ITask{Chat Noir}
& \PTask{Knots} & \ITask{Untangle}
& \PTask{Reason} & \ITask{Plan} \\
\midrule
& GPT-6 Astra
& \bestscore{99.33} & \bestscore{86.67} & \bestscore{99.33}
& \bestscore{74.67} & \bestscore{95.33}
& \bestscore{52.00} & \bestscore{83.33} & \bestscore{100.00}
& \bestscore{59.33} & \bestscore{66.67} & \bestscore{96.67}
& \bestscore{88.67} & \bestscore{18.00}
& \bestscore{76.33} & \bestscore{81.87} \\
& GPT-5.6-Sol
& \secondscore{60.00} & 18.67 & \secondscore{22.00}
& \secondscore{34.67} & \secondscore{18.67}
& \secondscore{8.00} & \secondscore{18.67} & \secondscore{80.67}
& 55.33 & \thirdscore{52.67} & \secondscore{70.67}
& 30.00 & 4.00
& \secondscore{34.75} & \secondscore{39.20} \\
& GPT-5.5
& 5.33 & 13.33 & 0.00
& \secondscore{34.67} & 0.00
& 0.00 & \thirdscore{9.33} & \thirdscore{39.33}
& 39.33 & \secondscore{54.67} & 25.33
& 36.67 & \secondscore{9.33}
& 24.17 & \thirdscore{14.80} \\
& GPT-5.4-mini
& 2.00 & 4.00 & 0.00
& 20.67 & 0.00
& 0.00 & 0.00 & 0.00
& 10.00 & 0.67 & 0.00
& 7.33 & 3.33
& 5.58 & 0.67 \\
& Gemini-3.1-Flash-Lite
& 0.67 & 2.67 & 0.00
& 25.33 & 0.00
& 0.00 & 1.33 & 0.00
& 26.67 & 3.33 & 0.67
& 18.00 & 0.67
& 9.75 & 0.27 \\
& Gemini-3.1-Pro
& \thirdscore{6.67} & \secondscore{27.33} & 0.00
& \thirdscore{30.00} & 0.00
& 0.00 & 1.33 & 2.00
& \secondscore{56.67} & 41.33 & 27.33
& \secondscore{64.00} & \thirdscore{6.00}
& \thirdscore{28.42} & 7.07 \\
& Claude-Haiku-5.5
& 0.67 & \thirdscore{24.67} & 0.00
& 29.33 & 0.00
& \thirdscore{0.67} & 3.33 & 0.00
& \thirdscore{56.00} & 48.67 & \thirdscore{44.00}
& \thirdscore{38.67} & 2.67
& 25.25 & 9.33 \\
& Nemotron-Nano-12B-VL-v2
& 4.00 & 2.67 & 0.00
& 26.00 & 0.00
& 0.00 & 0.00 & 0.00
& 6.67 & 0.67 & 0.00
& 3.33 & 0.00
& 5.42 & 0.00 \\
& Qwen3.5-397B-A17B
& 2.00 & 8.00 & 0.00
& 20.67 & 0.00
& 0.00 & 0.00 & 0.00
& 0.00 & 3.33 & 1.33
& 5.33 & 0.67
& 4.92 & 0.40 \\
& Llama-4-Maverick-17B-128E
& 4.67 & 2.67 & 0.00
& 25.33 & 0.00
& 0.00 & 2.00 & 0.00
& 4.00 & 0.00 & 0.00
& 1.33 & 2.00
& 5.00 & 0.40 \\
& Ministral-3-14B-Instruct-2512
& 5.33 & 4.00 & 0.00
& 23.33 & 0.00
& 0.00 & 0.00 & 0.00
& 2.67 & 0.00 & 0.00
& 3.33 & 2.67
& 4.83 & 0.53 \\
& InternVL3.5-241B
& 2.67 & 6.67 & 0.00
& 17.33 & 0.00
& 0.00 & 0.00 & 0.00
& 8.67 & 0.67 & 0.00
& 7.33 & 0.00
& 5.42 & 0.00 \\
& Gemma-4-31B-IT
& 4.00 & 6.00 & 0.00
& 17.33 & 0.00
& 0.00 & 0.00 & 0.00
& 2.67 & 0.00 & 2.67
& 2.00 & 0.67
& 4.00 & 0.67 \\
\midrule
& \textbf{Human}
& 98.33 & 95.33 & 100.00
& 83.33 & 100.00
& 93.00 & 96.00 & 100.00
& 96.33 & 99.56 & 99.33
& 92.00 & 100.00
& 94.23 & 99.87 \\
\bottomrule
\end{tabular}%
}
\caption{\textbf{Easy-level performance of humans and models (\%).} Model
scores cover the $150$ Easy instances of each task. Shading marks the best, second, and third model score in each
column.}
\label{tab:easy_level_human}
\end{table*}

\begin{table*}[!htbp]
\centering
\setlength{\tabcolsep}{7.55pt}
\renewcommand{\arraystretch}{1.15}
\resizebox{\textwidth}{!}{%
\begin{tabular}{@{}p{3mm} l *{13}{c}}
\toprule
& \multirow{2}{*}{\textbf{Tier}}
& \multicolumn{3}{c}{\textbf{Continuity}}
& \multicolumn{2}{c}{\textbf{Separation}}
& \multicolumn{3}{c}{\textbf{Order}}
& \multicolumn{3}{c}{\textbf{Enclosure}}
& \multicolumn{2}{c}{\textbf{Knots}} \\
\cmidrule(lr){3-5}\cmidrule(lr){6-7}\cmidrule(lr){8-10}\cmidrule(lr){11-13}\cmidrule(l){14-15}
&
& \PTask{2D Maze} & \PTask{3D Maze} & \ITask{Pipe}
& \PTask{Assembly} & \ITask{One Stroke}
& \PTask{Bead} & \PTask{Origami Point} & \ITask{Swap}
& \PTask{Sheep} & \PTask{Hole} & \ITask{Chat Noir}
& \PTask{Knots} & \ITask{Untangle} \\
\midrule
& easy
& 98.67 & 98.00 & 100.00
& 97.00 & 99.00
& 98.33 & 95.00 & 100.00
& 99.33 & 99.67 & 100.00
& 99.67 & 100.00 \\
& medium
& 98.00 & 97.00 & 100.00
& 94.67 & 100.00
& 95.67 & 95.00 & 100.00
& 97.67 & 99.67 & 99.33
& 90.33 & 100.00 \\
& hard
& 98.33 & 95.33 & 100.00
& 83.33 & 100.00
& 93.00 & 98.33 & 100.00
& 96.33 & 99.33 & 99.33
& 92.00 & 100.00 \\
\midrule
& \textbf{Total}
& 98.33 & 96.67 & 100.00
& 93.67 & 99.67
& 95.67 & 96.00 & 100.00
& 97.67 & 99.67 & 99.67
& 94.67 & 100.00 \\
\bottomrule
\end{tabular}%
}
\caption{Human accuracy on reasoning tasks and episode success rate on planning
tasks (\%) by difficulty tier of the previous release. Total is computed over
all three tiers.}
\label{tab:app_human_difficulty}
\end{table*}

\subsubsection{Inter-Annotator Agreement}
\label{app:iaa}

To measure inter-annotator agreement (IAA), three annotators independently
evaluated a shared subset of 200 examples across both reasoning and planning
tasks under the same instructions as the models. After the same answer
canonicalization used by the benchmark scorer, agreement among the three
annotators is $0.89$.

For reasoning tasks, agreement is computed by exact match after answer canonicalization. For planning tasks, agreement is computed under the task's deterministic evaluation protocol, so two annotations agree only when they reach the same evaluated outcome. Disagreements are not resolved by majority vote for the purpose of IAA calculation. Instead, they are reviewed separately to identify potential ambiguity, interface issues, or annotation errors.

\subsection{Full Results by Difficulty}
\label{app:full_results_difficulty}

Table~\ref{tab:topology_benchmark_refreshed} reports each task on 300 instances.
Tables~\ref{tab:app_difficulty_continuity} to \ref{tab:app_difficulty_knots} split
every score into its two levels of 150 instances, grouped by topological property.
Each score in
Table~\ref{tab:topology_benchmark_refreshed} is the mean of its two columns.

\definecolor{difficultyblue}{HTML}{5B8DB8}
\providecommand{\diffscore}[2]{\cellcolor{difficultyblue!#1}#2}

\begin{table}[H]
\centering
\scriptsize
\setlength{\tabcolsep}{3pt}
\renewcommand{\arraystretch}{1.05}
\begin{tabularx}{\textwidth}{@{}p{3mm} l *{6}{C}@{}}
\toprule
& \multirow{2}{*}{\textbf{Model}}
& \multicolumn{2}{c}{\PTask{2D Maze}} & \multicolumn{2}{c}{\PTask{3D Maze}} & \multicolumn{2}{c}{\ITask{Pipe}} \\
\cmidrule(lr){3-4}\cmidrule(lr){5-6}\cmidrule(l){7-8}
&
& \textbf{\scriptsize Easy} & \textbf{\scriptsize Hard} & \textbf{\scriptsize Easy} & \textbf{\scriptsize Hard} & \textbf{\scriptsize Easy} & \textbf{\scriptsize Hard} \\
\midrule
\multicolumn{8}{>{\columncolor{proprietary}}l}{\textit{\scriptsize Proprietary Models}} \\
& GPT-6 Astra & \bestscore{99.33} & \bestscore{42.00} & \bestscore{86.67} & \bestscore{76.00} & \bestscore{99.33} & \bestscore{10.00} \\
& GPT-5.6-Sol & \secondscore{60.00} & \diffscore{6}{6.00} & \diffscore{10}{18.67} & \secondscore{9.33} & \secondscore{22.00} & \diffscore{4}{0.00} \\
& GPT-5.5 & \diffscore{6}{5.33} & \diffscore{8}{12.00} & \diffscore{8}{13.33} & \diffscore{5}{2.00} & \diffscore{4}{0.00} & \diffscore{4}{0.00} \\
& GPT-5.4-mini & \diffscore{5}{2.00} & \diffscore{8}{13.33} & \diffscore{5}{4.00} & \diffscore{4}{1.33} & \diffscore{4}{0.00} & \diffscore{4}{0.00} \\
& Gemini-3.1-Flash-Lite & \diffscore{4}{0.67} & \diffscore{8}{12.67} & \diffscore{5}{2.67} & \diffscore{4}{0.00} & \diffscore{4}{0.00} & \diffscore{4}{0.00} \\
& Gemini-3.1-Pro & \diffscore{6}{6.67} & \secondscore{14.67} & \secondscore{27.33} & \diffscore{4}{0.67} & \diffscore{4}{0.00} & \diffscore{4}{0.00} \\
& Claude-Haiku-5.5 & \diffscore{4}{0.67} & \diffscore{4}{1.33} & \diffscore{12}{24.67} & \diffscore{5}{2.00} & \diffscore{4}{0.00} & \diffscore{4}{0.00} \\
\midrule
\multicolumn{8}{>{\columncolor{openweight}}l}{\textit{\scriptsize Hosted Open-Weight Models}} \\
& Nemotron-Nano-12B-VL-v2 & \diffscore{5}{4.00} & \diffscore{5}{4.00} & \diffscore{5}{2.67} & \diffscore{4}{0.00} & \diffscore{4}{0.00} & \diffscore{4}{0.00} \\
& Qwen3.5-397B-A17B & \diffscore{5}{2.00} & \diffscore{8}{11.33} & \diffscore{7}{8.00} & \diffscore{4}{1.33} & \diffscore{4}{0.00} & \diffscore{4}{0.00} \\
& Llama-4-Maverick-17B-128E & \diffscore{6}{4.67} & \diffscore{8}{11.33} & \diffscore{5}{2.67} & \diffscore{6}{4.67} & \diffscore{4}{0.00} & \diffscore{4}{0.00} \\
& Ministral-3-14B-Instruct-2512 & \diffscore{6}{5.33} & \diffscore{6}{4.67} & \diffscore{5}{4.00} & \diffscore{4}{0.00} & \diffscore{4}{0.00} & \diffscore{4}{0.00} \\
& InternVL3.5-241B & \diffscore{5}{2.67} & \diffscore{4}{1.33} & \diffscore{6}{6.67} & \diffscore{4}{0.00} & \diffscore{4}{0.00} & \diffscore{4}{0.00} \\
& Gemma-4-31B-IT & \diffscore{5}{4.00} & \diffscore{4}{0.67} & \diffscore{6}{6.00} & \diffscore{4}{0.67} & \diffscore{4}{0.00} & \diffscore{4}{0.00} \\
\bottomrule
\end{tabularx}
\caption{\textbf{Continuity results by difficulty level.} Accuracy (\%) for reasoning tasks and episode success rate (\%) for planning tasks on the Easy and Hard levels, 150 instances each. Each Table~\ref{tab:topology_benchmark_refreshed} score is the mean of the two columns. Dark and light lavender mark the highest and second highest positive score per column, and blue intensity grows with the score.}
\label{tab:app_difficulty_continuity}
\end{table}

\begin{table}[H]
\centering
\scriptsize
\setlength{\tabcolsep}{3pt}
\renewcommand{\arraystretch}{1.05}
\begin{tabularx}{\textwidth}{@{}p{3mm} l *{4}{C}@{}}
\toprule
& \multirow{2}{*}{\textbf{Model}}
& \multicolumn{2}{c}{\PTask{Assembly}} & \multicolumn{2}{c}{\ITask{One Stroke}} \\
\cmidrule(lr){3-4}\cmidrule(l){5-6}
&
& \textbf{\scriptsize Easy} & \textbf{\scriptsize Hard} & \textbf{\scriptsize Easy} & \textbf{\scriptsize Hard} \\
\midrule
\multicolumn{6}{>{\columncolor{proprietary}}l}{\textit{\scriptsize Proprietary Models}} \\
& GPT-6 Astra & \bestscore{74.67} & \bestscore{67.33} & \bestscore{95.33} & \bestscore{46.00} \\
& GPT-5.6-Sol & \secondscore{34.67} & \secondscore{37.33} & \secondscore{18.67} & \diffscore{4}{0.00} \\
& GPT-5.5 & \secondscore{34.67} & \diffscore{14}{31.33} & \diffscore{4}{0.00} & \diffscore{4}{0.00} \\
& GPT-5.4-mini & \diffscore{11}{20.67} & \diffscore{12}{24.67} & \diffscore{4}{0.00} & \diffscore{4}{0.00} \\
& Gemini-3.1-Flash-Lite & \diffscore{12}{25.33} & \diffscore{12}{23.33} & \diffscore{4}{0.00} & \diffscore{4}{0.00} \\
& Gemini-3.1-Pro & \diffscore{14}{30.00} & \diffscore{12}{24.00} & \diffscore{4}{0.00} & \diffscore{4}{0.00} \\
& Claude-Haiku-5.5 & \diffscore{14}{29.33} & \diffscore{14}{30.67} & \diffscore{4}{0.00} & \diffscore{4}{0.00} \\
\midrule
\multicolumn{6}{>{\columncolor{openweight}}l}{\textit{\scriptsize Hosted Open-Weight Models}} \\
& Nemotron-Nano-12B-VL-v2 & \diffscore{13}{26.00} & \diffscore{10}{17.33} & \diffscore{4}{0.00} & \diffscore{4}{0.00} \\
& Qwen3.5-397B-A17B & \diffscore{11}{20.67} & \diffscore{12}{24.00} & \diffscore{4}{0.00} & \diffscore{4}{0.00} \\
& Llama-4-Maverick-17B-128E & \diffscore{12}{25.33} & \diffscore{12}{22.67} & \diffscore{4}{0.00} & \diffscore{4}{0.00} \\
& Ministral-3-14B-Instruct-2512 & \diffscore{12}{23.33} & \diffscore{12}{24.67} & \diffscore{4}{0.00} & \diffscore{4}{0.00} \\
& InternVL3.5-241B & \diffscore{10}{17.33} & \diffscore{12}{23.33} & \diffscore{4}{0.00} & \diffscore{4}{0.00} \\
& Gemma-4-31B-IT & \diffscore{10}{17.33} & \diffscore{7}{8.67} & \diffscore{4}{0.00} & \diffscore{4}{0.00} \\
\bottomrule
\end{tabularx}
\caption{\textbf{Separation results by difficulty level.} Same layout as Table~\ref{tab:app_difficulty_continuity}.}
\label{tab:app_difficulty_separation}
\end{table}

\begin{table}[H]
\centering
\scriptsize
\setlength{\tabcolsep}{3pt}
\renewcommand{\arraystretch}{1.05}
\begin{tabularx}{\textwidth}{@{}p{3mm} l *{6}{C}@{}}
\toprule
& \multirow{2}{*}{\textbf{Model}}
& \multicolumn{2}{c}{\PTask{Bead}} & \multicolumn{2}{c}{\PTask{Origami Point}} & \multicolumn{2}{c}{\ITask{Swap}} \\
\cmidrule(lr){3-4}\cmidrule(lr){5-6}\cmidrule(l){7-8}
&
& \textbf{\scriptsize Easy} & \textbf{\scriptsize Hard} & \textbf{\scriptsize Easy} & \textbf{\scriptsize Hard} & \textbf{\scriptsize Easy} & \textbf{\scriptsize Hard} \\
\midrule
\multicolumn{8}{>{\columncolor{proprietary}}l}{\textit{\scriptsize Proprietary Models}} \\
& GPT-6 Astra & \bestscore{52.00} & \bestscore{54.67} & \bestscore{83.33} & \bestscore{55.33} & \bestscore{100.00} & \bestscore{2.67} \\
& GPT-5.6-Sol & \secondscore{8.00} & \secondscore{29.33} & \secondscore{18.67} & \diffscore{4}{0.67} & \secondscore{80.67} & \diffscore{4}{0.00} \\
& GPT-5.5 & \diffscore{4}{0.00} & \diffscore{9}{14.67} & \diffscore{7}{9.33} & \secondscore{11.33} & \diffscore{17}{39.33} & \diffscore{4}{0.00} \\
& GPT-5.4-mini & \diffscore{4}{0.00} & \diffscore{8}{10.67} & \diffscore{4}{0.00} & \diffscore{6}{4.67} & \diffscore{4}{0.00} & \diffscore{4}{0.00} \\
& Gemini-3.1-Flash-Lite & \diffscore{4}{0.00} & \diffscore{8}{12.67} & \diffscore{4}{1.33} & \diffscore{5}{4.00} & \diffscore{4}{0.00} & \diffscore{4}{0.00} \\
& Gemini-3.1-Pro & \diffscore{4}{0.00} & \diffscore{8}{12.00} & \diffscore{4}{1.33} & \diffscore{6}{4.67} & \diffscore{5}{2.00} & \diffscore{4}{0.00} \\
& Claude-Haiku-5.5 & \diffscore{4}{0.67} & \diffscore{11}{21.33} & \diffscore{5}{3.33} & \diffscore{6}{5.33} & \diffscore{4}{0.00} & \diffscore{4}{0.00} \\
\midrule
\multicolumn{8}{>{\columncolor{openweight}}l}{\textit{\scriptsize Hosted Open-Weight Models}} \\
& Nemotron-Nano-12B-VL-v2 & \diffscore{4}{0.00} & \diffscore{6}{5.33} & \diffscore{4}{0.00} & \diffscore{4}{0.67} & \diffscore{4}{0.00} & \diffscore{4}{0.00} \\
& Qwen3.5-397B-A17B & \diffscore{4}{0.00} & \diffscore{7}{9.33} & \diffscore{4}{0.00} & \diffscore{5}{2.67} & \diffscore{4}{0.00} & \diffscore{4}{0.00} \\
& Llama-4-Maverick-17B-128E & \diffscore{4}{0.00} & \diffscore{7}{9.33} & \diffscore{5}{2.00} & \diffscore{5}{4.00} & \diffscore{4}{0.00} & \diffscore{4}{0.00} \\
& Ministral-3-14B-Instruct-2512 & \diffscore{4}{0.00} & \diffscore{9}{14.00} & \diffscore{4}{0.00} & \diffscore{5}{3.33} & \diffscore{4}{0.00} & \diffscore{4}{0.00} \\
& InternVL3.5-241B & \diffscore{4}{0.00} & \diffscore{6}{6.00} & \diffscore{4}{0.00} & \diffscore{6}{7.33} & \diffscore{4}{0.00} & \diffscore{4}{0.00} \\
& Gemma-4-31B-IT & \diffscore{4}{0.00} & \diffscore{4}{0.00} & \diffscore{4}{0.00} & \diffscore{4}{0.00} & \diffscore{4}{0.00} & \diffscore{4}{0.00} \\
\bottomrule
\end{tabularx}
\caption{\textbf{Order results by difficulty level.} Same layout as Table~\ref{tab:app_difficulty_continuity}.}
\label{tab:app_difficulty_order}
\end{table}

\begin{table}[H]
\centering
\scriptsize
\setlength{\tabcolsep}{3pt}
\renewcommand{\arraystretch}{1.05}
\begin{tabularx}{\textwidth}{@{}p{3mm} l *{6}{C}@{}}
\toprule
& \multirow{2}{*}{\textbf{Model}}
& \multicolumn{2}{c}{\PTask{Sheep}} & \multicolumn{2}{c}{\PTask{Hole}} & \multicolumn{2}{c}{\ITask{Chat Noir}} \\
\cmidrule(lr){3-4}\cmidrule(lr){5-6}\cmidrule(l){7-8}
&
& \textbf{\scriptsize Easy} & \textbf{\scriptsize Hard} & \textbf{\scriptsize Easy} & \textbf{\scriptsize Hard} & \textbf{\scriptsize Easy} & \textbf{\scriptsize Hard} \\
\midrule
\multicolumn{8}{>{\columncolor{proprietary}}l}{\textit{\scriptsize Proprietary Models}} \\
& GPT-6 Astra & \bestscore{59.33} & \secondscore{11.33} & \bestscore{66.67} & \diffscore{4}{0.00} & \bestscore{96.67} & \bestscore{6.67} \\
& GPT-5.6-Sol & \diffscore{22}{55.33} & \diffscore{6}{7.33} & \diffscore{22}{52.67} & \diffscore{4}{0.00} & \secondscore{70.67} & \diffscore{4}{0.00} \\
& GPT-5.5 & \diffscore{17}{39.33} & \diffscore{7}{8.00} & \secondscore{54.67} & \diffscore{4}{0.00} & \diffscore{12}{25.33} & \diffscore{4}{0.00} \\
& GPT-5.4-mini & \diffscore{7}{10.00} & \diffscore{4}{0.00} & \diffscore{4}{0.67} & \diffscore{4}{0.00} & \diffscore{4}{0.00} & \diffscore{4}{0.00} \\
& Gemini-3.1-Flash-Lite & \diffscore{13}{26.67} & \diffscore{5}{2.00} & \diffscore{5}{3.33} & \diffscore{4}{0.00} & \diffscore{4}{0.67} & \diffscore{4}{0.00} \\
& Gemini-3.1-Pro & \secondscore{56.67} & \diffscore{5}{4.00} & \diffscore{18}{41.33} & \diffscore{4}{0.00} & \diffscore{13}{27.33} & \diffscore{4}{0.00} \\
& Claude-Haiku-5.5 & \diffscore{23}{56.00} & \bestscore{28.67} & \diffscore{20}{48.67} & \bestscore{2.00} & \diffscore{19}{44.00} & \diffscore{4}{0.00} \\
\midrule
\multicolumn{8}{>{\columncolor{openweight}}l}{\textit{\scriptsize Hosted Open-Weight Models}} \\
& Nemotron-Nano-12B-VL-v2 & \diffscore{6}{6.67} & \diffscore{4}{0.00} & \diffscore{4}{0.67} & \diffscore{4}{0.00} & \diffscore{4}{0.00} & \diffscore{4}{0.00} \\
& Qwen3.5-397B-A17B & \diffscore{4}{0.00} & \diffscore{4}{0.67} & \diffscore{5}{3.33} & \diffscore{4}{0.00} & \diffscore{4}{1.33} & \diffscore{4}{0.00} \\
& Llama-4-Maverick-17B-128E & \diffscore{5}{4.00} & \diffscore{4}{0.00} & \diffscore{4}{0.00} & \diffscore{4}{0.00} & \diffscore{4}{0.00} & \diffscore{4}{0.00} \\
& Ministral-3-14B-Instruct-2512 & \diffscore{5}{2.67} & \diffscore{4}{0.00} & \diffscore{4}{0.00} & \diffscore{4}{0.00} & \diffscore{4}{0.00} & \diffscore{4}{0.00} \\
& InternVL3.5-241B & \diffscore{7}{8.67} & \diffscore{4}{0.00} & \diffscore{4}{0.67} & \diffscore{4}{0.00} & \diffscore{4}{0.00} & \diffscore{4}{0.00} \\
& Gemma-4-31B-IT & \diffscore{5}{2.67} & \secondscore{11.33} & \diffscore{4}{0.00} & \diffscore{4}{0.00} & \diffscore{5}{2.67} & \diffscore{4}{0.00} \\
\bottomrule
\end{tabularx}
\caption{\textbf{Enclosure results by difficulty level.} Same layout as Table~\ref{tab:app_difficulty_continuity}.}
\label{tab:app_difficulty_enclosure}
\end{table}

\begin{table}[H]
\centering
\scriptsize
\setlength{\tabcolsep}{3pt}
\renewcommand{\arraystretch}{1.05}
\begin{tabularx}{\textwidth}{@{}p{3mm} l *{4}{C}@{}}
\toprule
& \multirow{2}{*}{\textbf{Model}}
& \multicolumn{2}{c}{\PTask{Knots}} & \multicolumn{2}{c}{\ITask{Untangle}} \\
\cmidrule(lr){3-4}\cmidrule(l){5-6}
&
& \textbf{\scriptsize Easy} & \textbf{\scriptsize Hard} & \textbf{\scriptsize Easy} & \textbf{\scriptsize Hard} \\
\midrule
\multicolumn{6}{>{\columncolor{proprietary}}l}{\textit{\scriptsize Proprietary Models}} \\
& GPT-6 Astra & \bestscore{88.67} & \bestscore{35.33} & \bestscore{18.00} & \bestscore{23.33} \\
& GPT-5.6-Sol & \diffscore{14}{30.00} & \diffscore{11}{20.67} & \diffscore{5}{4.00} & \diffscore{5}{2.00} \\
& GPT-5.5 & \diffscore{16}{36.67} & \secondscore{24.67} & \secondscore{9.33} & \diffscore{6}{6.00} \\
& GPT-5.4-mini & \diffscore{6}{7.33} & \diffscore{10}{17.33} & \diffscore{5}{3.33} & \diffscore{5}{3.33} \\
& Gemini-3.1-Flash-Lite & \diffscore{10}{18.00} & \diffscore{10}{18.00} & \diffscore{4}{0.67} & \diffscore{6}{5.33} \\
& Gemini-3.1-Pro & \secondscore{64.00} & \diffscore{10}{19.33} & \diffscore{6}{6.00} & \diffscore{6}{6.67} \\
& Claude-Haiku-5.5 & \diffscore{17}{38.67} & \secondscore{24.67} & \diffscore{5}{2.67} & \secondscore{7.33} \\
\midrule
\multicolumn{6}{>{\columncolor{openweight}}l}{\textit{\scriptsize Hosted Open-Weight Models}} \\
& Nemotron-Nano-12B-VL-v2 & \diffscore{5}{3.33} & \diffscore{10}{18.67} & \diffscore{4}{0.00} & \diffscore{4}{1.33} \\
& Qwen3.5-397B-A17B & \diffscore{6}{5.33} & \diffscore{10}{16.67} & \diffscore{4}{0.67} & \diffscore{4}{0.00} \\
& Llama-4-Maverick-17B-128E & \diffscore{4}{1.33} & \diffscore{10}{18.67} & \diffscore{5}{2.00} & \diffscore{5}{3.33} \\
& Ministral-3-14B-Instruct-2512 & \diffscore{5}{3.33} & \diffscore{9}{14.00} & \diffscore{5}{2.67} & \diffscore{5}{3.33} \\
& InternVL3.5-241B & \diffscore{6}{7.33} & \diffscore{11}{22.00} & \diffscore{4}{0.00} & \diffscore{4}{0.00} \\
& Gemma-4-31B-IT & \diffscore{5}{2.00} & \diffscore{9}{15.33} & \diffscore{4}{0.67} & \diffscore{4}{0.67} \\
\bottomrule
\end{tabularx}
\caption{\textbf{Knots results by difficulty level.} Same layout as Table~\ref{tab:app_difficulty_continuity}.}
\label{tab:app_difficulty_knots}
\end{table}

\FloatBarrier

\subsection{Generative Models as Topological World Models}
\label{app:generative_world_model}

\paragraph{Setup.}
We evaluate generative models as auxiliary world models rather than as direct
answerers. When the planner calls a generator, it first writes an image or
video prompt describing a plausible topological rollout from the current
observation. The
generator then produces an imagined future observation, and the same planner
commits to an answer or action from the original observation plus the generated
artifact. Image experiments use edit-mode generation anchored to the current
PNG observation, with \texttt{gpt-image-2} as the image backend. Video experiments use image-to-video rollouts with Wan2.2-I2V-A14B, Seedance-2.0-Mini, or Veo-3.1-Lite as the replay backend. Controls keep the planner, initial
observation, task prompt, and action budget fixed across no-imagination and
imagination variants.

\paragraph{Imagined-rollout protocol.}
For continuity tasks, the imagined artifact is asked to show a path opening,
pipe flow, or maze traversal after the candidate manipulation. For separation,
the rollout visualizes whether a stroke or object remains one connected piece.
For order, it shows the post-fold or post-swap ordering. For enclosure, it
shows escape paths, boundary closure, or hole visibility after a viewpoint or
state change. For knots, it asks the generator to simulate an endpoint move or
strand deformation while preserving over/under crossings. The planner may
call the image generator at every step, capped at ten calls per episode. The
end-to-end video setting uses one video per episode. Per-task success rates
for the interleaved-image and video variants appear in
Table~\ref{tab:interleaved_results}. Generated conditions and the GPT-5.6-Luna planning baselines use $105$
episodes per task, $35$ from each of three generator difficulty settings, so
they include instances outside the Easy and Hard benchmark levels. The three probing analyses draw on different samples. The
error annotation in Figure~\ref{fig:probing_gen_errors} comes from a separate
run with GPT-5.4-mini as the planner and covers $464$ GPT-Image-2 images and a
set of Wan2.2-I2V-A14B rollouts, with $60$ videos targeted for One Stroke and
$60$ for Untangle. One Untangle video was unavailable, leaving 60 scored
One Stroke videos and 59 scored Untangle videos (119 total).
The CV diagnostic audit in Table~\ref{tab:cv_metric_results} uses a third
sample of $105$ videos for each pairing of a task with one of the six
generators, $1{,}890$ videos in total.

\paragraph{Failure modes.}
Imagined rollouts often look locally plausible while violating the exact
topological invariant that the task tests. Our annotations track three
diagnostics, namely dynamics, invariance, and cross-frame consistency. Across the $119$ annotated Wan2.2-I2V-A14B video
rollouts, $116$ contain a dynamics error, $93$ an invariance error, and $90$ a
consistency error. Qualitative manifestations include identity drift, silent crossings,
hallucinated openings, and visually smooth but impossible transitions.

\subsubsection{Computer-Vision Video Diagnostics}
\label{app:cv_video_diagnostics}

\begin{figure*}[!htbp]
    \centering
    \includegraphics[width=\textwidth]{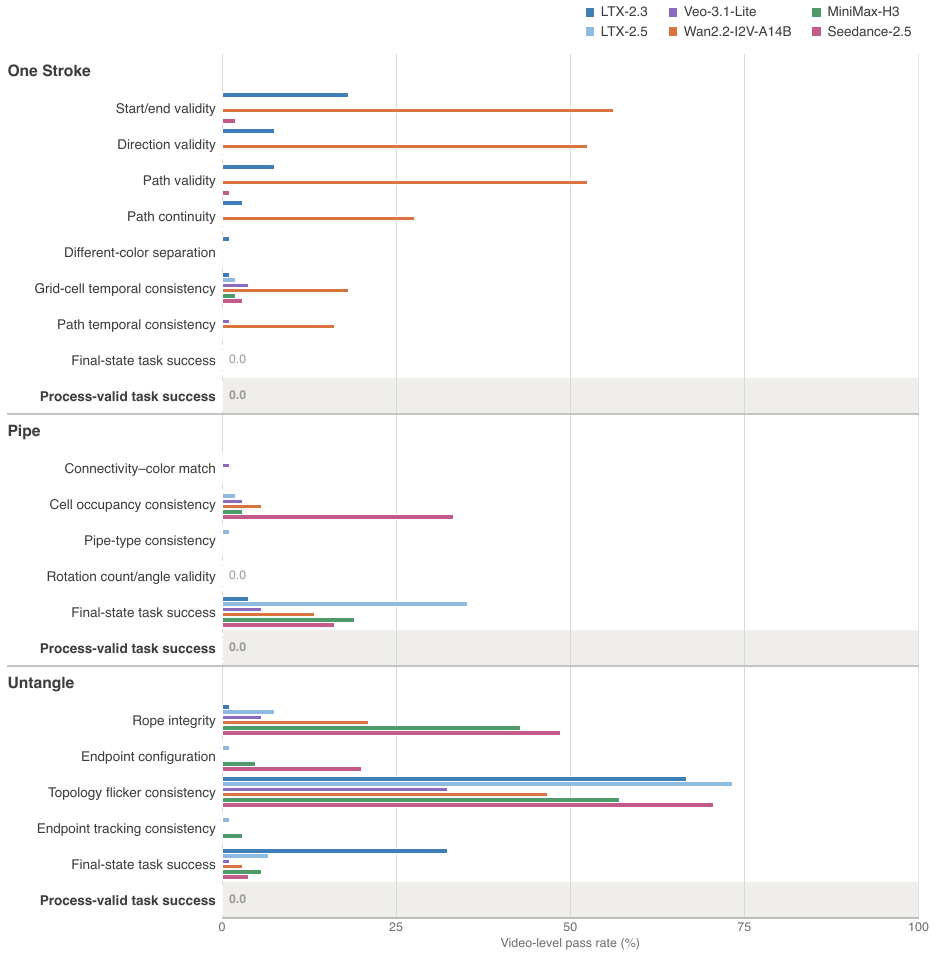}
    \caption{\textbf{Per-check video-level pass rates from Table~\ref{tab:cv_metric_results}, ordered loosest to strictest.} The shaded \emph{process-valid task success} row combines final-state success with every applicable static and dynamic check. The two per-group aggregates are omitted here for readability.}
    \label{fig:cv_metric_results}
\end{figure*}

\paragraph{Verifier and aggregation.}
We apply task-specific computer-vision verifiers to One Stroke, Pipe, and
Untangle rollouts from LTX-2.3, LTX-2.5, Veo-3.1-Lite, Wan2.2-I2V-A14B, MiniMax-H3, and Seedance-2.5. Each task and
generator cell contains 105 videos, for 1{,}890 videos in total, and an episode
without a generated video counts as failing every check. Static checks test whether each
selected parsed frame preserves the task state. Dynamic checks test whether consecutive
frames form a valid transition. The reported CV audit uses the default
\texttt{sample\_41} mode, which samples 41 evenly spaced frames from each
video, including the first and last frames. A video passes a check only when every
applicable frame or transition passes. Unscorable videos and videos with no
applicable instance of a check count as failures, so every rate in
Table~\ref{tab:cv_metric_results} (Figure~\ref{fig:cv_metric_results}) is a count out of 105.

\paragraph{Terminal-state parser status and scorability.}
Parser labels describe detector completeness. They are task-specific and are converted to the common binary
\emph{terminal state scorable} gate only after parsing. For the three tasks in
Table~\ref{tab:cv_metric_results}, \texttt{ok} and \texttt{degraded} are
readable, whereas \texttt{failed} and \texttt{grid\_not\_found} are unreadable.
Thus, a degraded frame can remain scorable when it contains enough state for
the task metrics, while a failed frame is counted as a failure in outcome,
static, dynamic, and overall video-level aggregates. Table~\ref{tab:cv_parser_status}
states the exact triggers and observed terminal-frame counts.

\paragraph{Operational metric definitions.}
Tables~\ref{tab:cv_metric_defs_one_stroke}
and~\ref{tab:cv_metric_defs_pipe_untangle} define every task-specific metric in
Table~\ref{tab:cv_metric_results}. Static metrics are evaluated on four evenly
spaced frames on which they apply, and the One Stroke color-separation and
end-marker checks use the final frame. Terminal outcome metrics use the final
sampled frame. Dynamic metrics compare adjacent sampled frames.
Only rows explicitly marked \texttt{not\_applicable} are omitted from a check's
denominator. Low-confidence rows remain applicable and their pass/fail decision
is retained. At video level, \emph{static validity} and \emph{dynamic validity}
are strict logical conjunctions over their respective applicable rows.
\emph{Final-state task success} applies the task-specific terminal oracle, and
\emph{process-valid task success} requires terminal scorability, task success,
static validity, and dynamic validity simultaneously.

\begin{figure*}[!htbp]
    \centering
    \includegraphics[width=\textwidth]{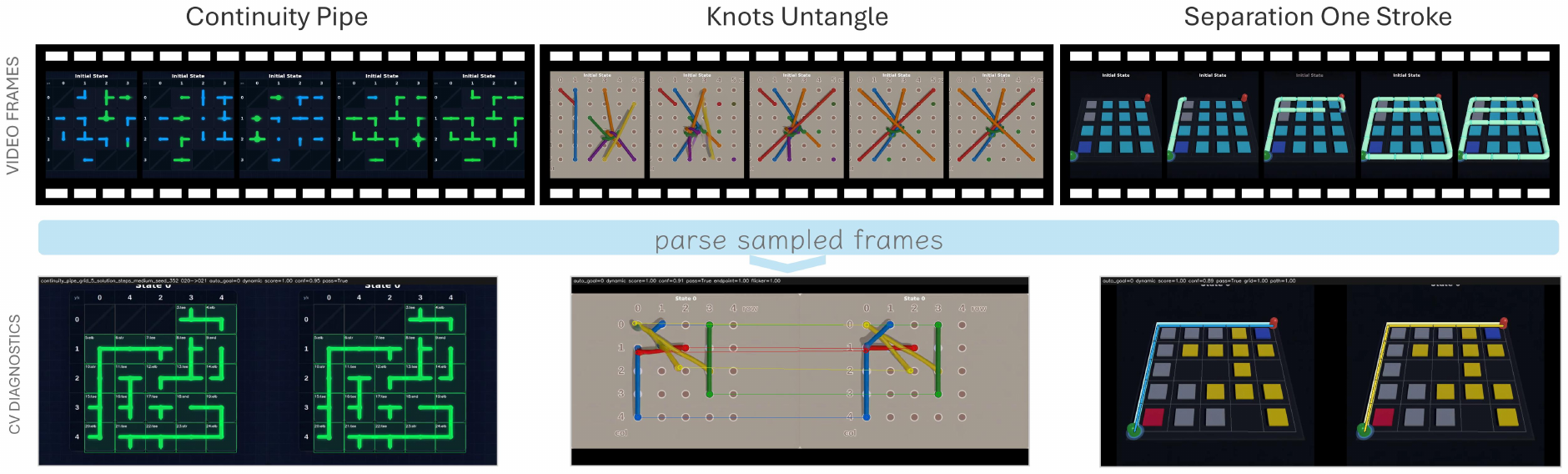}
    \caption{\textbf{Video sampling and computer-vision transition diagnostics.}
    For Pipe, Untangle, and One Stroke, the top filmstrips show five sampled
    frames from representative generated rollouts, and the bottom panels show
    task-specific overlays for adjacent-frame checks.}
    \label{fig:video_frames_and_cv}
\end{figure*}

\paragraph{Qualitative examples.}
Figure~\ref{fig:video_frames_and_cv} connects sampled generated-video frames
to the task-specific CV overlays used for dynamic verification. Its three
columns cover Pipe, Untangle, and One Stroke. The overlays illustrate the
cross-frame correspondences defined in
Tables~\ref{tab:cv_metric_defs_one_stroke}
and~\ref{tab:cv_metric_defs_pipe_untangle}.

\begin{table*}[!t]
\centering
\scriptsize
\setlength{\tabcolsep}{4pt}
\renewcommand{\arraystretch}{1.08}
\begin{tabularx}{\textwidth}{@{}l l X c r r@{}}
\toprule
\textbf{Environment} & \textbf{Raw terminal status} &
\textbf{Operational trigger} & \textbf{Scorable} &
\textbf{LTX-2.3} & \textbf{Wan2.2-I2V-A14B} \\
\midrule
\multirow{3}{*}{One Stroke}
 & \texttt{ok} & At least one colored cell and both start and end markers are detected. & Yes & 87 & 102 \\
 & \texttt{degraded} & Colored cells are detected, but either the start or end marker is missing. & Yes & 11 & 3 \\
 & \texttt{failed} & No colored cells are detected in the terminal frame. & No & 7 & 0 \\
\midrule
\multirow{2}{*}{Pipe}
 & \texttt{ok} & At least one active pipe cell is detected on the fitted board grid. & Yes & 97 & 105 \\
 & \texttt{grid\_not\_found} & No active pipe cell is detected, so no pipe state can be constructed. & No & 8 & 0 \\
\midrule
\multirow{3}{*}{Untangle}
 & \texttt{ok} & The hole lattice is fitted and every detected rope has exactly two connected, unambiguous endpoints resolved to lattice holes. & Yes & 2 & 3 \\
 & \texttt{degraded} & The hole lattice and at least one expected rope are detected, but endpoint evidence is incomplete, extra, isolated, ambiguous, or not fully resolved. & Yes & 96 & 102 \\
 & \texttt{failed} & The hole lattice cannot be fitted, or no expected rope mask survives the minimum-area detector threshold. & No & 7 & 0 \\
\bottomrule
\end{tabularx}
\caption{\textbf{Task-specific terminal parser statuses and their mapping to terminal-state scorability.}
Counts are terminal frames among the 105 videos in each cell of environment and generator.
Status labels detector completeness, and \texttt{failed} marks a terminal
frame whose puzzle state cannot be read.}
\label{tab:cv_parser_status}
\end{table*}

\begin{table*}[!t]
\centering
\scriptsize
\setlength{\tabcolsep}{5pt}
\renewcommand{\arraystretch}{1.08}
\begin{tabularx}{\textwidth}{@{}l l X@{}}
\toprule
\textbf{Type} & \textbf{Metric} & \textbf{Operational pass condition} \\
\midrule
Static & Different-color separation & On the final frame, the detected path partitions the board so that no connected region contains cells of more than one color. This isolates the separation constraint from the full task oracle. \\
Static & Direction validity & The detected arrowhead lies at the advancing path endpoint, points away from the start, and points toward the end. An orientable path without reliable arrowhead evidence is retained as a low-confidence pass. \\
Static & Path continuity & The stroke mask and skeleton each have one significant component, the snapped legal-edge graph is connected, fitted segment gaps remain below the grid-scaled tolerance, and combined confidence is at least 0.60. \\
Static & Path validity & The connected stroke snaps to one or more legal up/down/left/right board edges with snap confidence at least 0.55 and sufficient evidence that it does not cut through colored cells. A straight diagonal fails. \\
Static & Start/end validity & An intermediate path remains anchored at the start. On the final frame, one connected stroke must touch both the start and end markers. \\
Dynamic & Grid-cell temporal consistency & Adjacent frames preserve the number and color of cells and match each cell to a same-color position within the grid-scaled spatial tolerance. \\
Dynamic & Path temporal consistency & The current path retains the preceding legal-edge set except for edges removed by legal backtracking, adds no disconnected branch, and keeps sufficient adjacent-frame overlap. When edge snapping is unavailable, path area retention must be at least 0.75 and pixel overlap at least 0.45. Either branch is also gated by per-frame path continuity. \\
Outcome & Final-state task success & The final frame contains one connected start-to-end path on legal grid edges and passes the environment's exact region-constraint solver. \\
\bottomrule
\end{tabularx}
\caption{\textbf{Operational definitions of the One Stroke CV metrics.}
Early frames before any path is drawn are marked not applicable for path-based
checks rather than counted as passes.}
\label{tab:cv_metric_defs_one_stroke}
\end{table*}

\begin{table*}[!t]
\centering
\scriptsize
\setlength{\tabcolsep}{5pt}
\renewcommand{\arraystretch}{1.08}
\begin{tabularx}{\textwidth}{@{}l l l X@{}}
\toprule
\textbf{Environment} & \textbf{Type} & \textbf{Metric} & \textbf{Operational pass condition} \\
\midrule
\multirow{5}{*}{Pipe}
 & Static & Connectivity and color match & For every detected pipe cell, green means reachable from the source under the parsed arm connections and blue means unreachable. The source must be green. \\
 & Dynamic & Cell occupancy consistency & The set of occupied pipe-grid cells is identical in adjacent frames, and no pipe cell appears or disappears. \\
 & Dynamic & Pipe-type consistency & Every occupied cell shared by adjacent frames keeps the same rotation-invariant pipe shape. Orientation may change, but endpoint, straight, elbow, and tee identities may not. \\
 & Dynamic & Rotation count/angle validity & At most one pipe changes orientation in an adjacent transition, and it moves to a registered legal orientation no more than one quarter-turn away. The pixel-angle fallback likewise requires an absolute rotation below $90^{\circ}$. \\
 & Outcome & Final-state task success & Every ground-truth active pipe cell is connected to the source in the parsed terminal configuration. \\
\midrule
\multirow{5}{*}{Untangle}
 & Static & Endpoint configuration & Every detected rope has nonzero mask area, exactly two selected and resolved endpoints, no extra endpoint candidate, and no isolated endpoint cap. \\
 & Static & Rope integrity & Each rope mask has sufficient area and is not visibly fragmented, meaning that its largest component covers at least 45\% of rope area or it has at most two large components (each at least 12\% of rope area). At least 90\% of ropes must pass. \\
 & Dynamic & Endpoint tracking consistency & Each rope persists across adjacent frames and its two endpoints can be matched within half the nearest hole spacing. Missing ropes, unreadable endpoint counts, or larger jumps fail, while ambiguity is retained as low confidence rather than an automatic hard failure. \\
 & Dynamic & Topology flicker consistency & Crossing count may not change while endpoint-to-hole assignments are fixed or endpoints are nearly stationary, and endpoint-to-hole assignments may not change while endpoints are nearly stationary. A topology change accompanied by visible endpoint motion is not applicable to this flicker check. \\
 & Outcome & Final-state task success & The terminal crossing count is readable and equals zero. \\
\bottomrule
\end{tabularx}
\caption{\textbf{Operational definitions of the Pipe and Untangle CV metrics.}
Dynamic checks are evaluated on every adjacent pair of sampled frames for
which the check is applicable.}
\label{tab:cv_metric_defs_pipe_untangle}
\end{table*}

\paragraph{Human validation.}
Manual review audits the CV evaluator on LTX-2.3 and Wan2.2-I2V-A14B videos.
The evaluator validation set is a balanced sample of $180$ videos, with $30$
videos in every task and generator cell. For each video, a single
reviewer inspects the video and detection overlays, then records whether the visible task
state is scorable, whether the CV parse matches it exactly or at least
partially, and whether the resulting automatic metric decision agrees with
manual review.

Panel A of Table~\ref{tab:cv_human_validation} shows that every sampled parse is
usable (partial or exact) and that all 180 video-level metric decisions agree
with the manual audit. These results support using the CV verifier to compute
the diagnostic pass rates in Table~\ref{tab:cv_metric_results}. Exact parses
reach 61.1\% overall. Panel B evaluates the terminal-state scorability gate,
which accepts 136 of the 138 human-scorable videos and rejects 4 of the 42
human-unscorable videos, for 77.8\% accuracy and 54.0\% balanced accuracy.

\begin{table*}[!t]
\centering
\scriptsize
\setlength{\tabcolsep}{3pt}
\renewcommand{\arraystretch}{1.08}

\begin{tabular*}{\textwidth}{@{\extracolsep{\fill}}l l r r r r r@{}}
\toprule
\multicolumn{7}{@{}l}{\textbf{Panel A. Parse quality and CV metric agreement by environment and generator}} \\
\midrule
\textbf{Environment} & \textbf{Generator} & \textbf{$N$} &
\textbf{Human-scorable view} & \textbf{Exact CV parse} &
\textbf{Usable CV parse} & \textbf{CV metric agreement} \\
\midrule
Pipe & LTX-2.3 & 30 & 76.7\% & 96.7\% & 100.0\% & 100.0\% \\
Pipe & Wan2.2-I2V-A14B & 30 & 100.0\% & 100.0\% & 100.0\% & 100.0\% \\
One Stroke & LTX-2.3 & 30 & 76.7\% & 43.3\% & 100.0\% & 100.0\% \\
One Stroke & Wan2.2-I2V-A14B & 30 & 100.0\% & 100.0\% & 100.0\% & 100.0\% \\
Untangle & LTX-2.3 & 30 & 6.7\% & 3.3\% & 100.0\% & 100.0\% \\
Untangle & Wan2.2-I2V-A14B & 30 & 100.0\% & 23.3\% & 100.0\% & 100.0\% \\
\midrule
\textbf{Overall} & \textbf{Both} & \textbf{180} & \textbf{76.7\%} &
\textbf{61.1\%} & \textbf{100.0\%} & \textbf{100.0\%} \\
\bottomrule
\end{tabular*}

\vspace{0.8em}

\renewcommand{\arraystretch}{1.12}
\begin{tabular*}{\textwidth}{@{\extracolsep{\fill}}p{0.38\textwidth}p{0.27\textwidth}p{0.27\textwidth}@{}}
\toprule
\multicolumn{3}{@{}l}{\textbf{Panel B. Human versus automatic terminal-state scorability}} \\
\midrule
 & \textbf{Human scorable} & \textbf{Human unscorable} \\
\midrule
\textbf{Parser scorable} & 75.6\% & 21.1\% \\
\textbf{Parser unscorable} & 1.1\% & 2.2\% \\
\midrule
\multicolumn{3}{@{}l}{\textbf{Derived gate metrics}} \\
\textbf{Accuracy} & \multicolumn{2}{l}{77.8\% (140/180 correctly classified videos)} \\
\textbf{Scorable recall} & \multicolumn{2}{l}{98.6\% (136/138 human-scorable videos accepted)} \\
\textbf{Unscorable recall} & \multicolumn{2}{l}{9.5\% (4/42 human-unscorable videos rejected)} \\
\textbf{Balanced accuracy} & \multicolumn{2}{l}{54.0\% (mean of scorable and unscorable recall)} \\
\bottomrule
\end{tabular*}

\caption{\textbf{Human audit of the computer-vision verifier and terminal-state scorability gate.}
A balanced set of 180 videos spans three environments, the LTX-2.3 and
Wan2.2-I2V-A14B generators, and three generator difficulty settings (10 per environment, generator, and difficulty setting).
\textbf{Panel A} reports parse quality and agreement between automatic
video-level metric decisions and manual review. A usable parse may be partial
but retains enough evidence for the diagnostic decision. \textbf{Panel B}
compares automatic terminal parser acceptance with the human judgment of whether
the visible state is scorable, and each cell is a percentage of all 180 videos.}
\label{tab:cv_human_validation}
\end{table*}

\subsection{Camera and Viewpoint Controls}
\label{app:camera}

\paragraph{Controlled camera factors.}
Static generators store the scene seed and render camera metadata, so the
same underlying topology can be rerendered under a top or oblique view, a wider
field of view, a tighter crop, or a paired second view. For planning
environments, the browser viewport is fixed and scene-only screenshots are
captured at each step. Table~\ref{tab:app_camera} summarizes what each
controlled camera factor changes.

\begin{table*}[!htbp]
\centering
\scriptsize
\renewcommand{\arraystretch}{1.12}
\begin{tabular*}{\textwidth}{@{\extracolsep{\fill}}p{0.20\textwidth}p{0.72\textwidth}@{}}
\toprule
\textbf{Camera factor} & \textbf{What the factor changes} \\
\midrule
Top vs.\ oblique & Changes the Euclidean projection while connectivity,
enclosure, and knot over/under relations stay fixed. \\
Field of view & Changes how much of the scene and its boundary segments
appears in the frame. \\
Zoom/crop & Changes the image scale of small gaps, thin walls, and subtle
through-holes. \\
Multi-view pair & Adds a second view that can expose depth or occlusion cues
missing from the first. \\
\bottomrule
\end{tabular*}
\caption{Controlled camera factors in scene generation. The same seed can be
rerendered under controlled view changes while keeping the topological ground
truth fixed.}
\label{tab:app_camera}
\end{table*}

\subsection{Shortcut and Input Representation Controls}
\label{app:shortcut_controls}

\paragraph{Protocol.}
We test whether performance depends on scene structure that determines the
answer or on cues that remain after this structure is removed. Full input
contains the rendered observation and the complete task prompt. Text-only
removes the rendered observation while retaining the question for each
instance. Answer-prior also removes the information specific to each question
and therefore measures the signal carried by the answer distribution.
Appearance-only retains visual style cues without the original scene
structure. Within each row, all conditions use the same instances and decoding
configuration. Table~\ref{tab:shortcut_controls} reports these controls for
the task and model pairs on which all three were evaluated. We additionally
probe the input representation with a symbolic condition that replaces the
rendering with a symbolic description of the same state.
Table~\ref{tab:symbolic_input} compares it with the full input on six tasks
for both models.

\begin{table*}[tbp]
\centering
\scriptsize
\setlength{\tabcolsep}{4.2pt}
\renewcommand{\arraystretch}{1.12}
\begin{tabular*}{\textwidth}{@{\extracolsep{\fill}}llrcccc@{}}
\toprule
& & & \textbf{Full} & \multicolumn{3}{c}{\textbf{Shortcut controls}} \\
\cmidrule(l){5-7}
\textbf{Task} & \textbf{Model} & \textbf{N} & \textbf{input} & \textbf{Text only} & \textbf{Answer prior} & \textbf{Appearance only} \\
\midrule
\multicolumn{7}{l}{\textit{Reasoning tasks}} \\
\multirow{2}{*}{2D Maze}
& Gemini-3.1-Flash-Lite & 100 & 21.00 & 8.00 & 8.00 & 36.00 \\
& InternVL3.5-241B & 100 & 10.00 & 14.00 & 8.00 & 6.00 \\
\addlinespace[2pt]
\multirow{2}{*}{Sheep}
& Gemini-3.1-Flash-Lite & 100 & 61.00 & 56.00 & 56.00 & 43.00 \\
& InternVL3.5-241B & 100 & 28.00 & 23.00 & 28.00 & 18.00 \\
\midrule
\multicolumn{7}{l}{\textit{Planning tasks}} \\
\multirow{2}{*}{Pipe}
& Gemini-3.1-Flash-Lite & 60 & 0.00 & 0.00 & 0.00 & 0.00 \\
& InternVL3.5-241B & 60 & 1.67 & 0.00 & 0.00 & 0.00 \\
\addlinespace[2pt]
\multirow{2}{*}{One Stroke}
& Gemini-3.1-Flash-Lite & 60 & 0.00 & 0.00 & 0.00 & 0.00 \\
& InternVL3.5-241B & 60 & 0.00 & 0.00 & 0.00 & 0.00 \\
\bottomrule
\end{tabular*}
\caption{Shortcut controls. Each row evaluates one model on N instances of one task, and all conditions in a row use the same instances. Full input denotes the rendered scene with the complete prompt. Values are accuracy (\%) for reasoning tasks and episode success rate (\%) for planning tasks.}
\label{tab:shortcut_controls}
\end{table*}

\begin{table*}[tbp]
\centering
\scriptsize
\setlength{\tabcolsep}{4.2pt}
\renewcommand{\arraystretch}{1.12}
\begin{tabular*}{\textwidth}{@{\extracolsep{\fill}}llrcc@{}}
\toprule
\textbf{Task} & \textbf{Model} & \textbf{N} & \textbf{Full input} & \textbf{Symbolic input} \\
\midrule
\multicolumn{5}{l}{\textit{Reasoning tasks}} \\
\multirow{2}{*}{2D Maze}
& Gemini-3.1-Flash-Lite & 100 & 21.00 & 27.00 \\
& InternVL3.5-241B & 100 & 10.00 & 7.00 \\
\addlinespace[2pt]
\multirow{2}{*}{Sheep}
& Gemini-3.1-Flash-Lite & 100 & 61.00 & 9.00 \\
& InternVL3.5-241B & 100 & 28.00 & 4.00 \\
\midrule
\multicolumn{5}{l}{\textit{Planning tasks}} \\
\multirow{2}{*}{Pipe}
& Gemini-3.1-Flash-Lite & 60 & 0.00 & 0.00 \\
& InternVL3.5-241B & 60 & 1.67 & 0.00 \\
\addlinespace[2pt]
\multirow{2}{*}{One Stroke}
& Gemini-3.1-Flash-Lite & 60 & 0.00 & 0.00 \\
& InternVL3.5-241B & 60 & 0.00 & 0.00 \\
\addlinespace[2pt]
\multirow{2}{*}{Swap}
& Gemini-3.1-Flash-Lite & 60 & 13.33 & 46.67 \\
& InternVL3.5-241B & 60 & 13.33 & 8.33 \\
\addlinespace[2pt]
\multirow{2}{*}{Chat Noir}
& Gemini-3.1-Flash-Lite & 60 & 25.00 & 40.00 \\
& InternVL3.5-241B & 60 & 6.67 & 3.33 \\
\bottomrule
\end{tabular*}
\caption{Symbolic input probe. The rendered scene is replaced by a symbolic description of the same state, and the rest of the prompt is fixed. Each row evaluates one model on N instances of one task, and both conditions in a row use the same instances. Values are accuracy (\%) for reasoning tasks and episode success rate (\%) for planning tasks.}
\label{tab:symbolic_input}
\end{table*}

\paragraph{Prompt and answer priors are concentrated in Sheep.}
Of the two reasoning tasks probed, Sheep carries the stronger prior.
Gemini-3.1-Flash-Lite scores $61.0$ with the full Sheep input and retains $56.0$
under both text-only and answer-prior input. InternVL3.5-241B shows the same
pattern. Its full-input score is $28.0$ and its answer-prior score is $28.0$.
Symbolic input lowers Sheep accuracy for both models, from $61.0$ to $9.0$ and
from $28.0$ to $4.0$. The 2D Maze results differ. Removing the scene lowers
Gemini-3.1-Flash-Lite from $21.0$ to $8.0$, while the appearance-only control
reaches $36.0$.

\paragraph{Explicit structure helps selected planning tasks.}
Symbolic input raises Gemini-3.1-Flash-Lite by $33.3$ percentage points on Swap
($13.33$ to $46.67$) and by $15.0$ points on Chat Noir ($25.00$ to $40.00$), while
InternVL3.5-241B gains on neither task.

\paragraph{Symbolic input does not lift Pipe or One Stroke.}
Pipe and One Stroke remain at zero or near zero under every tested input
condition, including symbolic input.

\subsection{Environment-Specific Error Signatures}
\label{app:biases}

Percentages in this subsection are shares of the labeled failures of the two
representative models described in Appendix~\ref{app:error_analysis}, pooled
over both models ($70$ failures per environment) and labeled with the
seven-category taxonomy of Appendix~\ref{app:error_categories}.

\paragraph{Connectivity grounding.}
In 2D and 3D Maze errors, adjacent rooms, bars, or cells are treated as
connected despite a blocking wall, and in planning runs illegal transitions
attempt to move through occupied, blocked, or out-of-bounds cells. Across the two Maze
reasoning environments, $52.9\%$ of labeled failures are perception-grounding
errors. Another $40.7\%$ are instruction-following failures, most of them
InternVL3.5-241B responses that are cut off before a committed answer. Pipe
planning shows the same pattern in an interactive setting, where $90.0\%$ of
labeled failures are perception-grounding errors in which end pieces are read
as straight pipes or the source is given openings it does not have, so legal
rotations are spent on a misread network.

\paragraph{Hole-type grounding.}
Most Hole failures are perception-grounding errors ($64.3\%$), in which a
depression, tunnel, or opening is read as the wrong kind of structure. Task
understanding follows ($20.0\%$), mostly counting an opening spanned by a bar
as a single hole, and then instruction following ($15.7\%$). None of the
labeled Hole errors receive the topological-invariance label.

\paragraph{Knot-structure classification.}
Knots is the only reasoning environment whose failures include
explicit topological-invariance labels, but these account for just $11.4\%$ of
labeled Knots failures. Perception grounding accounts for $52.9\%$, as models
miss crossings or segment one rope into several rings, and instruction
following for another $32.9\%$, often a final answer that contradicts the
model's own reasoning. Untangle has a different profile, with instruction
following accounting for $58.6\%$ of labeled failures, including every
InternVL3.5-241B failure, and action planning for $21.4\%$. Table~\ref{tab:app_biases} summarizes these
environment-specific error signatures.

\begin{table*}[!htbp]
\centering
\scriptsize
\renewcommand{\arraystretch}{1.12}
\begin{tabular*}{\textwidth}{@{\extracolsep{\fill}}>{\raggedright\arraybackslash}p{0.20\textwidth}>{\raggedright\arraybackslash}p{0.26\textwidth}>{\raggedright\arraybackslash}p{0.44\textwidth}@{}}
\toprule
\textbf{Error signature} & \textbf{Where it appears} & \textbf{Evidence} \\
\midrule
Connectivity grounding & 2D Maze, 3D Maze, Pipe & Perception grounding accounts
for $52.9\%$ of Maze and $90.0\%$ of Pipe failures. \\
Hole-type grounding & Hole & Perception grounding accounts
for $64.3\%$, and topological invariance accounts for $0.0\%$. \\
Knot-structure classification & Knots & Perception grounding accounts
for $52.9\%$, and topological invariance for $11.4\%$. \\
Environment-specific planning & Swap, One Stroke, Chat Noir, Untangle & The
dominant label shifts by environment, to task understanding ($67.1\%$), dynamic
violations ($38.6\%$), action planning ($58.6\%$), and instruction
following ($58.6\%$), respectively. \\
\bottomrule
\end{tabular*}
\caption{Environment-specific error signatures in the labeled failures.
Percentages are shares of the labeled failures of both models in each
environment, sampled as described in Appendix~\ref{app:error_analysis}.}
\label{tab:app_biases}
\end{table*}

\section{Error Analysis}
\label{app:error_analysis}

\subsection{Annotation Framework}
\label{app:error_methodology}

\paragraph{Motivation.}
\benchmarkname evaluates whether foundation models can ground visual scenes,
identify topological relations, predict state changes under
manipulation, and plan valid actions. A single split into perception and
reasoning is therefore too coarse, because an incorrect answer may come from a bad
output format, a mistaken task objective, a visual grounding failure, a
wrong belief about topological invariance, an incorrect final-state
prediction, an impossible transition, or a poor action plan. We use the
following seven primary categories to separate these failure sources.

\paragraph{Top-level categories.}
Table~\ref{tab:app_error_categories} summarizes the seven categories.
Full definitions, subcategories, and labeled examples for each category
appear in Appendix~\ref{app:error_categories}.

\begin{table}[!htbp]
\centering
\small
\renewcommand{\arraystretch}{1.12}
\begin{tabularx}{\linewidth}{@{}>{\raggedright\arraybackslash}p{0.36\linewidth}X@{}}
\toprule
\textbf{Category} & \textbf{Definition} \\
\midrule
0. Instruction Following (Appendix~\ref{app:err_0}) & The response is unparseable or violates the protocol. \\
1. Task Understanding (Appendix~\ref{app:err_1}) & The model solves the wrong task or applies the wrong rule. \\
2. Perception Grounding (Appendix~\ref{app:err_2}) & The response relies on a wrong visual fact. \\
3. Topological Invariance (Appendix~\ref{app:err_3}) & The model misjudges whether a relation survives an allowed transformation. \\
4. Feature State Prediction (Appendix~\ref{app:err_4}) & The model predicts a wrong post-action state of a relevant feature. \\
5. Dynamic (Appendix~\ref{app:err_5}) & The model assumes an impossible transition during motion. \\
6. Action Planning (Appendix~\ref{app:err_6}) & The model selects a poor legal action or multi-step plan. \\
\bottomrule
\end{tabularx}
\caption{The seven primary error categories.}
\label{tab:app_error_categories}
\end{table}

\subsection{Error Category Definitions}
\label{app:error_categories}
\subsubsection{0. Instruction Following Error}
\label{app:err_0}

The model fails to provide a response that can be evaluated under the
prompt's requested protocol. This category captures answer-format and
response-compliance failures rather than visual or topological
capability failures. We assign category~0 when the fallback parser cannot
recover a single valid answer, when the response breaks an explicit output
constraint (0c), or when the parsed answer contradicts the model's own stated
conclusion (0d). Other parseable predictions are analyzed under
categories~1--6.

\paragraph{0a. Invalid Answer Format.}
The output cannot be parsed into the required type or schema.
Typical cases include prose returned when the prompt asks for
\texttt{\{"answer": 3\}}, a missing JSON key, a cell index returned as a
sentence, and a list returned when a scalar is required.

\paragraph{0b. Multiple, Hedged, or Missing Final Answer.}
The task expects one answer but the model gives several candidates,
hedges between options, or never states a final answer.
Typical cases include returning ``A or C'' for a multiple-choice task,
listing two candidate moves, and explaining the image without an answer.

\paragraph{0c. Prompt-Protocol Violation.}
The model ignores explicit output constraints even though it appears to
understand the broad task.
Typical cases include emitting chain-of-thought when only the final
answer is requested, using a free-form action when the prompt requires a
fixed enum, and adding extra fields that break the evaluator.

\paragraph{0d. CoT/Output Internal Inconsistency.}
The chain-of-thought reaches one conclusion but the structured answer
field reports a different value, so the parser commits a verdict that
contradicts the model's own stated reasoning. This is distinct from
0a, 0b, and 0c because the response is well-formed and a single answer is
emitted. The failure lies purely in the link between rationale and final
field.
Typical cases include a fence-repair trace concluding ``there are zero
gaps; done'' followed by an answer field of $1$, a hole-count
walk-through that enumerates three holes followed by an answer field of
$4$, and a knot-classification thought block stating ``this is an unknot''
followed by the knot label B.

\paragraph{0e. Output-Budget Truncation.}
The model devotes its output budget to deliberation and is cut off
before any structured answer is produced. We assign this label only when
the fallback parser also fails to recover an answer from the available text. This
differs from 0c (intentional protocol violation), because the model is
trying to comply but never reaches the answer.
Typical cases include a long endpoint enumeration on an Untangle step
that hits the response cap mid-sentence and a multi-hundred-word
deliberation on a separation question that ends without a final letter.

\subsubsection{1. Task Understanding Error}
\label{app:err_1}

The response is syntactically usable, but the model has misunderstood
the task objective, the permitted operation, or the relevant rule set.
This category differs from category~0, because the answer may be
parseable but is generated for the wrong problem.

\paragraph{1a. Wrong Objective.}
The model optimizes or answers the wrong target.
Typical cases include counting objects when the task asks whether an
object is enclosed, reporting the current state when the prompt asks for
the post-action state, and selecting the visually closest option when the
task requires topological equivalence.

\paragraph{1b. Wrong Rule Semantics.}
The model applies an incorrect rule for the environment.
Typical cases include treating diagonal grid contact as connectivity
when only four-neighbor connectivity is allowed, assuming a pipe piece can
connect through a closed side, and misunderstanding whether a rope can be
lifted over another rope in the specified setting.

\paragraph{1c. Wrong Entity or Reference Binding.}
The model answers for a different referenced object or option than the
one requested by the prompt.
Typical cases include comparing option B to the target when the prompt
asks about option C, moving the red rope when the instruction refers to the
blue rope, and using the goal state as the current state in Swap.

\paragraph{1d. Multi-Turn Task Drift.}
Specific to interactive tasks. The model identifies the task correctly
on the first turn, but in subsequent turns its reasoning context drifts
to a different problem framing, causing all later decisions to optimize
the wrong objective. It differs from 1a and 1b because the failure is not a
one-shot misreading of the prompt but a loss of task identity across
turns despite the original instructions still being supplied.
Typical cases include a One Stroke planner recognizing the rules at
step 0 and then describing the same board as a robotic-arm pick-and-place
scene from step 1 onward, a cat-and-mouse encloser switching to
hypothesizing prime-number patterns in the cell labels by the second move,
and a Pipe puzzle reframed as ``Unblock Me'' or as a sliding-block puzzle
once intermediate states are shown.

\subsubsection{2. Perception Grounding Error}
\label{app:err_2}

The model fails at the visual-grounding stage. The relevant task is
understood, but the response relies on a wrong visual fact.

\paragraph{2a. Object, Color, or Attribute Misidentification.}
A visible element is detected but its identity or attribute is wrong.
Typical cases include confusing the blue and red ropes, misreading a
through-hole as filled, treating a blocked cell as free, and mistaking an
open pipe side for a closed side.

\paragraph{2b. Location or Relative-Position Grounding Error.}
The model misreads where elements are or how they are positioned
relative to each other.
Typical cases include reversing the relative position of A and B,
assigning an object to the wrong grid cell, missing that one endpoint lies
inside a loop, and confusing left/right or above/below relations in the
rendered scene.

\paragraph{2c. Missed or Hallucinated Visual Element.}
A relevant element is absent from or added to the model's internal scene
representation, including fine-grained material or depth cues that
distinguish a real opening from a surface marking.
Typical cases include missing a hole, sheep, rope crossing, wall
segment, or pipe piece. Others hallucinate an extra obstacle or connection,
mistake a printed symbol or shaded depression for a through-hole, or
misread interior shading as a solid floor.

\paragraph{2d. Multi-Image Correspondence Error.}
The model parses individual images but fails to align corresponding
objects, views, or states.
Typical cases include mismatching the current and goal states in Swap,
aligning the wrong point across 3D Maze views, and confusing a target
object with a candidate option in separation tasks.

\subsubsection{3. Topological Invariance Understanding Error}
\label{app:err_3}

The model grounds the scene, but misjudges whether the relevant
topological relation is preserved or changed. This category targets
topological concepts such as connectivity, enclosure, separation,
overlap, crossing, linking, and inside/outside relations under
transformations. A typical failure relies on visible coordinates or shape
similarity and misses that topology has changed, or infers a topological
change from a purely geometric deformation.

\paragraph{3a. False Invariance.}
The model says a topological relation is unchanged when it has changed.
A typical case concludes that two configurations are equivalent because
endpoints or object positions look unchanged, even though a rope
crossing, overlap, enclosure, or connectivity relation has changed.

\paragraph{3b. False Change.}
The model says a topological relation changes under a transformation
that preserves it.
A typical case treats translation, rotation, stretching, or a viewpoint
change as breaking connectivity, enclosure, or linkage when the
topological relation remains the same.

\paragraph{3c. Wrong Topological Relation.}
The model reasons about the correct objects but assigns the wrong
topological relation.
Typical cases include linked vs.\ unlinked ropes, connected vs.\
separated pipe components, inside vs.\ outside a loop, and trapped vs.\
reachable in a maze-like enclosure.

\subsubsection{4. Feature State Prediction Error}
\label{app:err_4}

The model understands the instruction and initial scene, but predicts
the wrong final state of a feature after a specified manipulation. This
category concerns the resulting state, not whether the intermediate
motion was physically possible. For example, after the prompt describes
lifting a blue rope and placing it in the upper-left region, the model
may predict that the blue rope no longer overlaps the red rope, even
though the ground-truth final state still contains an overlap.

\paragraph{4a. Wrong Post-Action Topological Feature.}
The predicted final connectivity, overlap, crossing, enclosure, or
separation state is wrong.
Typical cases include predicting that a moved rope no longer overlaps
another rope when it still does, predicting that a rotated pipe network is
connected when one joint remains disconnected, and predicting that a
blocked cell traps the cat when an escape route remains.

\paragraph{4b. Wrong Post-Action Visual or Discrete State.}
The model applies the action to the wrong state variable.
Typical cases include a wrong post-rotation orientation of a pipe cell,
a wrong final slot after a swap, a wrong cell status after placing a block,
and a wrong endpoint location after a move.

\paragraph{4c. Incomplete State Update.}
The model updates one feature but leaves another dependent feature in
the old state.
Typical cases include moving a rope endpoint but not updating
crossings, rotating a pipe cell but not updating which cells become
source-connected, and blocking a cell but not updating reachability.

\subsubsection{5. Dynamic Error}
\label{app:err_5}

The model assumes an impossible intermediate transition during motion.
Unlike category~4, which concerns the final state after an action,
category~5 concerns whether the path from the initial state to the final
state violates the environment's dynamics or physical constraints.

\paragraph{5a. Barrier or Collision Violation.}
The model moves through an obstacle, wall, boundary, or occupied cell.
Typical cases include planning a maze step through a wall, sliding a
block through another block, routing a path through a forbidden cell, and
moving the cat through a blocked cell.

\paragraph{5b. Topological Barrier Violation.}
The model permits a transition that would require crossing, cutting, or
teleporting through a topological constraint that the task forbids.
Typical cases include passing a rope through another rope without an
allowed lift, untangling a knot by crossing strands through each other, and
separating linked components through an impossible motion.

\paragraph{5c. Unsupported Continuous Motion.}
The model predicts a valid-looking final arrangement but provides or
assumes no legal continuous path to reach it.
Typical cases include moving an object from one enclosed region to
another without crossing a boundary, rotating a component through a
collision, and placing a feature behind an occluding barrier as if it could
pass through it.

\subsubsection{6. Action Planning Error}
\label{app:err_6}

Specific to interactive tasks. The model follows the output protocol,
understands the task, grounds the relevant scene facts, and does not
assume an impossible transition, but still chooses a poor legal action
or action sequence.

\paragraph{6a. Legal but Irrelevant Action.}
The action is allowed but does not advance the objective.
Typical cases include rotating a non-critical pipe piece, blocking a
cell that is not on any escape route, moving a puzzle tile that leaves the
state equally far from the goal, and adjusting a rope segment that does
not reduce crossings.

\paragraph{6b. Short-Horizon or Trap-Inducing Plan.}
The action appears locally reasonable but causes later failure.
Typical cases include blocking a cell that lets the cat escape
elsewhere, forming a One Stroke path that cuts off an unvisited region,
making a swap that increases the minimum remaining distance, and
untangling one crossing while creating a worse downstream crossing.

\paragraph{6c. State-Tracking Planning Error.}
The model chooses a later action based on a stale but otherwise valid
state estimate.
Typical cases include forgetting a previously rotated pipe piece,
treating an already blocked cell as free, planning from the initial rather
than current puzzle state, and repeating an action that has already been
applied.

\subsection{Cross-Model and Cross-Task Distributions}
\label{app:error_distributions}

\paragraph{Sampling protocol.}
For each (model, task) pair we collected every incorrect prediction of the
model on the $300$ benchmark instances behind its cell in
Table~\ref{tab:topology_benchmark_refreshed}, $150$ on the Easy level and $150$
on the Hard level, using the same runs and the same scoring as that table. We
then drew $35$ of these failures uniformly at random with a fixed seed and
assigned each sampled prediction one primary category. Labels follow the
causal priority order instruction following (0), task understanding (1),
perception grounding (2), dynamic violation (5), feature-state prediction (4),
topological invariance (3), and action planning (6), so a failure receives the
earliest category in this order that explains it. Every pair has at least $175$ failures, more than the $35$ sampled, so each pair
contributes exactly $35$ labeled failures. This yields $560$ labeled reasoning
failures and $350$ labeled planning failures, of which $415$ lie on the Easy
level and $495$ on the Hard level. The reasoning split contains the 8
reasoning tasks (2D Maze, 3D Maze, Hole, Sheep, Knots, Bead, Origami
Point, and Assembly), and the planning split contains the 5 planning tasks
(Pipe, Chat Noir, Untangle, Swap, and One Stroke).

\begin{figure}[!htbp]
    \centering
    \includegraphics[width=0.78\linewidth]{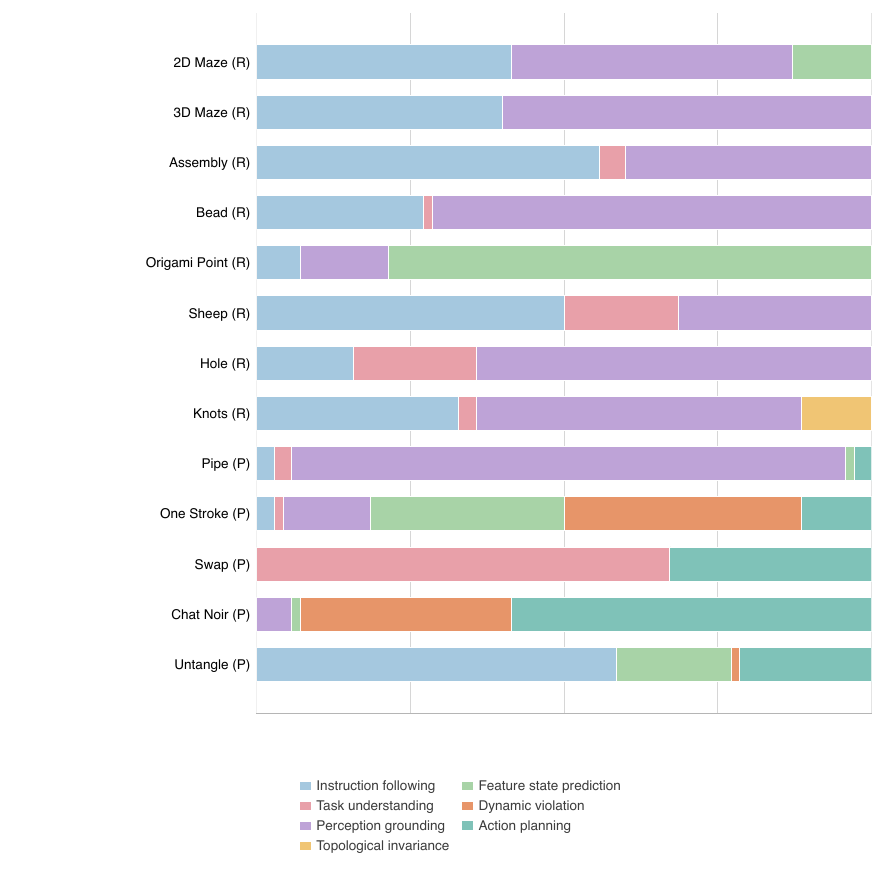}
    \caption{\textbf{Environment-level error distributions.}
    Each bar pools the $35$ labeled failures of Gemini-3.1-Pro and the $35$
    labeled failures of InternVL3.5-241B on one environment and shows the
    share of these $70$ failures assigned to each error category.}
    \label{fig:app_error_by_environment}
\end{figure}

\begin{figure}[!htbp]
    \centering
    \includegraphics[width=0.95\linewidth]{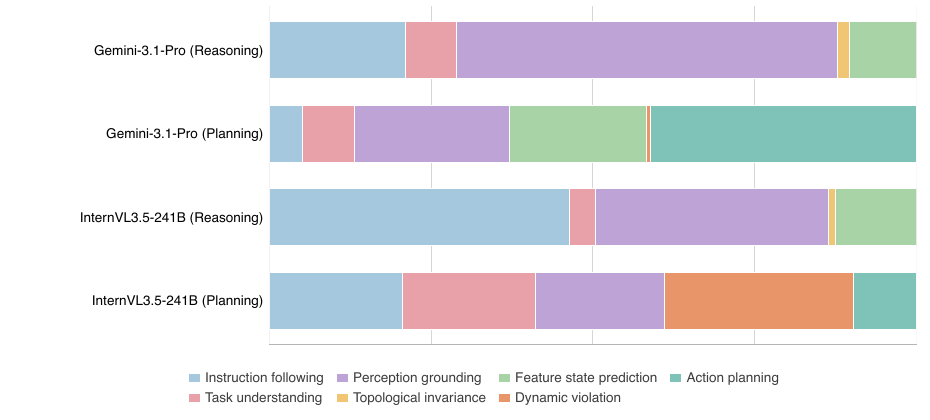}
    \caption{\textbf{Model-level error distributions across reasoning and planning.}
    Each bar shows the category shares of the labeled failures of one model
    on one split, which number $280$ for reasoning and $175$ for planning.}
    \label{fig:app_error_by_model}
\end{figure}

\paragraph{Aggregation.}
Every distribution in this appendix and in Section~\ref{sec:error_analysis}
is a share of the labeled failures. A distribution for a group of pairs, such
as one model on one split or one environment across both models, pools the
labeled failures of those pairs, so each pair enters with the same weight of
$35$ failures. Figures~\ref{fig:app_error_by_environment}
and~\ref{fig:app_error_by_model} provide the environment-level and
model-level breakdowns, and Tables~\ref{tab:app_err_perc_split}
and~\ref{tab:app_err_plan_split} give the labeled-sample counts for each model
and for both models combined.

\paragraph{Per-environment breakdown.}
Within a property, the dominant category can differ between tasks
(Figure~\ref{fig:app_error_by_environment}), for example feature-state
prediction in Origami Point ($78.6\%$) and task understanding in Swap
($67.1\%$), or perception grounding in Hole ($64.3\%$) and action planning in
Chat Noir ($58.6\%$).

\begin{table*}[ht]
\centering
\scriptsize
\begin{tabular*}{\textwidth}{@{\extracolsep{\fill}}lrrrrrr@{}}
\toprule
& \multicolumn{2}{c}{Gemini-3.1-Pro} & \multicolumn{2}{c}{InternVL3.5-241B} & \multicolumn{2}{c}{Combined} \\
\cmidrule(lr){2-3} \cmidrule(lr){4-5} \cmidrule(l){6-7}
Category & count & \% & count & \% & count & \% \\
\midrule
0. Instruction following     &   59 & 21.1 &  130 & 46.4 &  189 & 33.8 \\
1. Task understanding        &   22 &  7.9 &   11 &  3.9 &   33 &  5.9 \\
2. Perception grounding      &  165 & 58.9 &  101 & 36.1 &  266 & 47.5 \\
3. Topological invariance    &    5 &  1.8 &    3 &  1.1 &    8 &  1.4 \\
4. Feature state prediction  &   29 & 10.4 &   35 & 12.5 &   64 & 11.4 \\
5. Dynamic violation         &    0 &  0.0 &    0 &  0.0 &    0 &  0.0 \\
6. Action planning           &    0 &  0.0 &    0 &  0.0 &    0 &  0.0 \\
\midrule
Labeled failures             &  280 & 100 &  280 & 100 &  560 & 100 \\
\bottomrule
\end{tabular*}
\caption{Category distribution of the labeled reasoning failures. Each model contributes $35$ labeled failures on each of the $8$ reasoning tasks, and the combined columns pool the $560$ labeled failures of both models. Percentages are shares of the labeled failures in each column and may not sum to exactly $100$ because of rounding.}
\label{tab:app_err_perc_split}
\end{table*}

\begin{table*}[ht]
\centering
\scriptsize
\begin{tabular*}{\textwidth}{@{\extracolsep{\fill}}lrrrrrr@{}}
\toprule
& \multicolumn{2}{c}{Gemini-3.1-Pro} & \multicolumn{2}{c}{InternVL3.5-241B} & \multicolumn{2}{c}{Combined} \\
\cmidrule(lr){2-3} \cmidrule(lr){4-5} \cmidrule(l){6-7}
Category & count & \% & count & \% & count & \% \\
\midrule
0. Instruction following     &    9 &  5.1 &   36 & 20.6 &   45 & 12.9 \\
1. Task understanding        &   14 &  8.0 &   36 & 20.6 &   50 & 14.3 \\
2. Perception grounding      &   42 & 24.0 &   35 & 20.0 &   77 & 22.0 \\
3. Topological invariance    &    0 &  0.0 &    0 &  0.0 &    0 &  0.0 \\
4. Feature state prediction  &   37 & 21.1 &    0 &  0.0 &   37 & 10.6 \\
5. Dynamic violation         &    1 &  0.6 &   51 & 29.1 &   52 & 14.9 \\
6. Action planning           &   72 & 41.1 &   17 &  9.7 &   89 & 25.4 \\
\midrule
Labeled failures             &  175 & 100 &  175 & 100 &  350 & 100 \\
\bottomrule
\end{tabular*}
\caption{Category distribution of the labeled planning failures. Each model contributes $35$ labeled failures on each of the $5$ planning tasks, and the combined columns pool the $350$ labeled failures of both models. Percentages are computed as in Table~\ref{tab:app_err_perc_split}.}
\label{tab:app_err_plan_split}
\end{table*}

\section{Extended Related Work Discussion}
\label{app:extended_rw}

This appendix expands the Related Work in Section~\ref{sec:related} with a per-work discussion. Table~\ref{tab:comparison} compares \benchmarkname{} with related benchmarks.

\definecolor{groupgeneral}{HTML}{FFF2E6}
\definecolor{groupeuclid}{HTML}{EEF3FF}
\definecolor{grouptopo}{HTML}{E9F7EF}

\definecolor{tabgreen}{HTML}{1E8449}
\definecolor{tabred}{HTML}{C0392B}
\newcommand{\ymark}{\textcolor{tabgreen}{\ding{51}}}
\newcommand{\nmark}{\textcolor{tabred}{\ding{55}}}
\newcommand{\cmpgroup}[2]{%
  \addlinespace[2pt]%
  \multicolumn{10}{>{\columncolor{#1}}l}{\textit{#2}}\\[1pt]%
}

\begin{table*}[ht]
\centering
\resulttablefont
\setlength{\tabcolsep}{3pt}
\renewcommand{\arraystretch}{1.15}
\begin{tabular*}{\textwidth}{@{\extracolsep{\fill}}l l c c c c c c c c@{}}
\toprule
\textbf{Benchmark} & \textbf{Type} & \textbf{\#Topo.} & \textbf{\#Tasks} & \textbf{Size} & \textbf{QA} & \textbf{Inter.} & \textbf{Cog.} & \textbf{Diff.} & \textbf{Scale} \\
\midrule
\cmpgroup{groupgeneral}{General visual and abstract reasoning}
BlindTest$^{\dag}$~\cite{rahmanzadehgervi2024vision} & Visual perception & -- & 7   & 4,860       & \ymark & \nmark & \nmark & \ymark & \ymark \\
GameQA~\cite{tong2026gamerl}                         & Game reasoning    & -- & 158 & $\sim$140K & \ymark & \nmark & \nmark & \ymark & \ymark \\
PuzzleVQA~\cite{chia2024puzzlevqa}                   & Abstract patterns & -- & 20  & 2,000       & \ymark & \nmark & \ymark & \nmark & \ymark \\
\cmpgroup{groupeuclid}{Euclidean spatial reasoning}
CausalSpatial~\cite{ma2026causalspatial}             & Causal spatial    & -- & 4  & 1,012  & \ymark & \nmark & \nmark & \ymark & \ymark \\
3DSRBench~\cite{ma20253dsrbench}                     & Euclidean         & -- & 12 & 2,772  & \ymark & \nmark & \nmark & \nmark & \nmark \\
iVISPAR~\cite{mayer2025ivispar}                      & Interactive spatial & -- & 1 & 300 & \nmark & \ymark & \nmark & \ymark & \ymark \\
Mind the Gap~\cite{stogiannidis2025mind}             & Euclidean         & -- & 6  & 1,800  & \ymark & \nmark & \ymark & \ymark & \ymark \\
MindCube~\cite{wang2025mindcube}                     & Euclidean         & -- & 3  & 21,154 & \ymark & \nmark & \ymark & \nmark & \nmark \\
OmniSpatial$^{\dag}$~\cite{jia2025omnispatial}       & Euclidean         & -- & 50 & 8,400  & \ymark & \nmark & \ymark & \nmark & \nmark \\
SITE~\cite{wang2025site}                             & Euclidean         & -- & 6  & 8,068  & \ymark & \nmark & \ymark & \nmark & \nmark \\
SpatialMQA~\cite{liu2025can}                         & Euclidean         & -- & 6  & 5,392  & \ymark & \nmark & \nmark & \nmark & \nmark \\
SpatialVLM~\cite{chen2024spatialvlm}                 & Euclidean         & -- & 2  & 546    & \ymark & \nmark & \nmark & \nmark & \ymark \\
SpatialWorld~\cite{gao2026spatialworld}              & Interactive spatial & -- & 6 & 760 & \nmark & \ymark & \nmark & \ymark & \nmark \\
VSI-Bench~\cite{yang2025thinking}                    & Euclidean         & -- & 8  & 5,130  & \ymark & \nmark & \ymark & \nmark & \nmark \\
\cmpgroup{grouptopo}{Topological spatial reasoning}
CurveBench~\cite{mohseni2026curvebench}              & Topological       & 2 & 1 & 756 & \nmark & \nmark & \nmark & \ymark & \nmark \\
KnotGym~\cite{chen2025knot}                          & Topological       & 1 & 3 & Proc.  & \nmark & \ymark & \nmark & \ymark & \ymark \\
TopoBench~\cite{maniparambil2026topobench}          & Topological & 3 & 6 & 900 & \nmark & \nmark & \nmark & \ymark & \ymark \\
\midrule
\textbf{\benchmarkname~Core Set (ours)}                              & \textbf{Topological}       & \textbf{5} & \textbf{13} & \textbf{3,900} & \ymark & \ymark & \ymark & \ymark & \ymark \\
\bottomrule
\end{tabular*}
\caption{\textbf{Benchmark comparison.}
\#Topo.\ counts topological properties, Inter.\ marks interactive evaluation, Cog.\ marks an explicit human-cognition framework, Diff.\ marks difficulty levels, Scale marks automatic generation, and Proc.\ denotes procedural episodes.
$\dag$~Includes topology without making it the main focus.}
\label{tab:comparison}
\end{table*}

\noindent\textbf{Topological Cognition.}
Cognitive development positions topology as a primitive layer of spatial cognition that precedes projective and Euclidean reasoning. Piaget and Inhelder~\cite{piaget2013child} introduced the original five-class taxonomy of proximity, separation, order, enclosure, and continuity, with subsequent mathematical and constructionist extensions by Beth and Piaget~\cite{piaget1966mathematical} and Papert~\cite{papert1980mindstorms}. Adult perception shows a related sensitivity. Chen~\cite{chen1982topological,chen2005topological} showed that the visual system extracts topological invariants such as connectedness and number of holes ahead of feature-based attributes. Holes have been formalized as dependent entities by Casati and Varzi~\cite{casati1994holes}, with perceptual evidence on surrounded regions and hole shape from Nelson and Palmer~\cite{nelson2001holes} and Bertamini and Croucher~\cite{bertamini2003shape}. Knots constitute a distinct cognitive domain, characterized by Strohecker~\cite{strohecker1991knot} as the ``mother structure'' that coordinates the other relations, and shown by Croom and Firestone~\cite{croom2024tangled} to strain adult intuitive physics. Hatcher~\cite{hatcher2002algebraic} provides the algebraic-topology formalization, Martin~\cite{martin1976analysis} clarifies how Piagetian tasks map onto formal topological constructs, and G\"ardenfors~\cite{gardenfors2000conceptual} argues that natural categories rest on topological notions such as connectedness and convexity. \benchmarkname~translates this body of evidence into a unified visual taxonomy and asks whether modern foundation models exhibit the same topological primacy.

\noindent\textbf{Cognitively Grounded and Topology-Adjacent Benchmarks.}
A closer line of work either revisits developmental psychology or tests topology through controlled spatial tasks. Piaget-inspired benchmarks diagnose developmental gaps in foundation models. CogDevelop2K~\cite{li2024cogdevelop2k} and CogLM~\cite{wang2025coglm} stage tests of reversed cognitive development. ConserveBench~\cite{luo2024vision} and PerspectBench~\cite{gao2024vision} target conservation and perspective-taking. KiVA~\cite{yiu2024kiva} and BabyVision~\cite{chen2026babyvision} extend the developmental lens to visual analogy and pre-linguistic vision. These benchmarks evaluate abilities adjacent to rather than organized around topological invariants. Individual spatial tasks include path connectivity in AlphaMaze~\cite{dao2025alphamaze}, knots in KnotGym~\cite{chen2025knot}, folding in ORIGAMISPACE~\cite{xu2025origamispace} and GamiBench~\cite{spencer2025gamibench}, abstract positional reasoning in OPTiCAL~\cite{driggers2025optical}, and constrained-manifold puzzles in Thinking in Structures~\cite{yang2026thinking}. Related work also probes how vision models respond to topological and geometric concepts~\cite{wang2025computer}.

TopoBench is the closest recent benchmark of global topology-focused constraint solving~\cite{maniparambil2026topobench}. It contains 900 instances from six grid-puzzle families at three difficulty levels. The families cover path and network connectivity, loop closure, region partitioning under rotational symmetry, reflection-based visibility, and contiguity across intersecting axes. Puzzle-specific verifiers score complete solution grids. An analysis of 750 reasoning traces identifies recurring errors, and controlled interventions show that premature commitment and constraint forgetting directly reduce downstream accuracy. Additional experiments with cell-aligned representations and structured constraint tools attribute much of the remaining difficulty to extracting constraints from spatial representations. TopoBench presents symbolic grid inputs to reasoning LLMs and does not evaluate visual perception or closed-loop action. \benchmarkname~instead grounds four Piagetian properties and knots in rendered scenes and pairs reasoning questions with interactive planning environments.

\noindent\textbf{Spatial Reasoning Benchmarks for MLLMs.}
Most spatial-reasoning benchmarks for multimodal large language models target Euclidean and viewpoint-centric competence rather than topological invariance. Static VQA benchmarks measure metric properties such as distance, direction, and inter-object relations. SpatialVLM~\cite{chen2024spatialvlm} and SpatialMQA~\cite{liu2025can} ground this line. Broader aggregations including OmniSpatial~\cite{jia2025omnispatial}, SITE~\cite{wang2025site}, 3DSRBench~\cite{ma20253dsrbench}, SPATIAL-DISE~\cite{huang2025spatial}, and Mind the Gap~\cite{stogiannidis2025mind} expand coverage but stay within a metric grammar. CausalSpatial moves from static relations to object-centric causal anticipation~\cite{ma2026causalspatial}. Its 1,012 multiple-choice examples cover collision, occlusion, compatibility, and trajectory at two difficulty levels. Each question pairs a rendered 3D scene with a hypothetical motion that changes one or more objects. The accompanying Causal Object World model derives object trajectories and renders videos that provide additional evidence to the MLLM. CausalSpatial therefore tests whether a model can predict the visual consequence of a specified intervention, while its fixed-question protocol does not test closed-loop action selection. Cognitive-map and viewpoint-integration benchmarks such as Thinking in Space~\cite{yang2025thinking}, MindCube~\cite{wang2025mindcube}, and Theory of Space~\cite{zhang2026theory} probe spatial memory and limited-view inference. Interactive evaluations including iVISPAR~\cite{mayer2025ivispar}, VSP~\cite{wu2024vsp}, and RoboSpatial~\cite{song2025robospatial} test multi-step spatial control and affordance use.

SpatialWorld extends interactive spatial evaluation across heterogeneous environments~\cite{gao2026spatialworld}. It contains 760 human-annotated tasks from six scenario categories that are instantiated in eight simulation backends. Agents receive only egocentric RGB observations under partial observability and express high-level decisions through a shared text-based action interface. Each task includes a human-validated initial state, a reference trajectory, and a terminal-state verifier. The benchmark evaluates 15 multimodal agents, with the strongest reaching a task success rate of 17.4\%. Its categories organize application domains and action complexity rather than topological invariants. SpatialWorld therefore tests broad cross-environment interaction, while \benchmarkname~pairs reasoning and planning under a five-property topological taxonomy. Neither SpatialWorld nor the other spatial benchmarks isolate invariants under continuous deformation. Their tasks hinge on quantitative geometry, viewpoint, intervention outcomes, or application goals rather than qualitative spatial structure.

Mechanistic studies offer a complementary account of spatial reasoning failures. Attention analysis finds that errors in spatial relation questions often coincide with attention directed toward irrelevant objects~\cite{chen2025spatialattention}. The resulting ADAPTVIS method adjusts visual attention using model confidence. This work motivates examining visual grounding alongside task accuracy.

\noindent\textbf{World Models and Physical Reasoning.}
A separate line of work evaluates MLLMs and video models as world simulators or physical reasoners, with focus on physical fidelity rather than topology preservation. World-model benchmarks such as PhyGenBench~\cite{meng2024towards}, Physion-Eval~\cite{zhang2026physion}, EWMBench~\cite{yue2025ewmbench}, WorldArena~\cite{shang2026worldarena}, World-in-World~\cite{zhang2025world}, and PhysicsMind~\cite{mak2026physicsmind} score image and video rollouts on commonsense and dynamics realism. Reasoning-oriented evaluations including MMGR~\cite{cai2025mmgr}, TiViBench~\cite{chen2025tivibench}, and VideoThinkBench~\cite{tong2025thinking} extend this view to multi-modal generative reasoning, and ENACT~\cite{wang2025enact} casts embodied cognition as forward and inverse world modeling from egocentric interaction. Adjacent agentic benchmarks such as PhysBench~\cite{chow2025physbench}, DeepPHY~\cite{xu2026deepphy}, and CHAIN~\cite{wu2026perception} emphasize physics- and affordance-level reasoning. Phenomena such as objects passing through holes, surfaces being cut or sealed, and ropes tightening or untangling either appear only incidentally or are scored by metric and visual fidelity rather than by invariance under continuous deformation. \benchmarkname~complements this line with topology-grounded probes that score topology preservation separately from physical realism.

WorldAgen connects world modeling directly to action prediction through a shared model and adapts to new environments by learning from exploratory transitions at test time~\cite{wan2026worldagen}. Its evaluation on robotic manipulation tasks addresses how improved dynamics prediction supports control. Our rollout audit examines a complementary requirement by testing whether generated state transitions obey the structural constraints of each environment.

\section{Prompt Templates}
\label{app:prompts}

\subsection{Property-Level Prompt Templates}
\label{app:prompt_templates}

Representative property-level templates are shown below.

\begingroup
\setlength{\intextsep}{4pt plus 1pt minus 1pt}

\begin{figure}[H]
\centering
\begin{tcolorbox}[width=\linewidth,colback=black!2,colframe=black!20,boxrule=0.4pt,arc=2pt,left=4pt,right=4pt,top=4pt,bottom=4pt]
\footnotesize
\texttt{[Task]} You are solving a continuity task. Determine which
locations lie in the same connected component, or which single
manipulation reconnects two disconnected locations.\\
\texttt{[Rules]} Use only the provided image(s). Do not cross walls,
closed doors, blocking bars, or obstacle boundaries.\\
\texttt{[Question]} Return the requested reachable labels, removable
bars, or connection decision.\\
\texttt{[Answer Format]} Output exactly one JSON object:
\texttt{\{"answer": value\}}.
\end{tcolorbox}
\caption{Prompt for \emph{Continuity} tasks.}
\label{fig:app_prompt_continuity}
\end{figure}

\begin{figure}[H]
\centering
\begin{tcolorbox}[width=\linewidth,colback=black!2,colframe=black!20,boxrule=0.4pt,arc=2pt,left=4pt,right=4pt,top=4pt,bottom=4pt]
\footnotesize
\texttt{[Task]} You are solving a separation task. Identify whether
visible parts belong to one connected object/subassembly, or choose a
legal next action that keeps a single continuous stroke valid.\\
\texttt{[Rules]} Treat contact, overlap, and connected geometry according
to the task definition. Do not infer hidden connectors.\\
\texttt{[Question]} Select the matching option or next move.\\
\texttt{[Answer Format]} Output exactly one JSON object:
\texttt{\{"answer": value\}}.
\end{tcolorbox}
\caption{Prompt for \emph{Separation} tasks.}
\label{fig:app_prompt_separation}
\end{figure}

\begin{figure}[H]
\centering
\begin{tcolorbox}[width=\linewidth,colback=black!2,colframe=black!20,boxrule=0.4pt,arc=2pt,left=4pt,right=4pt,top=4pt,bottom=4pt]
\footnotesize
\texttt{[Task]} You are solving an order task. Read the sequence along a
continuous carrier, track point labels through a fold sequence, or plan
swaps that transform the current order into the goal order.\\
\texttt{[Rules]} Order is defined along the string, paper surface, or
puzzle track rather than by raw image coordinates.\\
\texttt{[Question]} Return the requested sequence, relation, visible
point set, or next action.\\
\texttt{[Answer Format]} Output exactly one JSON object:
\texttt{\{"answer": value\}}.
\end{tcolorbox}
\caption{Prompt for \emph{Order} tasks.}
\label{fig:app_prompt_order}
\end{figure}

\begin{figure}[H]
\centering
\begin{tcolorbox}[width=\linewidth,colback=black!2,colframe=black!20,boxrule=0.4pt,arc=2pt,left=4pt,right=4pt,top=4pt,bottom=4pt]
\footnotesize
\texttt{[Task]} You are solving an enclosure task. Decide which objects
are inside a closed boundary, whether an object can escape through gaps,
how many through-holes a solid has, or where to place a blocking action.\\
\texttt{[Rules]} A region is enclosed only when every continuous path to
the outside crosses a boundary. A hole must pass through the object.\\
\texttt{[Question]} Return the requested count, label set, repair choice,
or action.\\
\texttt{[Answer Format]} Output exactly one JSON object:
\texttt{\{"answer": value\}}.
\end{tcolorbox}
\caption{Prompt for \emph{Enclosure} tasks.}
\label{fig:app_prompt_enclosure}
\end{figure}

\begin{figure}[H]
\centering
\begin{tcolorbox}[width=\linewidth,colback=black!2,colframe=black!20,boxrule=0.4pt,arc=2pt,left=4pt,right=4pt,top=4pt,bottom=4pt]
\footnotesize
\texttt{[Task]} You are solving a knots task. Determine whether the
visible strand configuration is knotted, linked, unlinked, or composed of
multiple components, or choose a legal endpoint move that untangles it.\\
\texttt{[Rules]} Use crossing order to infer the topology of static
configurations. In Untangle, a legal move lifts one endpoint above the
other strands and lowers it into the target hole.\\
\texttt{[Question]} Return the requested class, component count, link
property, or next action.\\
\texttt{[Answer Format]} Output exactly one JSON object:
\texttt{\{"answer": value\}}.
\end{tcolorbox}
\caption{Prompt for \emph{Knots} tasks.}
\label{fig:app_prompt_knots}
\end{figure}

\endgroup

\subsection{Task-Level Prompt Cards}
\label{app:task_prompt_cards}

The following cards reproduce the complete task-level prompts and qualitative
examples, grouped by topological property and task. Each task's full
specification appears in Appendix~\ref{app:benchmark}.

\newsavebox{\promptcardbox}
\newcommand{\promptcard}[1]{%
  \sbox{\promptcardbox}{\includegraphics[width=\linewidth]{figures/prompt_templates/#1}}%
  \ifdim\ht\promptcardbox<0.6\textheight
    \includegraphics[width=\linewidth,height=0.43\textheight,keepaspectratio]{figures/prompt_templates/#1}%
  \else
    \includegraphics[width=\linewidth,height=0.91\textheight,keepaspectratio]{figures/prompt_templates/#1}%
  \fi}

\subsubsection{Continuity}

\begin{figure}[H]
\centering
\promptcard{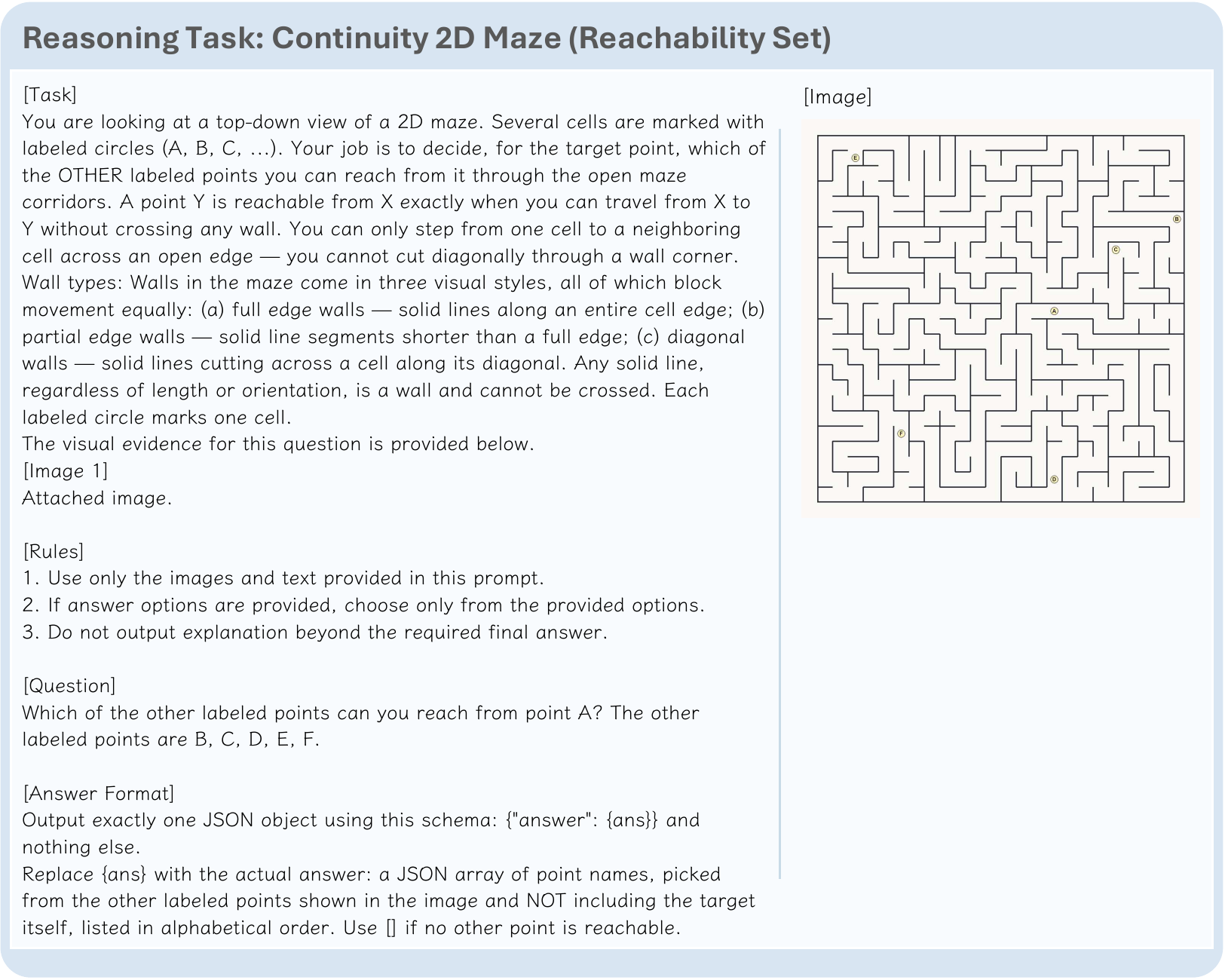}
\caption{\textbf{2D Maze} qualitative example (reachability set).}
\label{fig:app_task_2d_maze_examples}
\end{figure}

\begin{figure}[H]
\centering
\promptcard{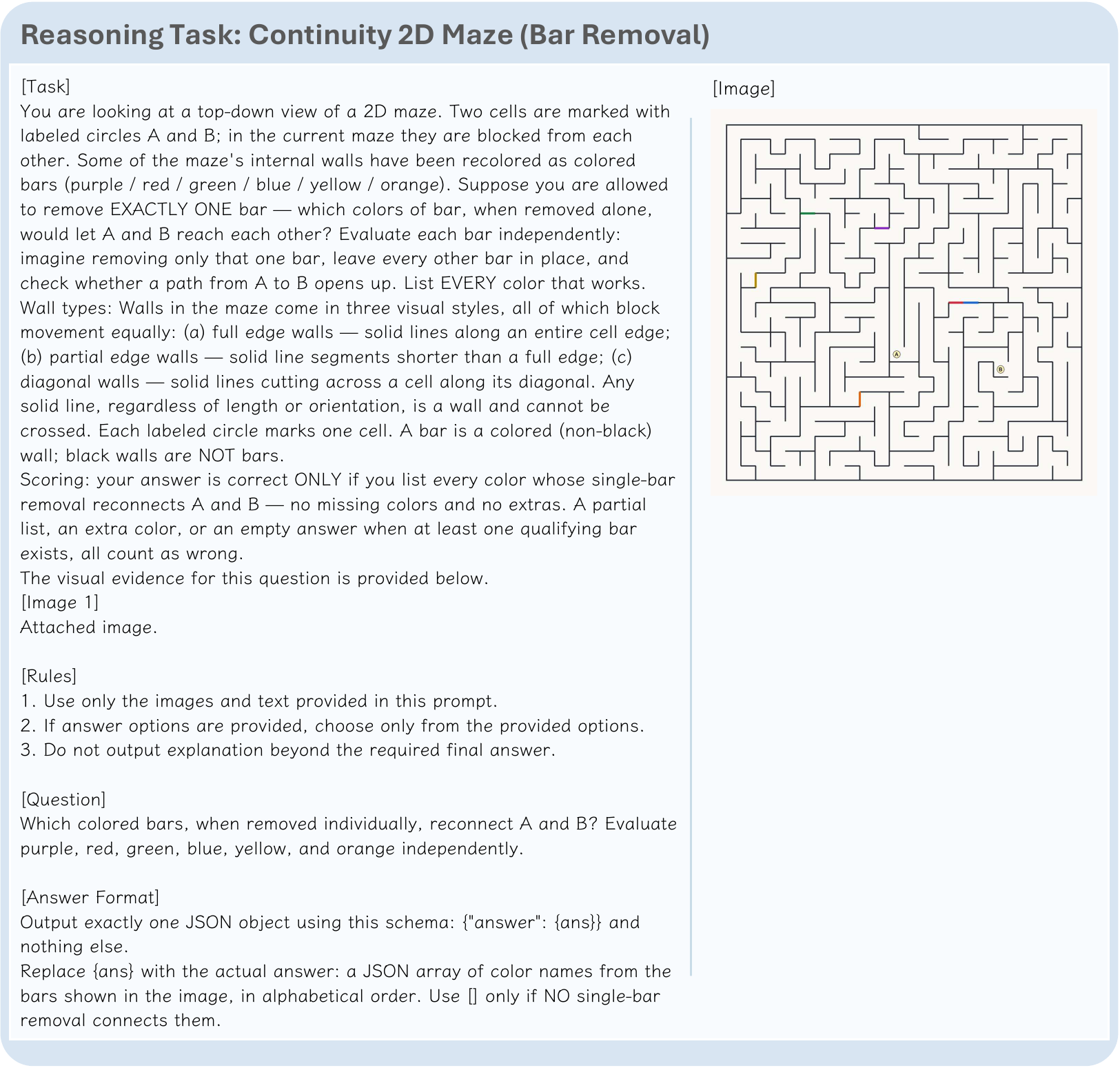}
\caption{\textbf{2D Maze} qualitative example (bar removal).}
\label{fig:app_task_2d_maze_examples_bar_removal}
\end{figure}

\begin{figure}[H]
\centering
\promptcard{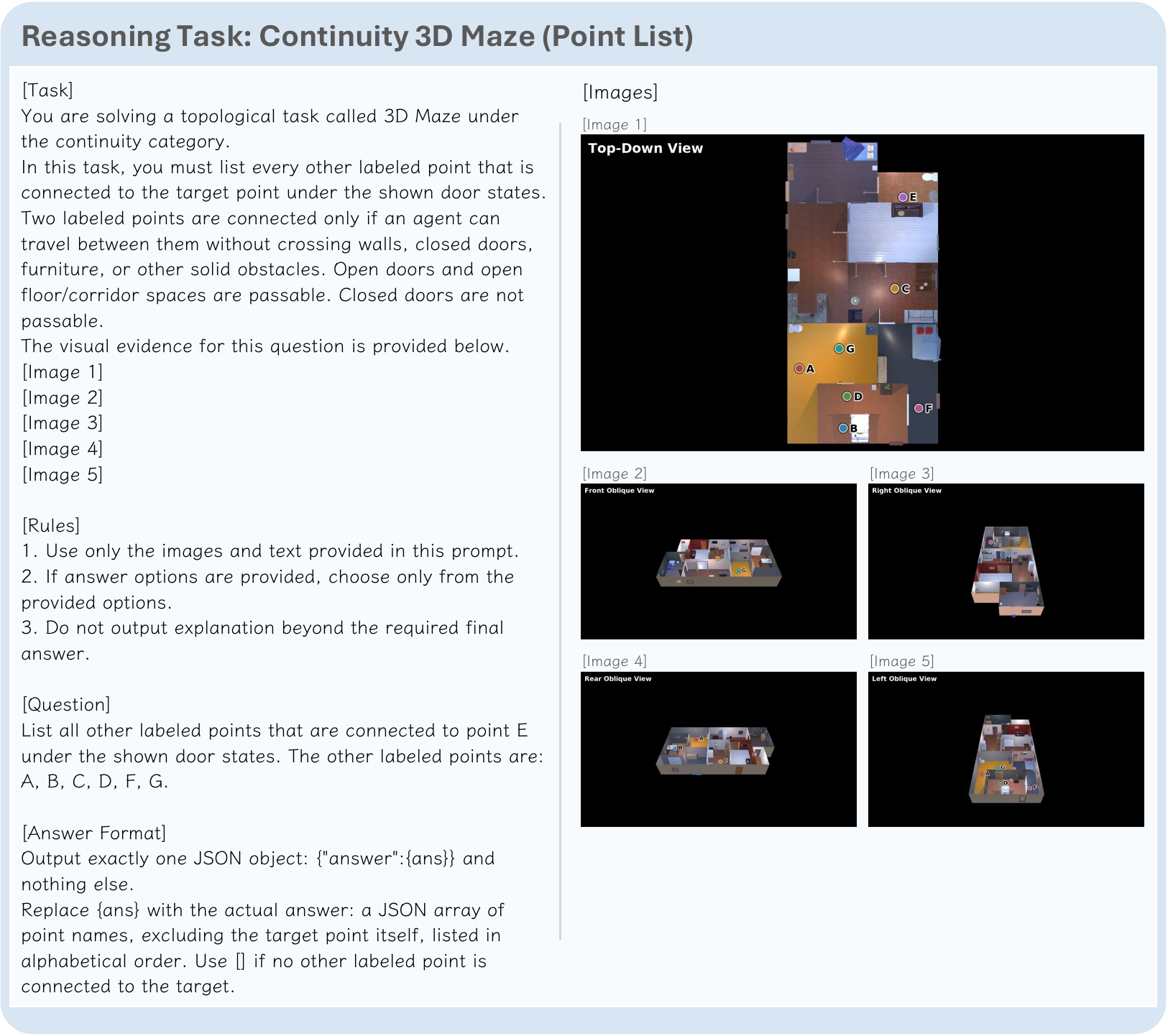}
\caption{\textbf{3D Maze} qualitative example (point list).}
\label{fig:app_task_3d_maze_examples}
\end{figure}

\begin{figure}[H]
\centering
\promptcard{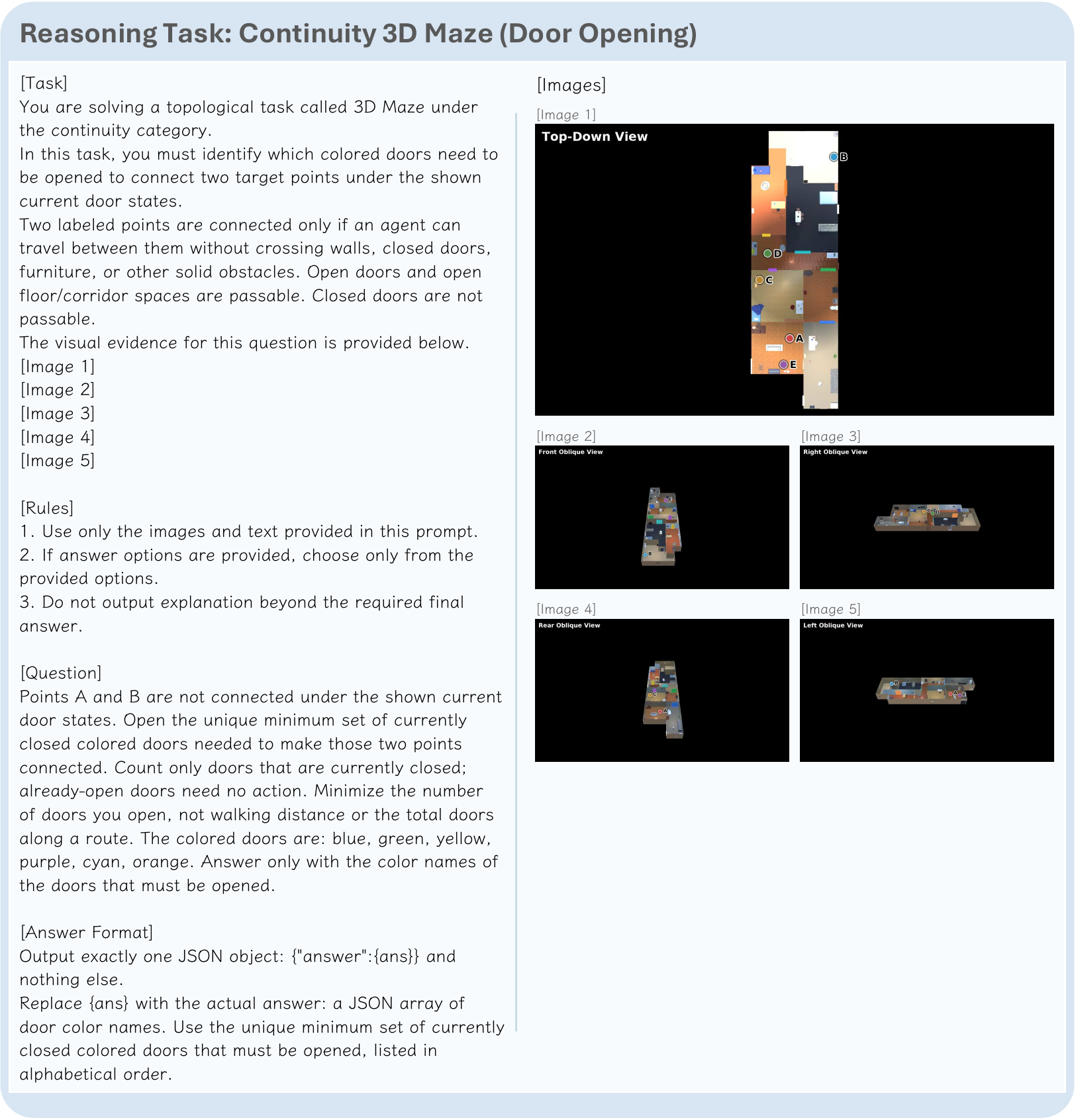}
\caption{\textbf{3D Maze} qualitative example (door opening).}
\label{fig:app_task_3d_maze_examples_door_open}
\end{figure}

\begin{figure}[H]
\centering
\promptcard{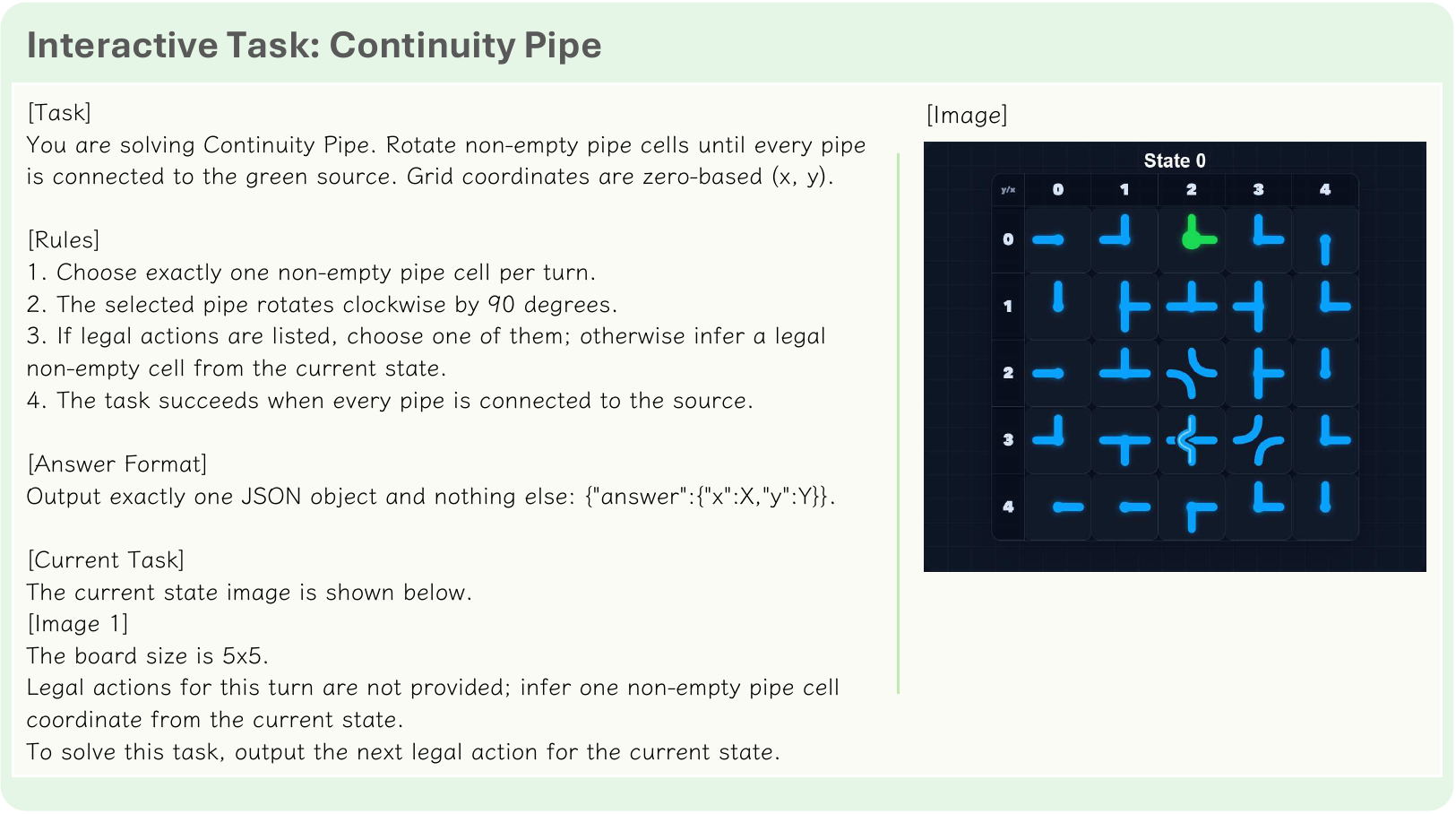}
\caption{\textbf{Pipe} qualitative example (first turn, $5{\times}5$ grid).}
\label{fig:app_task_pipe_examples}
\end{figure}

\subsubsection{Separation}

\begin{figure}[H]
\centering
\promptcard{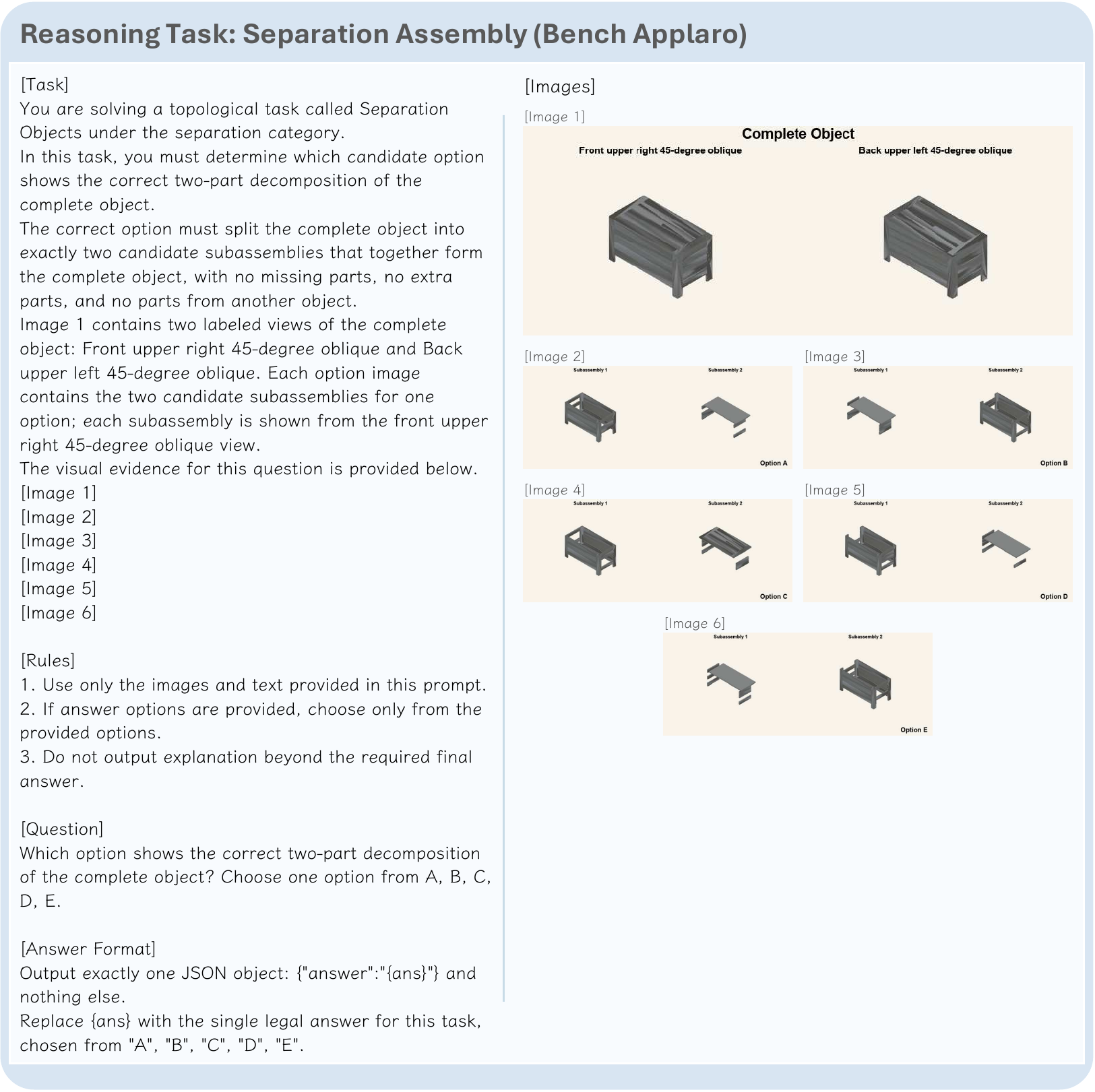}
\caption{\textbf{Assembly} qualitative example (bench, Applaro).}
\label{fig:app_task_separation_objects_examples_bench_sialland}
\end{figure}

\begin{figure}[H]
\centering
\promptcard{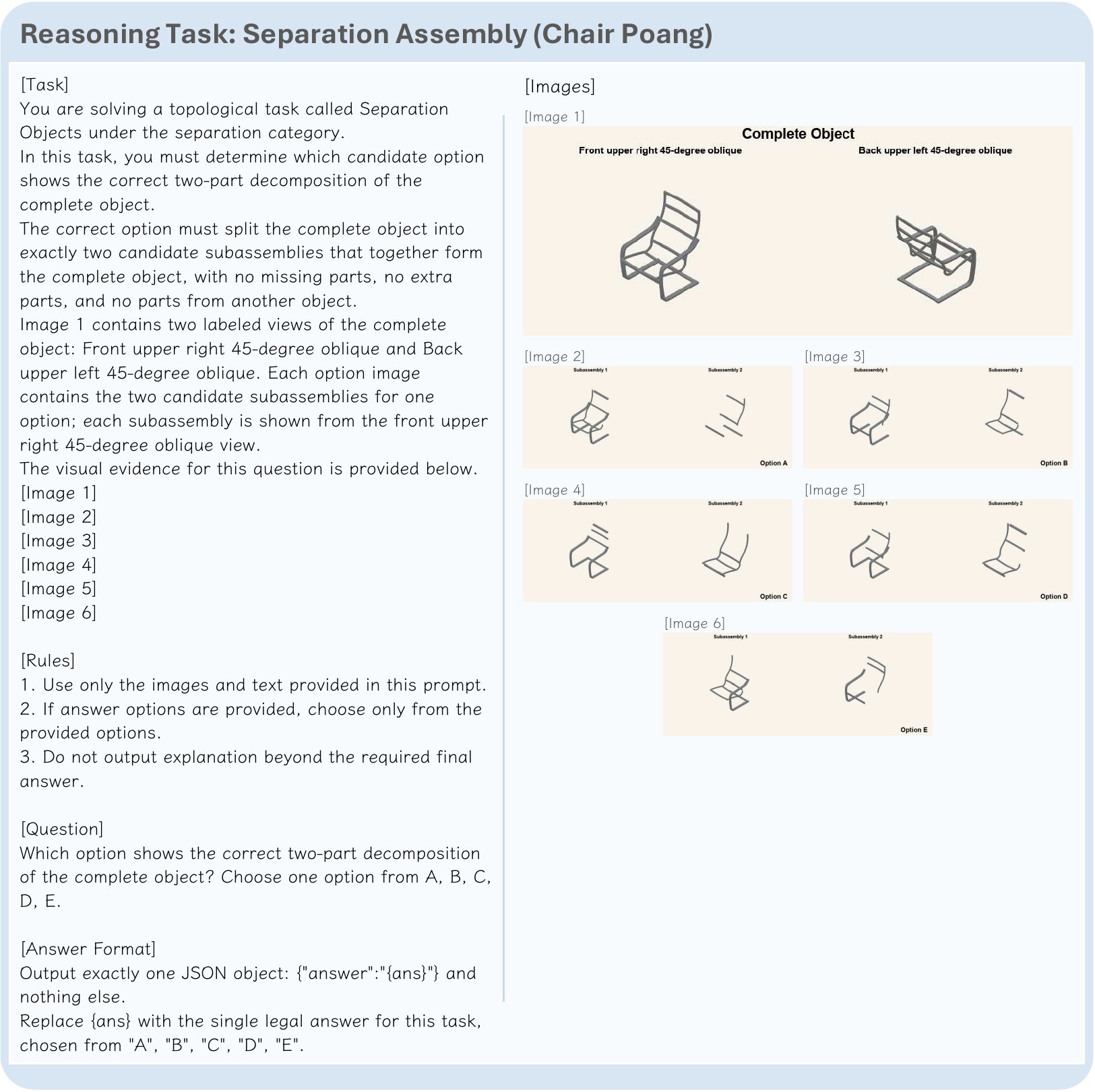}
\caption{\textbf{Assembly} qualitative example (chair, Poang).}
\label{fig:app_task_separation_objects_examples_bench_applaro}
\end{figure}

\begin{figure}[H]
\centering
\promptcard{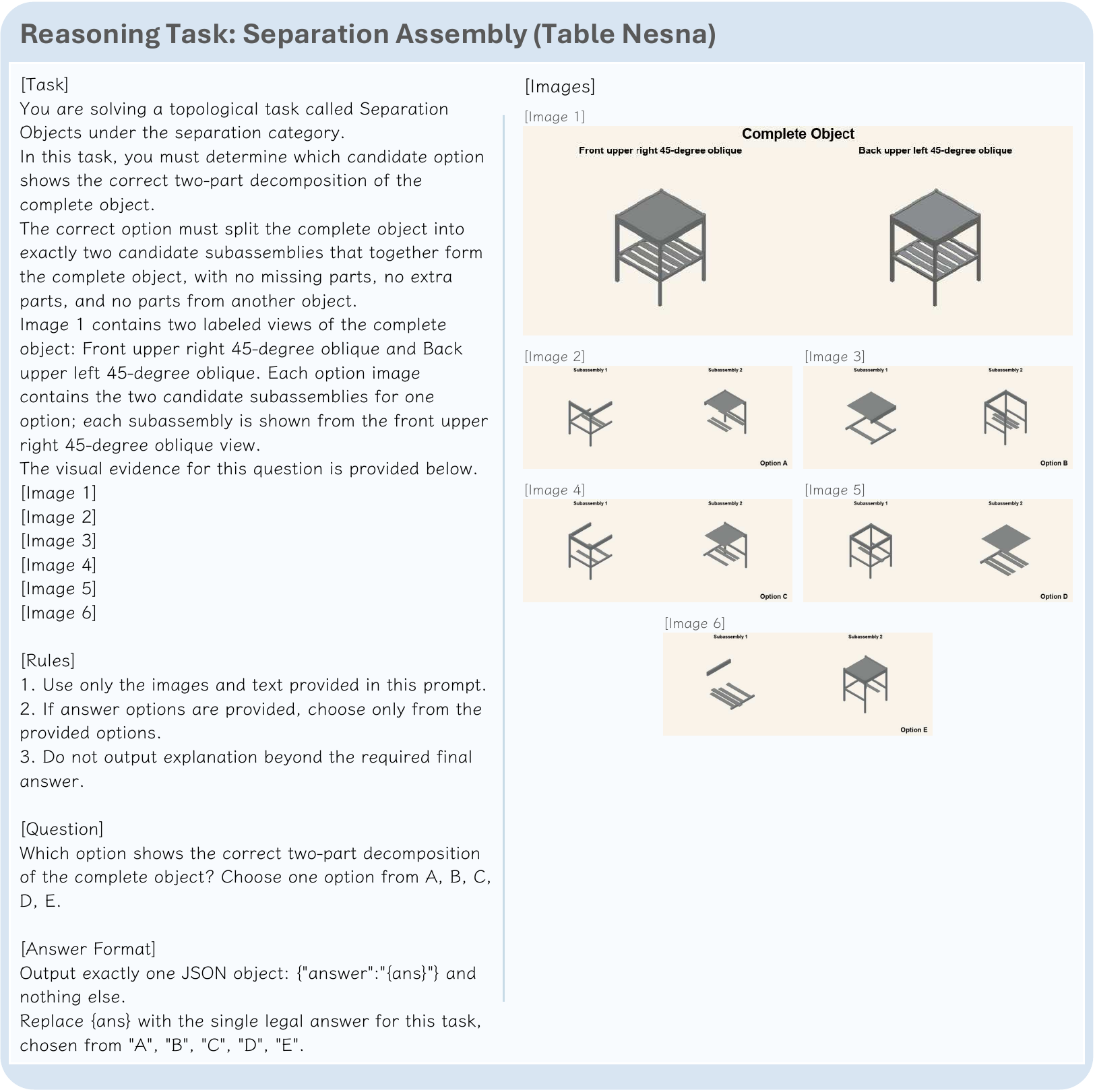}
\caption{\textbf{Assembly} qualitative example (table, Nesna).}
\label{fig:app_task_separation_objects_examples_chair_agam}
\end{figure}

\begin{figure}[H]
\centering
\promptcard{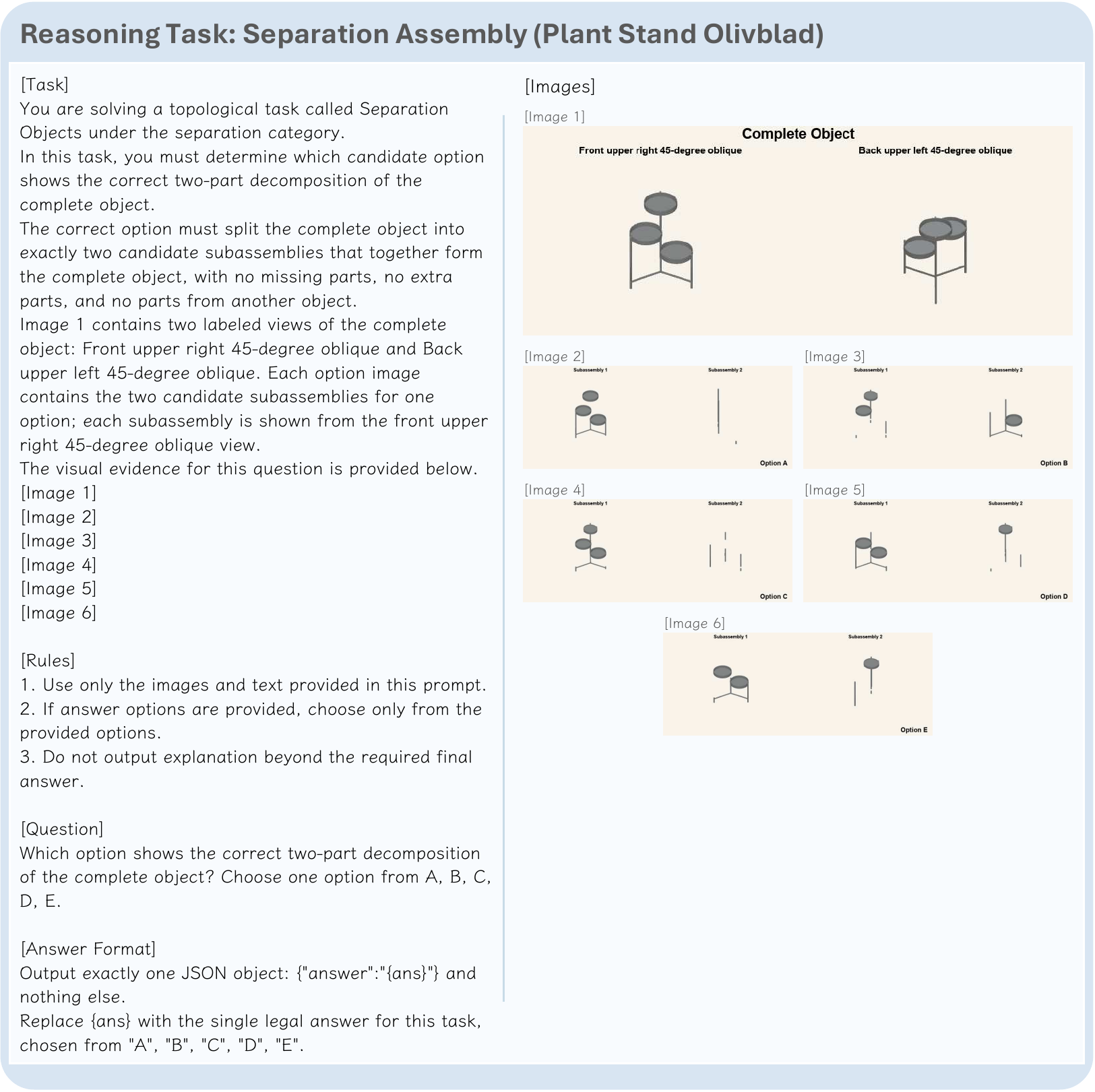}
\caption{\textbf{Assembly} qualitative example (plant stand, Olivblad).}
\label{fig:app_task_separation_objects_examples_chair_applaro}
\end{figure}

\begin{figure}[H]
\centering
\promptcard{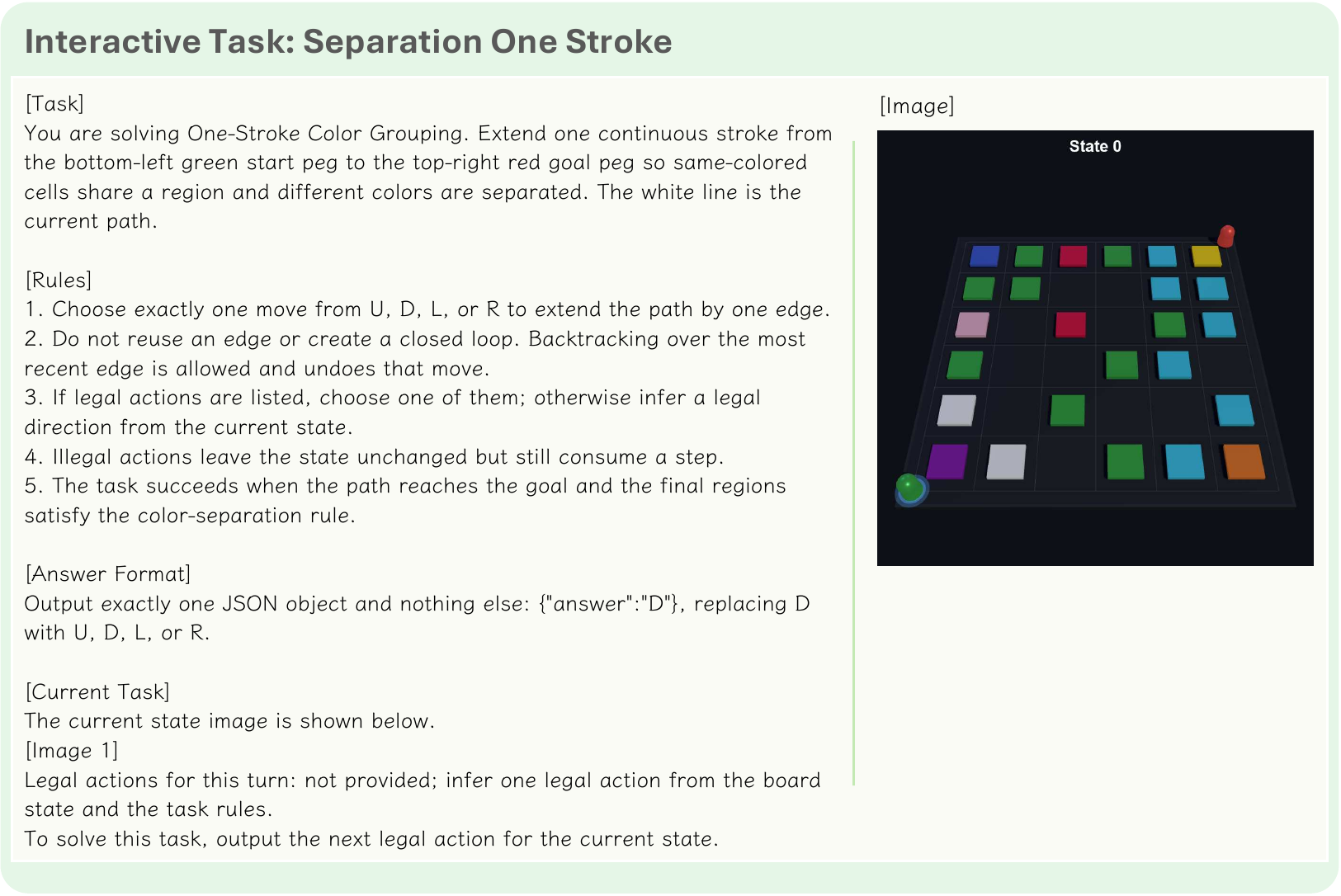}
\caption{\textbf{One Stroke} qualitative example (first turn, $6{\times}6$ board).}
\label{fig:app_task_one_stroke_examples}
\end{figure}

\subsubsection{Order}

\begin{figure}[H]
\centering
\promptcard{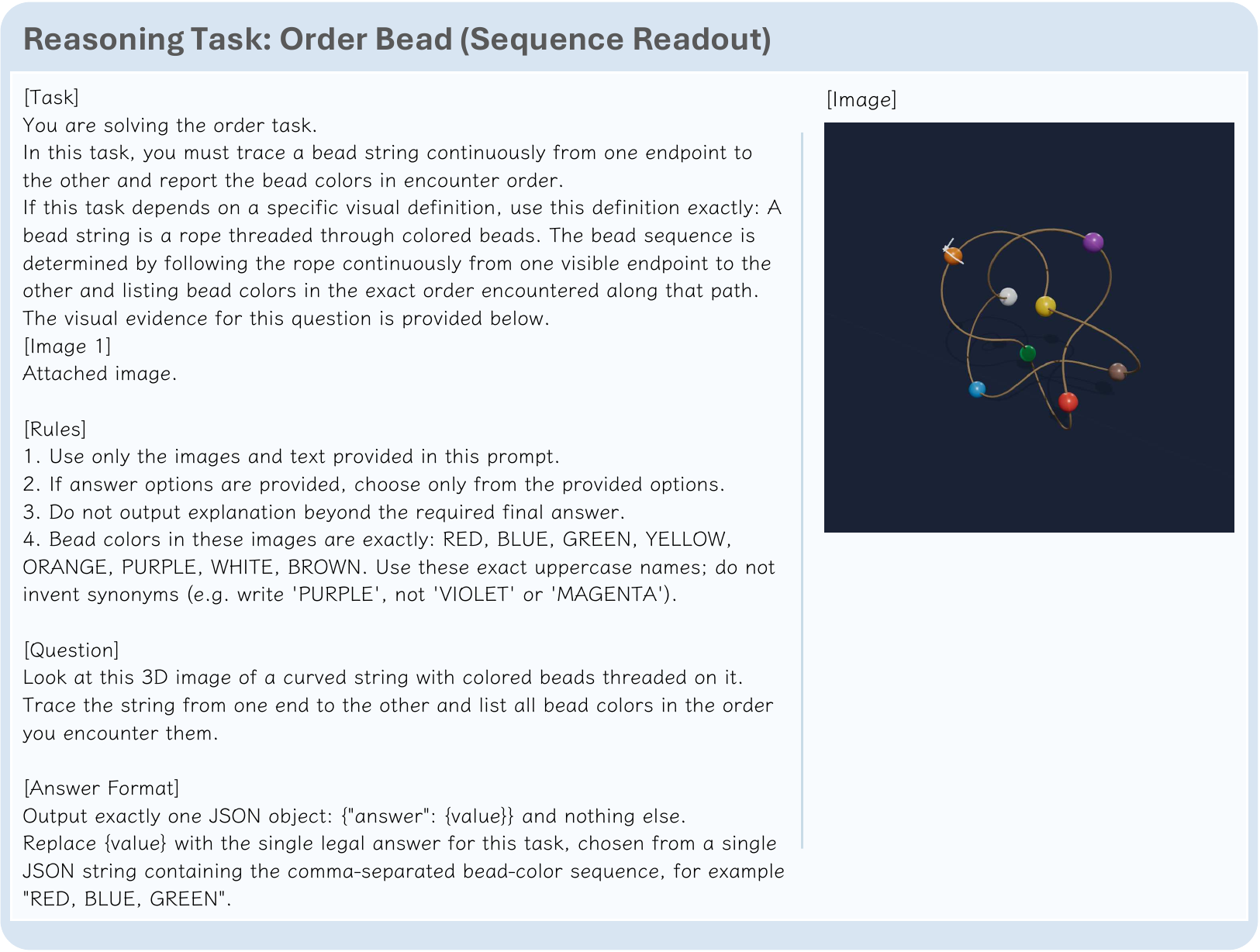}
\caption{\textbf{Bead} qualitative example (sequence readout).}
\label{fig:app_task_bead_string_examples_description}
\end{figure}

\begin{figure}[H]
\centering
\promptcard{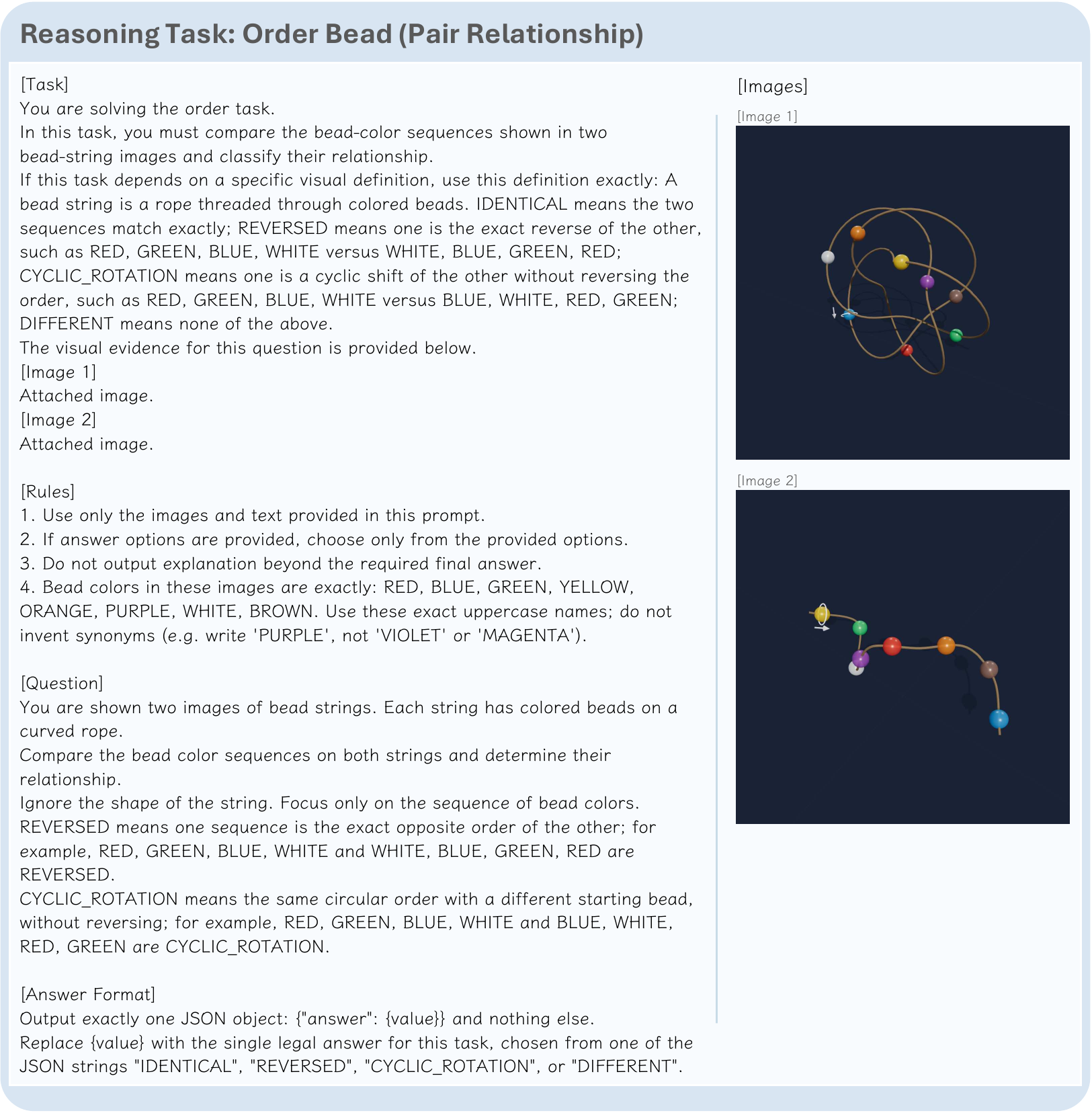}
\caption{\textbf{Bead} qualitative example (pair relationship).}
\label{fig:app_task_bead_string_examples_pair_relationship}
\end{figure}

\begin{figure}[H]
\centering
\promptcard{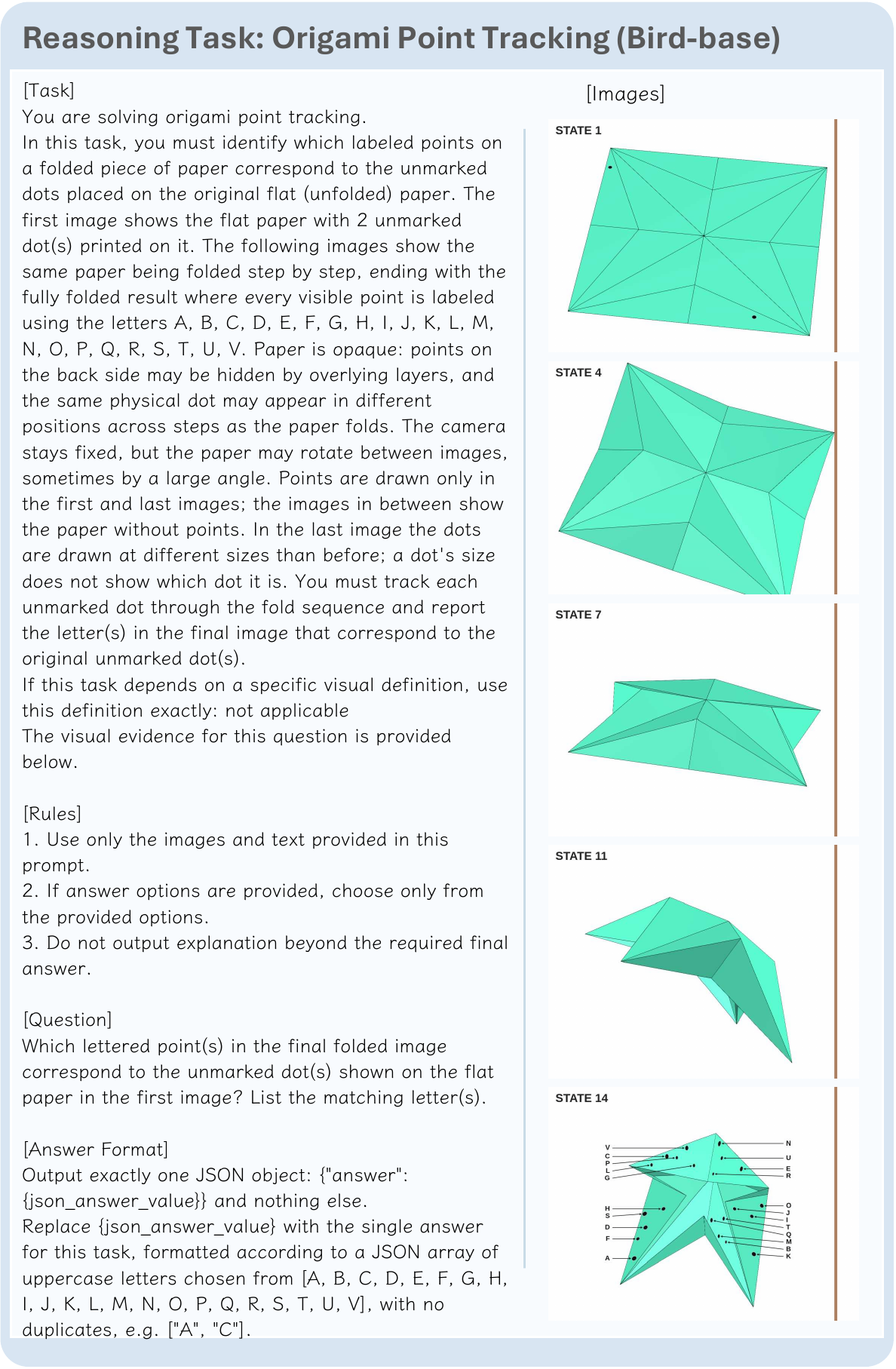}
\caption{\textbf{Origami Point} qualitative example (bird base).}
\label{fig:app_task_origami_examples_bird_base}
\end{figure}

\begin{figure}[H]
\centering
\promptcard{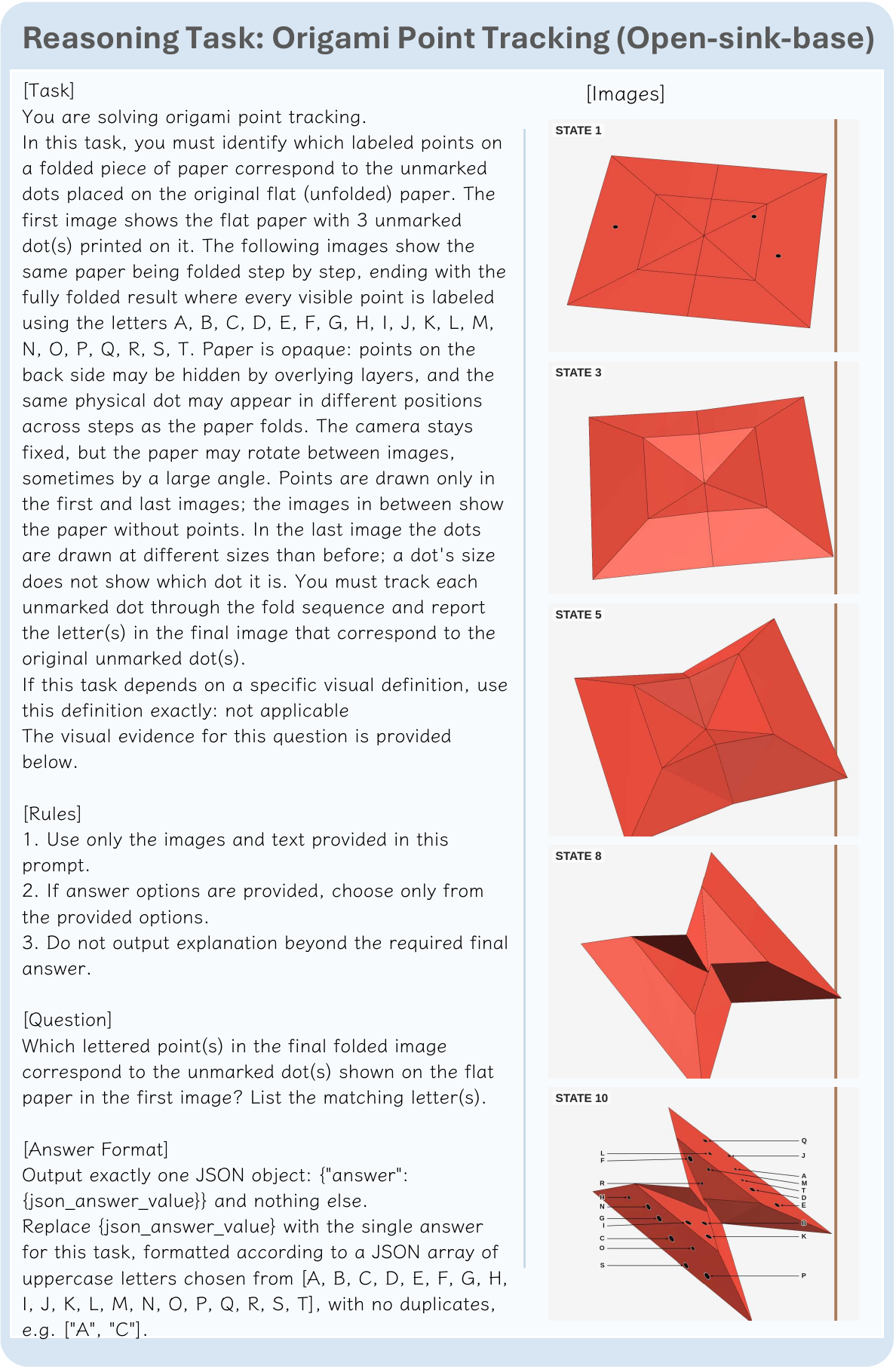}
\caption{\textbf{Origami Point} qualitative example (open-sink base).}
\label{fig:app_task_origami_examples_opensink_base}
\end{figure}

\begin{figure}[H]
\centering
\promptcard{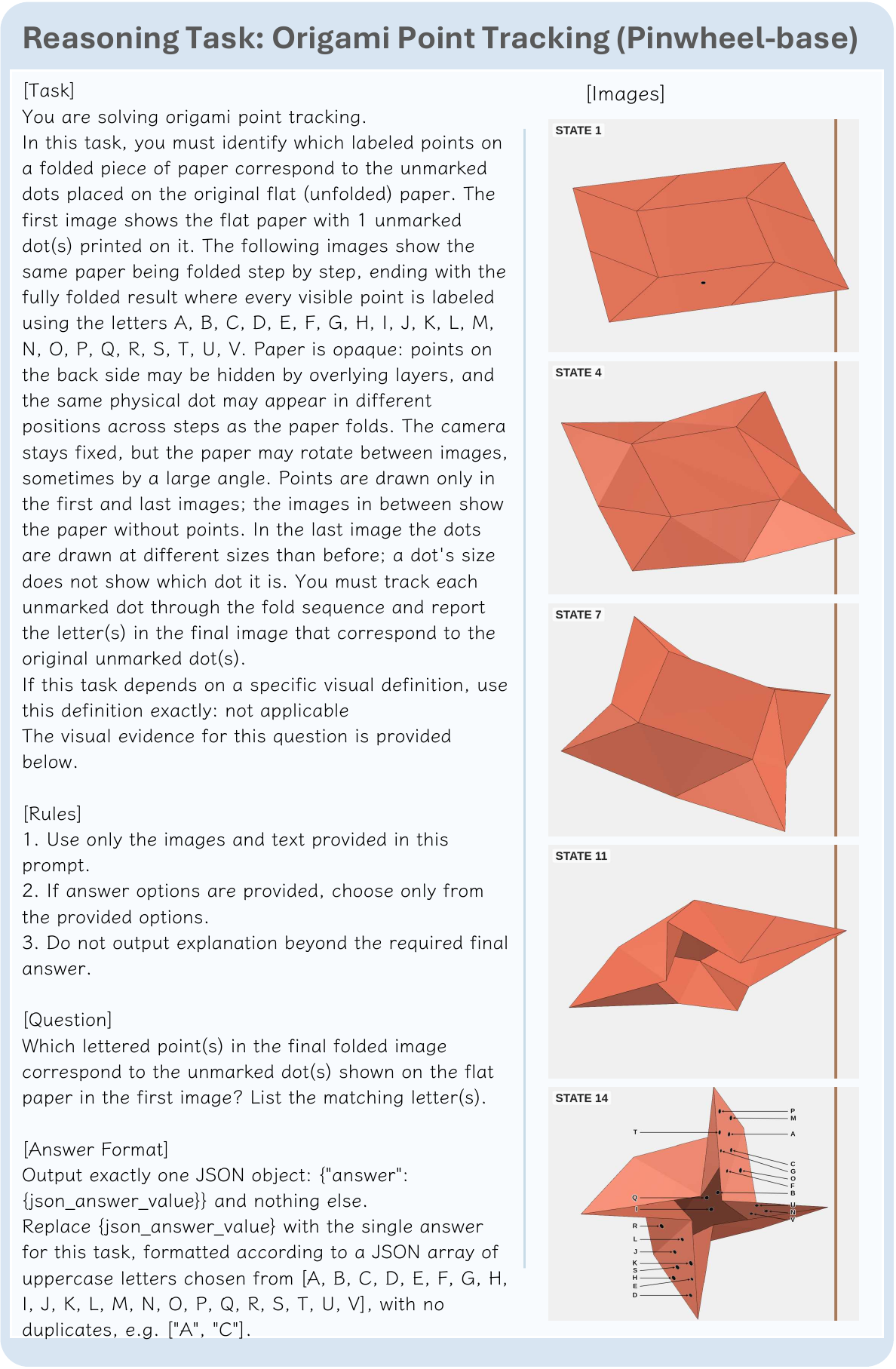}
\caption{\textbf{Origami Point} qualitative example (pinwheel base).}
\label{fig:app_task_origami_examples_pinwheel_base}
\end{figure}

\begin{figure}[H]
\centering
\promptcard{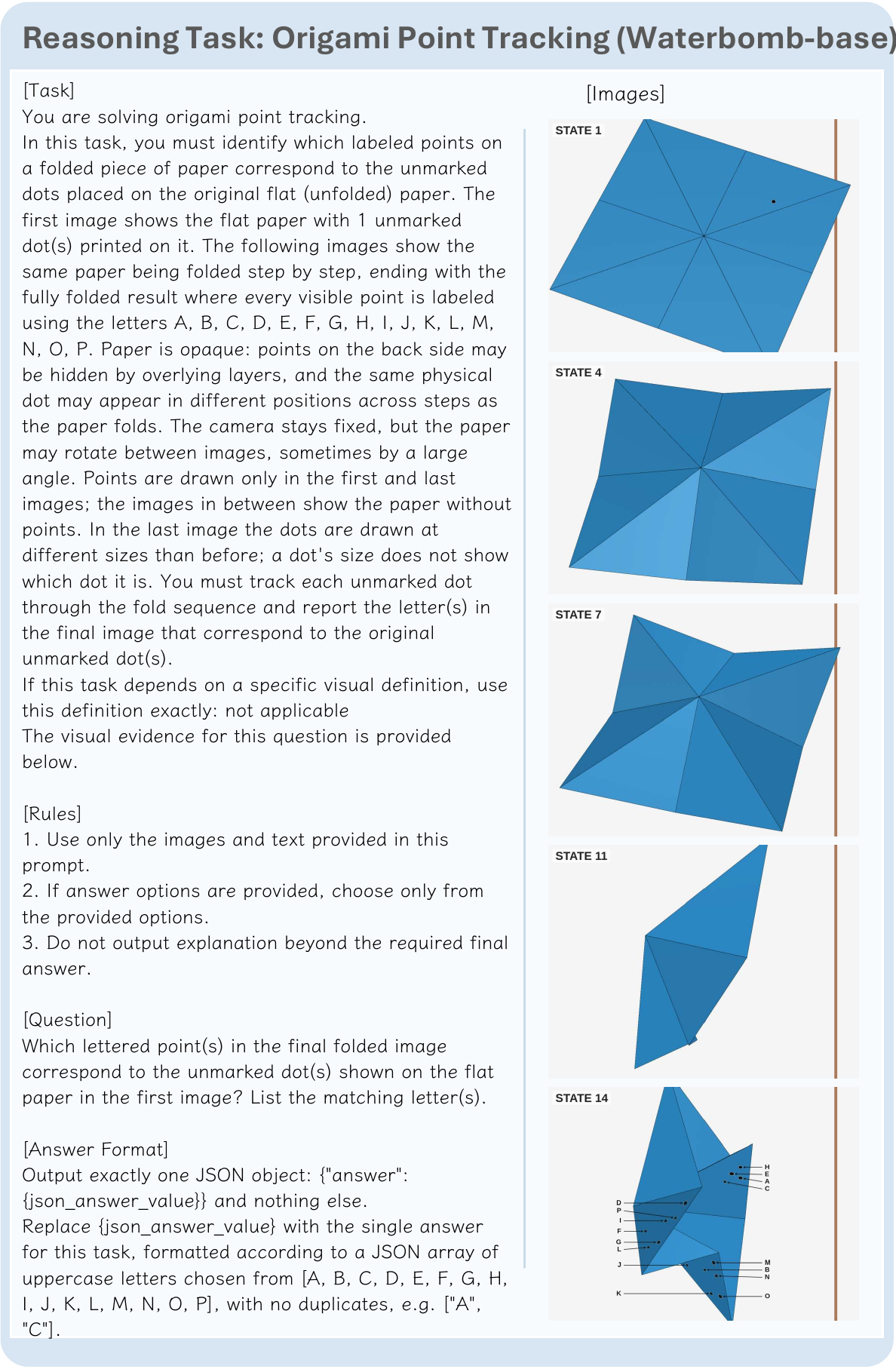}
\caption{\textbf{Origami Point} qualitative example (waterbomb base).}
\label{fig:app_task_origami_examples_waterbomb_base}
\end{figure}

\begin{figure}[H]
\centering
\promptcard{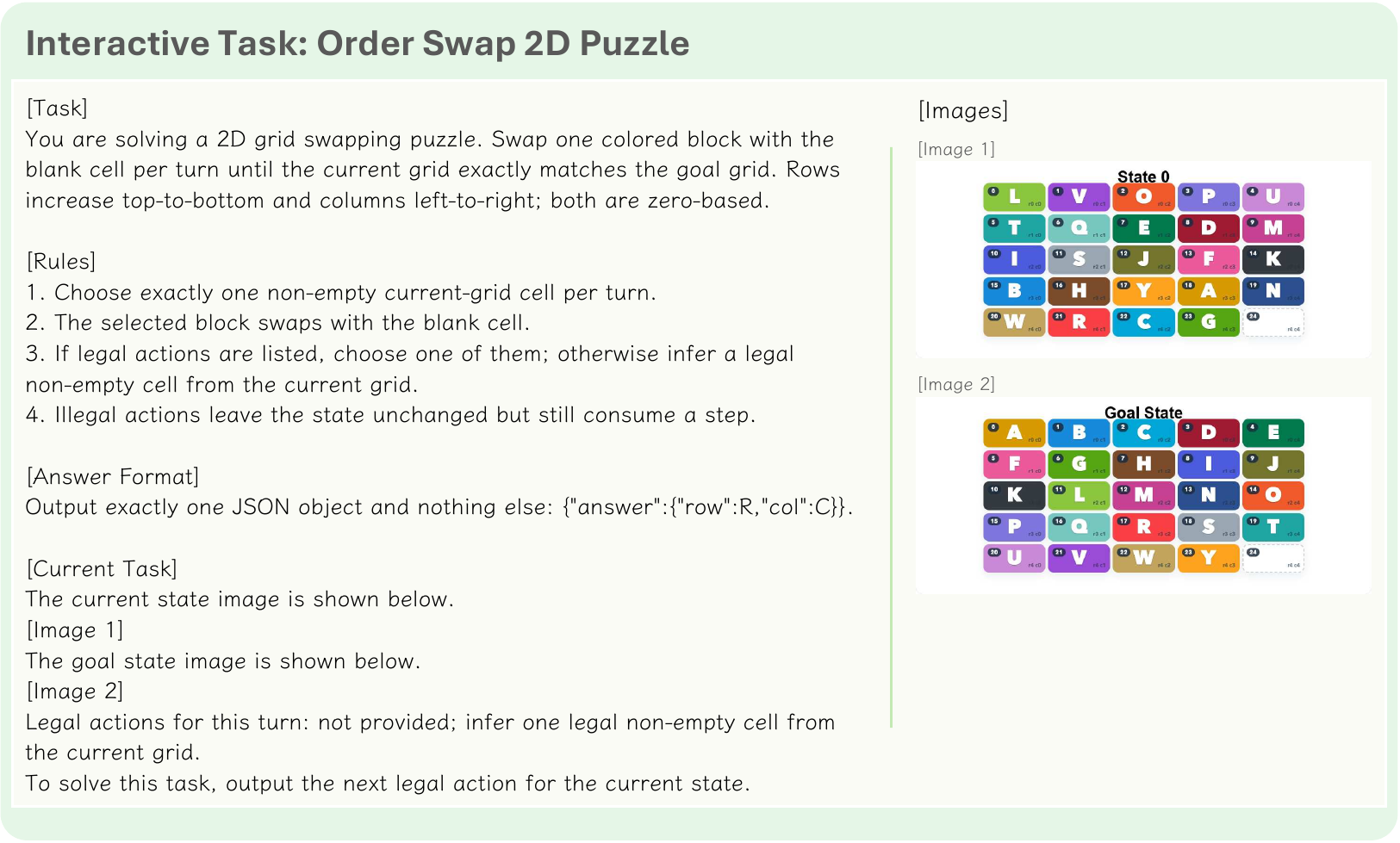}
\caption{\textbf{Swap} qualitative example (first turn, $5{\times}5$ grid).}
\label{fig:app_task_swap_puzzle_examples}
\end{figure}

\subsubsection{Enclosure}

\begin{figure}[H]
\centering
\promptcard{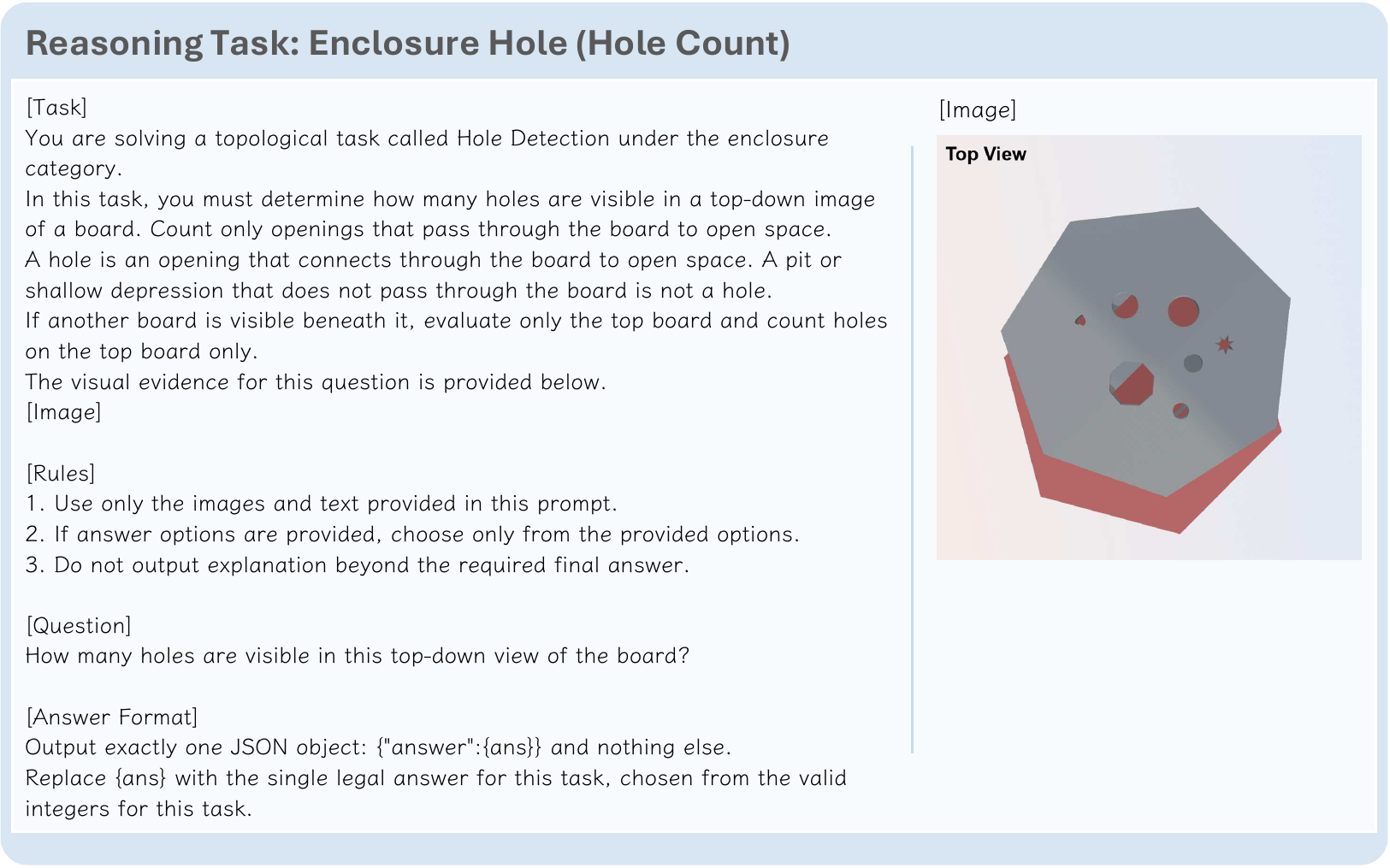}
\caption{\textbf{Hole} qualitative example (hole count).}
\label{fig:app_task_hole_detection_examples}
\end{figure}

\begin{figure}[H]
\centering
\promptcard{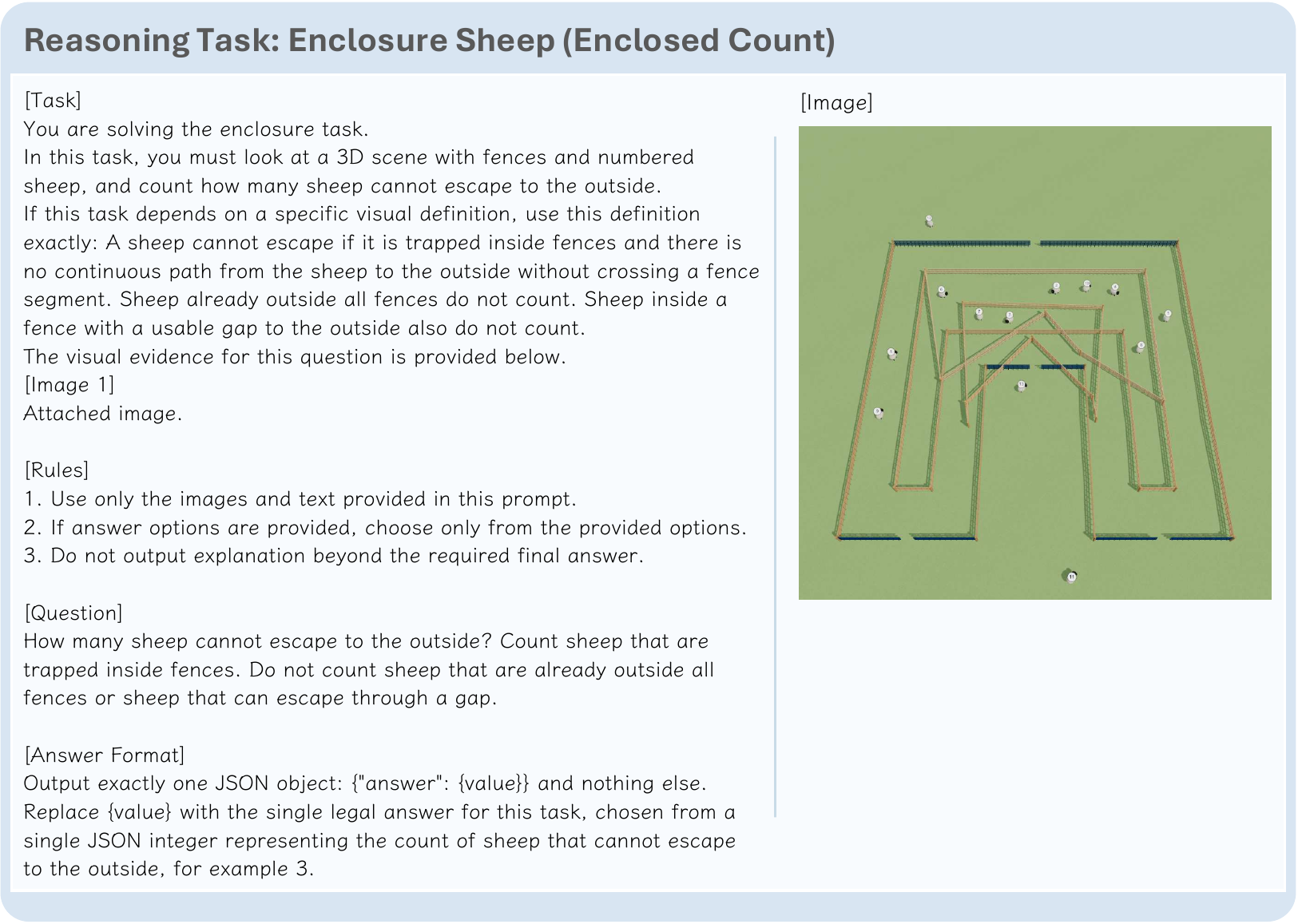}
\caption{\textbf{Sheep} qualitative example (enclosed count).}
\label{fig:app_task_sheep_examples_count_inside_and_escape_probability}
\end{figure}

\begin{figure}[H]
\centering
\promptcard{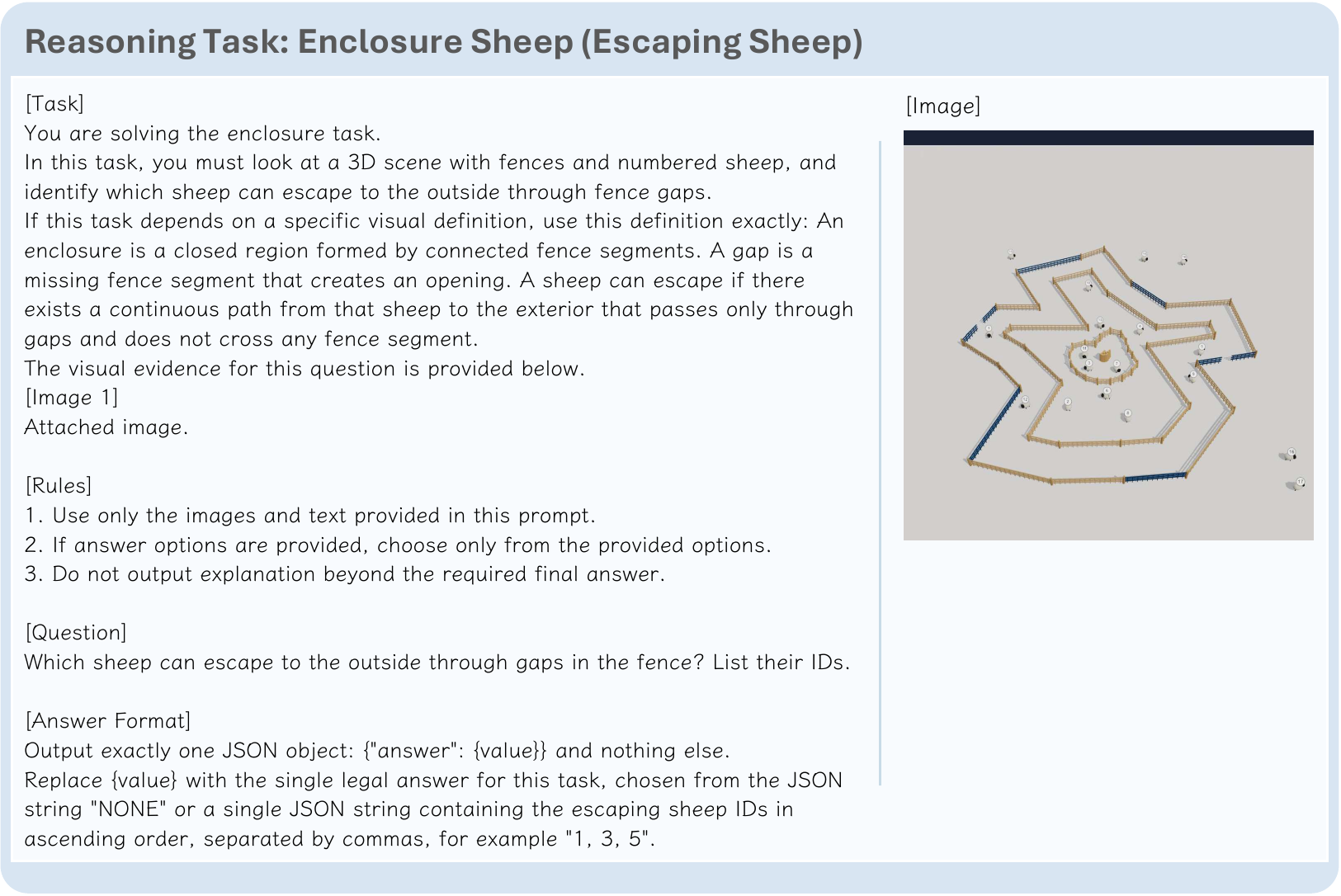}
\caption{\textbf{Sheep} qualitative example (escaping sheep).}
\label{fig:app_task_sheep_examples_escape_probability}
\end{figure}

\begin{figure}[H]
\centering
\promptcard{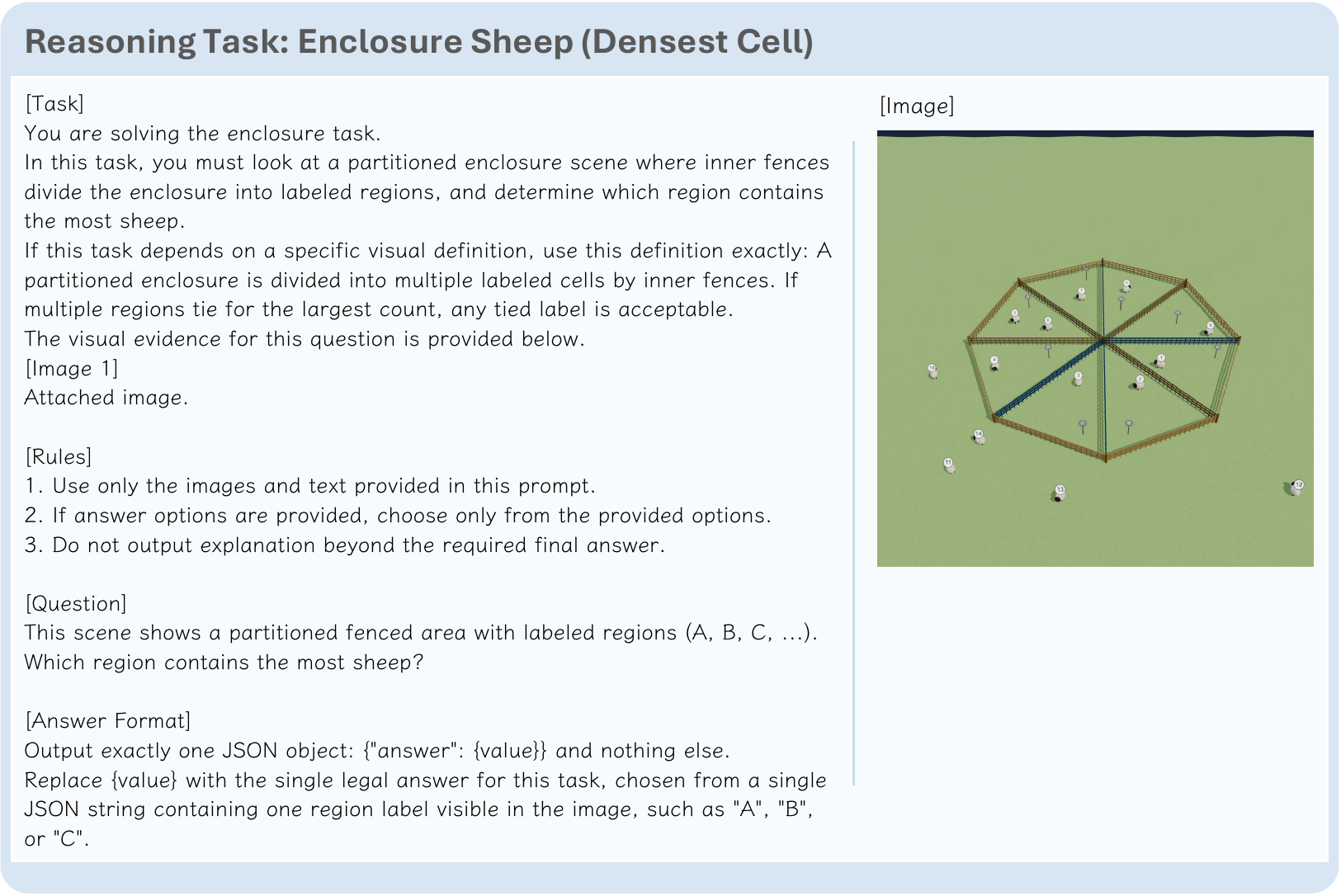}
\caption{\textbf{Sheep} qualitative example (densest cell).}
\label{fig:app_task_sheep_examples_max_cell_count_and_fence_repair}
\end{figure}

\begin{figure}[H]
\centering
\promptcard{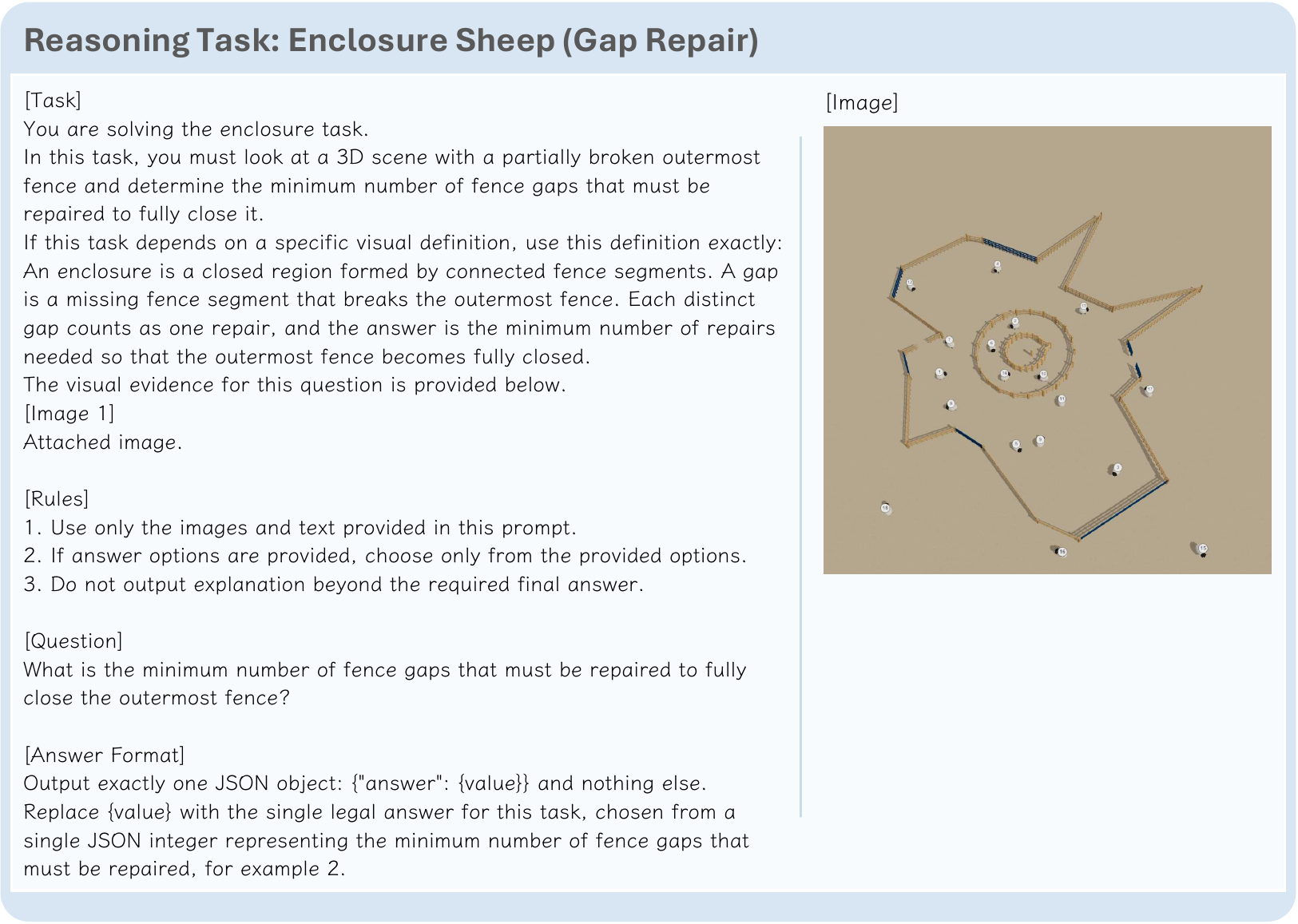}
\caption{\textbf{Sheep} qualitative example (gap repair).}
\label{fig:app_task_sheep_examples_fence_repair}
\end{figure}

\begin{figure}[H]
\centering
\promptcard{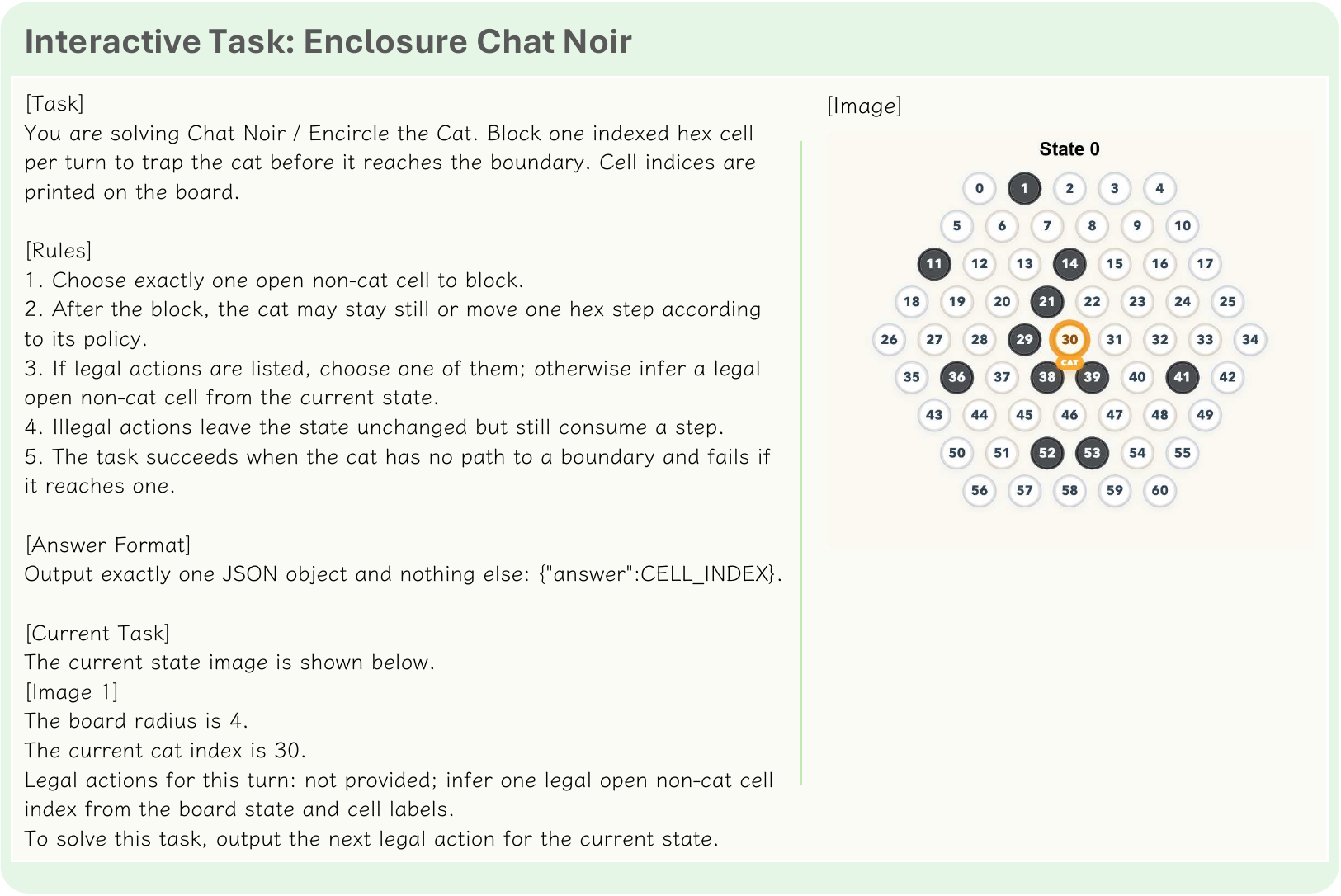}
\caption{\textbf{Chat Noir} qualitative example (first turn, board radius 4).}
\label{fig:app_task_chat_noir_examples}
\end{figure}

\subsubsection{Knots}

\begin{figure}[H]
\centering
\promptcard{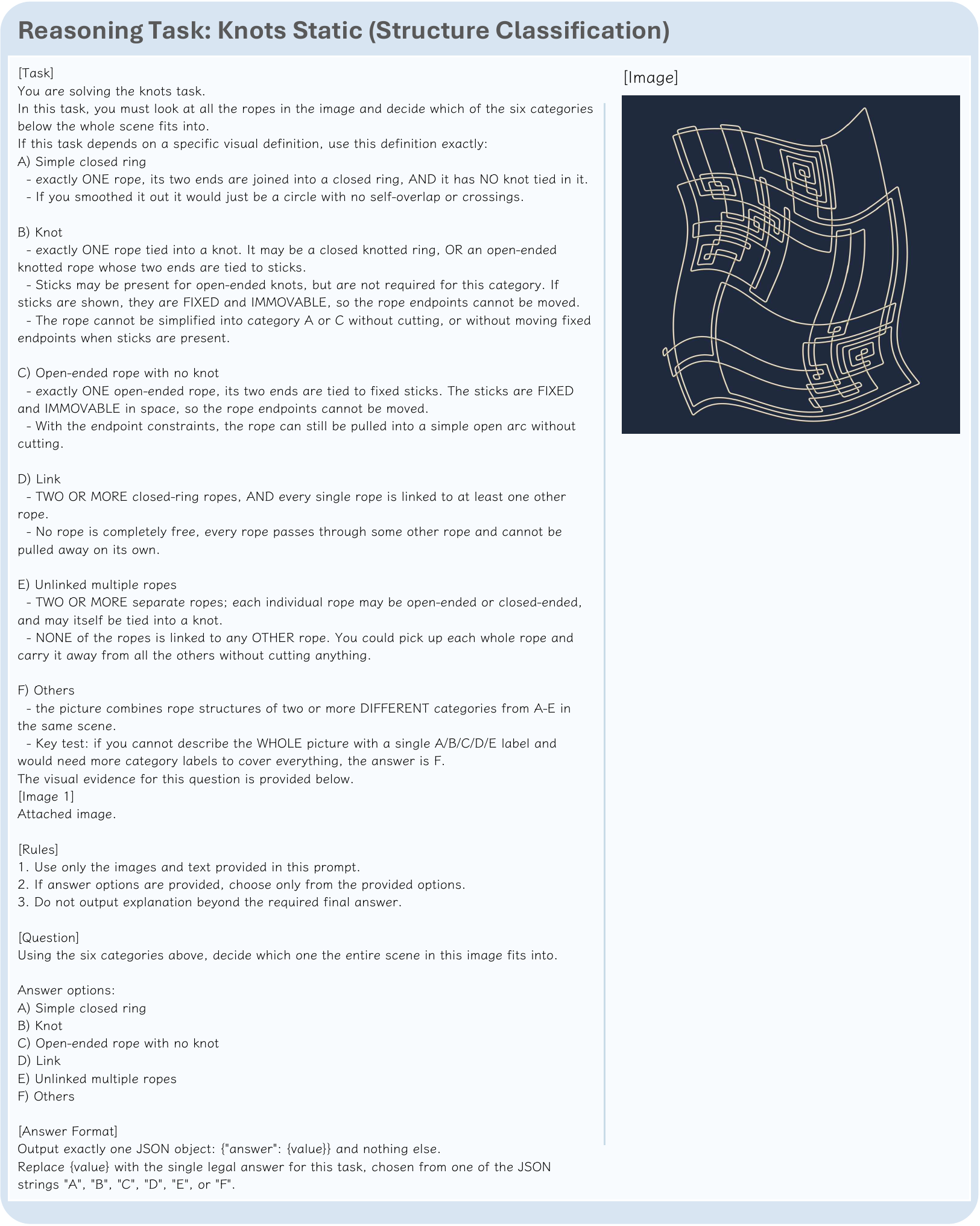}
\caption{\textbf{Knots} qualitative example (structure classification).}
\label{fig:app_task_knots_examples_structure_classification}
\end{figure}

\begin{figure}[H]
\centering
\promptcard{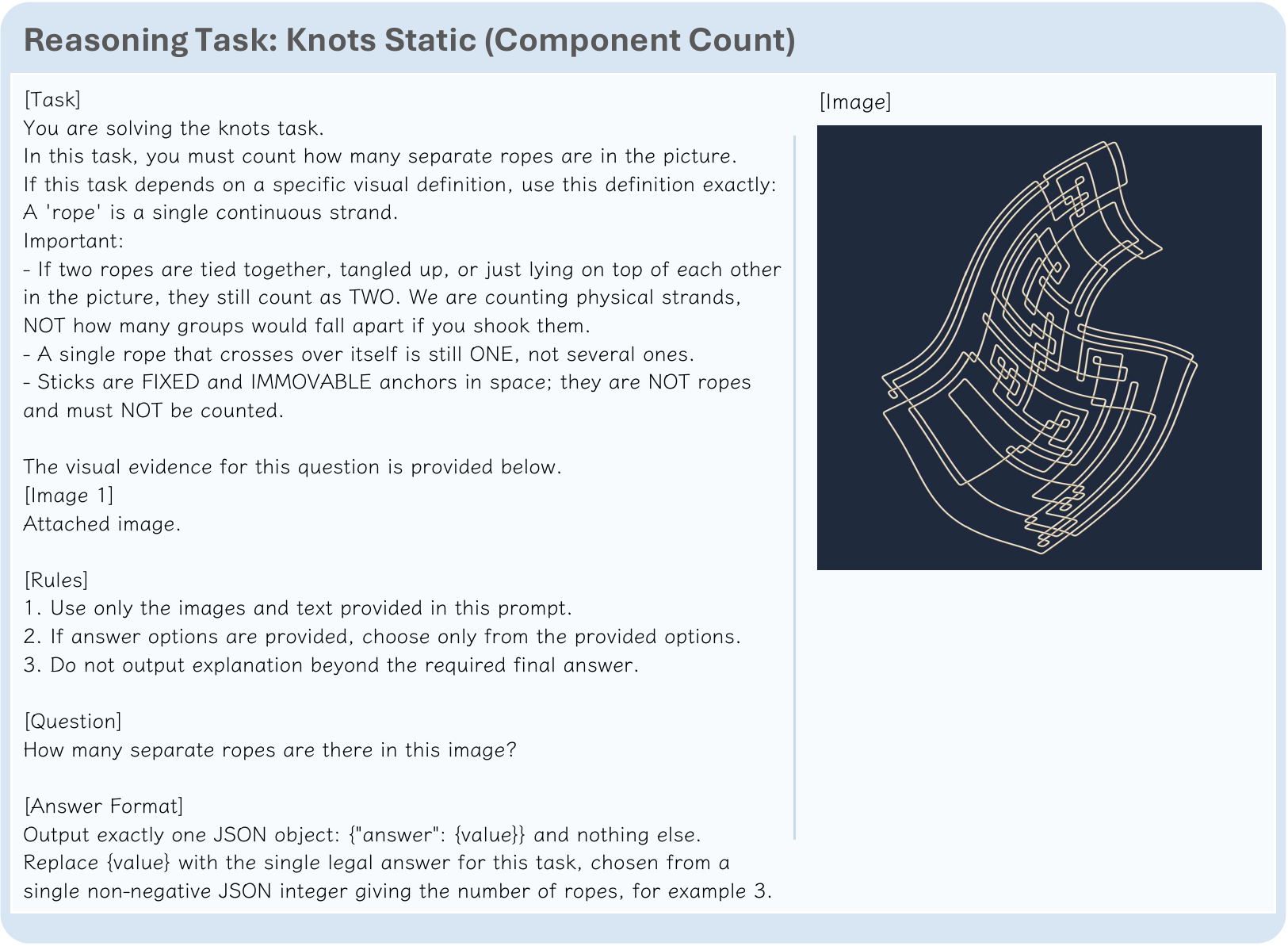}
\caption{\textbf{Knots} qualitative example (component count).}
\label{fig:app_task_knots_examples_component_count}
\end{figure}

\begin{figure}[H]
\centering
\promptcard{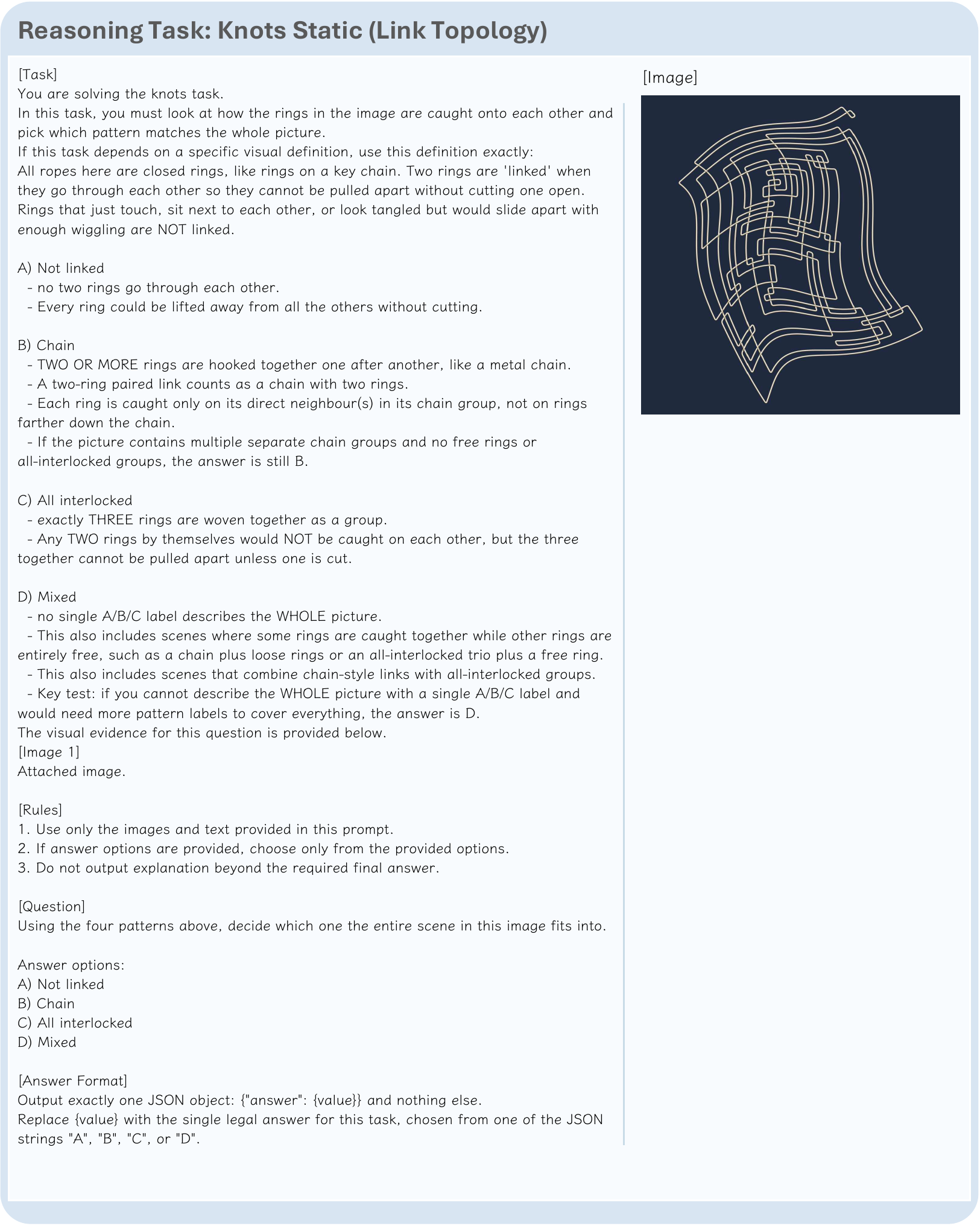}
\caption{\textbf{Knots} qualitative example (link topology).}
\label{fig:app_task_knots_examples_link_topology}
\end{figure}

\begin{figure}[H]
\centering
\promptcard{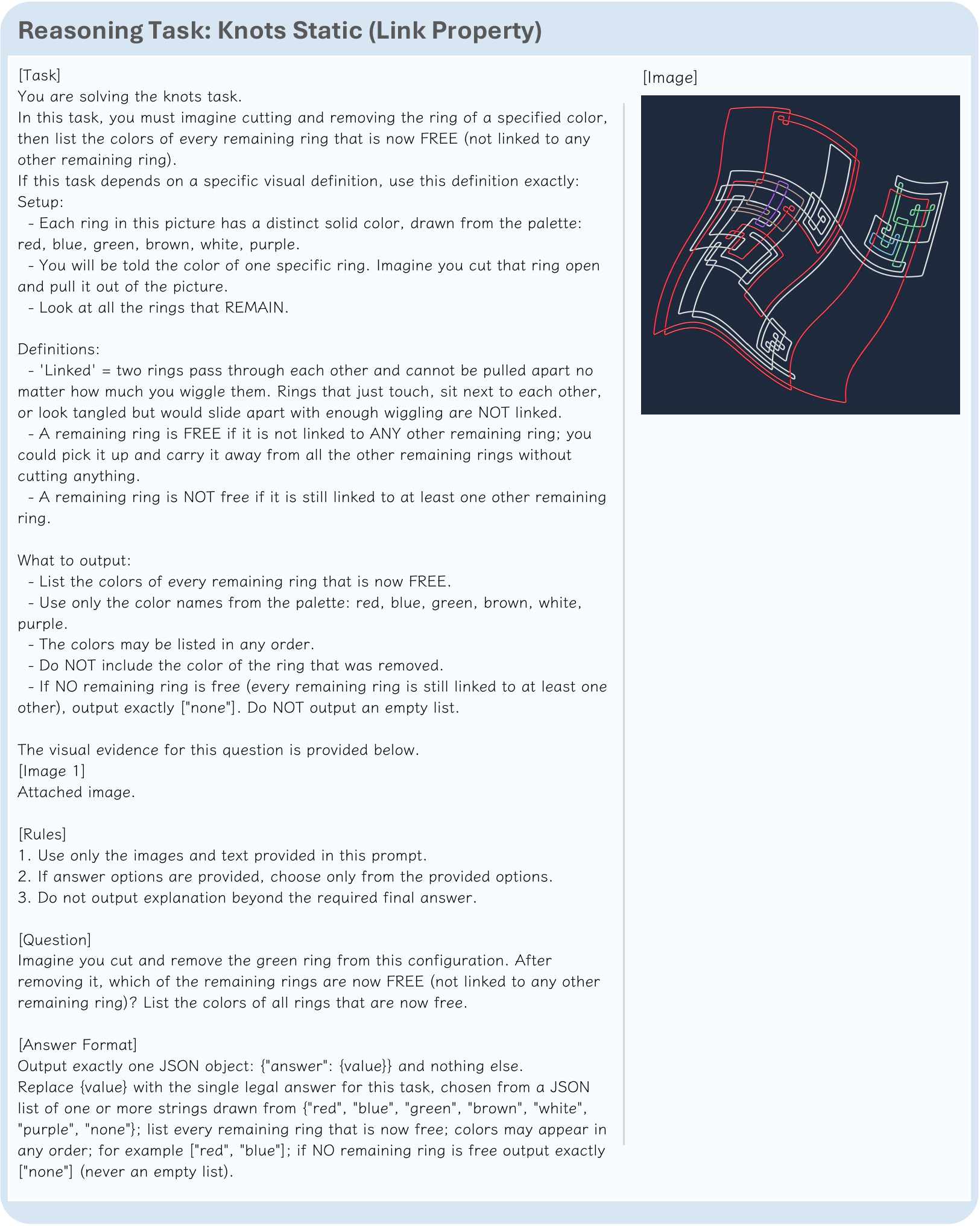}
\caption{\textbf{Knots} qualitative example (link property).}
\label{fig:app_task_knots_examples_link_property}
\end{figure}

\begin{figure}[H]
\centering
\promptcard{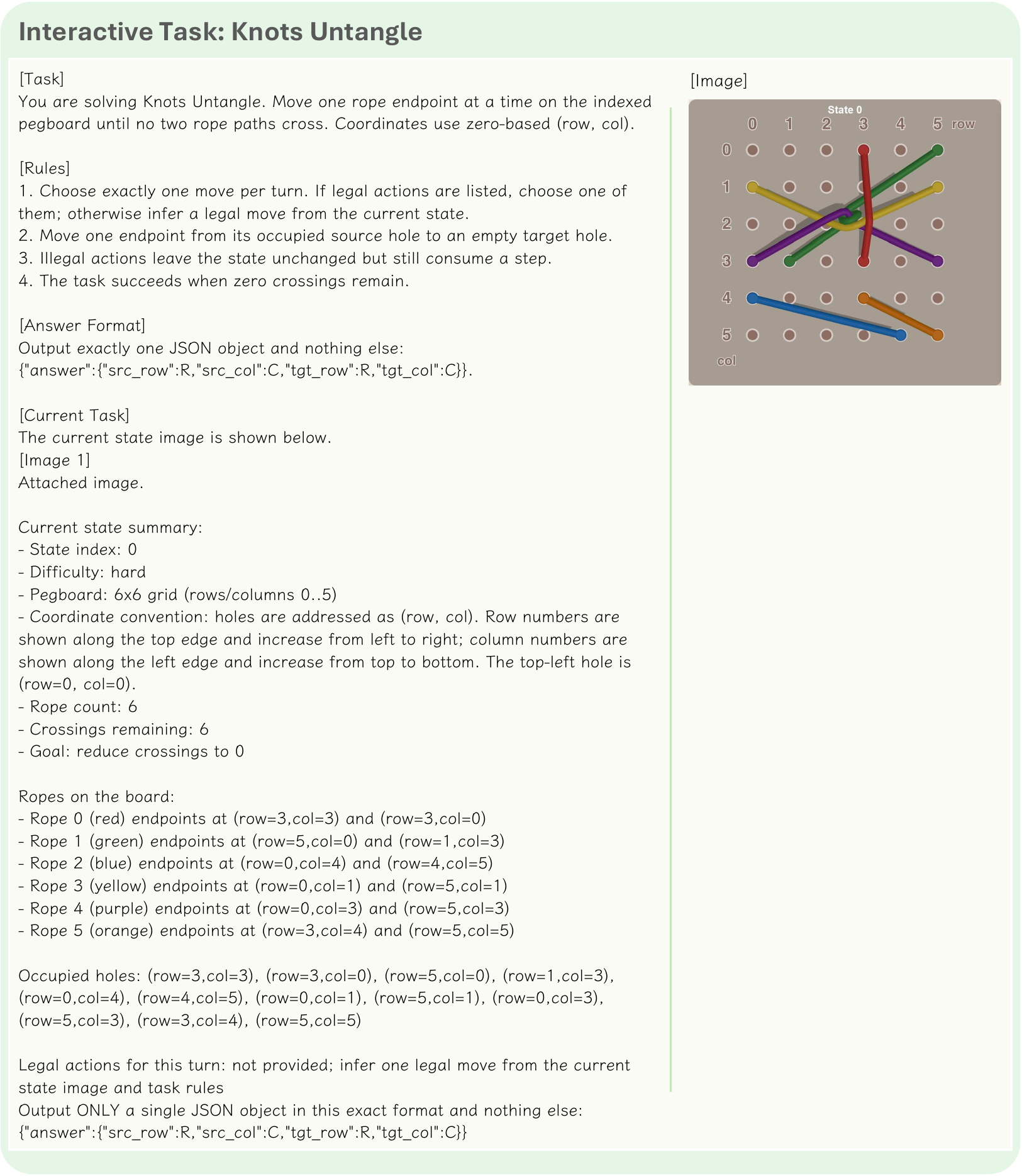}
\caption{\textbf{Untangle} qualitative example (first turn, $6{\times}6$ pegboard).}
\label{fig:app_task_knots_untangle_examples}
\end{figure}

\section{The Use of Large Language Models}
\label{app:llm_use}

We used large language models (LLMs), including Google's Gemini-3.1-Pro and OpenAI's GPT-5.5, as auxiliary tools to assist with
writing, editing, and conducting the literature review for this
manuscript. All content was critically revised and fact-checked by the
human authors to ensure its scientific validity and originality. The
authors are fully responsible for all statements and conclusions
presented in this paper.

\clearpage
\section{Extended Set}
\label{app:extended_data}

\paragraph{Composition.}
The Core Set contains 3{,}900 instances, with 150 Easy and 150 Hard instances
for every task. The Extended Set preserves the complete original dataset of
11{,}035 instances across the same 13 tasks in the code repository, and its
instances keep the difficulty labels of their source generators. The two sets
share 2{,}250 exact model inputs. Core also contains inputs absent from
Extended. For the Extended results, we use the 8{,}780-instance evaluation
subset defined below. Table~\ref{tab:release_composition} in the main paper
gives its per-task sizes alongside Core.

\paragraph{Evaluation subset.}
Starting from the complete Extended Set, we exclude the 2{,}250 inputs that
appear in Core and five repeated Swap inputs. The resulting evaluation subset
has no exact model-input overlap with Core. This filtering defines the
reported evaluation and leaves the complete Extended dataset in the code
repository intact. For reasoning, input equality requires identical
whitespace-normalized question text and an identical ordered sequence of
image-file SHA256 hashes. For planning, it requires an identical saved initial
scene and reset configuration together with the same action budget. Generation
seeds and difficulty tags do not affect this comparison. Shared identifiers do
not establish duplicate inputs: 150 Origami identifiers refer to different
questions and images and remain in Core and the Extended evaluation subset
under distinct manifest IDs.

\paragraph{Evaluation results.}
Table~\ref{tab:extended_results} in the main paper reports results on the
filtered evaluation subset. We recompute each score after excluding Core
overlaps and the five repeated Swap rows. The table includes the 11 models
whose excluded rows are recoverable from per-instance logs or matched Core subset results.

\paragraph{Traceability.}
The paper's Core and Extended evaluation manifests record the task, instance
identifier, source record, record hash, and input hash. The Core manifest also
records the Easy or Hard level. The Extended evaluation manifest selects
8{,}780 instances from the complete Extended dataset. The removal manifest
records each excluded row and whether it matched Core or repeated another
Extended input. The result provenance file records
the integer counts and source hashes used to recompute Table~\ref{tab:extended_results}.

\end{document}